\documentclass[letterpaper]{article} % DO NOT CHANGE THIS
\usepackage{aaai2026}  % DO NOT CHANGE THIS
\usepackage{times}  % DO NOT CHANGE THIS
\usepackage{helvet}  % DO NOT CHANGE THIS
\usepackage{courier}  % DO NOT CHANGE THIS
\usepackage[hyphens]{url}  % DO NOT CHANGE THIS
\usepackage{graphicx} % DO NOT CHANGE THIS
\usepackage{natbib}  % DO NOT CHANGE THIS AND DO NOT ADD ANY OPTIONS TO IT
\usepackage{caption} % DO NOT CHANGE THIS AND DO NOT ADD ANY OPTIONS TO IT
\usepackage{algorithm}
\usepackage{algorithmic}

\usepackage{newfloat}
\usepackage{listings}
\DeclareCaptionStyle{ruled}{labelfont=normalfont,labelsep=colon,strut=off} % DO NOT CHANGE THIS
\floatstyle{ruled}
\newfloat{listing}{tb}{lst}{}
\floatname{listing}{Listing}
\usepackage{multirow} 
\usepackage{subcaption}
\usepackage{mathtools}

\usepackage{amsmath}
\usepackage{amssymb}
\usepackage{dsfont}
\usepackage{booktabs}

\title{Conformal Constrained Policy Optimization for Cost-Effective LLM Agents}
\author{
    Wenwen Si, Sooyong Jang, Insup Lee, Osbert Bastani
}
\affiliations{
    Department of Computer and Information Science, University of Pennsylvania\\
    3330 Walnut St, Philadelphia, PA 19104 USA\\
    \{wenwens, sooyong, lee, obastani\}@seas.upenn.edu
}

\begin{document}

\maketitle

\begin{abstract}
While large language models (LLMs) have recently made tremendous progress towards solving challenging AI problems, they have done so at increasingly steep computational and API costs. We propose a novel strategy where we combine multiple LLM models with varying cost/accuracy tradeoffs in an agentic manner, where models and tools are run in sequence as determined by an orchestration model to minimize cost subject to a user-specified level of reliability; this constraint is formalized using conformal prediction to provide guarantees. To solve this problem, we propose Conformal Constrained Policy Optimization (CCPO),
a training paradigm that integrates constrained policy optimization with off-policy reinforcement learning and recent advances in online conformal prediction. CCPO jointly optimizes a cost-aware policy (score function) and an adaptive threshold.
Across two multi-hop question answering benchmarks, CCPO achieves up to a 30\% cost reduction compared to other cost-aware baselines and LLM-guided methods without compromising reliability. Our approach provides a principled and practical framework for deploying LLM agents that are significantly more cost-effective while maintaining reliability.
\end{abstract}

\section{Introduction}

While large language models (LLMs) have made tremendous progress towards solving challenging tasks, they often require significant cost to do so. Human decision-makers can reduce cost by employing a meta-level strategy where they monitor their own uncertainty, seek targeted help, and choose to try again precisely when the potential benefit outweighs the cost of additional attempts. Bringing this capability to LLM agents is essential for cost-effective deployment. 
\begin{figure}[!t]
\centering
\includegraphics[width=0.42\textwidth]{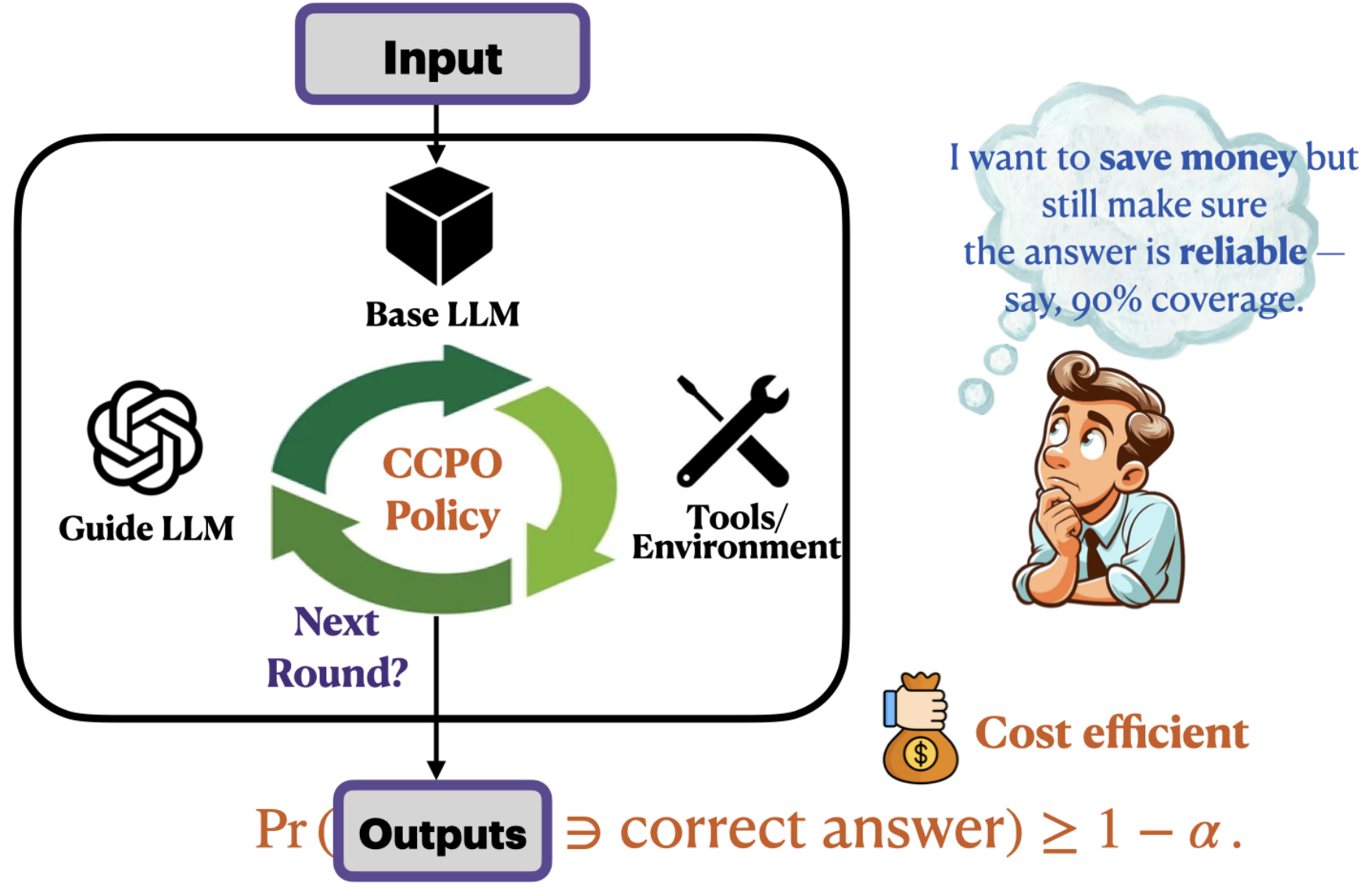}
\caption{We propose a framework for training a policy designed to orchestrate a fast, inaccurate LLM agent and a slow, accurate LLM to minimize cost while maintaining reliability (as formalized by conformal prediction).}
\label{fig:main}
\end{figure}
LLMs range from smaller LLMs that are cheap but inaccurate, to larger LLMs that can be far more accurate but significantly more costly. A natural strategy is to use a small, cheap LLM when it is confident, switching to the large LLM otherwise to ensure reliability. Beyond this simple strategy, we can consider more complex combinations of the two LLMs based on the intermediate results they produce.

Specifically, we consider an agentic setting, where a policy orchestrates a small LLM and a large LLM across a series of rounds to answer the question, with the goal of minimizing the expected cost subject to a user-specified reliability level. To formalize the reliability constraint, we use conformal prediction~\cite{shafer2008tutorial}, which provides theoretical guarantees by modifying the system to output sets of labels (called \emph{prediction sets}) instead of individual labels. Conformal prediction provides the \emph{coverage guarantee}, which says the prediction set contains the true label with high probability. In our setting, we can only cover the true answer if some orchestration strategy does so; thus, our reliability constraint says the following holds with high probability: if some sequence of actions  produces the correct answer, then our prediction set contains the true label.

Thus, the orchestration policy must select among LLM agents to satisfy the coverage guarantee while minimizing average cost (we implicitly constrain prediction set size). Prior work has addressed only parts of this goal. For instance, chain-of-thought prompting~\cite{wei2022chain}, ReAct~\cite{yao2023react}, debate protocols, and RL-based LLM agent frameworks~\cite{zhou2025sweet, bensal2025reflect} improve accuracy but ignore cost.
Alternatively, FrugalGPT~\cite{chen2024frugalgpt} reduces API expenses but does not guarantee reliability. Finally, uncertainty-aware methods~\cite{ren2023robots, han2024towards} calibrate model confidences but rely on ad-hoc techniques to perform model selection and do not seek to minimize cost. No existing approach unifies these desiderata.

We propose \emph{Conformal Constrained Policy Optimization (CCPO)}, which combines recent advances in online conformal prediction~\citep{angelopoulos2024online} with constrained policy optimization~\cite{achiam2017constrained}. Specifically, we seek to train a conformal policy that outputs a prediction set over actions to minimize cost subject to a conformal constraint; an optional penalty can also be imposed on prediction set size. However, training a set-based action policy is infeasible; instead, we convert the conformal policy into a stochastic one (i.e., uniform distribution over the action set) and train this policy instead. A key challenge is that the conformal policy visits a different state-action distribution compared to the stochastic policy; to correct for this shift, we use importance weighted V-trace targets and policy gradients~\cite{espeholt2018impala}. Finally, we apply online conformal prediction~\cite{angelopoulos2024online} to scale the action sets to guarantee coverage.

We empirically evaluate our approach on the free-form HotpotQA~\cite{yang2018hotpotqa} and multiple-choice MMLU~\cite{hendrycks2020measuring} datasets, showing that CCPO reduces total cost by up to 30\% while satisfying  target coverage guarantees, outperforming state-of-the-art cost-minimization baselines.

To summarize, our contributions are three-fold. First, we propose a formal framework for cost-effective and reliable LLM-agent deployment. Second, we apply V-trace off-policy corrections between the behavioral score function and the conformal-wrapped target policy, thereby sidestepping the exponential search over action sets. That is, rather than enumerating all set-valued mappings, we optimize stochastic surrogates that softly enforce both cost and coverage objectives and allow efficient gradient updates. Third, across diverse models and datasets, our method consistently achieves the lowest cost while meeting the target coverage level, demonstrating its practical effectiveness.

\section{Related Work}
    \paragraph{LLM Agents.} 
One way to improve the accuracy of LLM agents is to instruct them to perform reasoning. Chain-of-Thought (CoT) explicitly elicits step-by-step explanations \citep{wei2022chain}, and ReAct interleaves reasoning with tool calls \citep{yao2023react}. Extensions incorporate uncertainty quantification---e.g., UALA estimates predictive variance before committing to an answer \citep{han2024towards}. They also propose to reduce cost by making early-exit decisions in multi-turn chains \citep{lu2025runaway} and role-based multi-agent editing pipelines \citep{wan2025mamm}. 

Rather than using hand-written prompts, an alternative strategy is to train control policies. 
\citet{min2025self} first trains a Process Reward Model to score each intermediate thought and then learns a binary tabular policy---``request help'' vs. ``proceed''.
SWEET-RL \citep{zhou2025sweet} tackles multi-turn tasks by using external feedback as contextual states, yielding better credit assignment. \citet{bensal2025reflect} leverages a self-reflection mechanism for self-improvement. However, while these methods improve reliability or fluency, they do not provide any guarantees. In contrast, we aim to provide reliability guarantees using conformal prediction.

\paragraph{Safe \& Off-Policy RL.}  
There has been work on imposing safety constraints in reinforcement learning. TRPO constrains each update by a KL trust region \cite{schulman2015trust}; CPO augments this approach with an explicit cost bound for one-step feasibility \cite{achiam2017constrained}. Shielding methods block actions that violate specifications \cite{alshiekh2018safe}, while Lagrangian penalties trade reward and risk \citep{tessler2019reward}. \citet{bastani2021safe} combines a learned policy with a backup policy, sacrificing reward when necessary to ensure safety.

A complementary line of work tackles stable learning from off-policy data. Retrace clips importance ratios for stable Q-learning from replay \cite{munos2016safe}; IMPALA’s V-trace extends this to distributed actor–learner systems \cite{espeholt2018impala}. In purely offline data, CQL \cite{kumar2020conservative} and IQL \cite{kostrikov2021offline} suppress over-optimistic estimates with regularization.

\paragraph{Online Conformal Prediction.} 
Online conformal prediction extends classical, exchangeability‐based CP to sequential or adversarial data streams by tracking the conformity threshold, largely through online optimization. ACI \citep{gibbs2021adaptive} introduces an adaptive quantile estimator that preserves marginal coverage when the underlying distribution evolves. MVP \citep{bastani2022practical} achieves adversarially robust, group-conditional (``multivalid'') coverage efficiently with tight prediction sets. \citet{angelopoulos2024online} propose a weight-decayed quantile tracker that gives online conformal prediction ``best-of-both-worlds'' guarantees---$O(\sqrt{T})$ coverage regret for adversarial streams and near-optimal set sizes for i.i.d. data.

\paragraph{Learning Conformal Score Functions.} \citet{stutz2021learning} proposes a method to simulate split-conformal prediction within each mini-batch and backpropagate through the procedure, enabling the training of optimal score functions for classification problems. A final batch  calibration step restores the coverage guarantee, yielding tighter confidence sets than conventional post-hoc conformal prediction. However, optimizing the conformal score function remains unexplored in sequential settings such as ours.

\section{Cost-Effective LLM Agents}
We propose a collaborative LLM-agent framework where uncertainty-guided decisions govern interactions among multiple LLMs (possibly augmented with external tools) to solve a task. Specifically, we consider a question-answering task with question-answer pair distribution $(Q, Y^*) \sim \mathcal{D}$.

\paragraph{LLM Agent Orchestration.}
We assume given both (i) a \emph{base agent}, which is relatively cheap (e.g., a small open‐source LLM) but with weaker performance, and (ii) a \emph{guide agent}, which offers substantially better capabilities (e.g., more accurate answers, better reasoning skills) and uncertainty measurements at a higher cost. Intuitively, we want to switch to the guide agent if the base agent is uncertain. However, the base agent may not be effective at estimating its own uncertainty. Thus, we make use of the fact that input tokens tend to be substantially cheaper than output tokens, which enables us to have the guide agent assess confidence based on the base agent's reasoning trace. Specifically, we consider the following orchestration strategy: on each step (i) run the base agent to generate a reasoning trace and answer, (ii) run the guide agent to evaluate the base agent's generation and produce a correct answer, and (iii) have the policy to decide the next action (choose the base answer, guide answer, or continue for another round). Note that (ii) is cheap since it mostly uses input tokens and very few output tokens. This process is repeated until the policy chooses an answer or a maximum number of rounds is reached.

In more detail, the guide model is given the question along with the base model's reasoning chain and answer (for the current round), and outputs a binary judgment (yes/no) and corrected answer (without performing any reasoning itself). The base model's context includes the reasoning traces and answers from previous rounds and the guide model's corresponding output. When the base model is LLaMA-2-7B and the guide model is GPT-4o, we find that this strategy achieves comparable or even stronger performance than using GPT-4o with the chain-of-thought reasoning.

\paragraph{POMDP formulation.}
We formalize this strategy as a finite-horizon Partially Observable Markov Decision Process (POMDP) with $T$ steps. At each step, we run the base model on the context so far, and the policy $\pi$ receives observation $o_t \in \mathcal{O}$, which includes the base model's context, the guide model's judgment and uncertainty, the index of the current round, and cumulative guide model token usage. Then, the policy selects an action $a_t \in \mathcal{A}$, where
%\WS{8pt too wide}
\begin{align*}
\mathcal{A} = \{\text{guide answer}, \text{base answer}, \text{next round}\}.
\end{align*}
This process continues until the policy chooses an answer, at which point the episode terminates. The goal is to learn a policy that achieves a (i) \textit{coverage guarantee}---i.e., a set of answers generated by the orchestration process contains the true answer if there is some sequence of actions that produces the true answer, and (ii) \emph{cost minimization}---i.e., minimize the total monetary cost (e.g., API usage fees). 
To this end, we introduce the \emph{conformal policy} $C:\mathcal{O}\to2^{\mathcal{A}}$, which maps an observation $o_t\in\mathcal{O}$ to an action set $A_t=C(o_t)\subseteq\mathcal{A}$. Rather than producing a single rollout, this strategy produces a set of rollouts by taking every possible action $a\in A_t$, and recursively collecting rollouts across each. Since only the action ``next round'' results in additional choices, the maximum number of rollouts is $2T$ (where $T$ is the horizon), so we do not suffer exponential blowup.

Next, let $\mathcal{Y}(q)$ denote the set of answers that can arise from all possible action sequences---i.e., letting the base and guide answers on round $t$ be $\hat{y}_{\text{base}}^t , \hat{y}_{\text{guide}}^t \in \mathcal{Y}$, respectively, where $\mathcal{Y}$ is the answer space (here, we assume the policy proceeds to round $T$), then $\mathcal{Y}(q) = \{\hat{y}_{\mathrm{base}}^t(q)\}_{t=1}^T \cup \{\hat{y}_{\mathrm{guide}}^t(q)\}_{t=1}^T$.
Now, we aim to solve the following:
\begin{equation}\label{eq:goal}
\begin{aligned}
&\min_{C} \mathds{E}[J (C, Q ) + \lambda |C(Q)|] \\
&\text{s.t.}~\Pr[\mathds{1}\{Y^*\in C(Q) \vee Y^* \not\in \mathcal{Y}(Q)\}]\geq 1-\alpha,
\end{aligned}
\end{equation}
where $J(C, q)$ denotes the (random) cumulative reward 
of $C$ given question $q$ (in our setting, it is the cumulative API cost incurred in that episode),
$C(Q)\in2^{\mathcal{Y}}$ denotes the (random) set of answers obtained by applying $C$ to $Q$,
$\alpha$ is the user-specified reliability level,
and $\lambda$ is a hyperparameter. The objective optionally includes the added term $\lambda|C(Q)|$ to impose a penalty on larger prediction sets; since the structure of our problem imposes the constraint $|C(Q)|\le2T$ (where $T$ is the horizon), we can also take $\lambda=0$.
The constraint says that with high probability, if $Y^*\in\mathcal{Y}(Q)$, then $Y^*\in C(Q)$ as well---i.e., if there is any chance of getting the correct answer $Y^*$, then the prediction set $C(Q)$ includes $Y^*$.

\section{Conformal CPO}
We propose Conformal Constrained Policy Optimization (CCPO) to solve Eq.~\eqref{eq:goal}.
The key challenge is that directly searching over all conformal policies $C$ is challenging due to the combinatorial action space $2^{\mathcal{A}}$. Instead, we parameterize $C$ by a stochastic policy $\pi(a\mid o)$ and a threshold $\kappa \in [0,1]$; the corresponding conformal policy is
\begin{equation}\label{eq:conf}
C_{\pi, \kappa}(o)=\{a\in\mathcal{A}:\pi(a\mid o)\ge\kappa\}.
\end{equation}
Then,
by updating $\kappa$ using online conformal prediction and assuming $\pi$ converges, we guarantee that $C_{\pi, \kappa}$ will asymptotically achieve the desired $1-\alpha$ coverage. 

To update $\pi$, we adapt the trust‐region method from CPO~\cite{achiam2017constrained}, yielding consistent improvement in our objective while maintaining our coverage constraint. Taking into account the KL‐ball search in CPO and the contractive  validity of V-trace (see Section~\ref{sec:discussion}),
we optimize the stochastic conformal policy $S_{\pi,\kappa}$ as the target policy. Concretely, for any score function $\pi$ and threshold $\kappa$, we define
\begin{equation}\label{eq:sconf}
S_{\pi, \kappa}(a \mid o) = 
\frac{\mathds{1}\{\pi(a \mid o) \geq \kappa \}}{\sum_{a'\in\mathcal{A}} \mathds{1}\{\pi(a' \mid o) \geq \kappa\}  } 
\end{equation}
i.e., $S_{\pi,\kappa}$ places uniform probability over the actions in the conformal set $C_{\pi, \kappa}(o)$. 
We then employ an actor–critic architecture in which both the value function $V$ and constraint function $V_C$ (i.e., the value function with the reward replaced by the constraint value),
 and the policy updates are computed with respect to the stochastic conformal policy $S_{\pi,\kappa}$. 

\paragraph{Critic update.} To bridge the gap between $\pi$ (which is used to collect trajectories) and $S_{\pi,\kappa}$, we employ V‐trace off‐policy correction as in \citet{espeholt2018impala}. At each iteration, rollouts under $\pi$ produce tuples $(o_t,a_t,r_t,c_t)$, where $r_t$ and $c_t$ are the reward and constraint values on step $t$, respectively. In our POMDP in Eq.~\eqref{eq:goal}, we have reward values
\begin{align*}
r_t = J_t(C,q) + \mathds{1}\{t=t_f\} \cdot \lambda |C(q)|,
\end{align*}
where $J_t(C,q)$ is the API cost on step $t$ and $t_f$ is the step on which the policy $\pi$ selects an answer, and constraint values
\begin{align*}
c_t = \mathds{1}\{t=t_f\} \cdot \mathds{1}\{y^*\in C(q) \vee y^* \not\in \mathcal{Y}(q)\}.
\end{align*}
Now, given $\bar\rho>0$, we define truncated importance weights
\begin{align*}
\rho_t = \min\left\{ \bar{\rho}, \frac{S_{\pi, \kappa}(a_t \mid o_t)}{\pi(a_t \mid o_t)} \right\}
\end{align*}
to correct for the discrepancy between $\pi$ and $S_{\pi,\kappa}$. We use $\bar{\rho} = 1$, which clips large weights but not small ones; see Section~\ref{sec:discussion} for a discussion of our choice. Then, the V‐trace target for the value function estimate $V_{\theta}$ is
\begin{align}
\label{eq:impala}
v_t &= V_\theta(o_t) + \delta_t V + \rho_t (v_{t+1} - V_\theta(o_{t+1})) \\
\delta_t V &= \rho_t (r_t + V_\theta(o_{t+1}) - V_\theta(o_t)), \nonumber
\end{align}
and update $\theta$ via gradient descent on the $L_2$ loss to $v_t$:
\begin{equation}
\label{eq:gradientdescent}
\theta\gets\theta-\sum_{t=1}^T\nu(v_t - V_\theta(o_t))\nabla_\theta V_\theta(o_t),
\end{equation}
where $\nu$ is the learning rate. We apply the same gradient update rule to train a constraint function estimate $V_{C,\phi}$, where $r_t$ is replaced by $c_t$.
In addition, for each sampled action, we obtain the advantage function
\begin{equation}
\hat A^{S_{\pi,\kappa}}_t=r_t+ v_{t+1}-V_\theta(o_t),
\end{equation}
and analogously for the constraint advantage function $\hat A^{S_{\pi,\kappa}}_{C,t}$ by replacing $r_t$ by $c_t$ and $V_{\theta}$ with $V_{C,\phi}$.

\begin{algorithm}[t]
\caption{Conformal Constrained Policy Optimization}
\textbf{Input}: Initial behavior policy $\pi_0$, initial threshold $\kappa_0$, user-specified reliability level $\alpha$, dataset $D = \{(q_i, y_i)\}_i$. \\
%\textbf{Hyper-parameters}: Critic learning rates $r_c, r_v$, decay rate $\epsilon$. \\
\textbf{Output}: Conformal policy $C_{\pi, \kappa}$.
\begin{algorithmic}[1]
\FOR{$k\in\{0,1,...\}$}
\STATE Sample $(q, y) \sim D$ and collect rollouts with $\pi$
\STATE Update value/constraint function estimates via Eq.~\eqref{eq:gradientdescent}
\STATE Compute $\pi_{k+1}$ via Eq.~\eqref{eq:policyupdate}
\STATE Compute $\kappa_{k+1}$ via Eq.~\eqref{eq:ccpo_decay}
\ENDFOR
\end{algorithmic}
\end{algorithm}

\paragraph{Policy Update.} Next, we describe our trust–region policy update. We let $S_{k}=S_{\pi_k,\kappa_k}$ and $C_{k}=C_{\pi_k,\kappa_k}$ denote the stochastic and conformal policies, respectively, on the $k$th iteration of optimization.
Then, as in \citet{achiam2017constrained}, we iteratively solve the following approximation of Eq.~\eqref{eq:goal} (we describe how $\kappa$ is updated later):
\begin{alignat}{3}
\label{eq:policyupdate}
&\mathrlap{\min_{\pi} \mathds{E}\left[\sum_{t=1}^T \hat{A}^{S_{\pi,\kappa}}_t\right]} \\
&\text{s.t.}~&&
\bar{J}_{C}^{\pi,\kappa}\ge 1 - \alpha,
\quad D_{\mathrm{KL}}(S_{\pi, \kappa}\|S_k)\le\delta.
\nonumber
\end{alignat}
Here, we have used the fact that our original objective is equivalent to the sum of advantages $A_t^{S_{\pi,\kappa}}$; furthermore, the constraint is replaced by a constraint $\bar{J}_C^{\pi,\kappa}\ge1-\alpha$ and a KL constraint that keeps the new policy close to the current one, where
$\bar{J}_C^{\pi,\kappa}$ is an upper bound on the constraint value:
\begin{align*}
\bar{J}_C^{\pi,\kappa}\ge J_C^{\pi,\kappa}=\Pr[Y^*\in C_{\pi,\kappa}(Q) \vee Y^* \not\in \mathcal{Y}(Q)].
\end{align*}
To derive $\bar{J}_C^{\pi,\kappa}$, first note that
\begin{align*}
J_C^{\pi,\kappa}\le\Pr[Y^*\in C_{\pi,\kappa}(Q)]+\Pr[Y^*\notin\mathcal{Y}(Q)].
\end{align*}
The second term is easy to estimate; for the first, we have
\begin{align*}
&\Pr[Y^*\in C_{\pi,\kappa}(Q)] \\
&\le\mathds{E}_S\left[\left(\prod_{t=1}^T|C_{\pi,\kappa}(o_t)|\right)\mathds{1}\{S_{\pi,\kappa}(Q)=Y^*\}\right] \\
&\approx\mathds{E}_\pi\left[\left(\prod_{t=1}^T\rho_t\cdot|C_{\pi,\kappa}(o_t)|\right)\mathds{1}\{\pi(Q)=Y^*\}\right]
\end{align*}
where $S_{\pi,\kappa}(Q)$ is the random answer obtained by using stochastic policy $S_{\pi,\kappa}$ in the POMDP on question $Q$; the approximation in the second line comes from the fact that we are clipping our importance weights $\rho_t$. Thus, we use
\begin{align*}
\bar{J}_C^{\pi,\kappa}&= \bar{J}^{\pi,\kappa}_{C,0} +\Pr[Y^*\notin\mathcal{Y}(Q)] \\
\bar{J}^{\pi,\kappa}_{C,0}&=\mathds{E}_\pi\left[\left(\prod_{t=1}^T\rho_t\cdot|C_{\pi,\kappa}(o_t)|\right)\mathds{1}\{\pi(Q)=Y^*\}\right].
\end{align*}
As in CPO, we then compute the optimal Lagrange multiplier for the constraint, and use conjugate gradient descent to solve for the natural gradient step that satisfies the KL radius $\delta$; the resulting step gives the new policy $\pi_{k+1}$.

A remaining caveat is that the definition of $S_{\pi,\kappa}$ in Eq.~\eqref{eq:sconf} involves the indicator function, which is non-differentiable. Thus, we
replace it with a smooth sigmoid approximation; given $\epsilon>0$, we instead use a softmask function
\begin{equation}\label{eq:app}
\text{softmask}(a,o;\kappa)
=
\sigma\left(\frac{\pi(a\mid o)-\kappa}{\epsilon}\right),
\end{equation}
where $\sigma(z)=(1+e^{-z})^{-1}$ is the sigmoid function. As $\epsilon\to0$, $\text{softmask}(a,o;\kappa)$ recovers the binary indicator.

\paragraph{Threshold Calibration.} The threshold $\kappa$ is updated after each episode to provably maintain the coverage guarantee without relying on convergence of the policy. When solving for $\pi_{k+1}$ in Eq.~\eqref{eq:policyupdate}, we fix $\kappa=\kappa_k$. Afterwards, we update $\kappa_k$ based on the coverage of $C_{\pi_{k+1},\kappa_k}$ using the algorithm in  \citet{angelopoulos2024online}, that $\kappa_{k+1}$ is
\begin{equation}\label{eq:ccpo_decay}
 \kappa_k + \eta_k \left(1 - \mathds{1}\{Y^*\in C_{\pi,\kappa}(Q)\vee Y^*\not\in\mathcal{Y}(Q)\} - \alpha\right).
\end{equation}
Assuming  
$\sum_t\eta_t=\infty$ and $\sum_t\eta_t^{2}<\infty$, then \citet[Theorem 4 and Corollary 2]{angelopoulos2024online} show that in the i.i.d. setting, the coverage converges to $1-\alpha$ even though $\pi$ is being updated in a streaming fashion, as long as it converges.  If additionally $\eta_t \propto t^{-1/2 - \xi}$ for some $\xi \in (0, 1/2)$, then in the adversarial setting the long-run coverage error is bounded as $O(T^{-1/2 + \xi})$ \citep[Theorem~1]{angelopoulos2024online}. While we guarantee coverage for the final policy $\pi_K$ assuming $\pi_k$ converges, we perform a batch calibration on $\pi_K$ (i.e., use traditional conformal prediction~\citep{shafer2008tutorial} on a held-out calibration set to select a final value $\kappa_{K+1}$) to ensure coverage for $\pi_K$ even without this assumption.

\section{Experiments}

\subsection{Experimental Setup}

\paragraph{Environment.}
We set the horizon to $T=4$ since questions that cannot be solved in this time are generally too difficult. For conformal prediction, we use $\alpha \in \{0.1, 0.2\}$.

\paragraph{Training.}
All our policy networks are three-layer neural networks with 64 hidden units per layer. We train them from scratch for 1500-2000 steps, using a learning rate of $10^{-3}$ and batch size of 10. We set the KL divergence constraint to $\delta=0.01$, soft masking parameter $\epsilon=0.01$, and conformal decay step $\xi=0.1$. For optimizing $\kappa$, we perform a grid search over the threshold with a granularity of $10^{-6}$.

\paragraph{Datasets.} We evaluate our approach on two challenging question‐answering benchmarks. HotpotQA~\cite{yang2018hotpotqa} is a commonsense QA dataset requiring multi‐hop reasoning and often multiple Wikipedia searches to answer each question. MMLU~\cite{hendrycks2020measuring} is a multiple‐choice QA dataset spanning academic and professional subjects such as biology, mathematics, and physics. 
Following UALA~\cite{han2024towards}, we use Wikipedia search as a tool for HotpotQA. For MMLU, UALA uses the 
Google API as a tool. However, the cost of Google API is \$5 per 1,000 calls, which far exceeds our budget; thus, we apply CoT with deterministic decoding (temperature 0.0) in the first step and temperature 1.0 in the subsequent steps. For HotpotQA, we train on 1,000 examples, while for MMLU, we train on 560 examples. For both, we use 200 examples for batch calibration and 200 for testing.

\paragraph{Models.} We consider both LLaMA-2-7B (8-bit) and LLaMA-3.2-3B as base models, and use GPT-4o as the guide model. However, because LLaMA-3.2-3B achieves notably stronger performance in some cases, we deliberately set a stricter target ($\alpha = 0.05$) for it to avoid trivial results.

\begin{table}[!t]
\centering\small\setlength{\tabcolsep}{2pt}
\begin{tabular}{ccccccc}
\toprule
Policy    & \begin{tabular}[c]{@{}c@{}}Base-\\Guide \end{tabular} & $\alpha$-aware & \begin{tabular}[c]{@{}c@{}}Conf-\\ormal \end{tabular} & \begin{tabular}[c]{@{}c@{}}Point-\\wise \end{tabular} & RL & \begin{tabular}[c]{@{}c@{}}LLM\\Policy\end{tabular} \\ \hline
LLM-EXIT       &       &     &       &   \checkmark    &            &   \checkmark                          \\ \hline
Random       &      \checkmark      &       &              &           \checkmark        &      &   \\ \hline
\begin{tabular}[c]{@{}c@{}}GPT-4o/\\ LLaMA\end{tabular}        &     \checkmark       &   \checkmark    &           &      \checkmark   &            &                   \checkmark           \\ \hline
UALA          &      \checkmark      &       &         &           &        &          \\ \hline
CPO          &    \checkmark        &     \checkmark  &         &    \checkmark     &  \checkmark           &   \\ \hline
\begin{tabular}[c]{@{}c@{}}CPO batch/\\ online\end{tabular}   &     \checkmark       &     \checkmark  &        \checkmark      &         & \checkmark  &     \\ \hline
CCPO (ours)            &     \checkmark       &    \checkmark   &   \checkmark     &  &  \checkmark        &     \\ \bottomrule
\end{tabular}
\caption{Summary of key features of each approach.}\label{tbl:feature}
\end{table}

\begin{table}[!t]
\centering\small\setlength{\tabcolsep}{1pt}
\begin{tabular}{ccccc}
\toprule
Policy & Cost (cents) & Coverage & Avg. Len. & Set Size\\
\midrule
GPT-4o EXIT  & 827.0 &0.908 & 2.39 & 2.39 \\ 
LLaMA-2 EXIT & 0.000 &0.653 & 2.55 & 2.55 \\ \midrule
Random   & 6.811         & 0.578          & 1.44         & 1.00  \\ 
GPT-4o     & 18.65 & 0.780 &  1.20   &  1.00   \\ 
LLaMA-2-7B    & 4.551 & 0.675 &   1.03   &  1.00 \\ 
UALA   &    9.153        &    0.923            &        2.00       &     2.00     \\ 
CPO & 4.704 & 0.832 & 1.004 & 1.00 \\ \midrule
CPO batch &  7.835        & 0.905 & 1.70          & 2.19     \\ 
CPO online &  7.484        & 0.897 & 1.58          & 2.38   \\ 
CCPO ($\lambda=0$)    & \textbf{6.552}  & 0.902          & \textbf{1.34} & 2.35     \\
CCPO ($\lambda=2e^{-4}$)   & 7.026  & 0.903          & 1.48 & 2.22     \\ 
\bottomrule
\end{tabular}
\caption{HotpotQA results with LLaMA-2-7B, $\alpha = 0.1$.}\label{tbl:hot-llama2-01}
\end{table}

\begin{figure}[!t]
\centering
\includegraphics[width=0.43\textwidth]{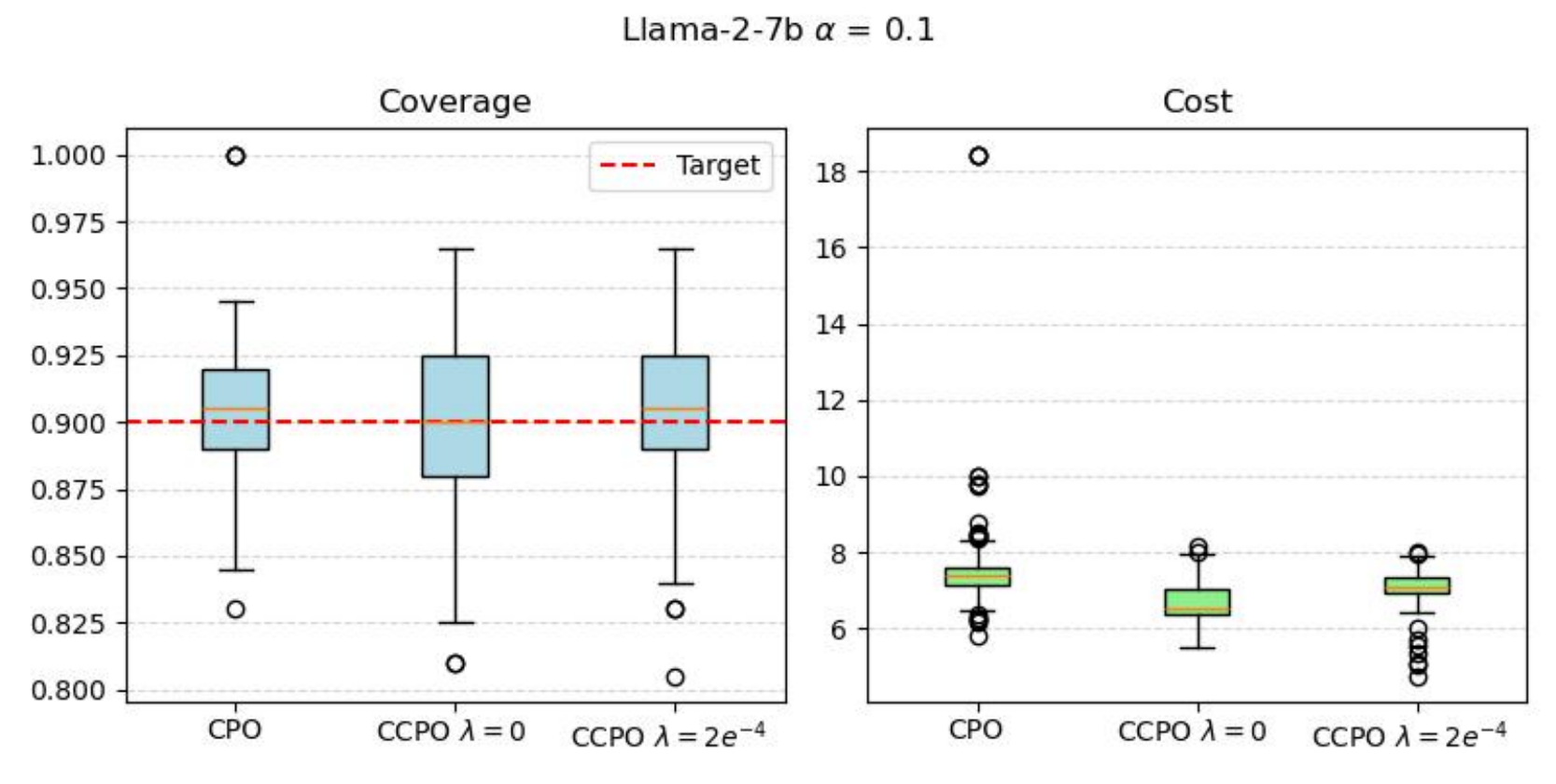}
\caption{HotpotQA coverage and cost with LLaMA-2-7B base model over 100 splits, $\alpha = 0.1$.}
\label{fig:hot-llama2-01}
\end{figure}

\paragraph{Metrics.}

We evaluate the performance of our method using four primary metrics. ``Cost'' measures the cumulative cost of API usage over the test set. ``Coverage'' measures the coverage rate $\Pr[Y^* \in C(Q) \vee Y^* \notin \mathcal{Y}(Q)]$. ``Avg. Len.'' is the average episode length (i.e., the average number of steps taken before the policy chooses an answer; for conformal policies, it is the maximum length across branches). ``Set Size'' is the average prediction set size $|C(q)|$.

\paragraph{Baselines.}

We consider two kinds of baselines: LLM agents and CPO methods. First, we summarize our LLM agent baselines (see Appendix~\ref{app:temp} for prompt templates):
\begin{itemize}
\item Random policy: We apply a uniform policy $\pi$ that randomly selects among the three actions in $\mathcal{A}$.
\item LLM policy: We replace the parameterized policy $\pi$ with an LLM, supplying it with the question, each model’s answer, and uncertainty scores derived from GPT-4o. We consider using both of the LLaMA models as well as GPT-4o; these are denoted as GPT-4o, LLaMA-3.2-3B, and LLaMA-2-7B, in our results.
\item LLM-EXIT~\cite{lu2025runaway}: This baseline uses a single LLM both to provide answers and to decide whether to proceed to the next round. \footnote{For exit policies, we report \textit{matched coverage} by summing the percentages of correct and unsolvable questions, ensuring a fair comparison.}
\item GPT-4o-guided UALA: In  UALA \citep{han2024towards}, both uncertainty estimates and answers are generated by the same model, which tend to be biased and inaccurate for smaller models. To align with our approach, we adopt GPT-4o to quantify uncertainty. Then, this baseline follows the same steps as our approach, except instead of using a learned policy, it applies fixed thresholds on uncertainty to choose the action.
\end{itemize}
Next, our policy learning baselines use CPO to solve
\begin{equation*}
\begin{aligned}
&\min_{\pi}  \mathds{E}[J(\pi,Q)] \\
&\text{s.t.}~
\Pr[ Y^*\in \pi(Q) \vee Y^* \not\in \mathcal{Y}(Q)] \ge 1-\alpha,
\end{aligned}
\end{equation*}
with the following variations:
\begin{itemize}
\item CPO: Vanilla CPO, where we train a stochastic policy using the CPO algorithm without any conformal prediction.
\item CPO batch: Vanilla CPO, but perform batch conformal prediction on the held-out calibration set to select a value of $\kappa$ for the final stochastic policy to obtain a conformal policy that guarantees coverage.
\item CPO online: Vanilla CPO, but where we use online conformal prediction to convert the stochastic policy to obtain a conformal policy that guarantees coverage (using the same episodes as in CCPO training).
\end{itemize}

Table~\ref{tbl:feature} summarizes key features of the different approaches: ``Base-Guide'' indicates whether it uses our strategy for composing a base and a guide LLM, ``$\alpha$-aware'' indicates whether it is given the user-specified reliability level $\alpha$, ``Conformal'' indicates whether it provides a coverage guarantee, ``Pointwise'' indicates whether it outputs a single label (instead of prediction set), ``RL'' indicates whether it uses reinforcement learning to train a policy, ``LLM Policy'' indicates whether it uses a prompted LLM as the policy.
Compared to CPO online/batch, CCPO (our approach) tightly integrates conformal prediction into CPO.

\subsection{Results and Discussion}

\paragraph{Results.}

\begin{figure}[!t]
    \centering
    \includegraphics[width=0.43\textwidth]{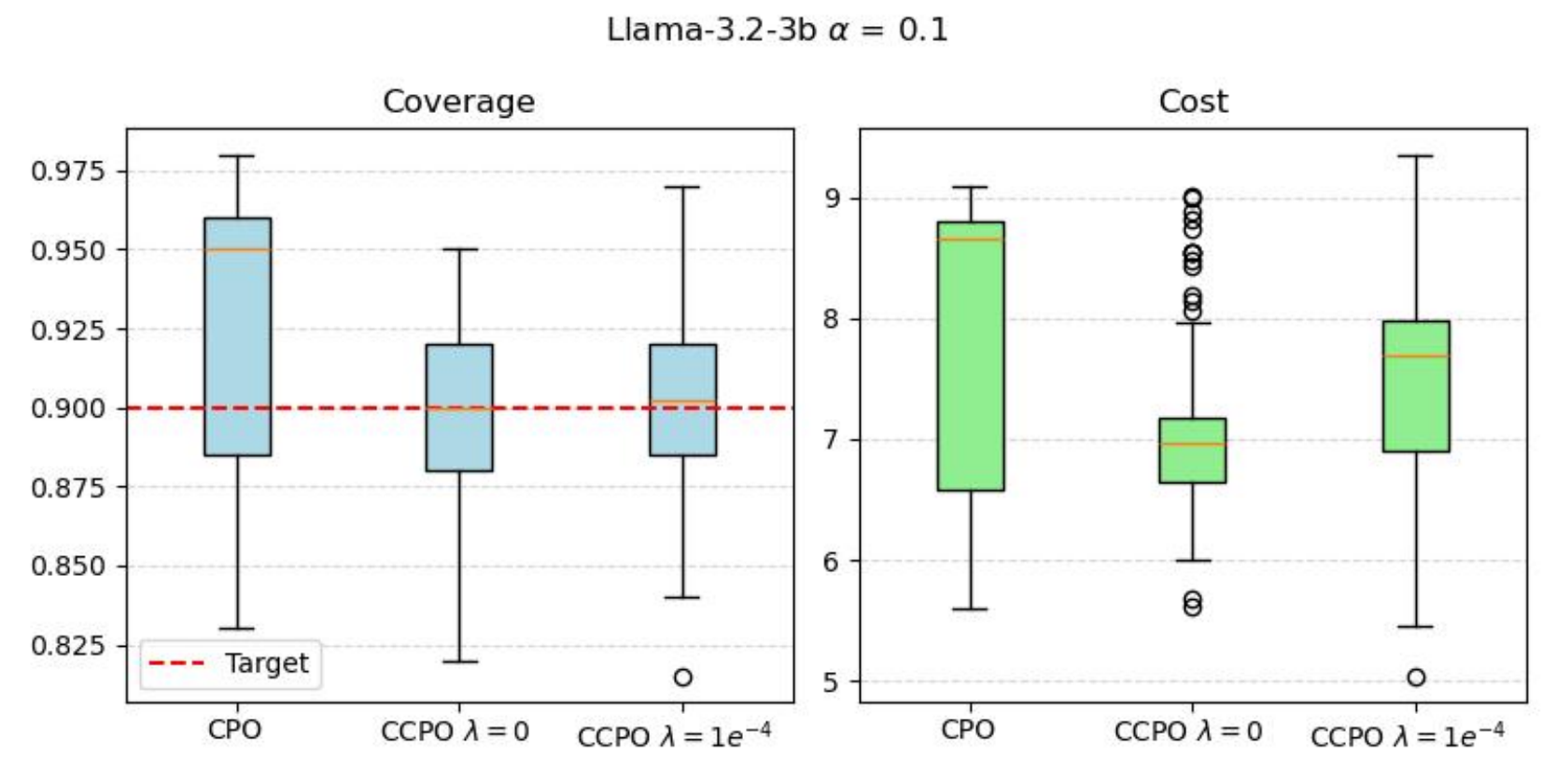}
    \caption{HotpotQA coverage and cost with LLaMA-3.2-3b base model over 100 splits, $\alpha = 0.1$.}
    \label{fig:hot-llama3-01}
\end{figure}

\begin{figure}[!t]
    \centering
    \includegraphics[width=0.43\textwidth]{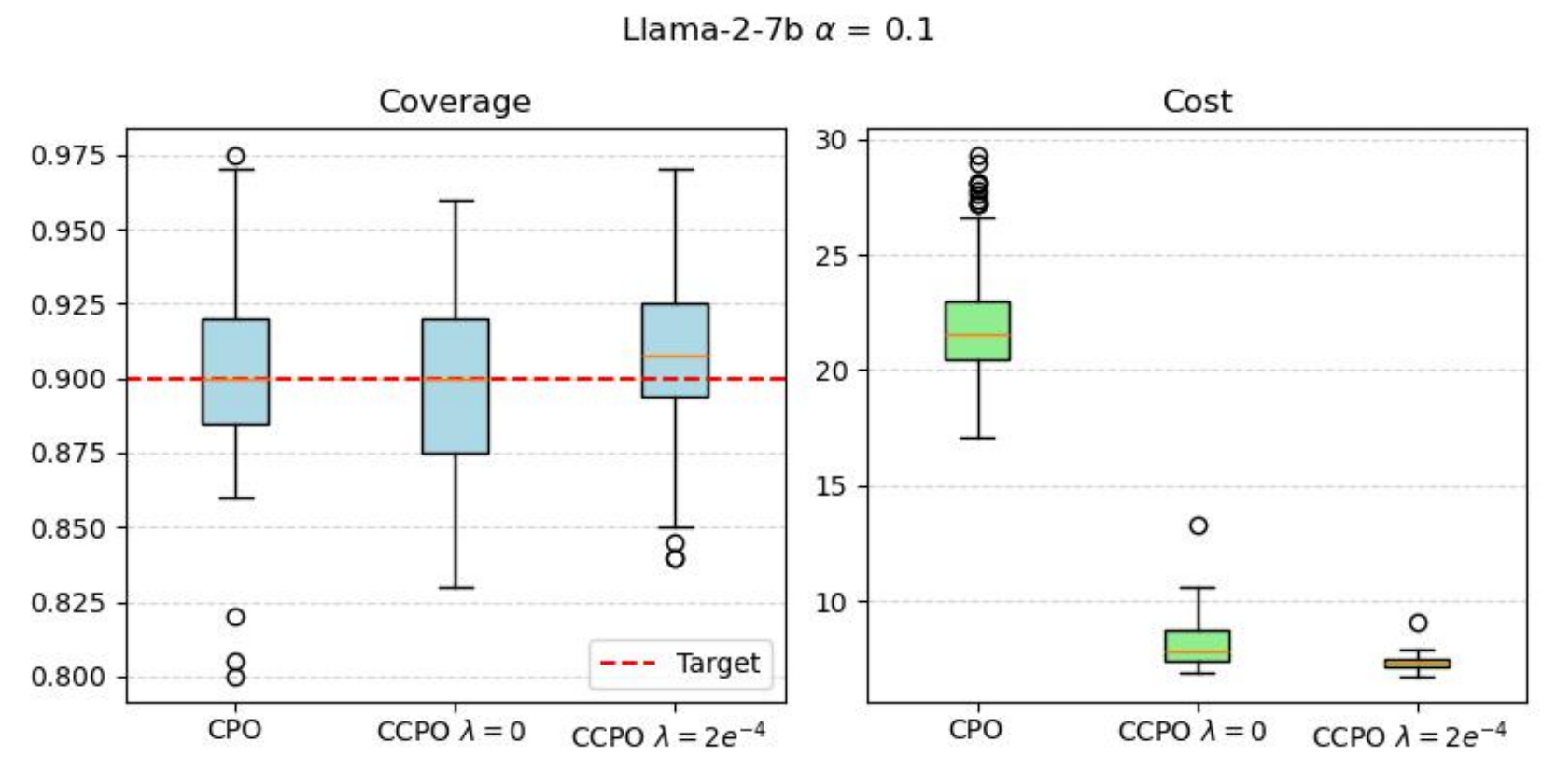}
    \caption{MMLU coverage and cost with LLaMA-2-7B base model over 100 splits, $\alpha = 0.1$.}
    \label{fig:mmlu-llama2-01}
\end{figure}

\begin{figure}[!t]
    \centering
        \includegraphics[width=0.43\textwidth]{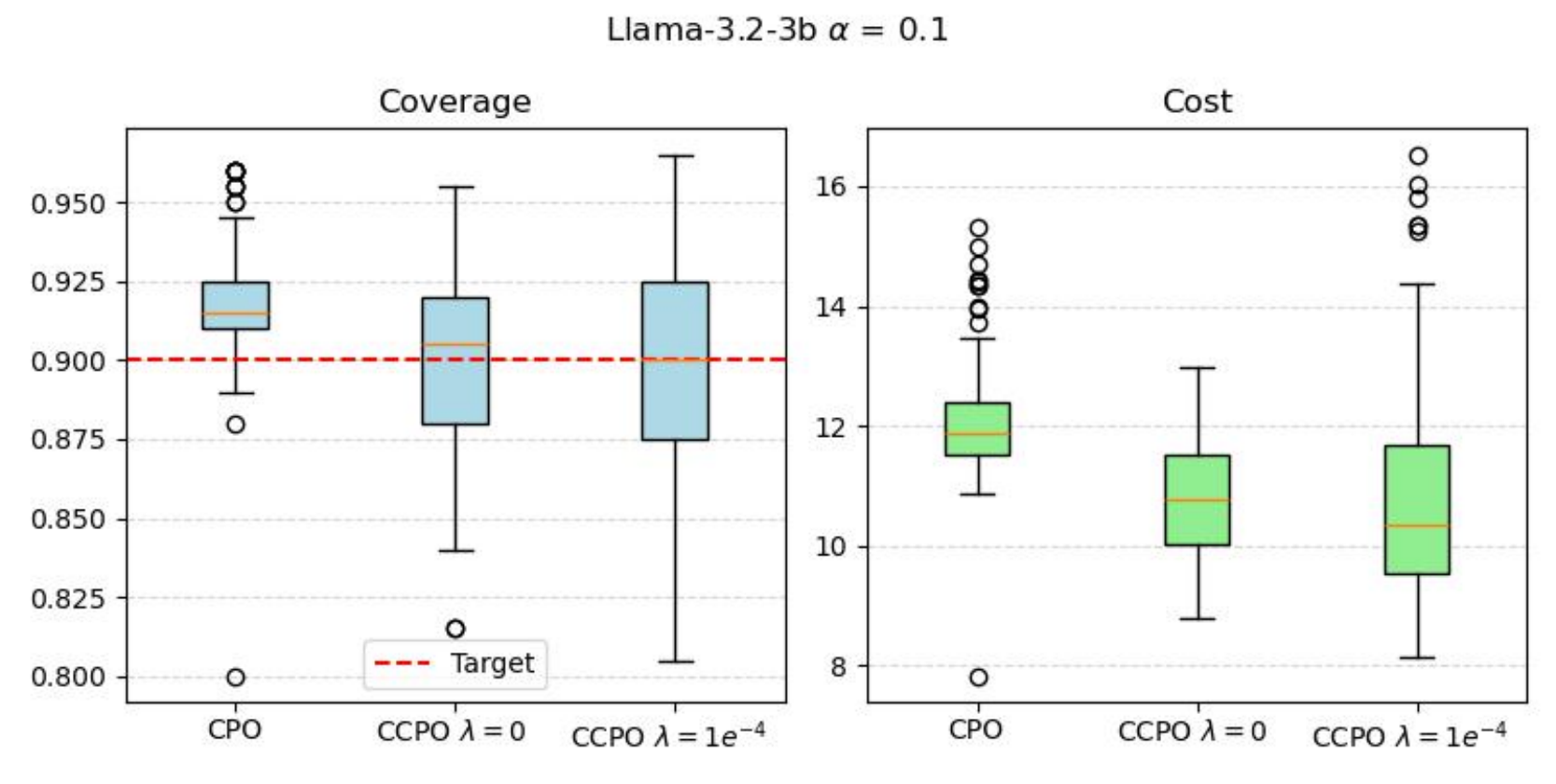}
    \caption{MMLU coverage and cost with LLaMA-3.2-3b base model over 100 splits, $\alpha = 0.1$.}
    \label{fig:mmlu-llama3-01}
\end{figure}

Our results on both HotpotQA and MMLU are shown in Tables~\ref{tbl:hot-llama2-01}, \ref{tbl:hot-llama3-01}, \ref{tbl:mmlu-llama2-01}, \&~\ref{tbl:mmlu-llama3-01}; in addition, results for an alternate choice of $\alpha$ are provided in Tables~\ref{tbl:hot-llama2-02}, \ref{tbl:hot-llama3-005}, \ref{tbl:mmlu-llama2-02}, \&~\ref{tbl:mmlu-llama3-02} in Appendix~\ref{app:more-res}. 
The first two methods do not use base-guide composition, the next five use this strategy but do not provide conformal guarantees, and the last four use this strategy and provide conformal guarantees. We also show box plots of coverage and cost for both CPO and CCPO, computed across 100 random evaluation splits, in Figures~\ref{fig:hot-llama2-01}, \ref{fig:hot-llama3-01} \ref{fig:mmlu-llama2-01}, \&~\ref{fig:mmlu-llama3-01}, and for an alternate choice of $\alpha$ in Figures~\ref{fig:hot-llama2-02}, \ref{fig:hot-llama3-005}, \ref{fig:mmlu-llama2-02}, \&~\ref{fig:mmlu-llama3-02} in Appendix~\ref{app:more-res}.

\paragraph{Discussion.}

First, cost-aware policy optimization efficiently minimizes cost while maintaining the desired coverage, achieving orders of magnitude cost reduction compared to LLM-based methods with similar coverage. The uncertainty-thresholding method with UALA is also effective, consistent with the findings of \citet{chen2024frugalgpt}. Nevertheless, policy optimization methods still offer a clear performance advantage, as they explicitly incorporate cost information and can potentially leverage additional contextual features for even stronger decision-making.

On the other hand, pointwise methods, whether LLM-guided or RL, fail to achieve higher coverage. Moreover, LLM-guided approaches built on small base models offer only limited control over agent collaboration and provide marginal improvements over a random policy. It is worth noting that GPT-4o serves as a strong judgment model, but inevitably incurs significantly higher inference cost.

Next, CCPO achieves the lowest cost at the desired coverage level. Its advantage is clear, even when compared to variants of CPO-based policy training. Notably, the CPO batch/online baselines, our strongest competitors, incorporate trust-region optimization to jointly enforce coverage and minimize cost with guarantees. Yet, CCPO reduces cost by 12\% to 27\% compared to this baseline. Additionally, CCPO outperforms the GPT-guided UALA on optimal batch-mode uncertainty calibration, further demonstrating its strong cost efficiency under coverage constraints.

\begin{table}[!t]
\centering\small\setlength{\tabcolsep}{1pt}
\begin{tabular}{ccccc}
\toprule
Policy &  Cost (cents) & Coverage & Avg. Len. & Set Size\\
\midrule
GPT-4o EXIT & 827.0 &0.908 & 2.39 & 2.39 \\ 
LLaMA-3 EXIT  & 0.000 &0.718 & 4.00 & 4.00 \\ \midrule
Random   & 1.411          & 0.61    & 1.51          &  1.00  \\ 
GPT-4o   & 18.09 & 0.850 & 1.15        &  1.00    \\ 
LLaMA-3.2-3b   & 4.449 & 0.615 & 1.00  &  1.00  \\ 
UALA    &       8.012         &    0.923     &        2.00       &   2.00       \\
CPO & 4.655 & 0.829 & 1.00 & 1.00 \\ \midrule
CPO batch &   8.007   & 0.932     & 1.79        & 3.58     \\ 
CPO online &   8.829   & 0.875  &  1.856     &  3.11 \\ 
CCPO ($\lambda=0$)    & \textbf{7.061} & 0.901     & \textbf{1.59} & 3.18     \\
CCPO ($\lambda=1e^{-4}$)  &  7.397   & 0.902 & 1.66 &  \textbf{2.78}  \\
\bottomrule
\end{tabular}\caption{HotpotQA results with LLaMA-3.2-3b, $\alpha = 0.1$.}\label{tbl:hot-llama3-01}
\end{table}

Finally, CCPO produces reasonable prediction set sizes. While cost efficiency and coverage are our primary goals, it is also important that the resulting answer sets remain non-trivial and informative. Compared to the CPO variants, CCPO achieves similar answer set sizes while offering significantly greater cost efficiency. Notably, even when compared to the more costly GPT-4o-based methods, CCPO maintains comparable set sizes, further underscoring the informativeness and quality of the learned policy.

\paragraph{Comparison to CPO.}
We find that CPO often fails to find an optimal policy, especially at higher coverage levels. Moreover, when CPO already struggles to satisfy the coverage constraint, it becomes particularly sensitive to issues such as noisy or limited data, which can lead to severe divergence during training. In contrast, CCPO aims for a conformal policy, allowing multiple actions to be taken at each stage and thus flexibly achieving any desired coverage level. Empirically, we also observe that CCPO delivers more stable training performance under these data challenges (see Appendix~\ref{sec:curves} for training curves).

Furthermore, since CPO optimizes for pointwise performance, successful training often produces sharp action distributions with extreme confidence. This can lead to overfitting, making the policy sensitive to noise and difficult to calibrate. In contrast, CCPO simultaneously updates the threshold $\kappa$, resulting in a more separable policy geometry and avoiding the abrupt coverage jumps often seen during calibration on sequential data. As a consequence, CCPO achieves tighter and more consistent coverage performance.

%%%%% MMLU Results %%%%
\begin{table}[!t]
\centering
\small
\setlength{\tabcolsep}{1pt}
\begin{tabular}{ccccc}
\toprule
Policy & Cost (cents) & Coverage & Avg. Len. & Set Size\\
\midrule
GPT-4o EXIT &104.6 & 0.960 & 1.39 & 1.39 \\
LLaMA-2 EXIT &0.000 & 0.790 & 3.14 & 3.14\\ \midrule
Random & 12.99 & 0.545 & 1.53 & 1.00 \\
GPT-4o & 25.35 & 0.930 & 1.08 & 1.00 \\ 
LLaMA-2-7B &  8.682 & 0.660 & 1.11 & 1.00 \\ 
UALA & 16.99 & 0.920 & 2.00 & 2.00\\
CPO & 9.174 & 0.70 & 1.195 & 1.00 \\ \midrule
CPO batch & 22.25 & 0.902 & 2.86 & 4.20 \\
CPO online & 19.07 & 0.901 & 2.48 & 3.46 \\
CCPO ($\lambda=0$) &  \textbf{7.949} & 0.920 & \textbf{1.08} & 2.14 \\
CCPO ($\lambda=2e^{-4}$)     &  8.071 & 0.907  &1.11  &  \textbf{1.48}  \\ 
\bottomrule
\end{tabular}
\caption{MMLU results with LLaMA-2-7B, $\alpha = 0.1$.}
\label{tbl:mmlu-llama2-01}
\end{table}

\paragraph{Comparison to LLM policies.}

LLMs have demonstrated great potential as language-based controllers. However, in our setting, large models such as GPT-4o are required for strong performance, but these  incur high costs. In contrast, smaller models like LLaMA-2-7B and LLaMA-3.2-3B exhibit limited ability to assess the correctness of responses, whether their own or from other models, often performing comparably to random. Among LLM policy strategies, LLM-EXIT performs best, likely due to the inherent strength of LLMs in binary decision-making tasks.

\begin{table}[!t]
\centering
\small
\setlength{\tabcolsep}{1pt}
\begin{tabular}{ccccc}
\toprule
Policy & Cost (cents) & Coverage & Avg. Len. & Set Size\\
\midrule
GPT-4o EXIT & 104.6 & 0.960 & 1.39 & 1.39 \\
LLaMA-3 EXIT & 0.000 & 0.840 & 3.97 & 3.97 \\ \midrule
Random  & 12.74 & 0.607 & 1.55 & 1.00\\
GPT-4o  & 26.87 & 0.913 & 1.19 & 1.00\\
LLaMA-3-2-3b & 8.540 & 0.580 & 1.03 & 1.00\\
UALA & 16.10 & 0.893  & 2.00 & 2.00\\
CPO & 6.846 & 0.860  &1.01   & 1.00  \\ \midrule
CPO batch & 12.12 &0.917& 1.51 & 2.36 \\
CPO online & 11.76 & 0.878 & 1.51 & 2.53 \\
CCPO ($\lambda=0$) & \textbf{10.82} & 0.901 & \textbf{1.34} & 2.66 \\
CCPO ($\lambda=1e^{-4}$)    & 10.90   & 0.900 & 1.35  &  2.70  \\
\bottomrule
\end{tabular}
\caption{MMLU results with LLaMA-3.2-3b, $\alpha = 0.1$.}\label{tbl:mmlu-llama3-01}
\end{table}

%%%%% END OF MMLU Results %%%%

\section{Discussion}
\label{sec:discussion}

\paragraph{Clipping.}

To guarantee convergence and a unique fixed point for the critic, we clip the importance weights so that the V-trace operator becomes a contraction mapping. Specifically, \citet[proof of Remark 3]{espeholt2018impala} show that with $\gamma \bar c < 1$, the V-trace operator $\mathcal{R}V$ is a contraction mapping in the sup-norm. Thus, setting $\bar c = \bar\rho = 1$ preserves the contraction property and introduces only negligible bias in the critic update. For policy updates, we apply clipping only to the reward gradient to reduce variance, while leaving the cost gradient unclipped. Clipping $g_c$ would introduce bias, so we avoid it. Overall, this clipping scheme ensures that the critic remains stable and that the CPO gradients are both low-variance and nearly unbiased.

\paragraph{Need for stochastic policy.}
We need a stochastic policy both for the CPO policy update and for the stability of the V-trace estimate. First, CPO imposes a trust-region constraint on policy updates via the Kullback–Leibler (KL) divergence
$D_{\mathrm{KL}}\bigl(\pi_{\text{new}} \,\|\, \pi_{\text{old}}\bigr) \le \delta$,
which ensures that each policy update remains within a local neighborhood where surrogate approximations are valid. However, KL divergence is only well-defined when $\pi_{\text{new}}$ is absolutely continuous with respect to $\pi_{\text{old}}$ and has the same support. Thus, the KL divergence for a conformal policy may become infinite or undefined. Using a stochastic policy ensures valid policy updates.

Second, \citet[proof of Remark 3]{espeholt2018impala} argue that for $\mathcal{R}V$ to be a contraction mapping, the policy must satisfy 
$\mathds{E}_{\mu} \left[ \frac{\pi_{\text{new}}(a_t \mid o_t)}{\pi_{\text{old}}(a_t \mid o_t)} \right] = 1$.
Using a stochastic policy ensures the importance weights are bounded, thereby preserving convergence of the V-trace method.

\section{Conclusion}

We have proposed Conformal Constrained Policy Optimization (CCPO), an algorithm that unifies constrained policy optimization and online conformal prediction to orchestrate LLM agents to minimize cost while achieving a user-specified reliability level $\alpha$. We have formalized the problem as a finite-horizon reinforcement learning problem where we jointly optimize a stochastic policy and a threshold. We solve this problem using a combination of constrained policy optimization and online conformal prediction. Experiments on multi-hop question answering benchmarks demonstrate that our approach can reduce total computational and API costs by up to 30\% compared to state-of-the-art cost-aware baselines at the target coverage levels. These results demonstrate that our approach is a promising strategy for reliable and cost-effective LLM agents.

\section*{Acknowledgements}
This work was supported in part by NIH R01EY037101 and NSF Award CCF-2338777.

\bibliography{aaai2026}

\section{Prompt Templates}\label{app:temp}

\subsection{Guide Model Template}

\begin{center}
\fbox{%
\begin{minipage}{0.95\linewidth}
\ttfamily
Evaluate the answer.\\
- If correct, reply "Yes".\\
- If incorrect, reply "No [correct answer]" (only the correct answer, nothing else).\\
Q: \{question\}\\
A: \{answer\}\\
Your response:
\end{minipage}%
}
\end{center}

\subsection{LLM EXIT Template}

\begin{center}
\fbox{%
\begin{minipage}{0.95\linewidth}
\ttfamily
You will be given a question and its historical answers. You must decide whether to exit. \\

Instructions: \\
Evaluate the current answer history and determine whether it meets any of the following conditions: \\
1. The recent responses are repetitive, or no valid answer has been given. \\
2. The most recent answer is correct, and no further attempts are necessary. \\

\textbf{Input:} \\
Question: \{question\} \\
Previous Answers: \{answers\} \\

\textbf{Output:} \\
Exactly one of [EXIT] or [CONTINUE], and nothing else. \\
Answer:
\end{minipage}%
}
\end{center}

\subsection{LLM-as-a-Policy Template}
\begin{center}
\fbox{%
\begin{minipage}{0.95\linewidth}
\ttfamily
\textbf{System Prompt:}\\
You are a decision-making assistant. When given: \\
• A question \\
• An answer from a base model \\
• A judgement from a GPT-4o model in the format ``agreement (yes/no), gpt\_answer'' \\
• An uncertainty score from the GPT-4o model \\

Your job is to pick one of these actions: \\
1. Use the base model’s answer. \\
2. Use the GPT-4o model’s answer. \\
3. Go to the next round (neither answer is acceptable). \\

Your goal is to maintain high decision quality and approximately 0.9 coverage over accepted answers. \\

Output only the action number (``1'', ``2'', or ``3'') and nothing else. \\

\vspace{1em}
\textbf{User Content:} \\
Question: \{question\} \\
\\
Base answer: \{base\_ans\} \\
\\
GPT-4o judgement (agreement, answer): \{gpt\_judgement\} \\
\\
GPT-4o uncertainty: \{gpt\_uncertainty\} \\
\\
Action:
\end{minipage}%
}
\end{center}

\clearpage
\section{Additional Results}
\label{app:more-res}

\begin{table}[H]
\centering\small\setlength{\tabcolsep}{1pt}
\begin{tabular}{ccccc}
\toprule
Policy & Cost (cents) & Coverage & Avg. Len. & Set Size\\
\midrule
Random   & 6.811         & 0.578          & 1.44         & 1.00  \\ 
GPT-4o     & 19.25 & 0.810 & 1.16   & 1.00   \\ 
LLaMA-2-7b   & 4.524 & 0.685 & 1.02 & 1.00   \\ 
UALA   & 4.635          &       0.850  &     1.00         &    1.00     \\ 
CPO & 4.711 & 0.840 & 1.00 & 1.00 \\ \midrule
CPO batch   & 4.663          & 0.841    & 1.00          & 1.00      \\ 
CPO online   &   4.709   & 0.819  &     1.00     &  1.00  \\ 
CCPO ($\lambda=0$)  & 4.640 & 0.862    & 1.00 & 1.00      \\ 
CCPO ($\lambda=1e^{-4}$) &4.647 & 0.815 & 1.00 & 1.00 \\
\bottomrule
\end{tabular}
\caption{HotpotQA results with LLaMA-2-7B, $\alpha = 0.2$.}\label{tbl:hot-llama2-02}
\end{table}

\begin{figure}[H]
    \centering
    \includegraphics[width=0.43\textwidth]{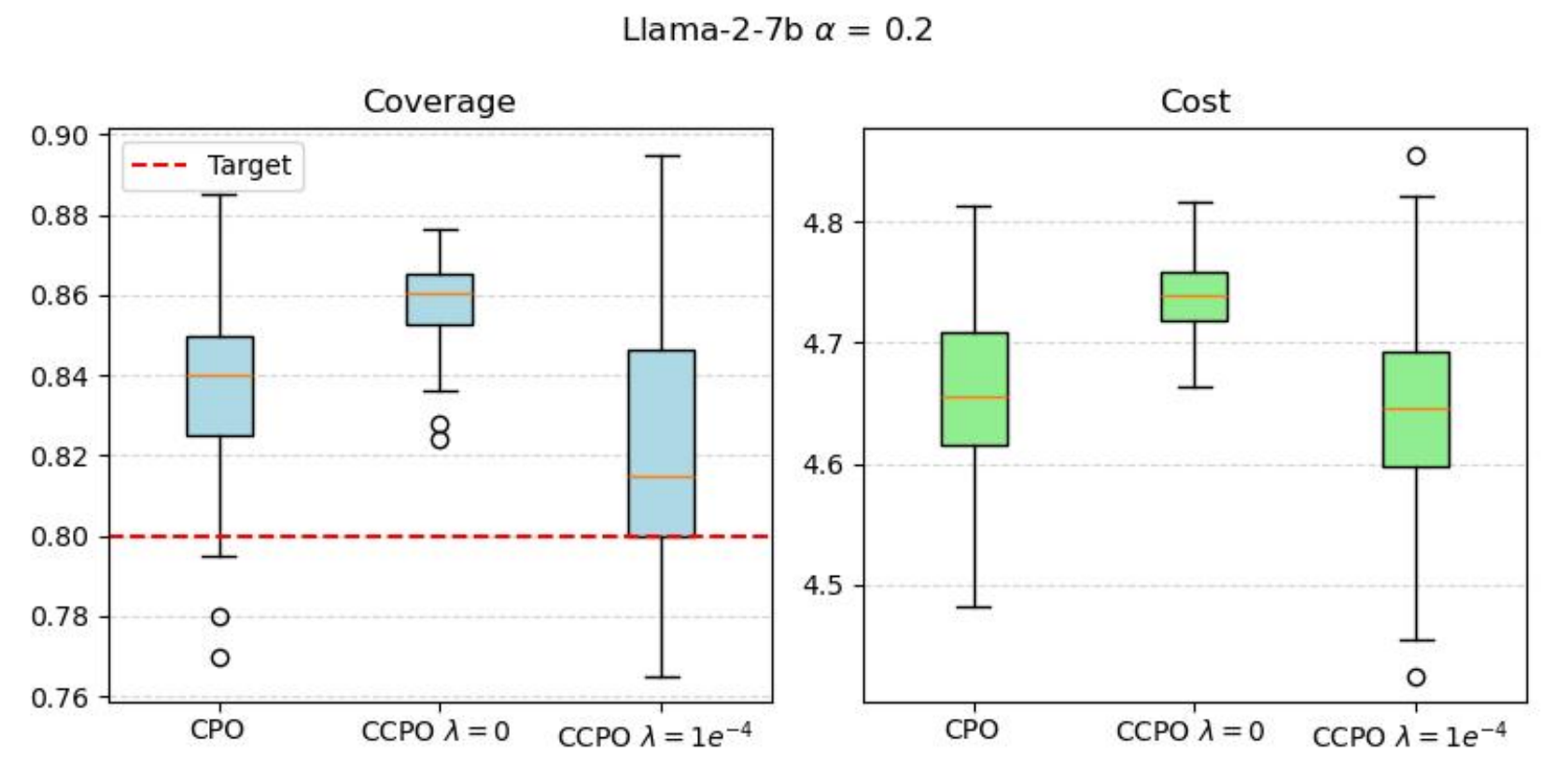}
    \caption{HotpotQA coverage and cost with LLaMA-2-7B base model over 100 splits, $\alpha = 0.2$.}
    \label{fig:hot-llama2-02}
\end{figure}

\begin{table}[H]
\centering\small\setlength{\tabcolsep}{1pt}
\begin{tabular}{ccccc}
\toprule
Policy & Cost (cents) & Coverage & Avg. Len. & Set Size\\
\midrule
Random     & 1.411          & 0.610& 1.51           & 1.00     \\ 
GPT-4o  &  18.77 & 0.845 & 1.12
 & 1.00   \\ 
LLaMA-3.2-3b   & 4.495 & 0.605 & 1.00   & 1.00   \\ 
UALA     &     13.25          &        0.970       &        3.00        &     3.00     \\ 
CPO & 4.722 & 0.846 & 1.02 & 1.00 \\ \midrule
CPO batch   &        8.717        &         0.979      &        1.97        &   3.66       \\ 
CPO online   &        10.02     &     0.894       &     2.11         &  3.68     \\ 
CCPO ($\lambda=0$)    & \textbf{8.126}  & 0.954       & \textbf{1.86} & 3.39    \\
CCPO ($\lambda=1e^{-4}$)  & 8.575 & 0.951 &  1.95 & \textbf{3.20} \\ 
\bottomrule
\end{tabular}
\caption{HotpotQA results with LLaMA-3.2-3b, $\alpha = 0.05$. }\label{tbl:hot-llama3-005}
\end{table}

\begin{figure}[H]
    \centering        \includegraphics[width=0.43\textwidth]{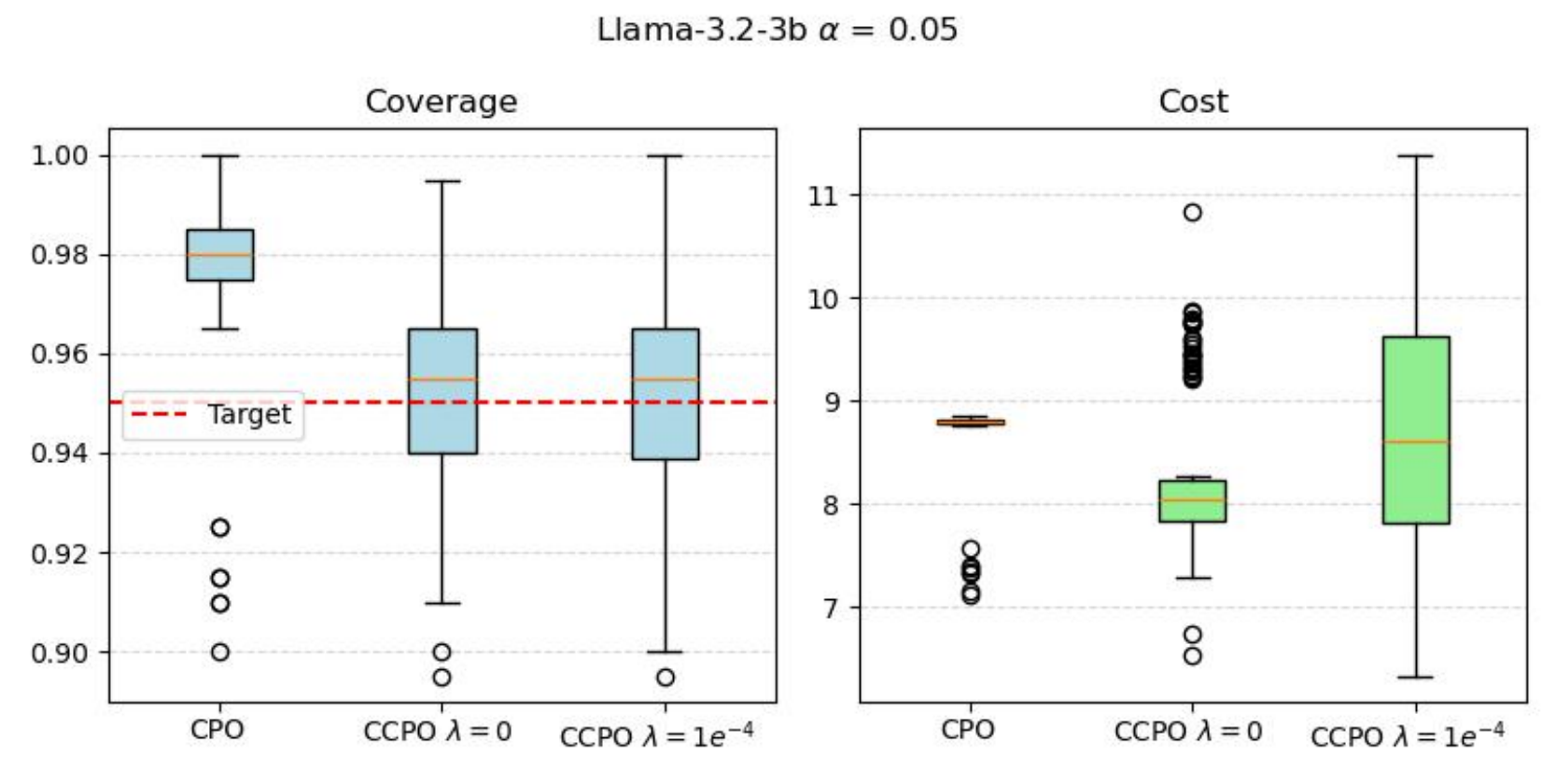}
    \caption{HotpotQA coverage and cost with LLaMA-3.2-3b base model over 100 splits, $\alpha = 0.05$.}
    \label{fig:hot-llama3-005}
\end{figure}

\begin{table}[H]
\centering\small\setlength{\tabcolsep}{1pt}
\begin{tabular}{ccccc}
\toprule
Policy &  Cost (cents) & Coverage & Avg. Len. & Set Size\\
\midrule
Random & 12.99 & 0.545 & 1.53 & 1.00 \\
GPT-4o &  27.15 & 0.887 & 1.31 & 1.00 \\
LLaMA-2-7b & 8.712 & 0.520 & 1.12 & 1.00\\
UALA & 7.612 & 0.889 & 1.00 & 1.00\\
% CPO & 11.15 & 0.760 & 1.44 & 1.00\\ \midrule
CPO &  7.521 & 0.860 & 1.00 & 1.00 \\ \midrule
CPO batch & 8.024  & 0.886 & 1.00 & 1.00 \\
CPO online & 7.643 & 0.828 & 1.00 & 0.95 \\
CCPO ($\lambda=0$) & 7.612 & 0.801  & 1.00  &0.90 \\
CCPO ($\lambda=2e^{-4}$)    & 7.670& 0.805 & 1.00 & 0.90 \\
\bottomrule
\end{tabular}
\caption{MMLU results with LLaMA-2-7B, $\alpha = 0.2$.}\label{tbl:mmlu-llama2-02}
\end{table}

\begin{figure}[H]
\centering
\includegraphics[width=\linewidth]{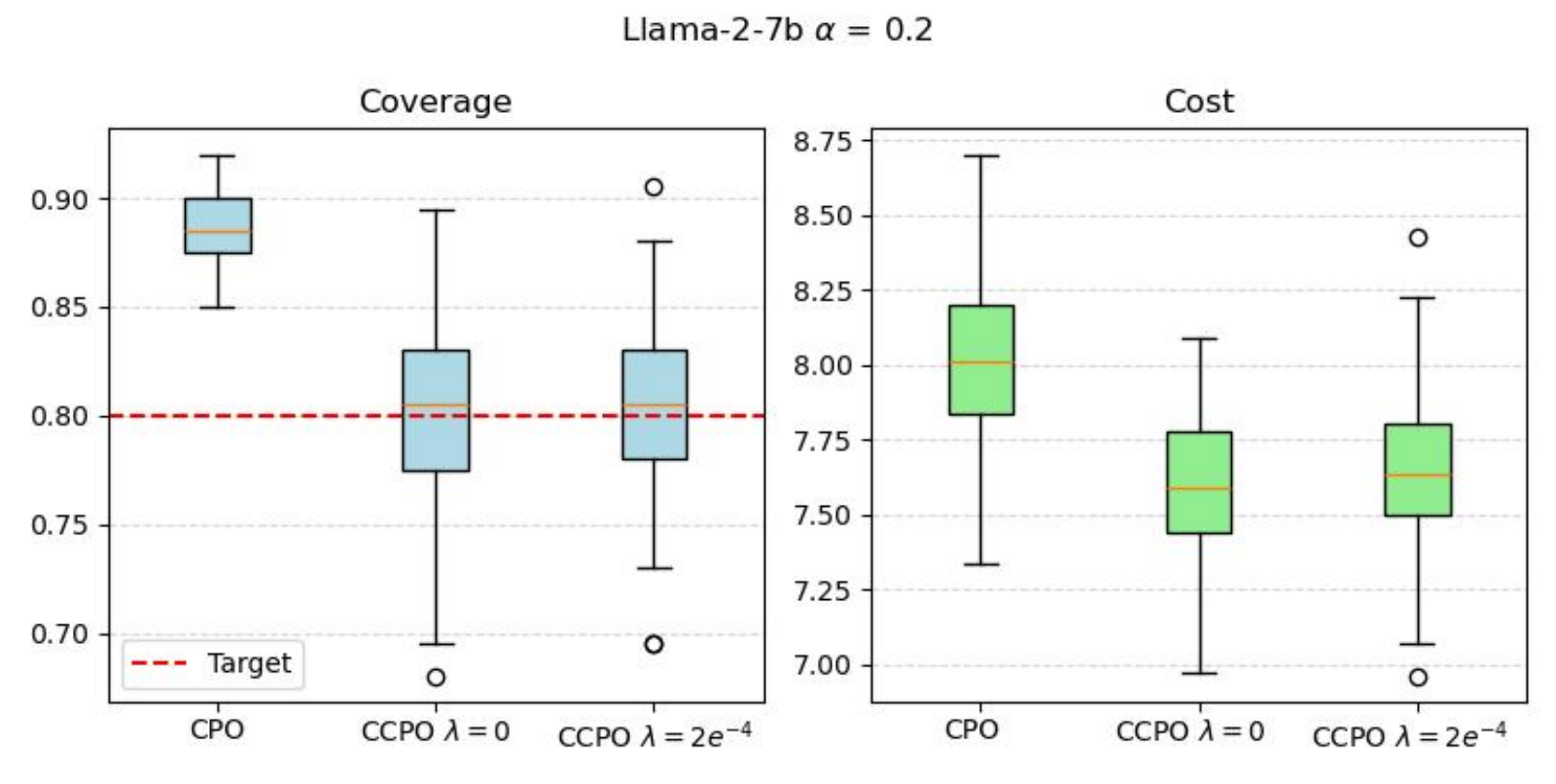}
\captionof{figure}{MMLU coverage and cost with LLaMA-2-7B base model over 100 splits, $\alpha = 0.2$.}
\label{fig:mmlu-llama2-02}
\end{figure}

\begin{table}[H]
\centering\small\setlength{\tabcolsep}{1pt}
\begin{tabular}{cccccc}
\toprule
Policy &  Cost (cents) & Coverage & Avg. Len. & Set Size\\
\midrule
Random  & 12.74 & 0.607 & 1.55 & 1.00\\
GPT-4o  & 26.87 & 0.913 & 1.19 & 1.00\\
LLaMA-3-2-3b &  8.547 & 0.580 & 1.03 & 1.00\\
UALA &  7.755 & 0.839& 1.00 & 1.00 \\
CPO & 7.121 & 0.805 & 1.00 & 1.00 \\ \midrule
CPO batch  & 7.767 & 0.805 & 1.00 & 0.91\\
CPO online  & 7.620 & 0.818 & 1.00 & 0.98\\
CCPO ($\lambda=0$)  & 7.735 & 0.806 & 1.00 & 1.02\\
CCPO ($\lambda=2e^{-4}$) & 7.742& 0.806 & 1.00 & 0.89 \\
\bottomrule
\end{tabular}
\caption{MMLU results with LLaMA-3.2-3b, $\alpha = 0.2$.}\label{tbl:mmlu-llama3-02}
\end{table}

\begin{figure}[H]
\centering
\includegraphics[width=\linewidth]{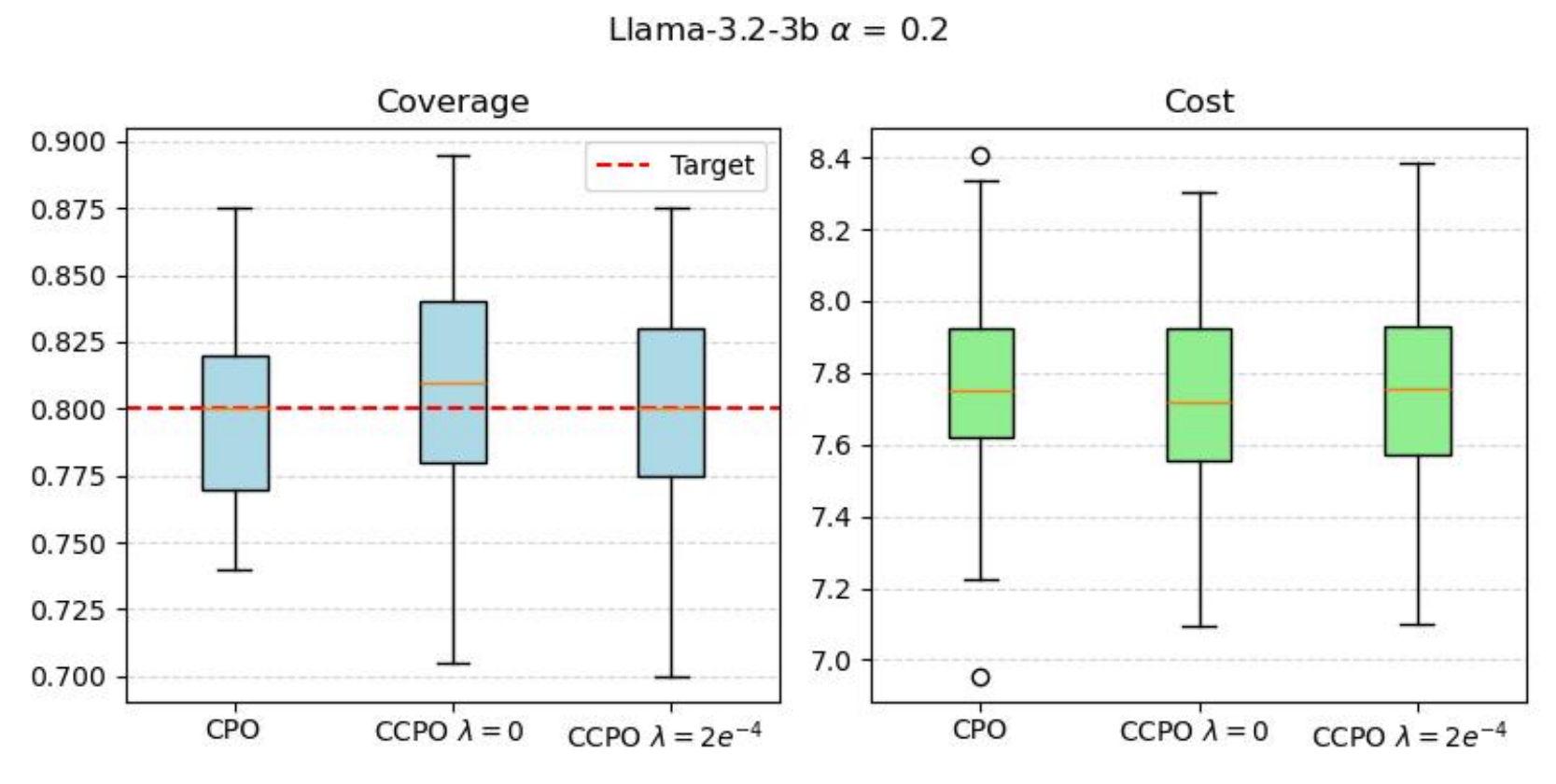}
\captionof{figure}{MMLU coverage and cost with LLaMA-3.2-3b base model over 100 splits, $\alpha = 0.2$.}
\label{fig:mmlu-llama3-02}
\end{figure}

\clearpage
\onecolumn
\section{Training Curves}
\label{sec:curves}
\begin{figure*}[htbp]
    \centering
    \vspace{1em}
    Costs\\
    \includegraphics[width=0.24\textwidth]{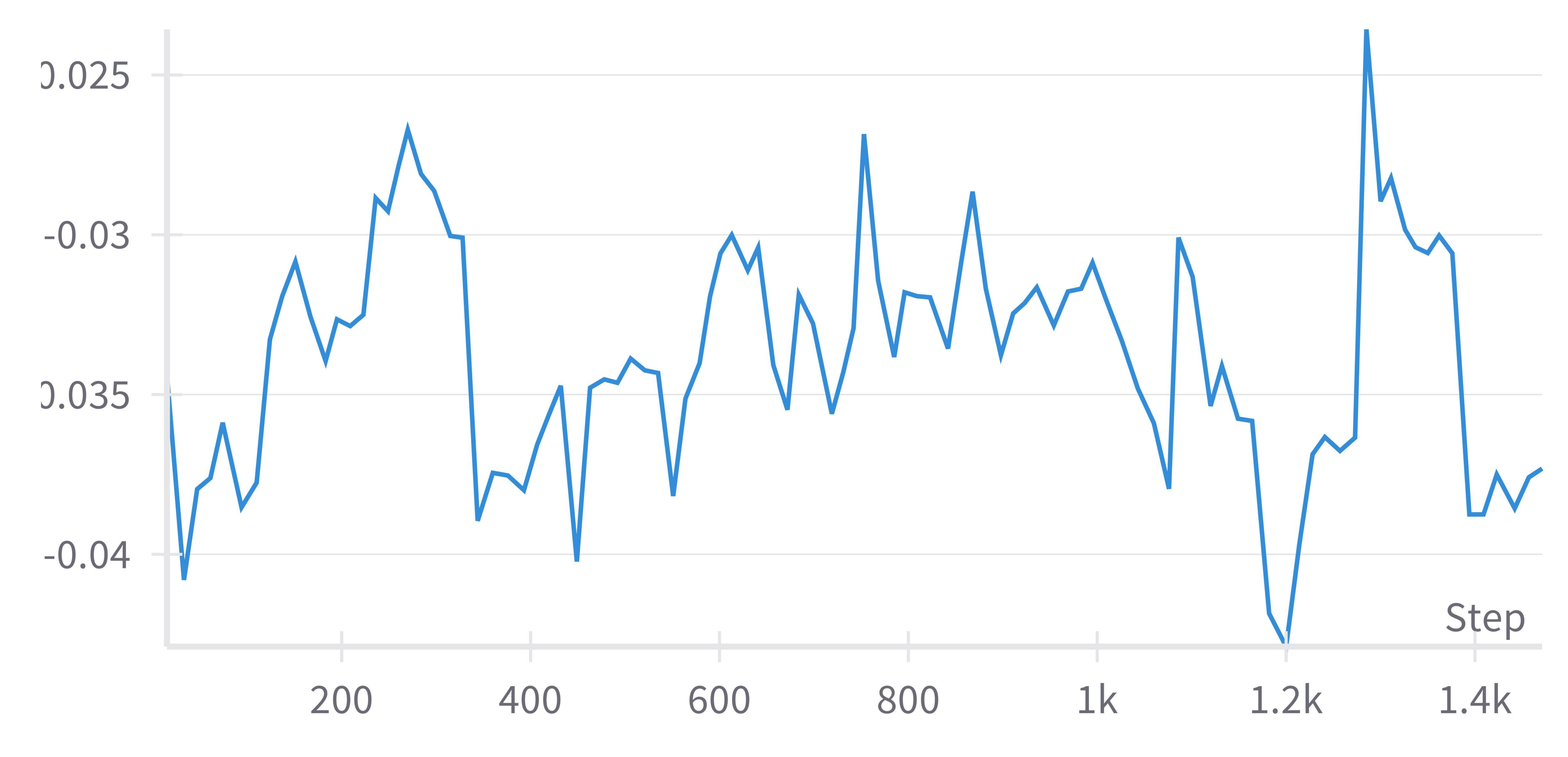}
    \includegraphics[width=0.24\textwidth]{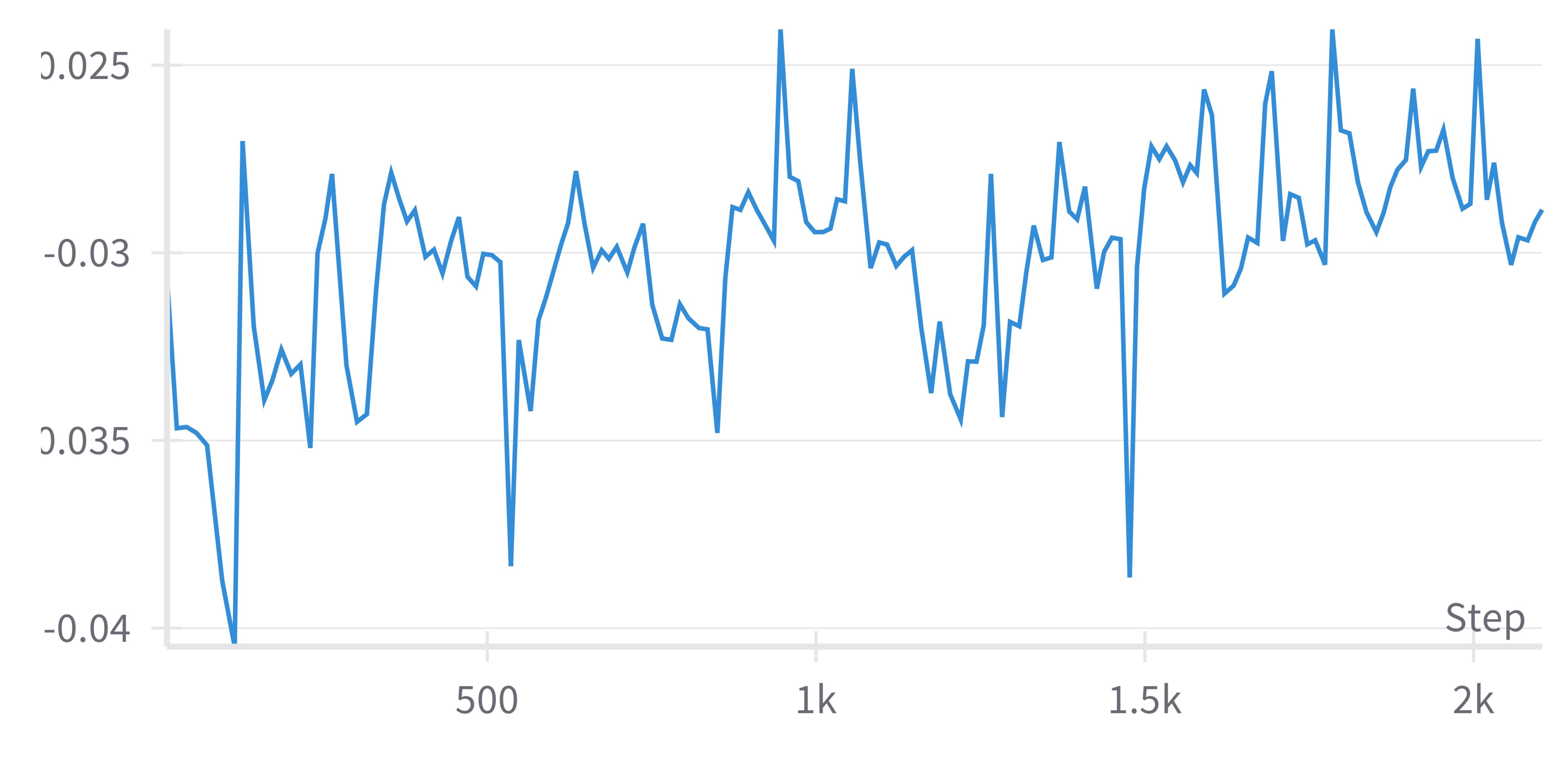}
    \includegraphics[width=0.24\textwidth]{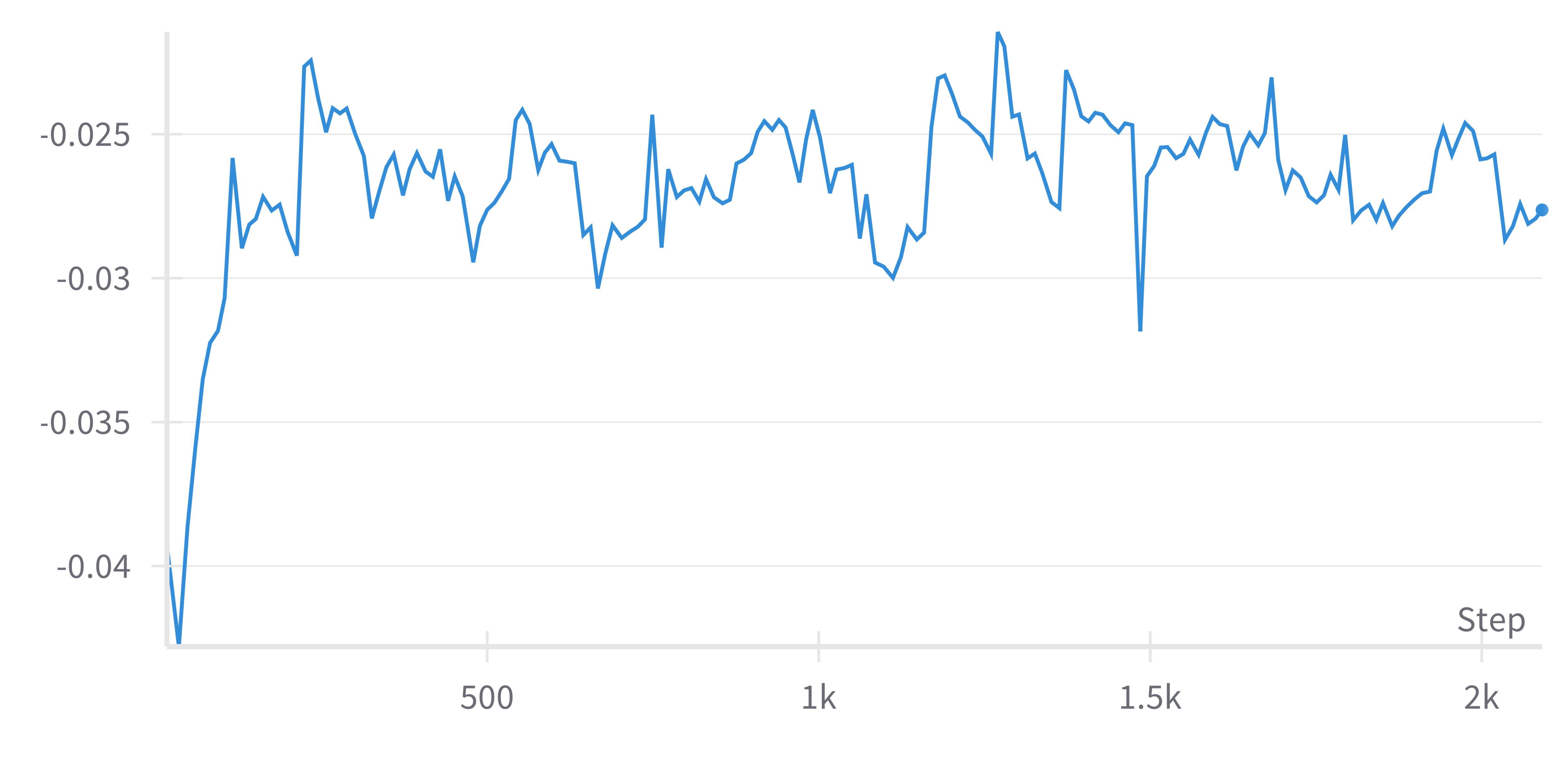}
    \includegraphics[width=0.24\textwidth]{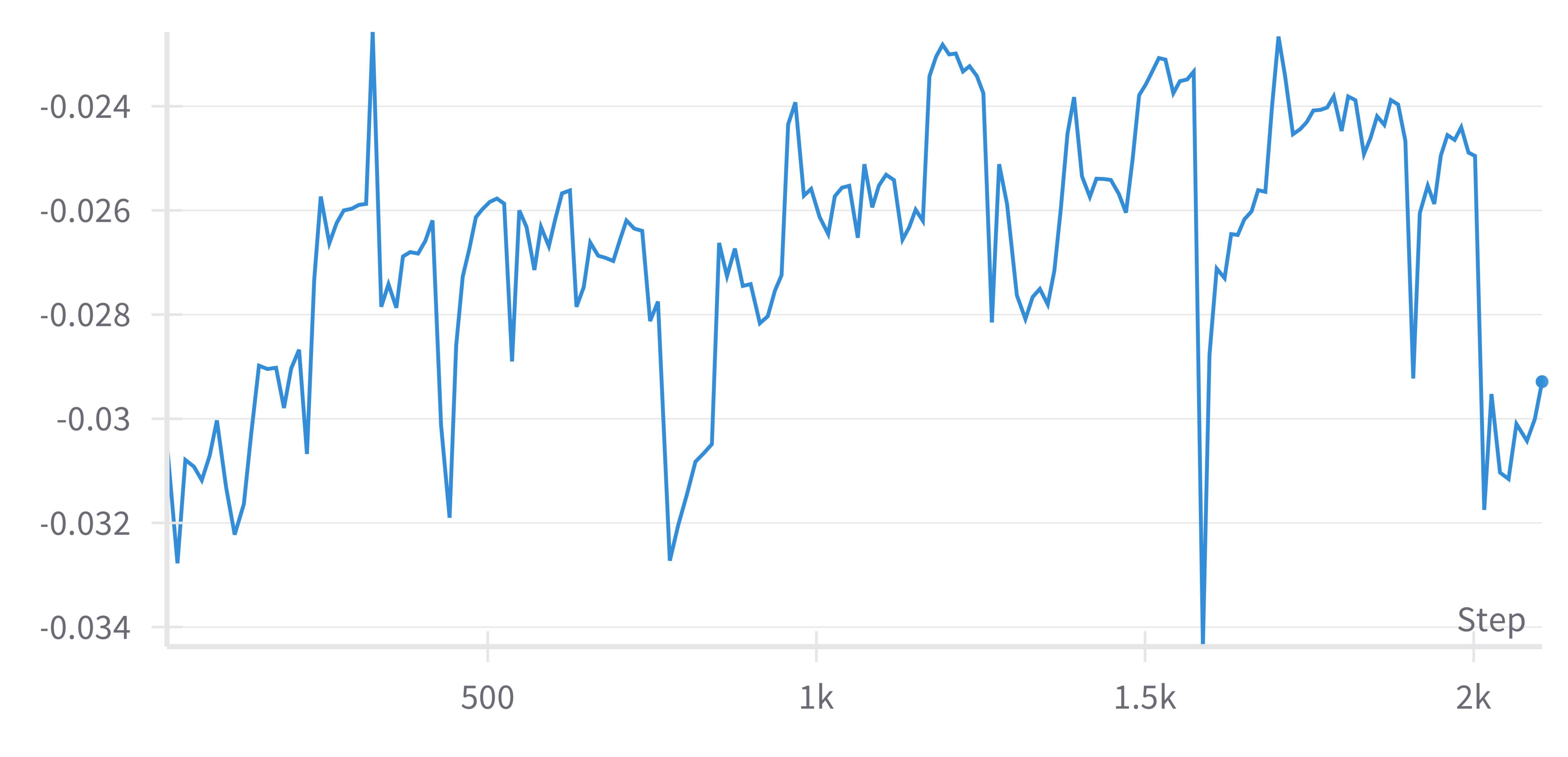}
    \vspace{1em}
    \text{Coverage Surrogate Violation}\\
    \includegraphics[width=0.24\textwidth]{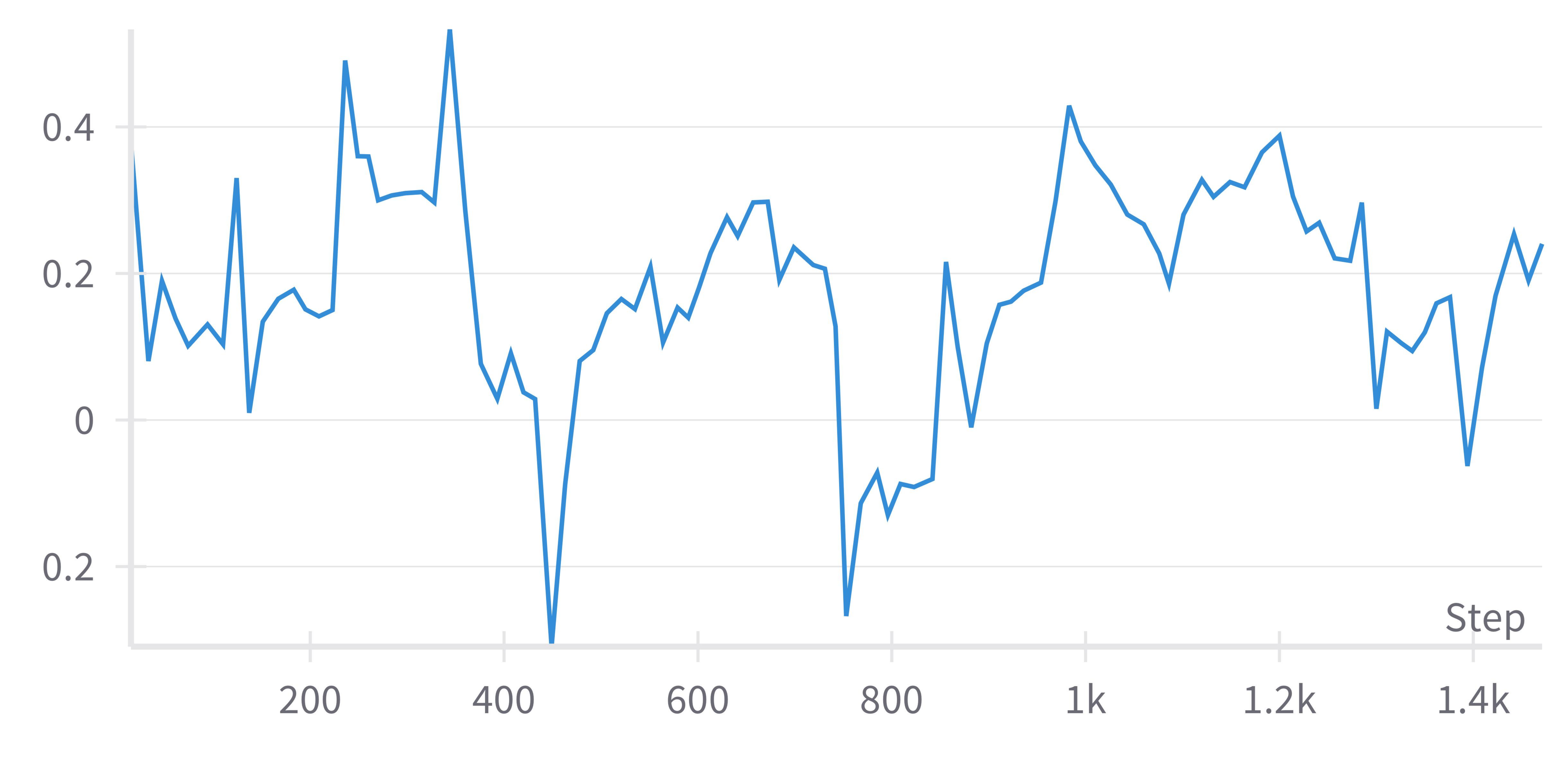}
    \includegraphics[width=0.24\textwidth]{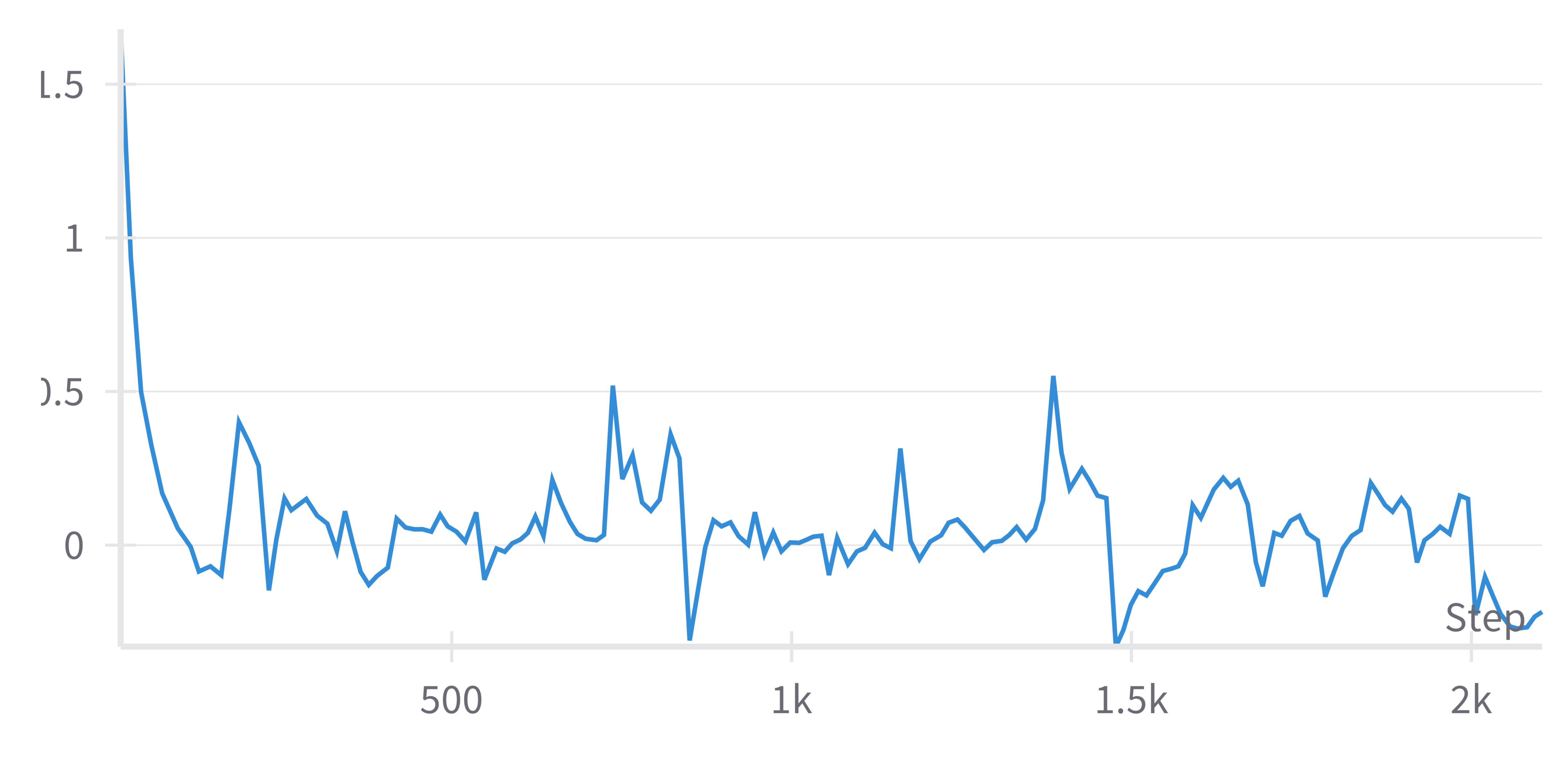}
    \includegraphics[width=0.24\textwidth]{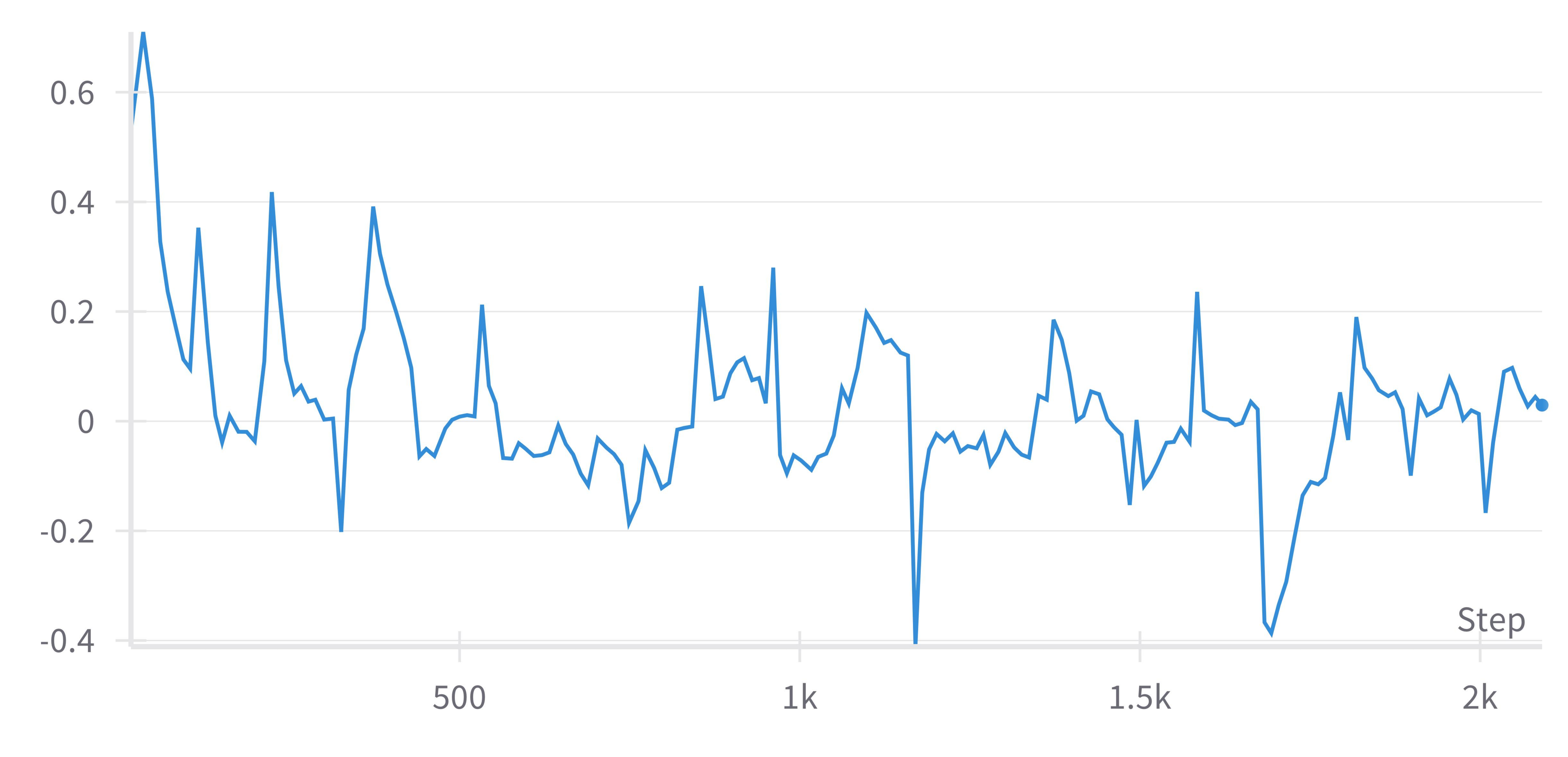}
    \includegraphics[width=0.24\textwidth]{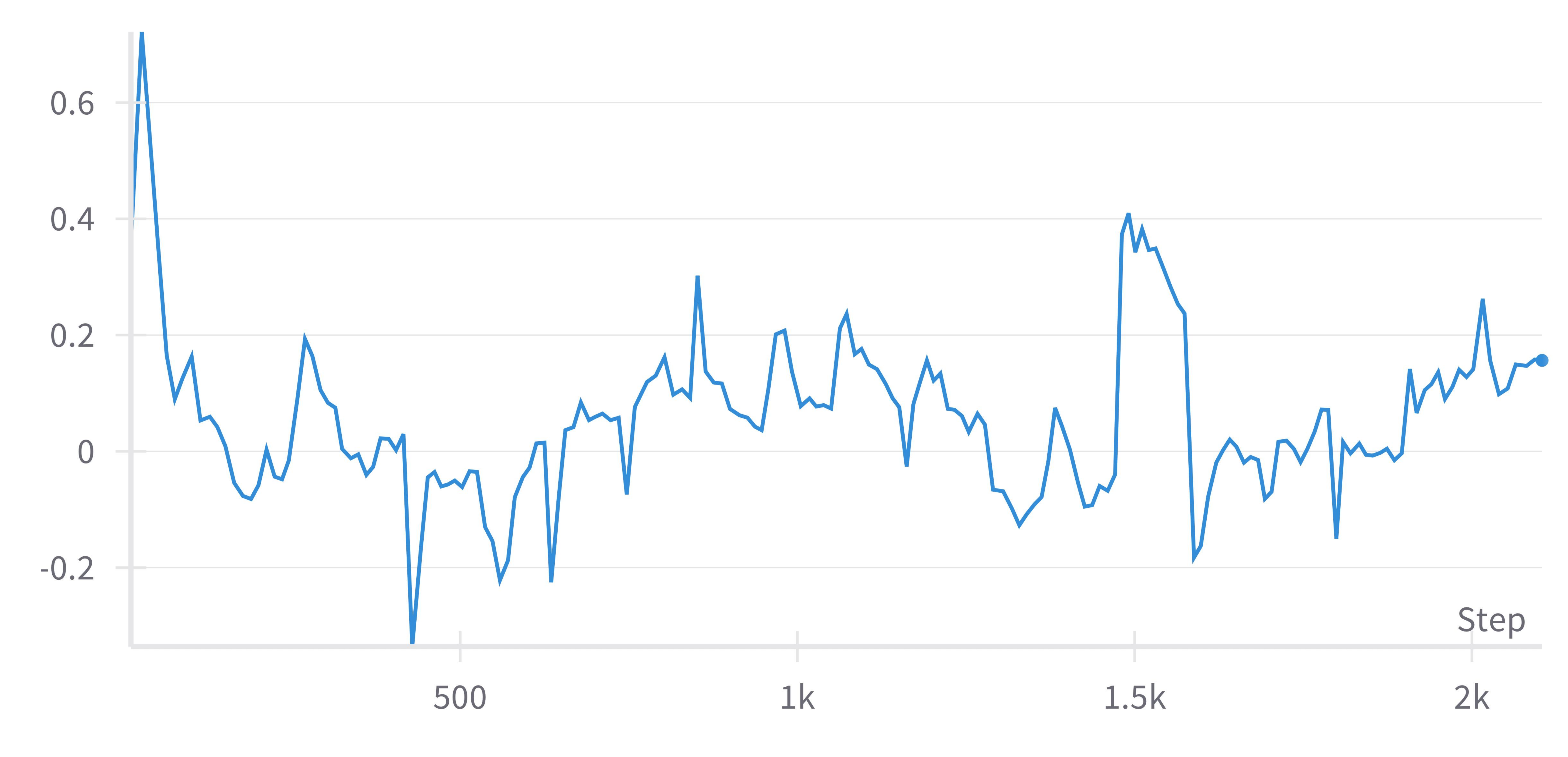}
    \vspace{1em}
    \text{Training Coverage}\\
    % Second row: 4 images with subcaptions
    \begin{subfigure}[b]{0.24\textwidth}
        \includegraphics[width=\textwidth]{ 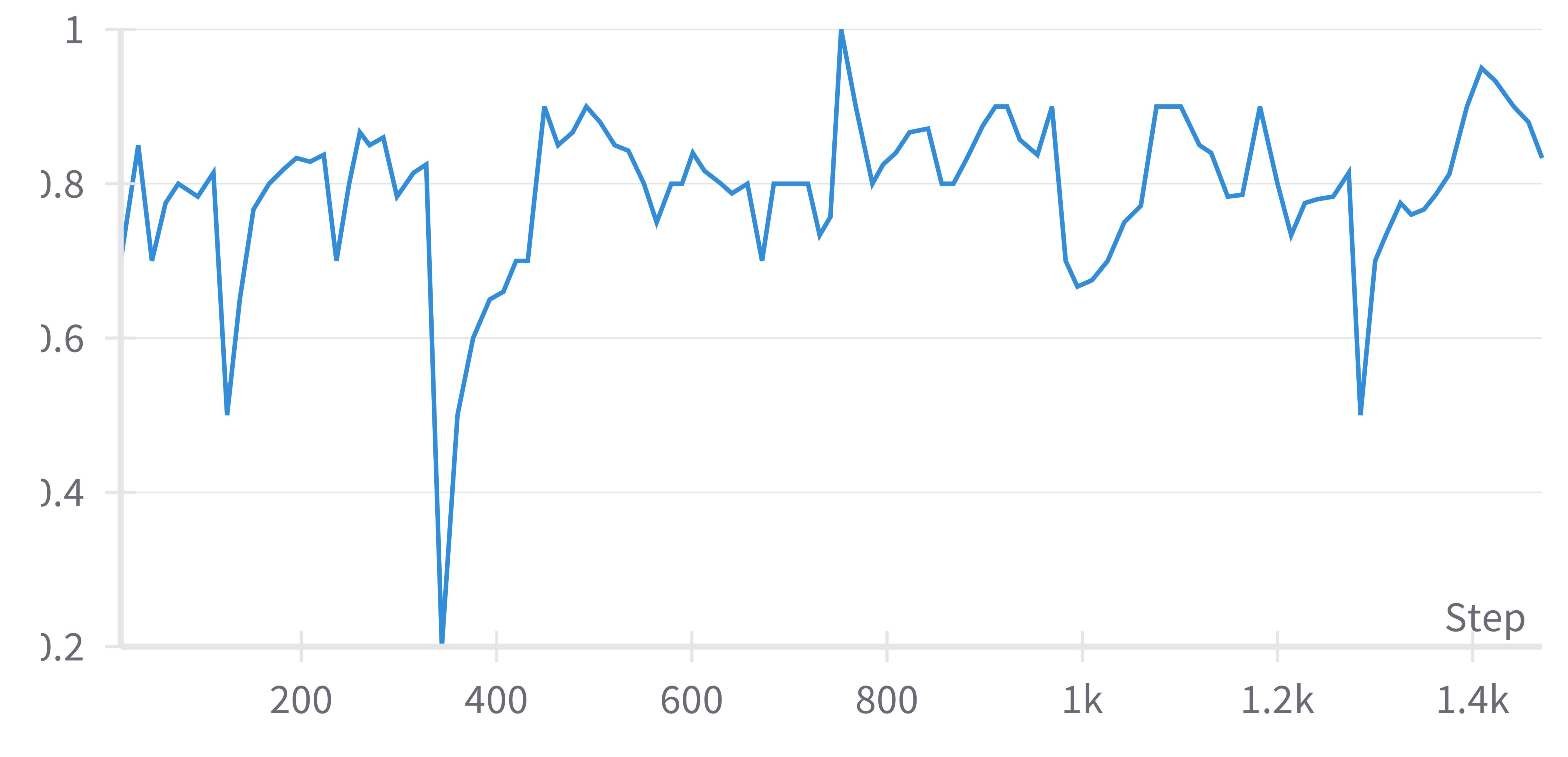}
        \caption{LLaMA-2-7b, $\alpha=0.2$}
    \end{subfigure}
    \begin{subfigure}[b]{0.24\textwidth}
        \includegraphics[width=\textwidth]{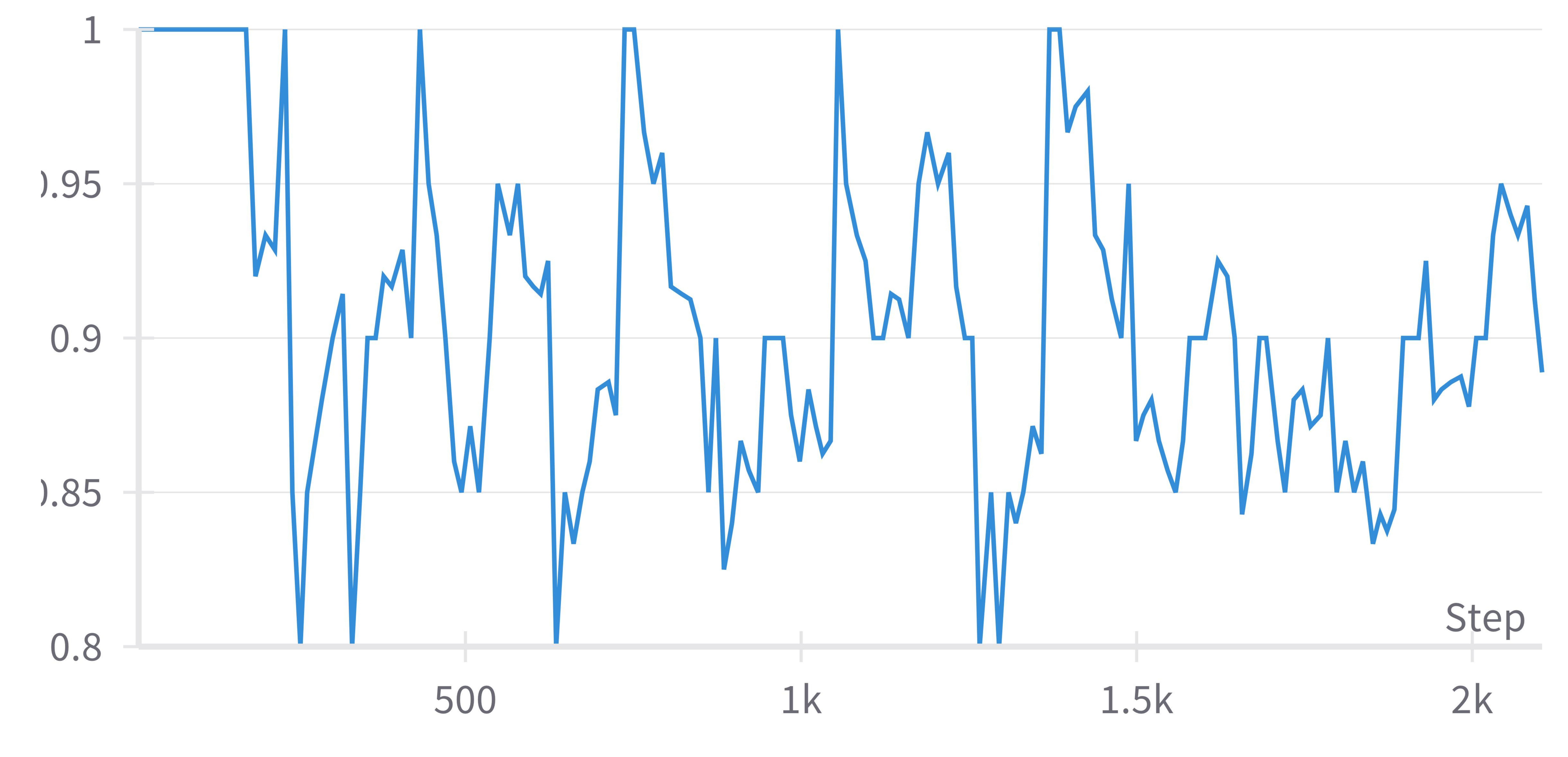}
        \caption{LLaMA-2-7b, $\alpha=0.1$}
    \end{subfigure}
    \begin{subfigure}[b]{0.24\textwidth}
        \includegraphics[width=\textwidth]{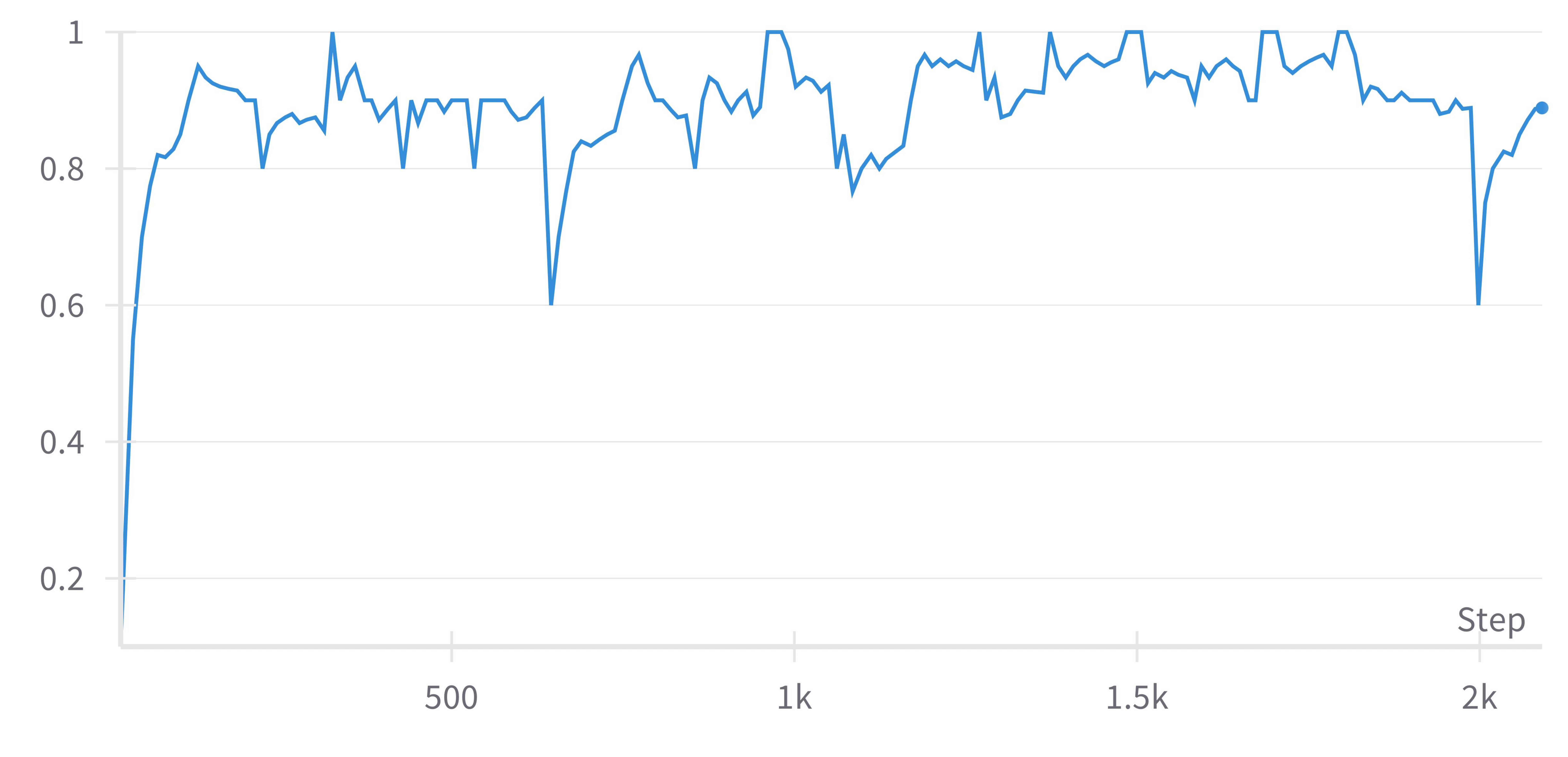}
        \caption{LLaMA-3.2-3b, $\alpha=0.1$}
    \end{subfigure}
    \begin{subfigure}[b]{0.24\textwidth}
        \includegraphics[width=\textwidth]{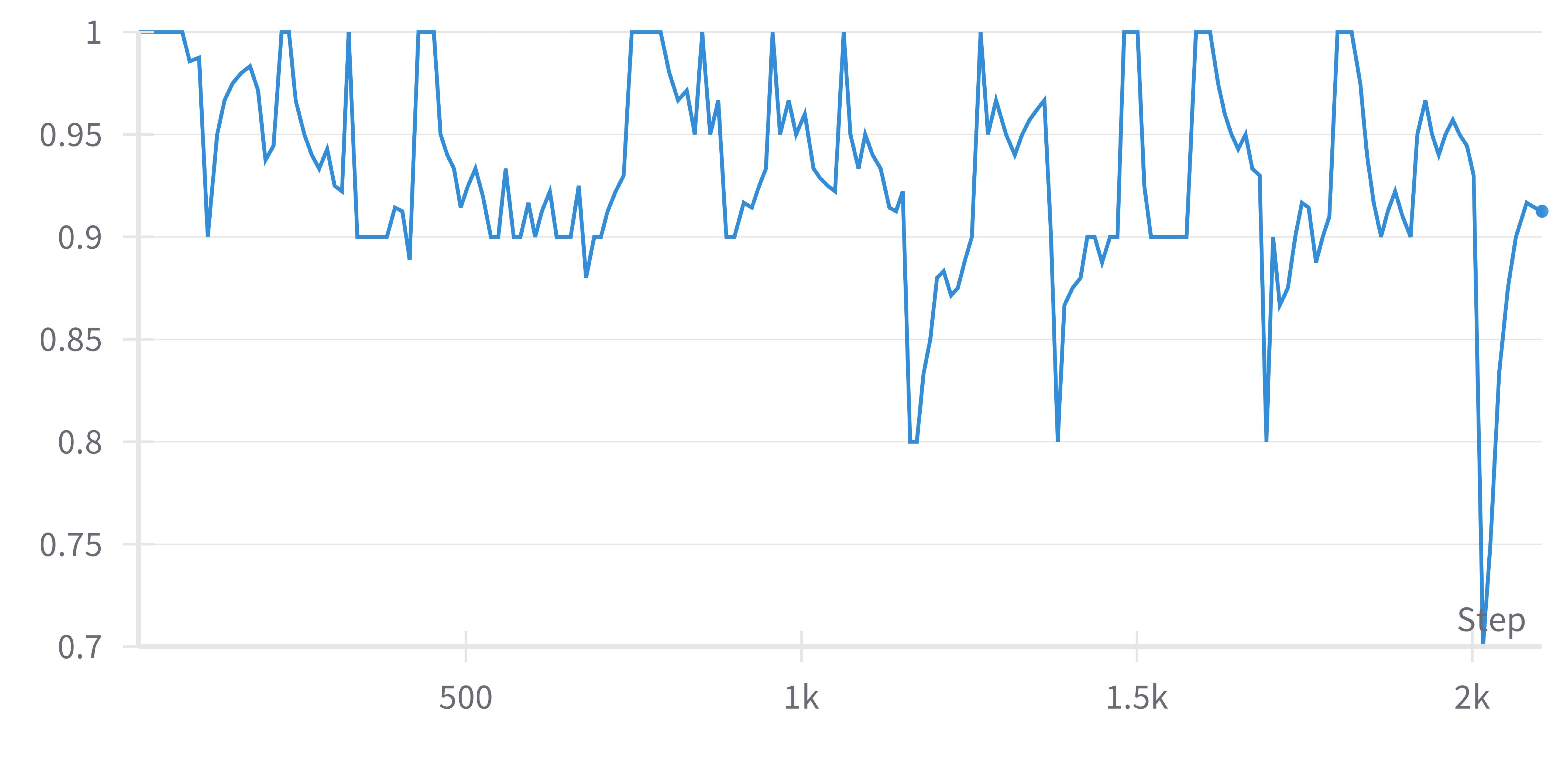}
        \caption{LLaMA-3.2-3b, $\alpha=0.05$}
    \end{subfigure}
    
    \caption{Overall CCPO training performance on HotpotQA dataset, $\lambda=0$.}
    \label{fig:perf}
\end{figure*}

\begin{figure*}[htbp]
    \centering
    \vspace{1em}
    Coverage\\
    \includegraphics[width=0.24\textwidth]{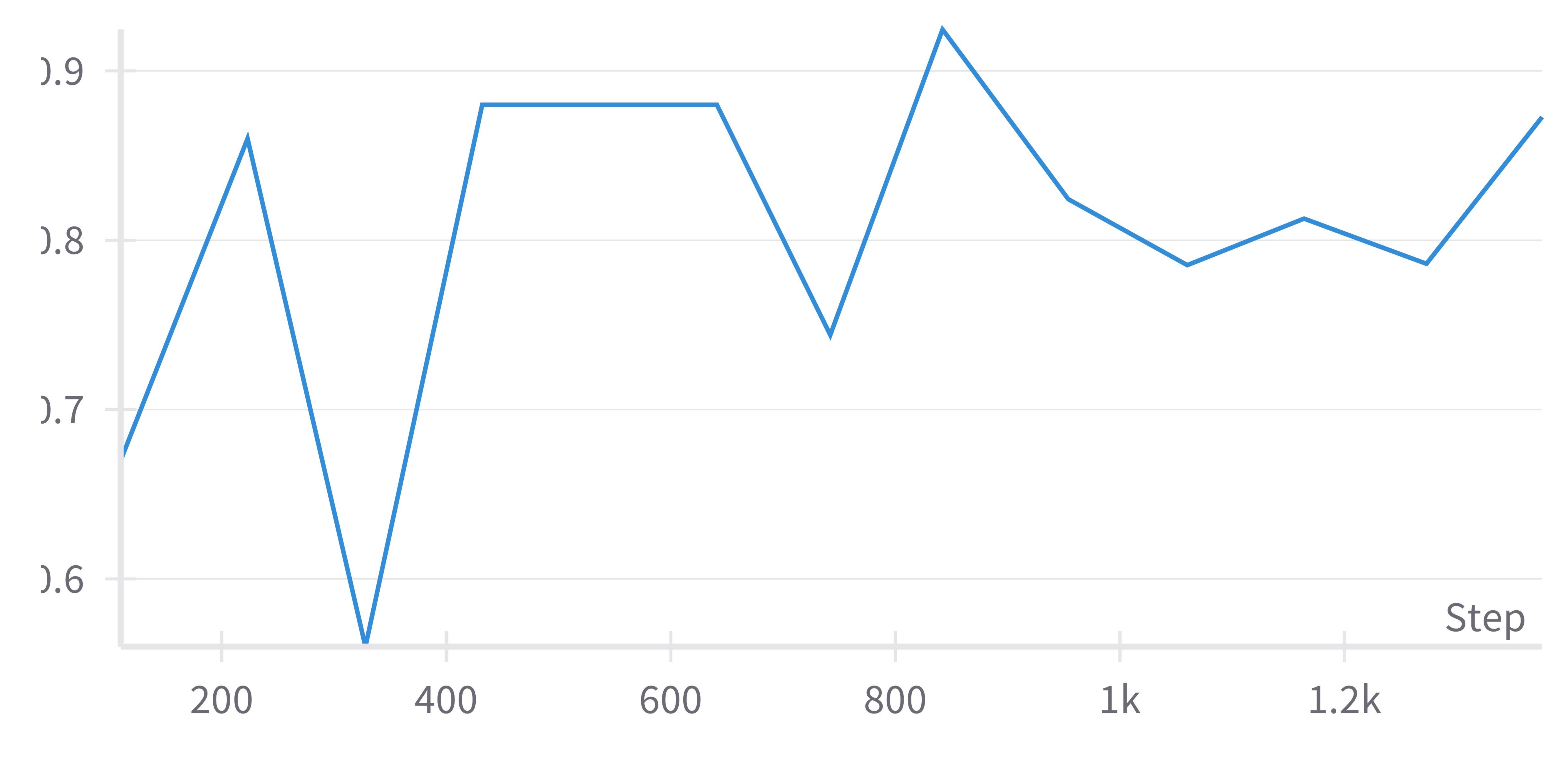}
    \includegraphics[width=0.24\textwidth]{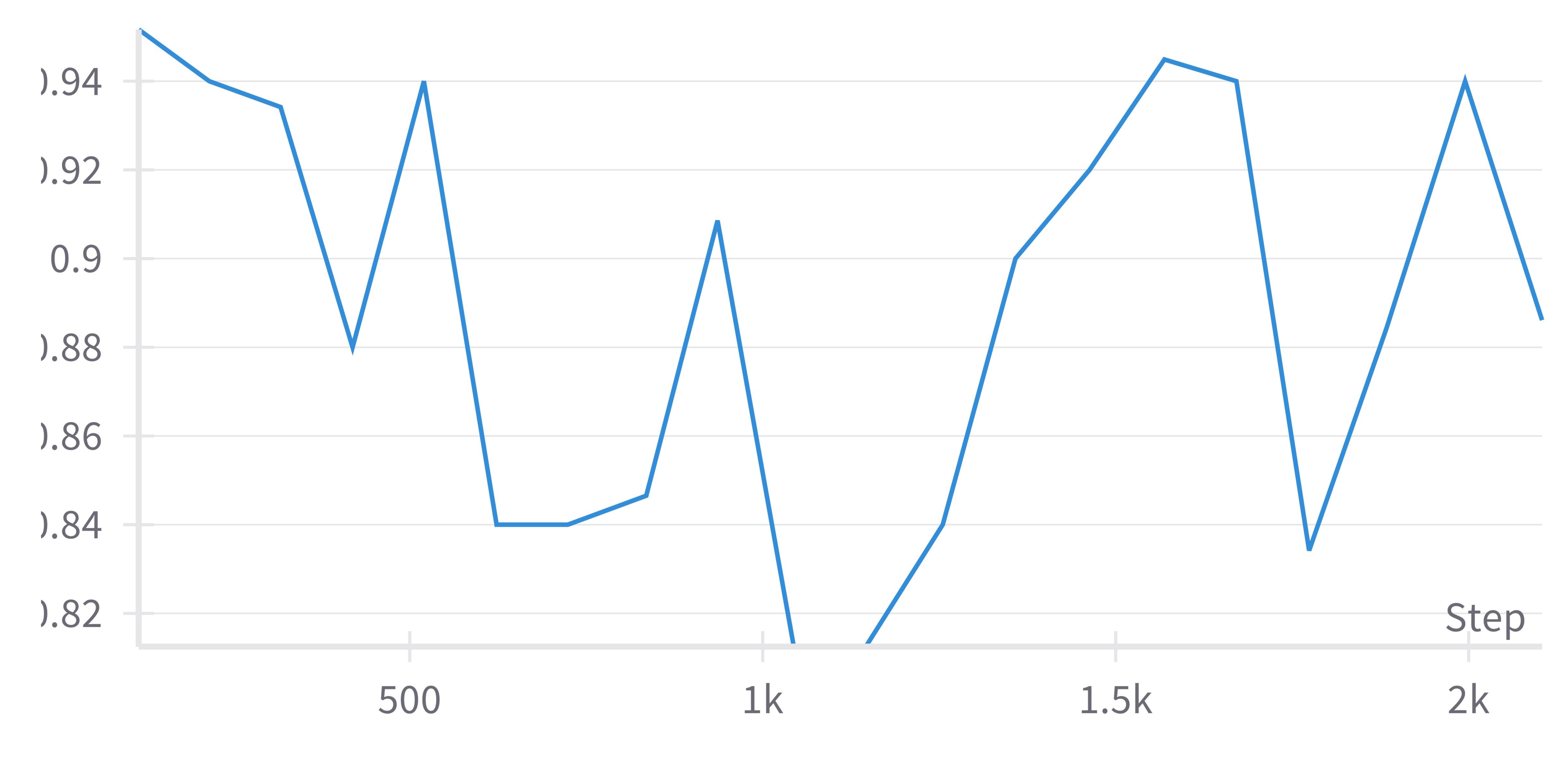}
    \includegraphics[width=0.24\textwidth]{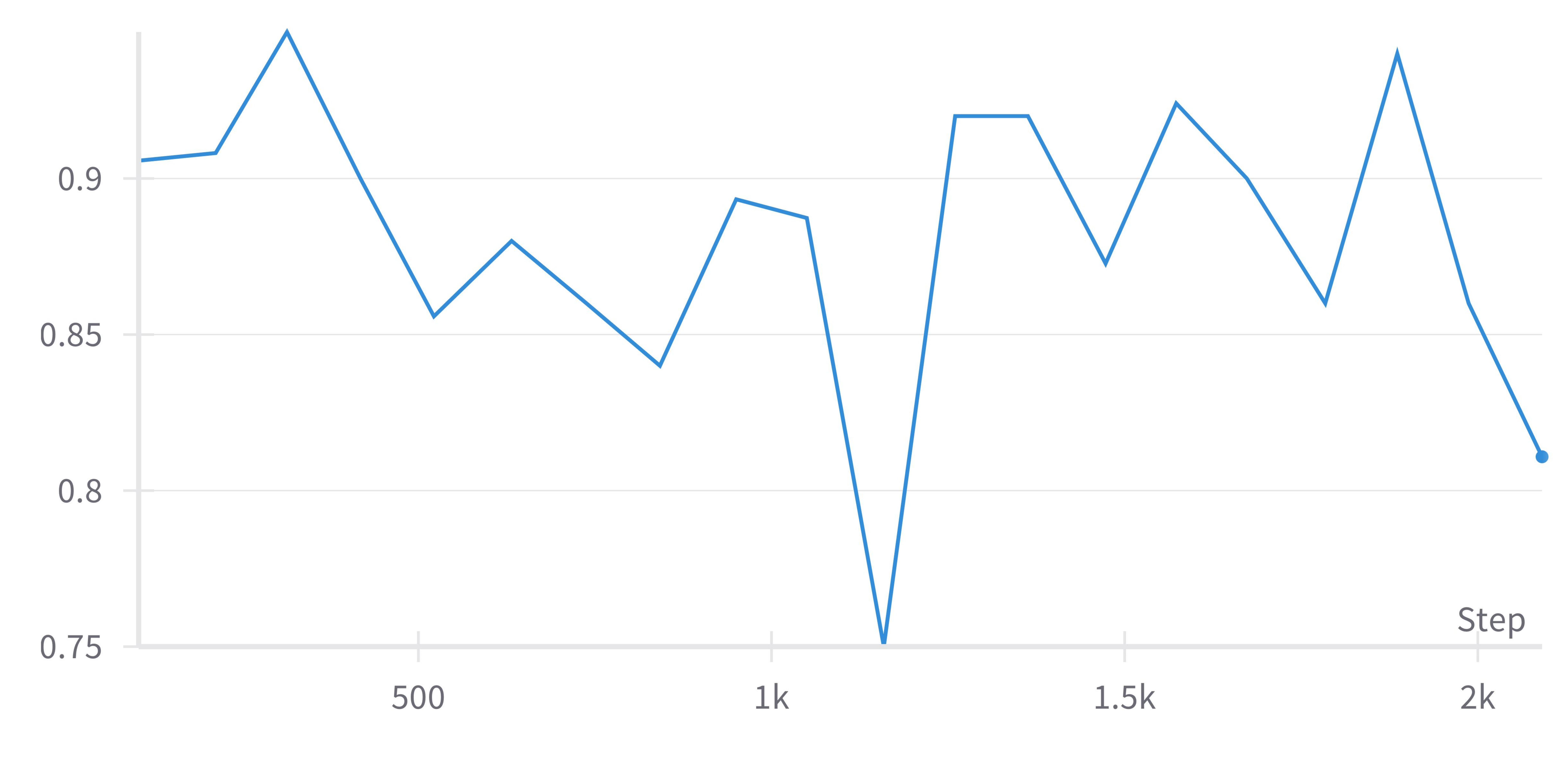}
    \includegraphics[width=0.24\textwidth]{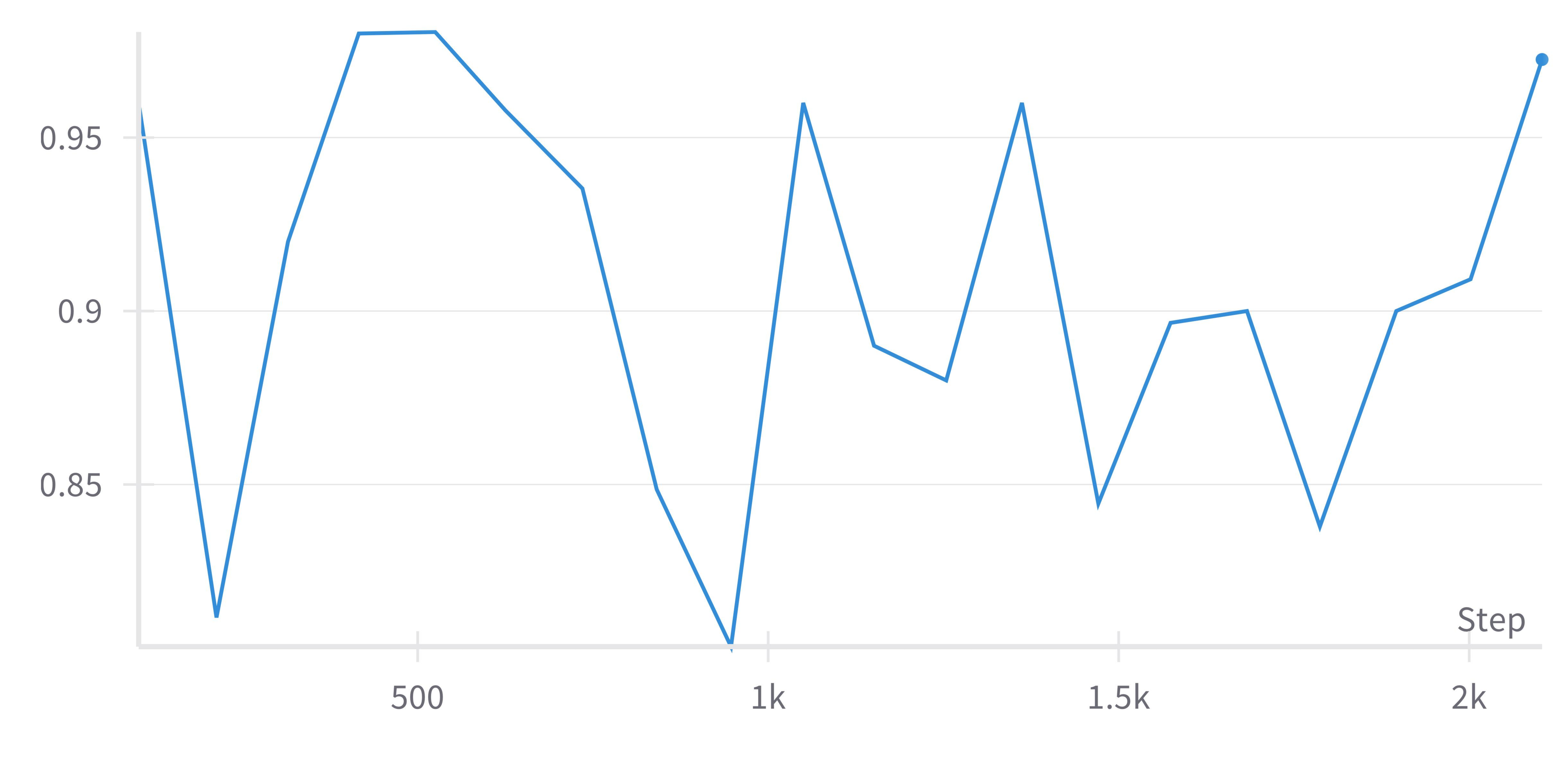}
    \vspace{1em}
    \text{Average Length}\\
    \begin{subfigure}[b]{0.24\textwidth}
        \includegraphics[width=\textwidth]{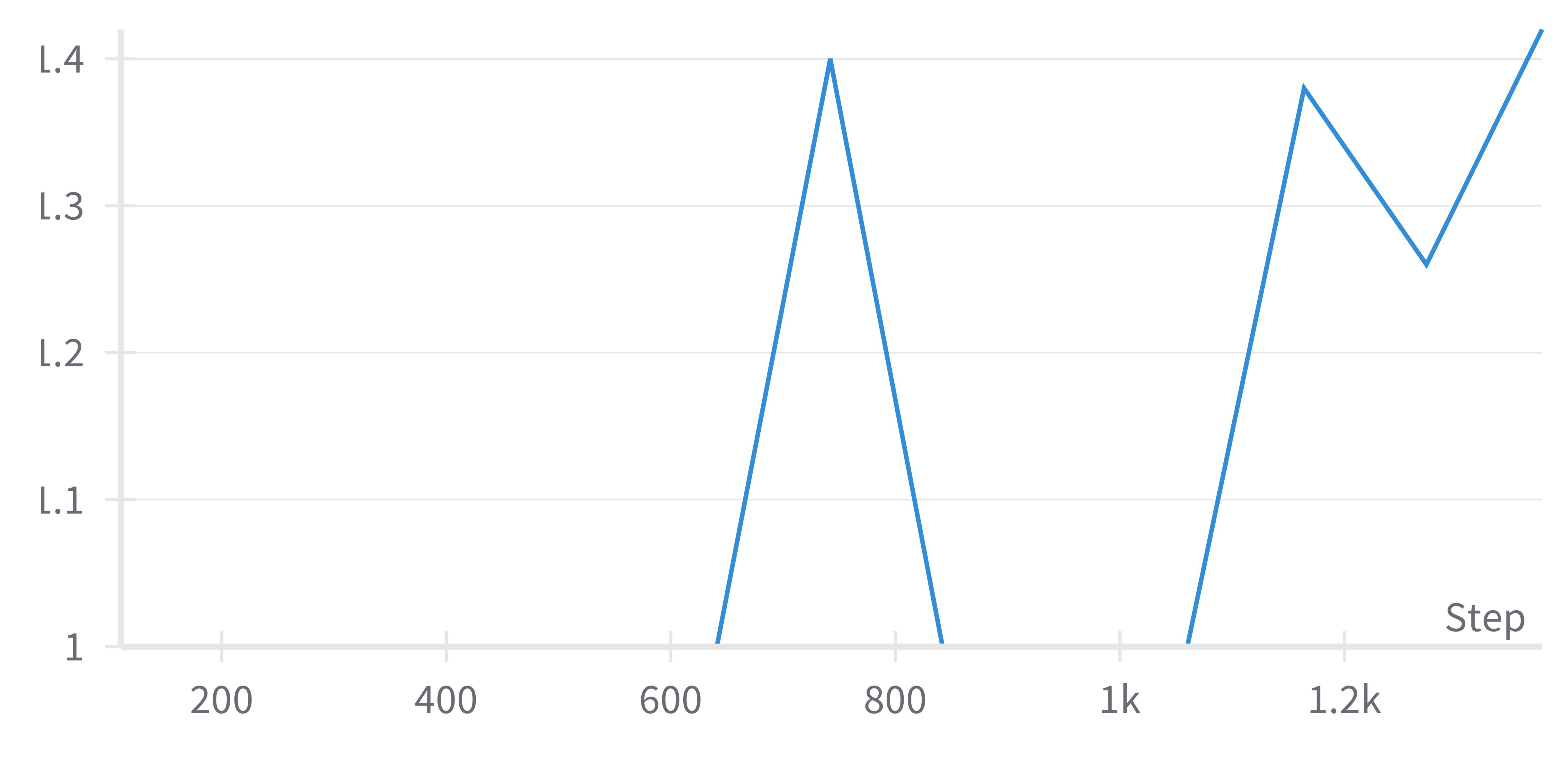}
        \caption{LLaMA-2-7b, $\alpha=0.2$}
    \end{subfigure}
    \begin{subfigure}[b]{0.24\textwidth}
        \includegraphics[width=\textwidth]{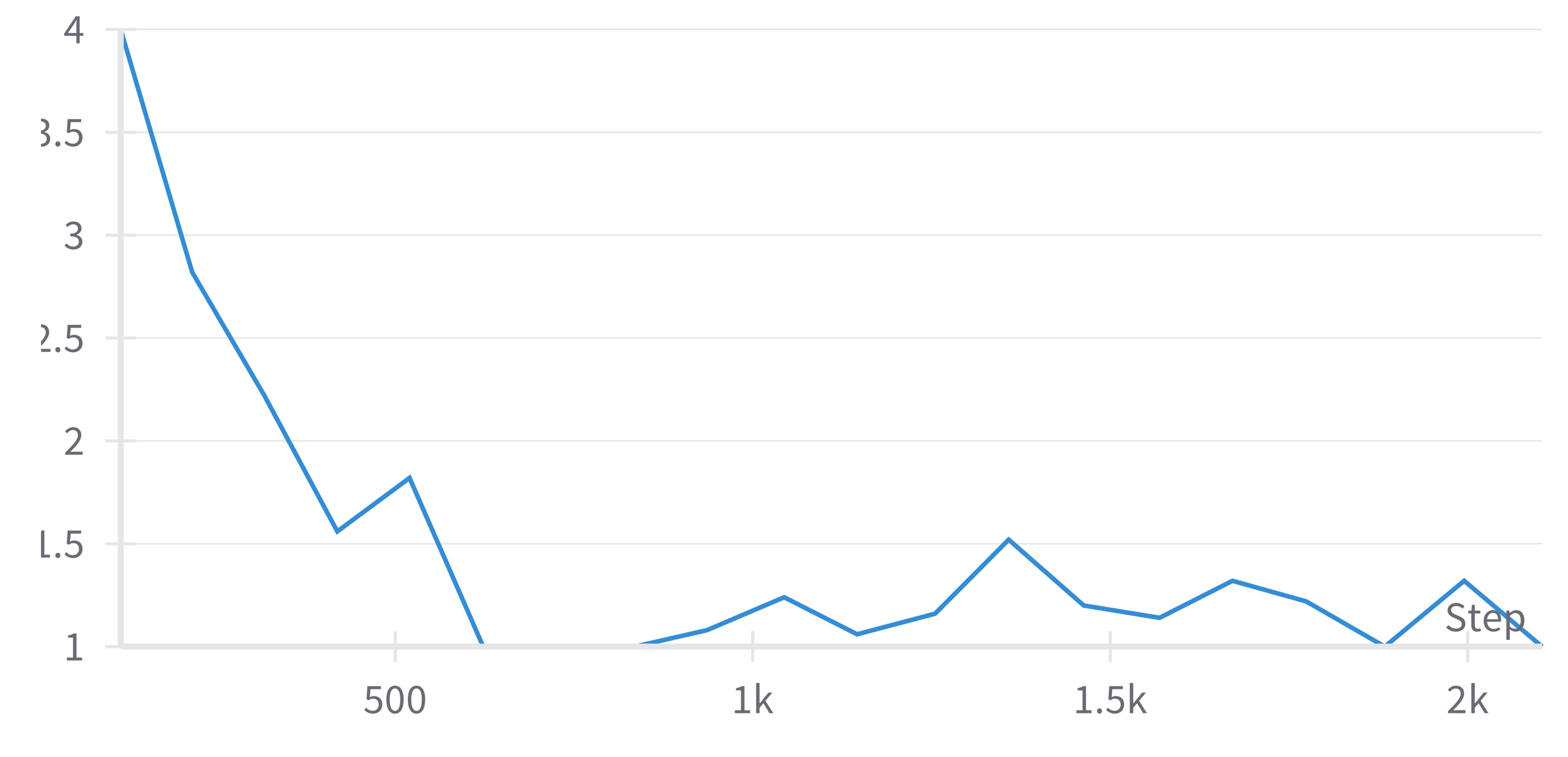}
        \caption{LLaMA-2-7b, $\alpha=0.1$}
    \end{subfigure}
    \begin{subfigure}[b]{0.24\textwidth}
        \includegraphics[width=\textwidth]{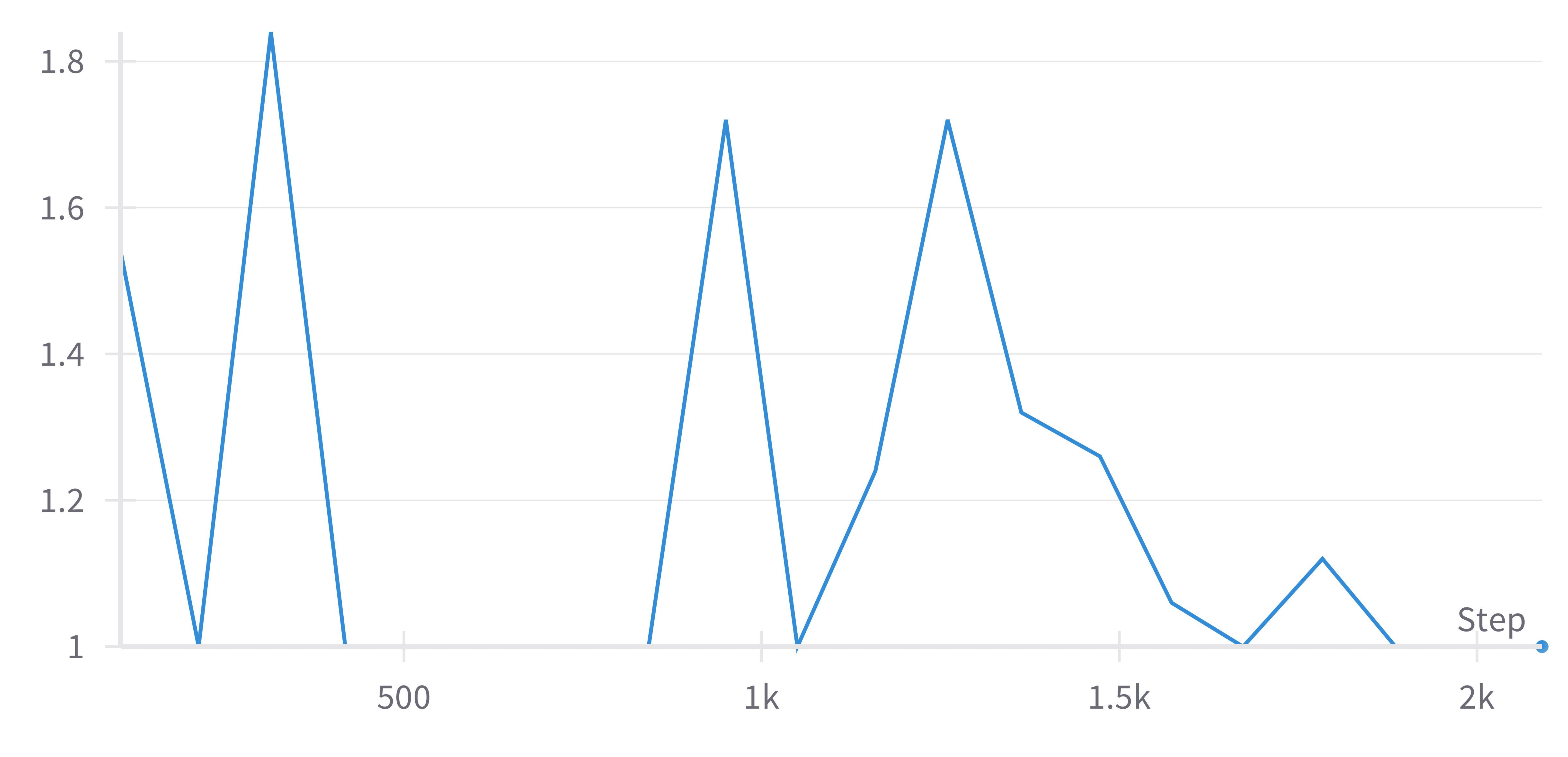}
        \caption{LLaMA-3.2-3b, $\alpha=0.1$}
    \end{subfigure}
    \begin{subfigure}[b]{0.24\textwidth}
        \includegraphics[width=\textwidth]{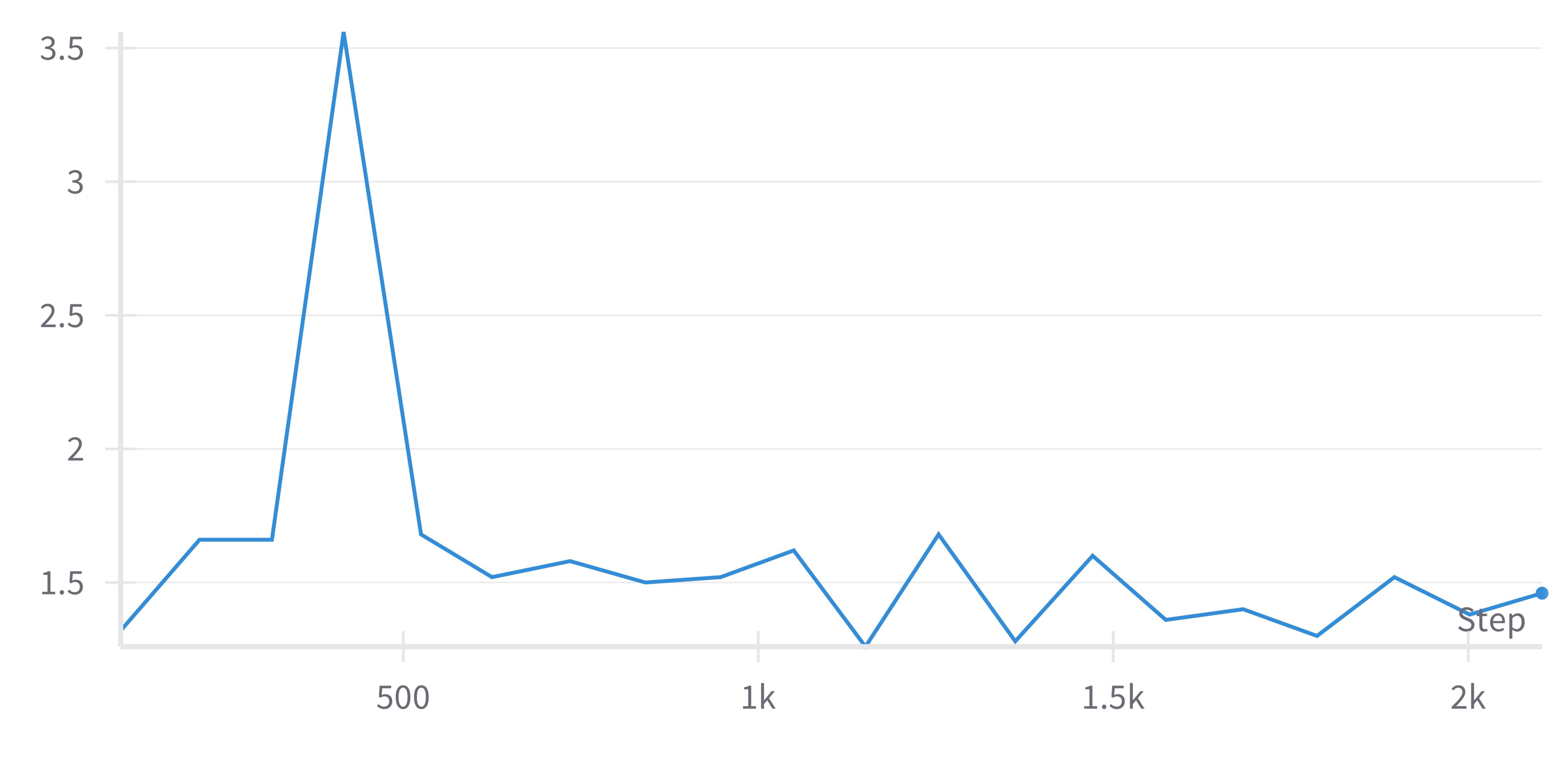}
        \caption{LLaMA-3.2-3b, $\alpha=0.05$}
    \end{subfigure}
    
    \caption{CCPO validation performance on HotpotQA dataset, $\lambda=0$.}
    \label{fig:val}
\end{figure*}

\begin{figure*}[htbp]
    \centering
    \vspace{1em}
    Train Coverage\\
    % First row: 4 images without subcaptions
    \includegraphics[width=0.24\textwidth]{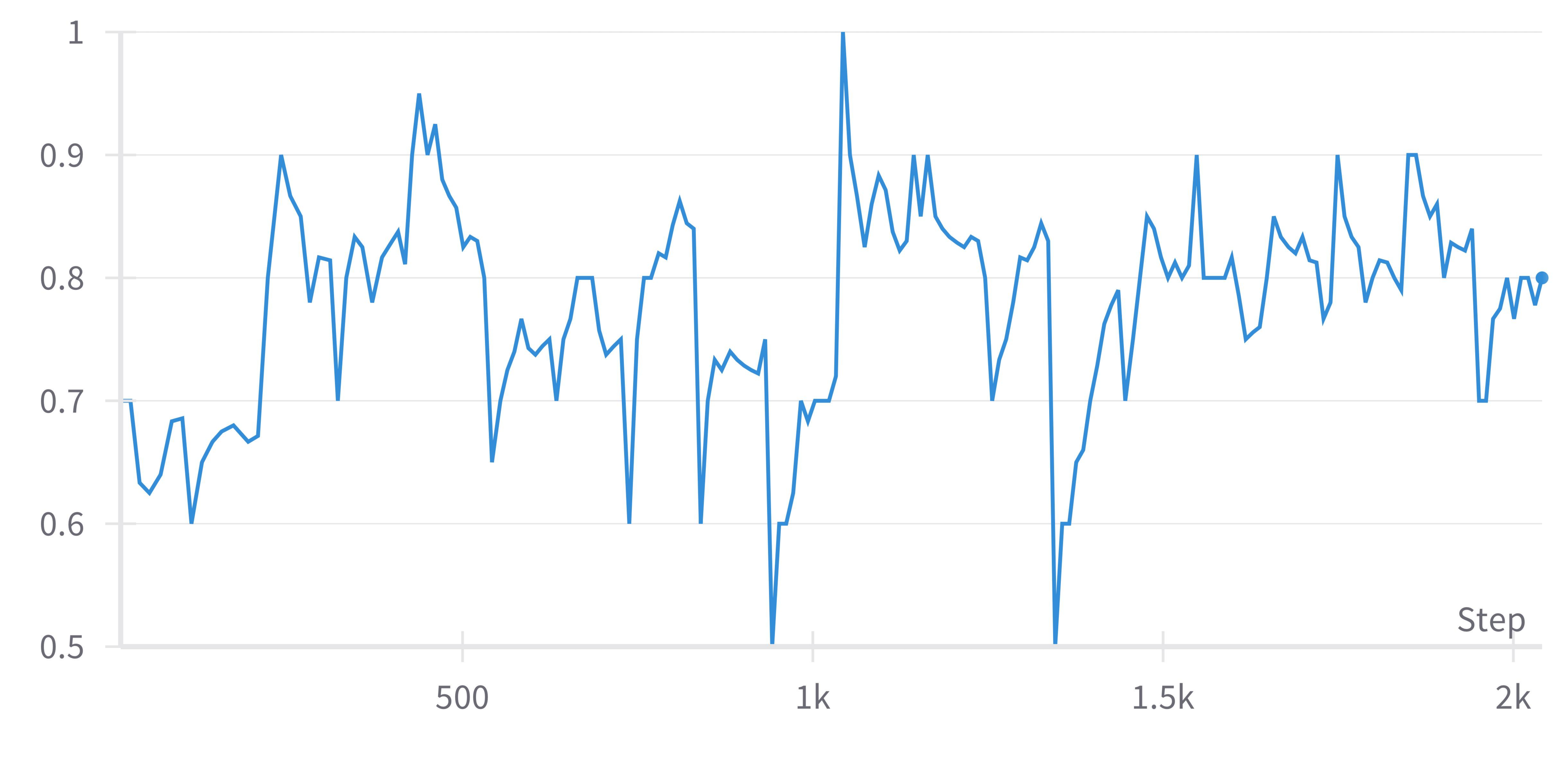}
    \includegraphics[width=0.24\textwidth]{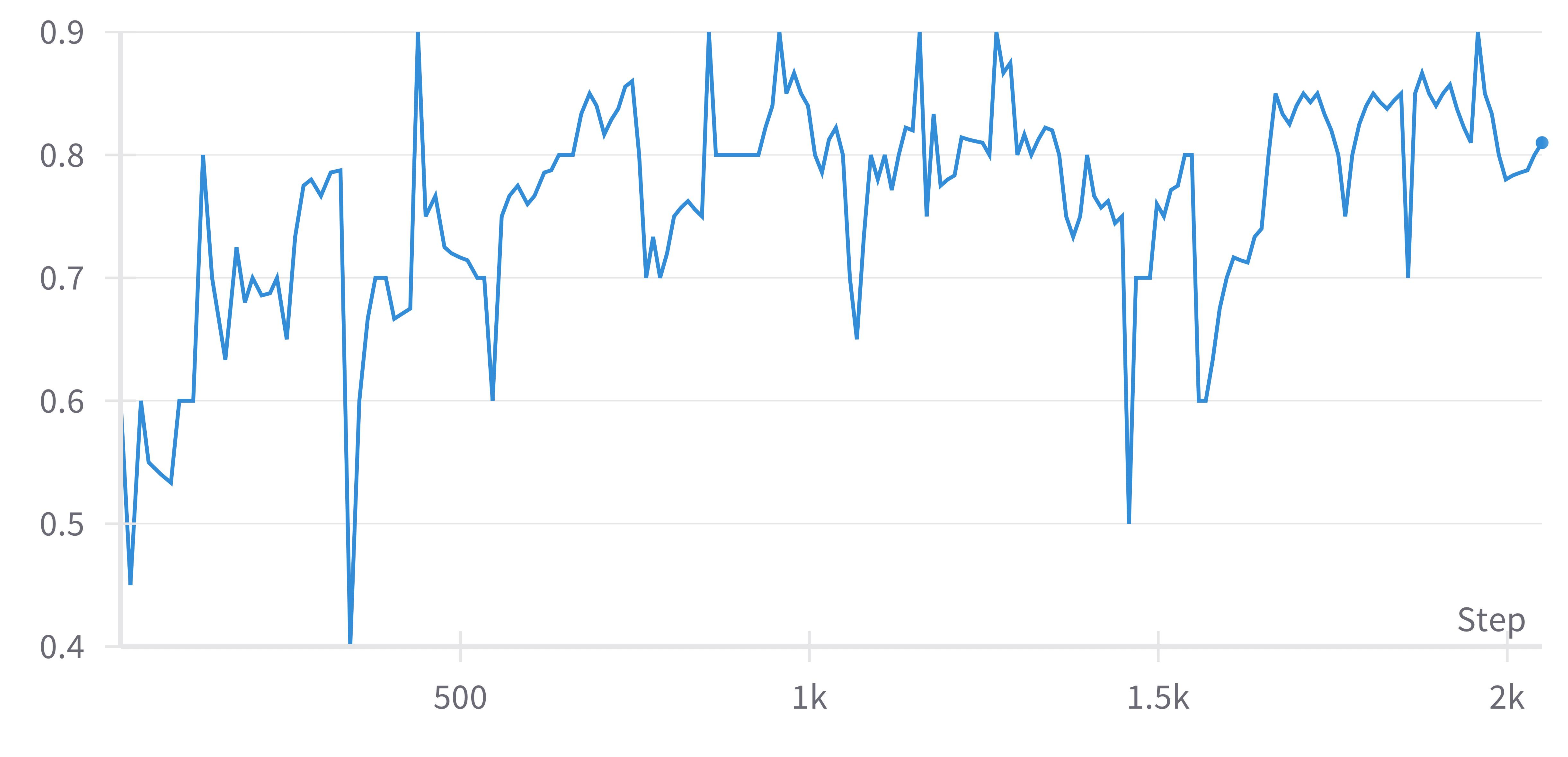}
    \includegraphics[width=0.24\textwidth]{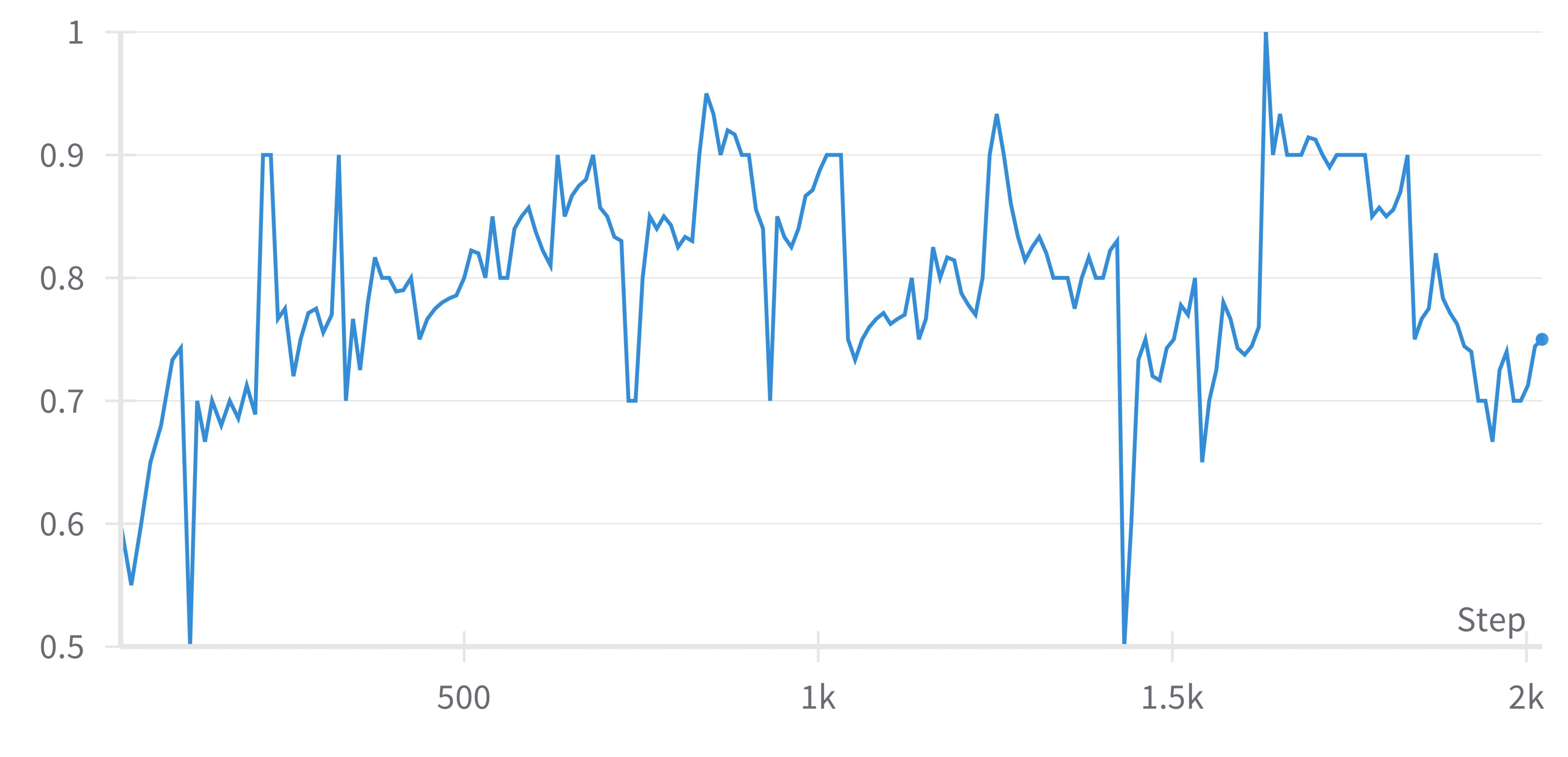}
    \includegraphics[width=0.24\textwidth]{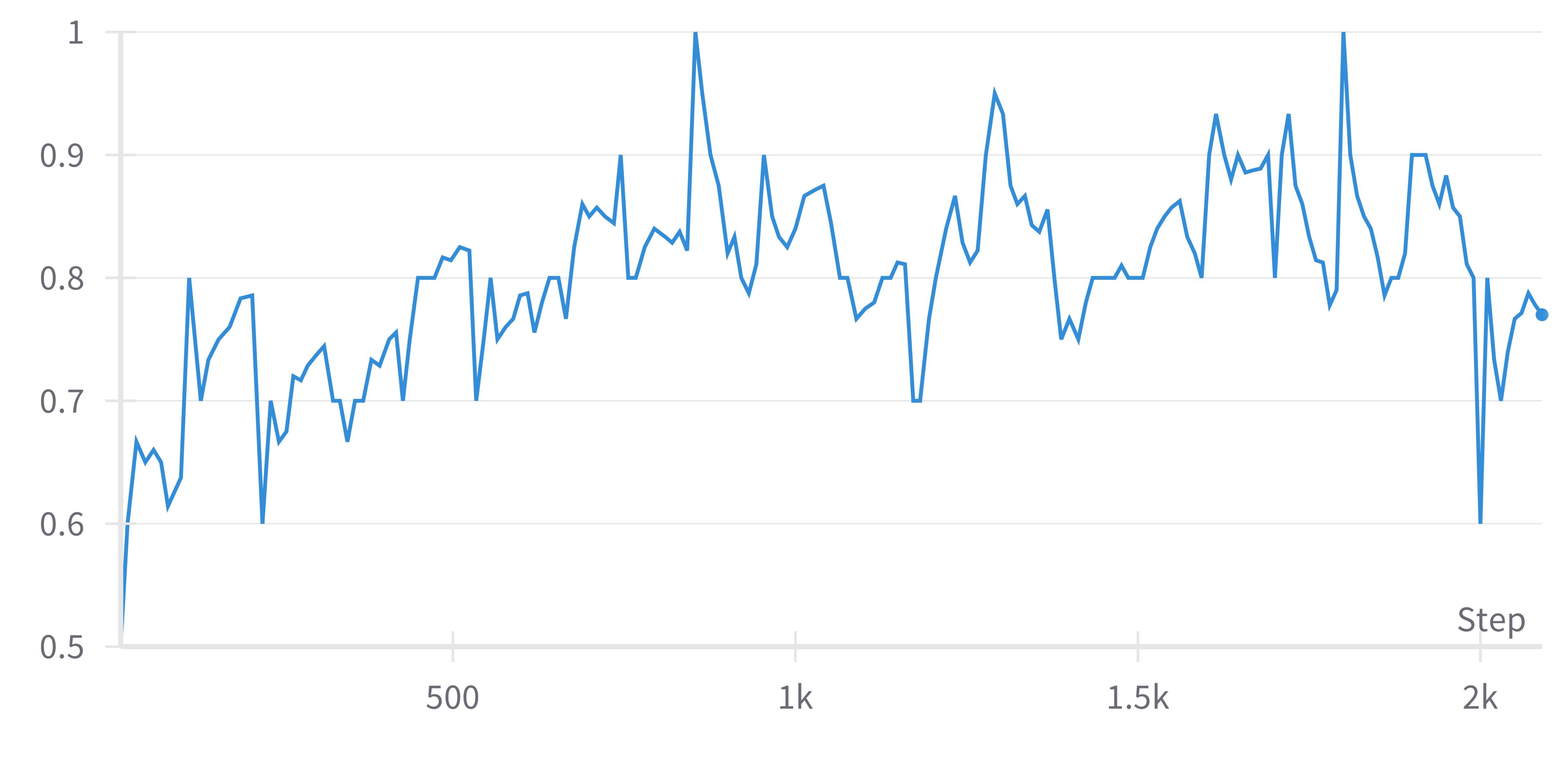}
    \vspace{1em}
    \text{Validation Coverage}\\
    % \vspace{0.5cm} % vertical space between rows
    \includegraphics[width=0.24\textwidth]{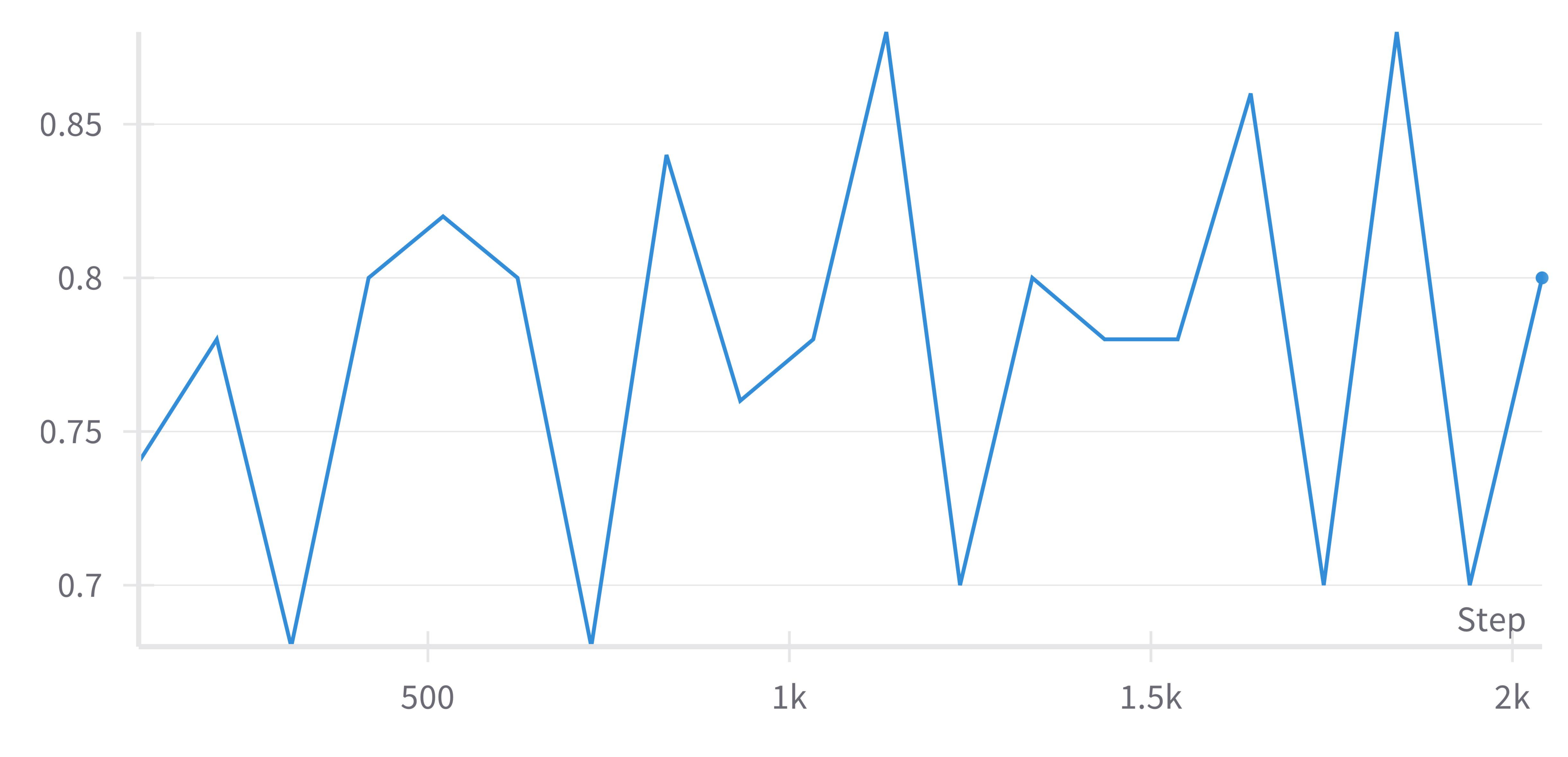}
    \includegraphics[width=0.24\textwidth]{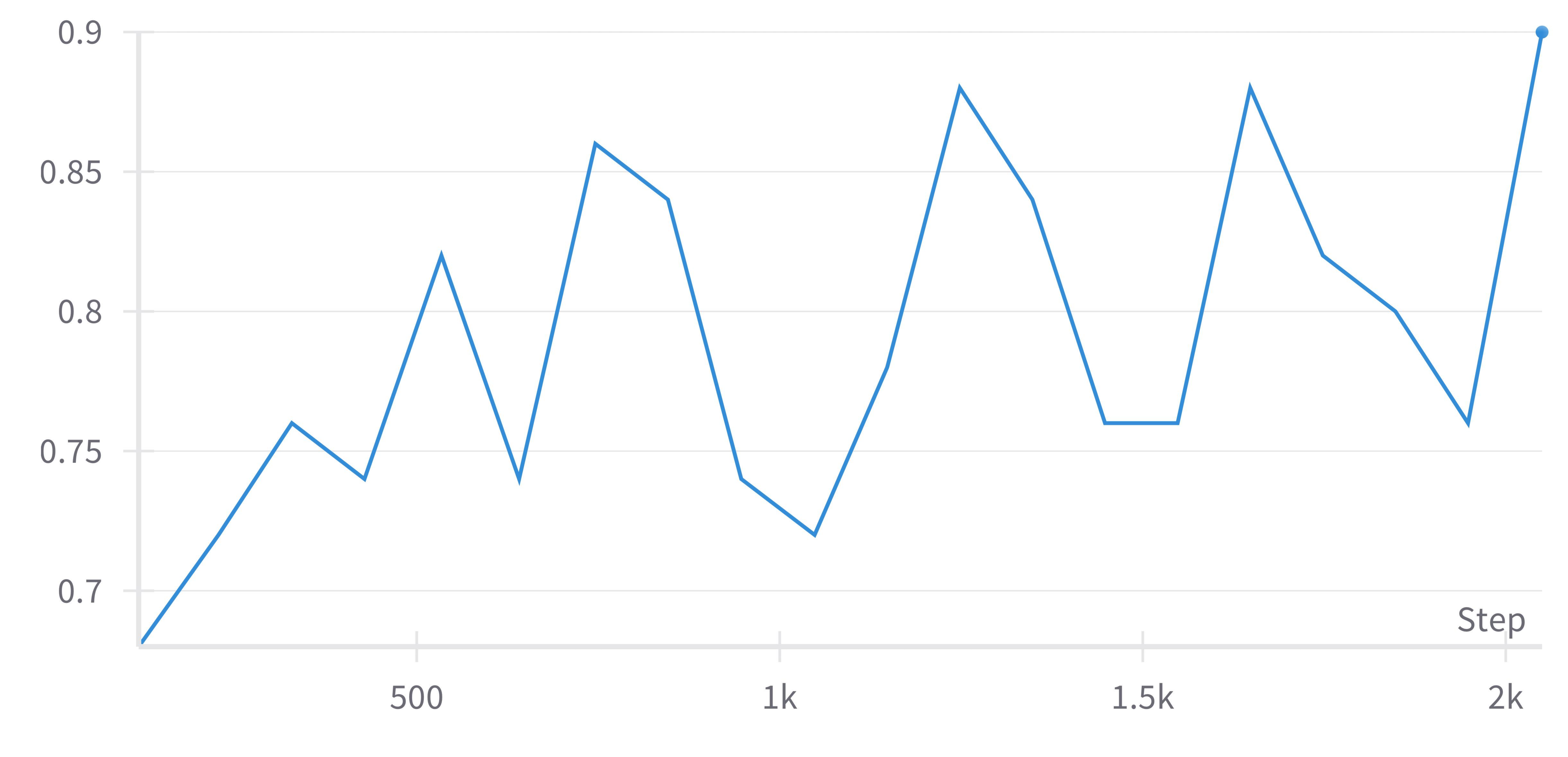}
    \includegraphics[width=0.24\textwidth]{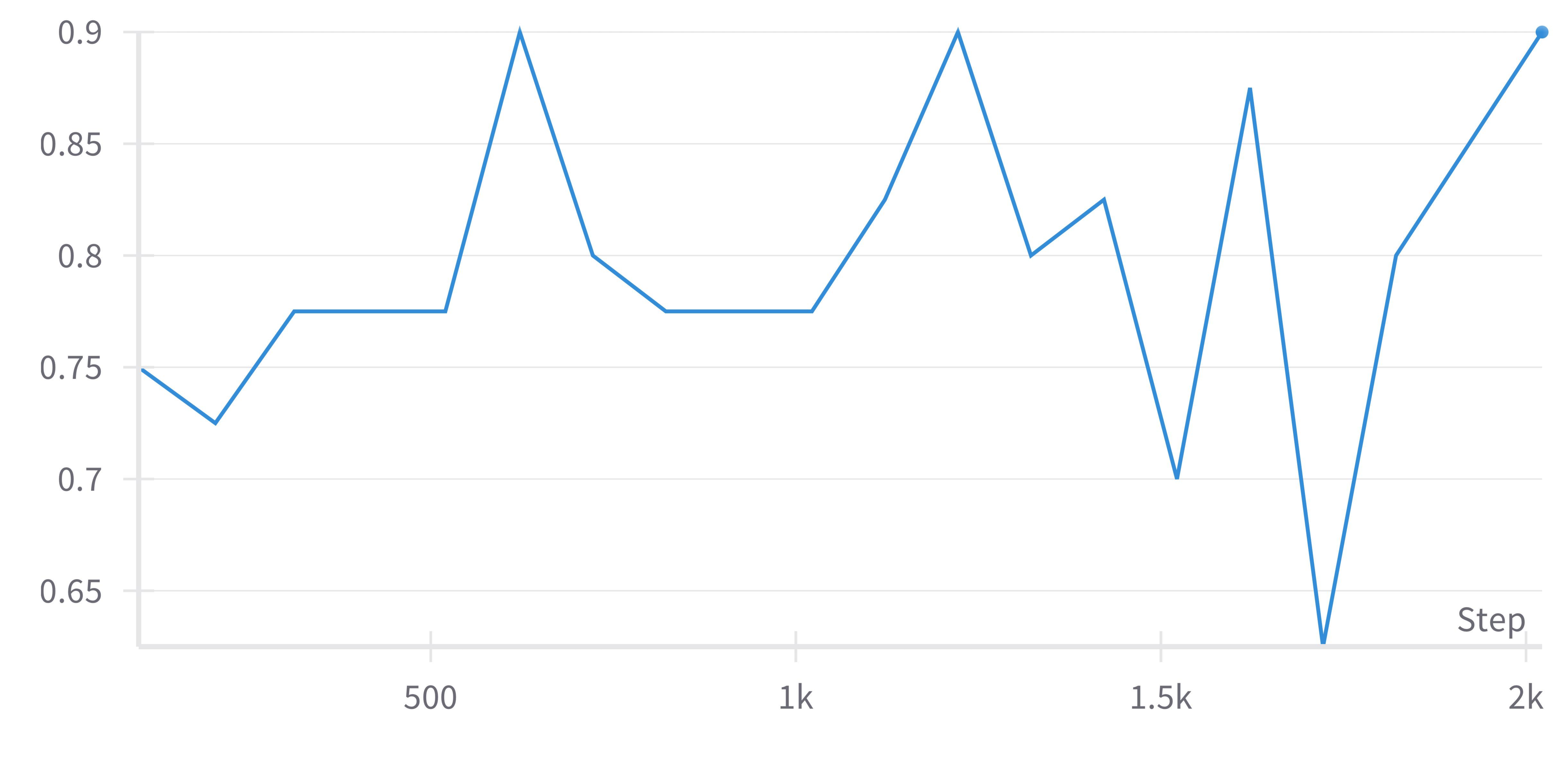}
    \includegraphics[width=0.24\textwidth]{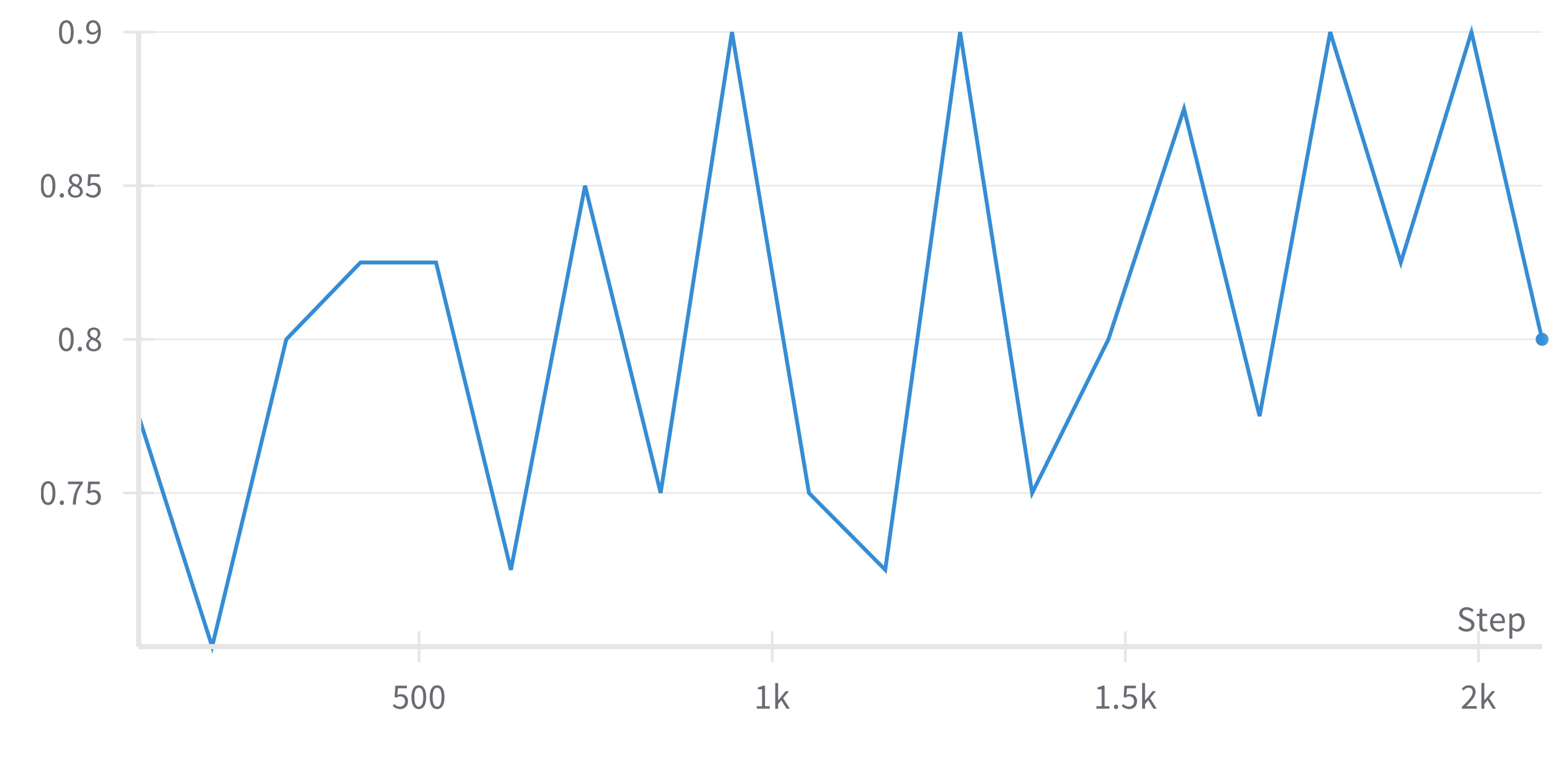}
    \vspace{1em}
    \text{Validation Length}\\
    % Second row: 4 images with subcaptions
    \begin{subfigure}[b]{0.24\textwidth}
        \includegraphics[width=\textwidth]{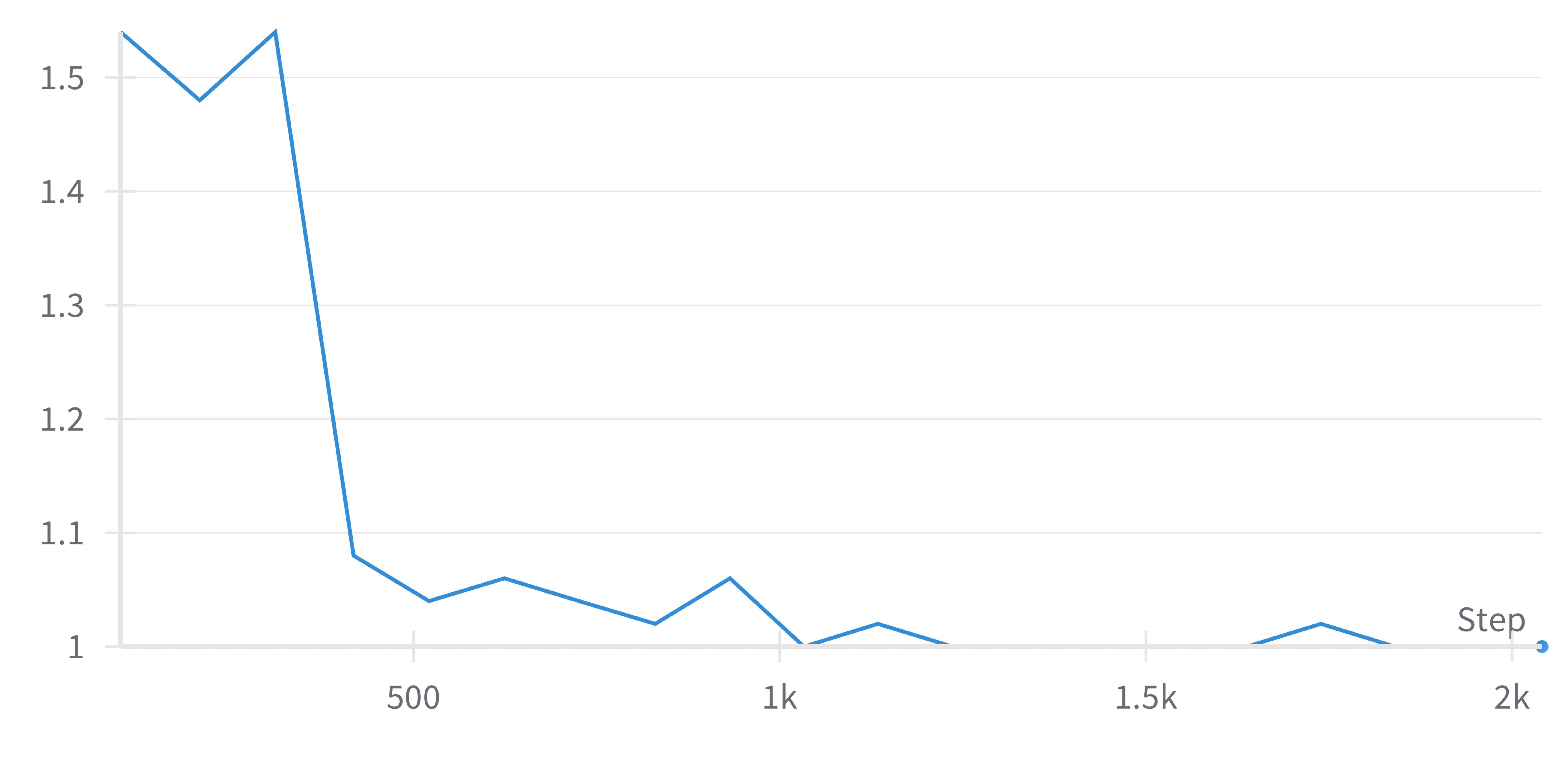}
        \caption{LLaMA-2-7b, $\alpha=0.2$}
    \end{subfigure}
    \begin{subfigure}[b]{0.24\textwidth}
        \includegraphics[width=\textwidth]{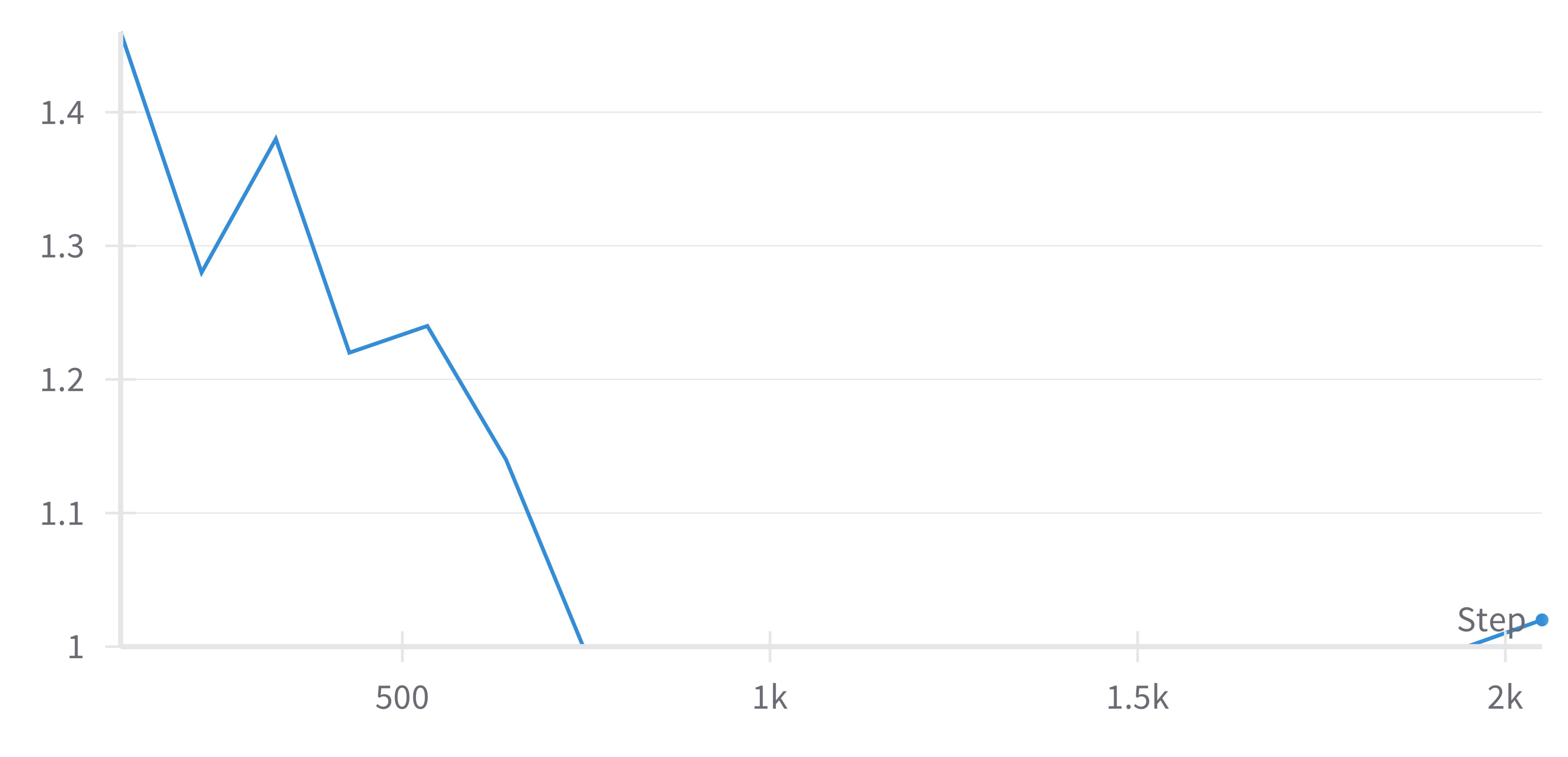}
        \caption{LLaMA-2-7b, $\alpha=0.1$}
    \end{subfigure}
    \begin{subfigure}[b]{0.24\textwidth}
        \includegraphics[width=\textwidth]{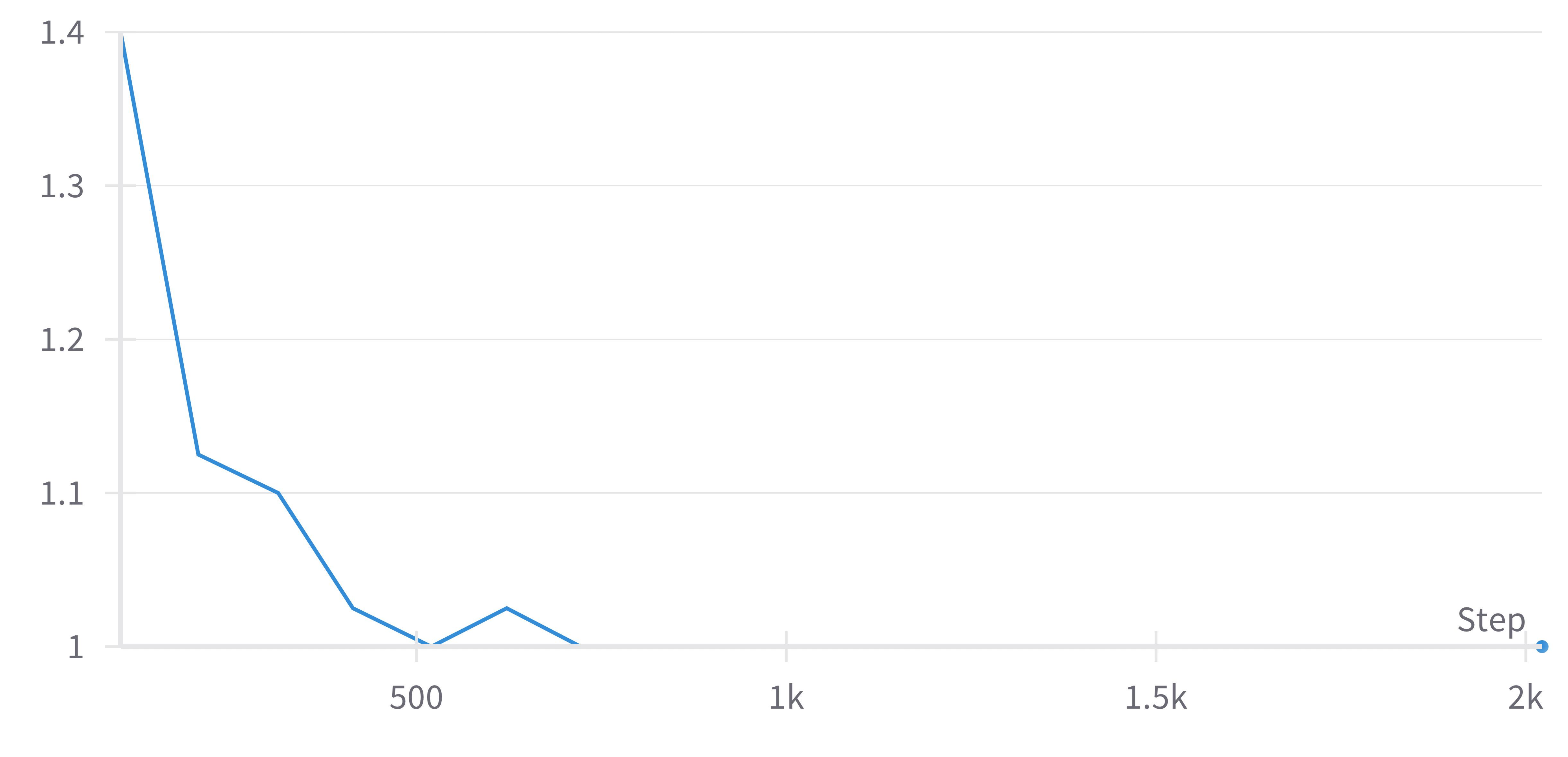}
        \caption{LLaMA-3.2-3b, $\alpha=0.1$}
    \end{subfigure}
    \begin{subfigure}[b]{0.24\textwidth}
        \includegraphics[width=\textwidth]{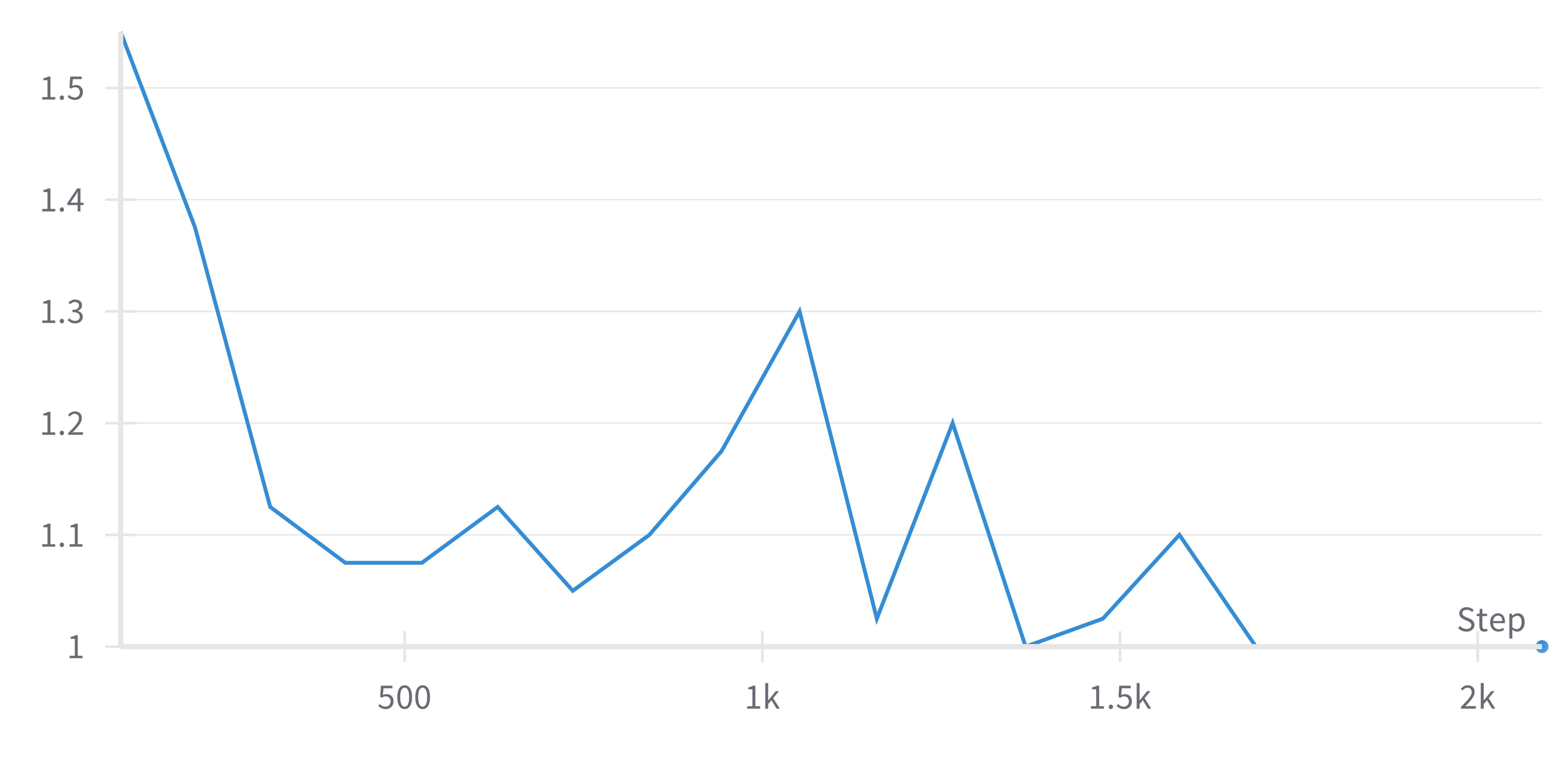}
        \caption{LLaMA-3.2-3b, $\alpha=0.05$}
    \end{subfigure}
    
    \caption{CPO performance on HotpotQA dataset.}
    \label{fig:cpo_perf}
\end{figure*}

\begin{figure*}[htbp]
    \centering
    \vspace{1em}
    Costs\\
    % First row: 4 images without subcaptions
    \includegraphics[width=0.24\textwidth]{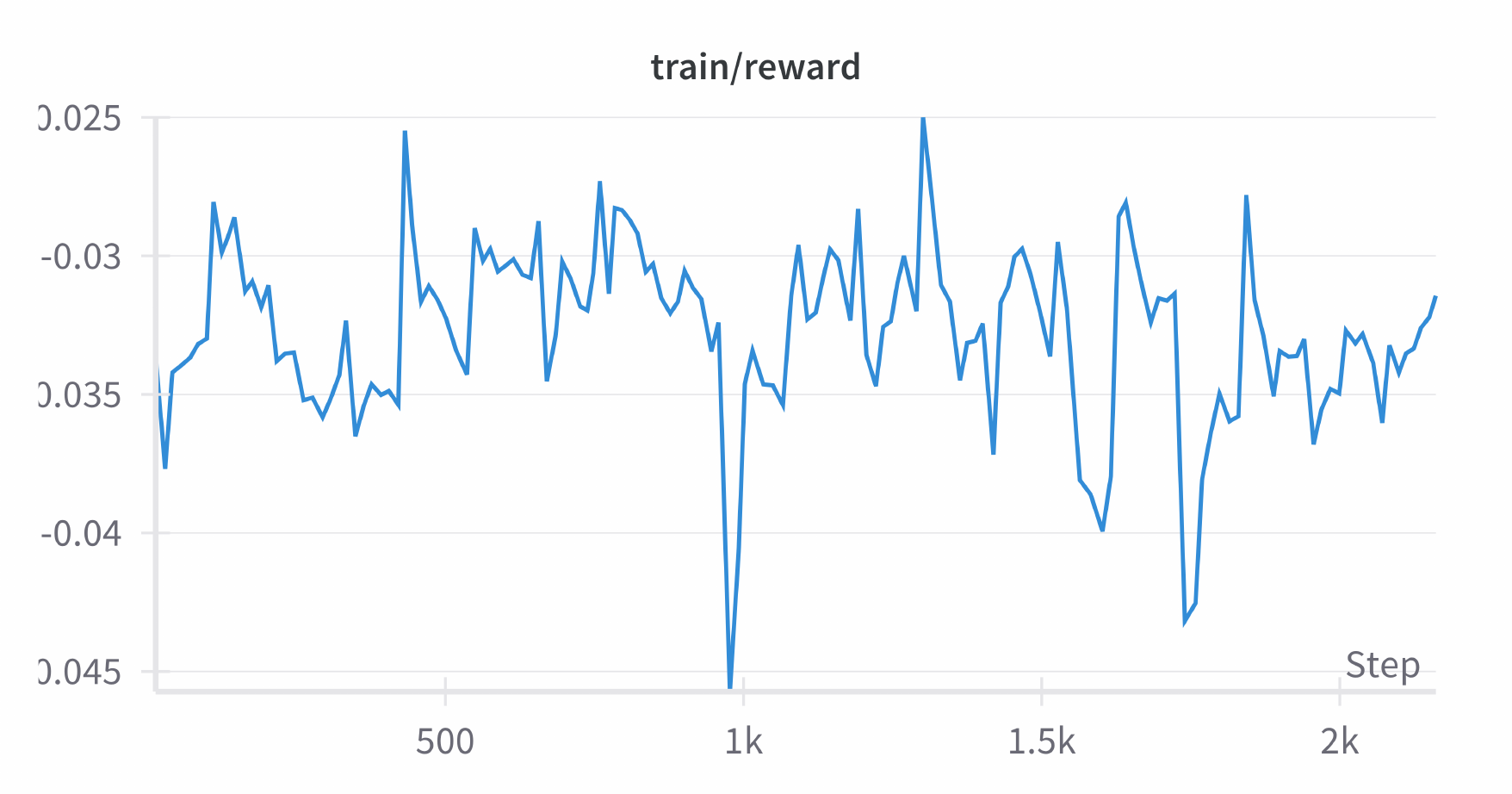}
    \includegraphics[width=0.24\textwidth]{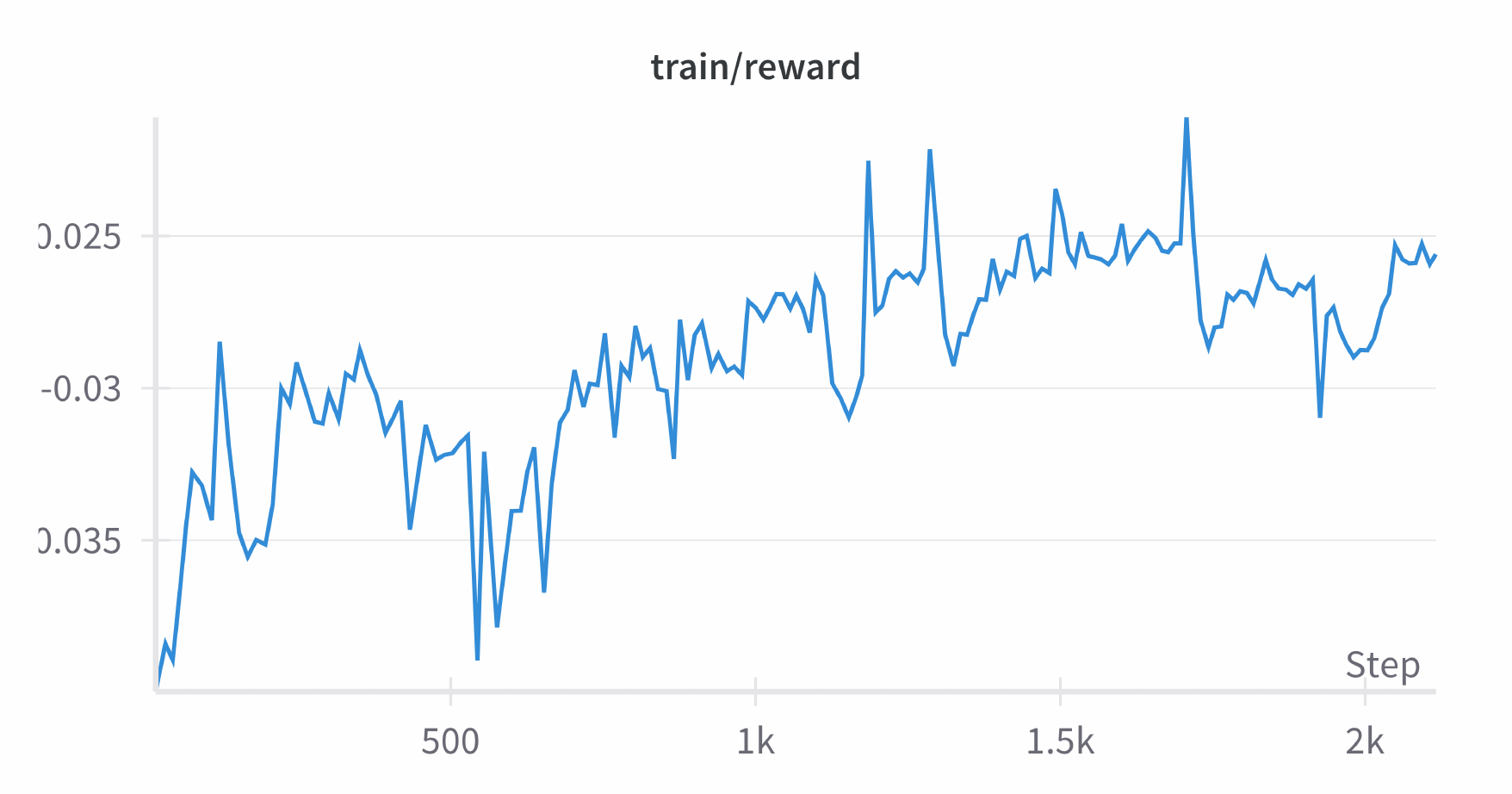}
    \includegraphics[width=0.24\textwidth]{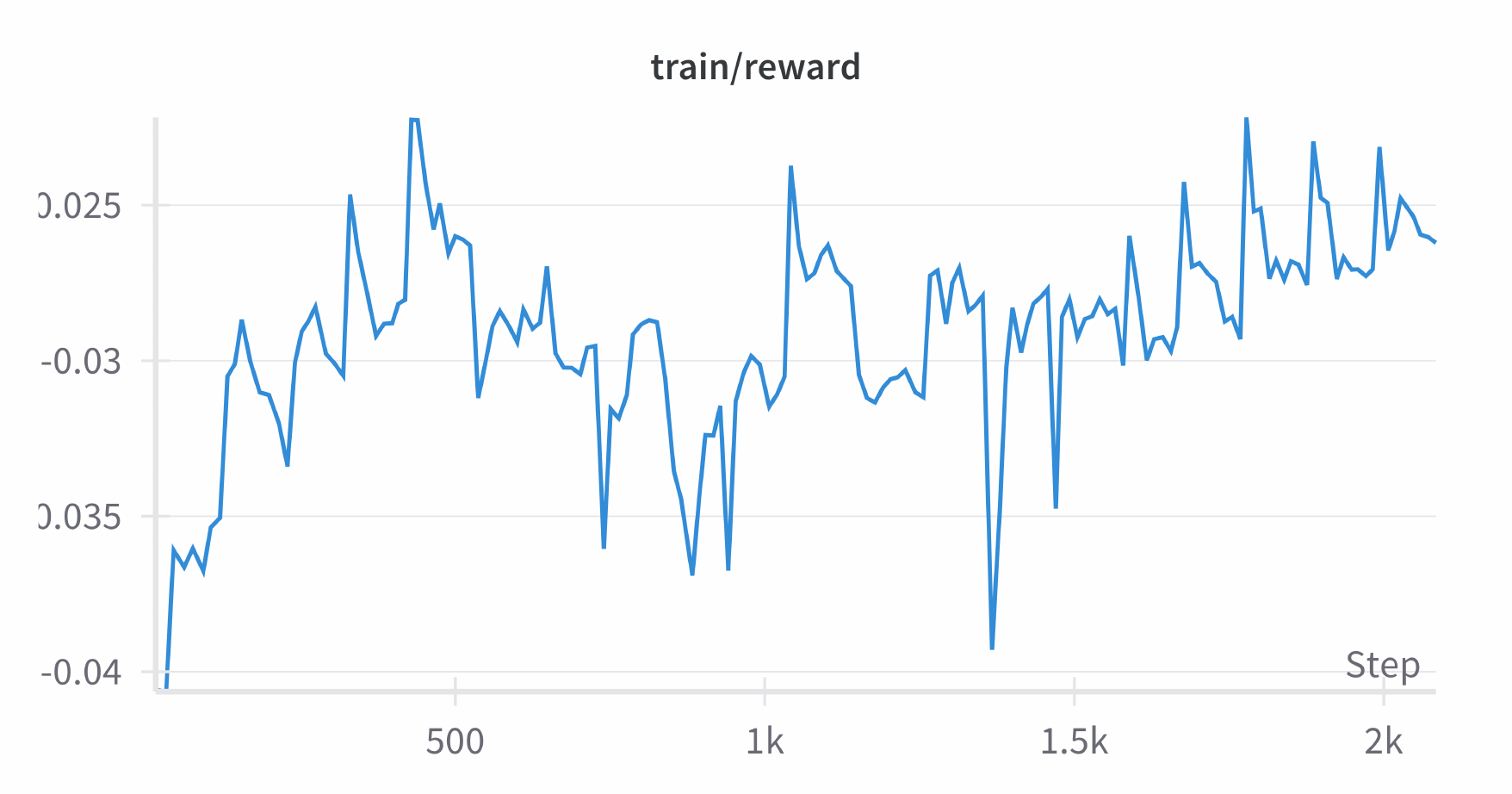}
    \includegraphics[width=0.24\textwidth]{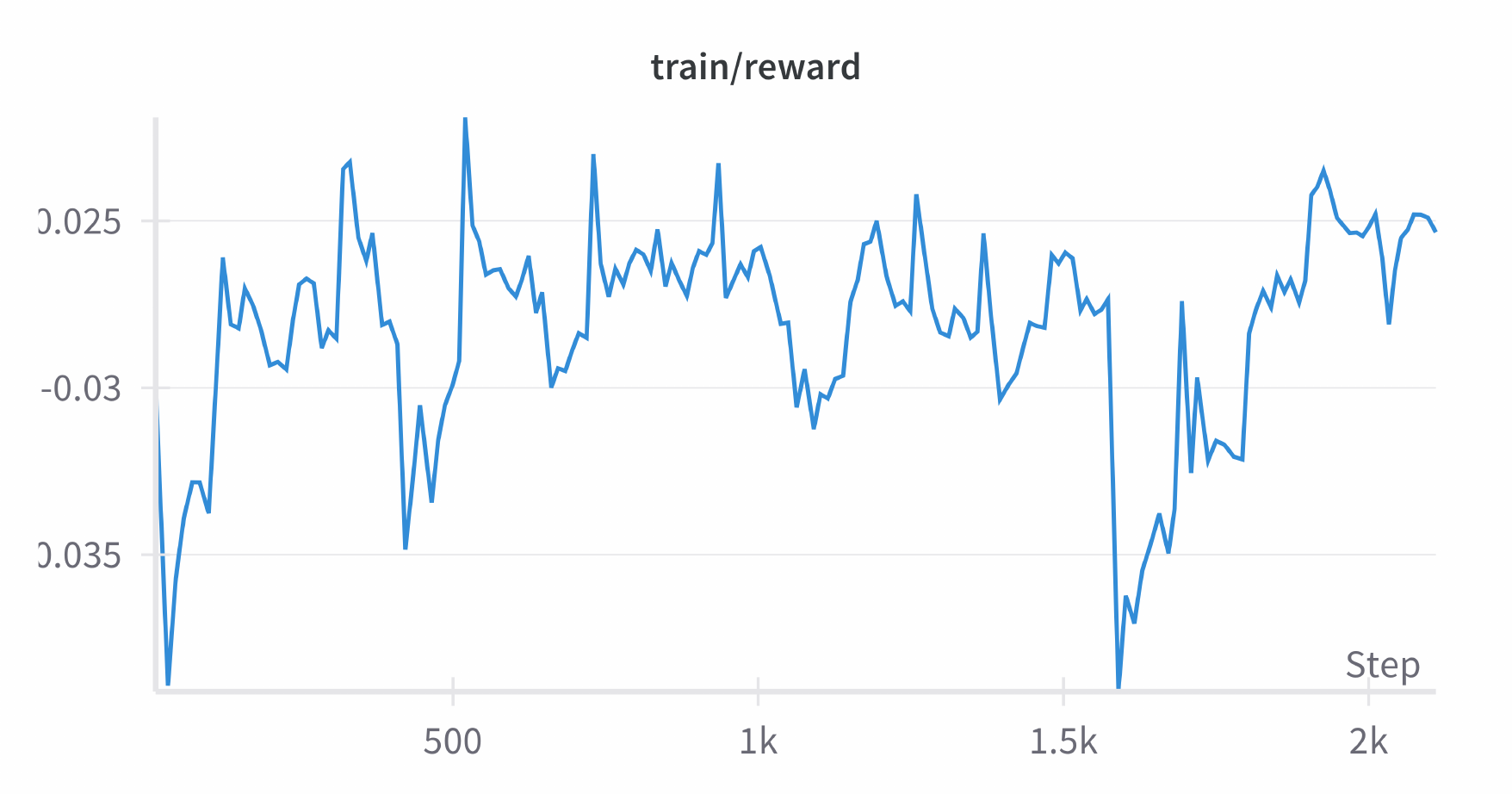}
    \vspace{1em}
    \text{Coverage Surrogate Violation}\\
    % \vspace{0.5cm} % vertical space between rows
    \includegraphics[width=0.24\textwidth]{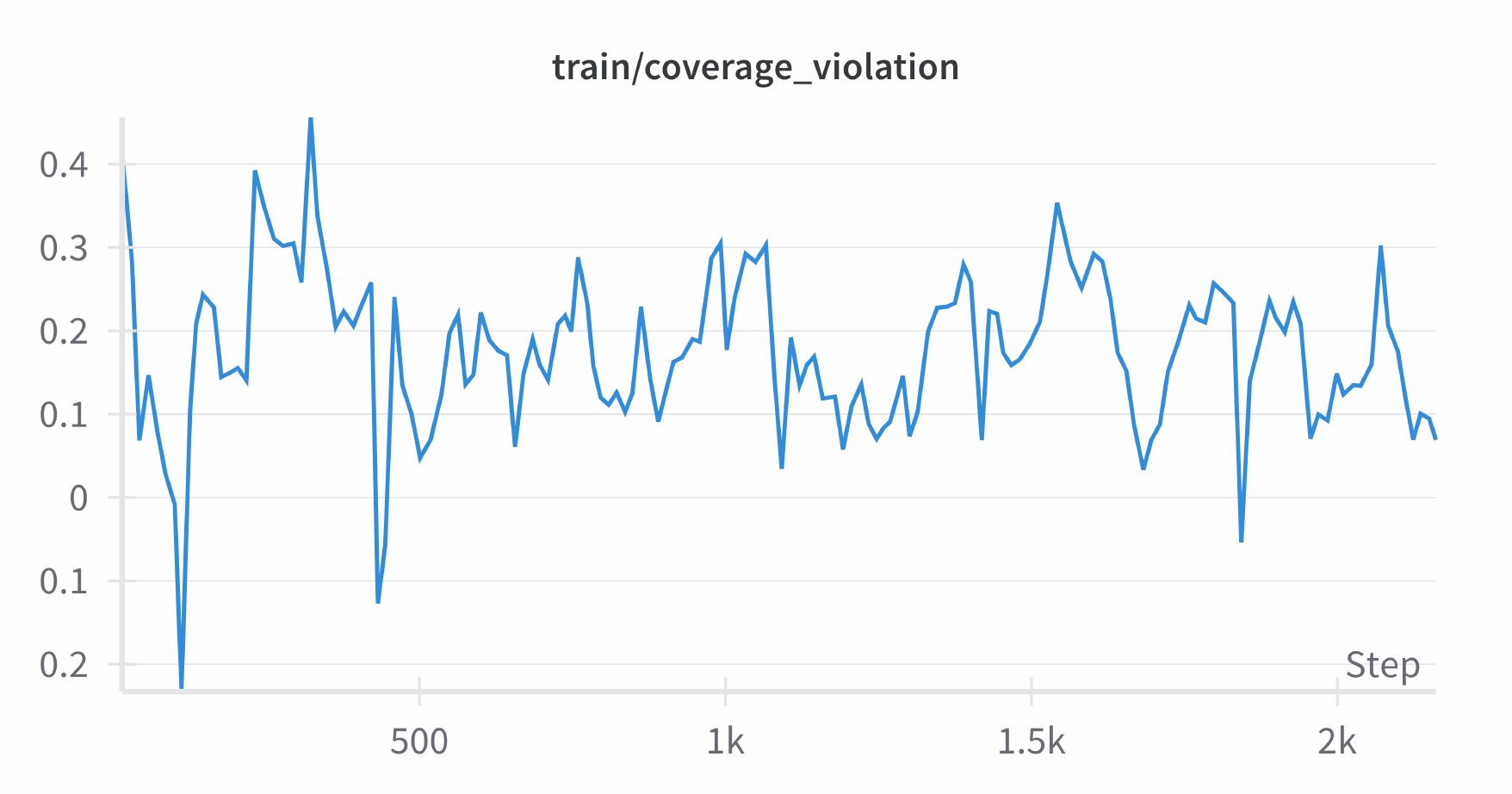}
    \includegraphics[width=0.24\textwidth]{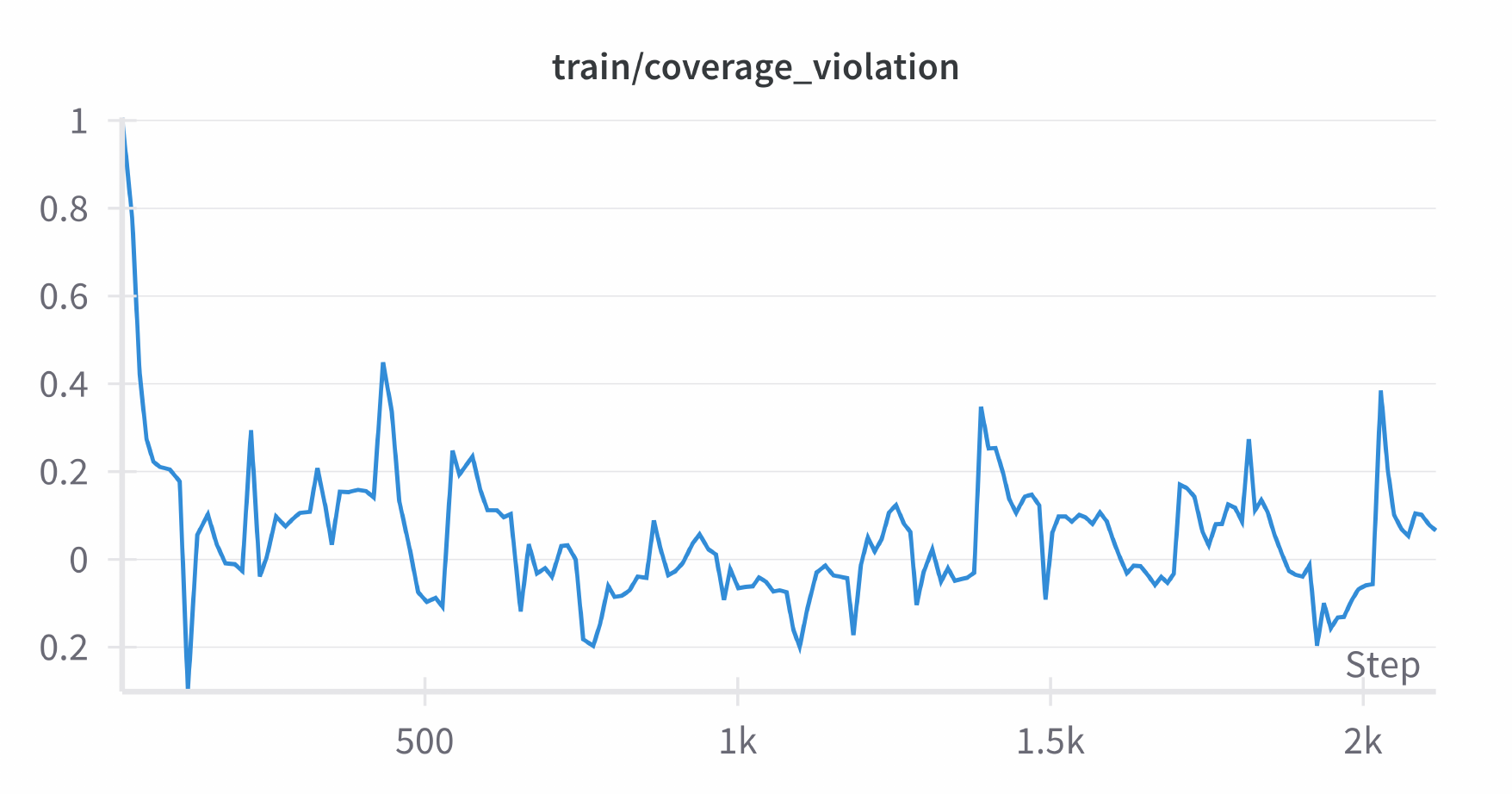}
    \includegraphics[width=0.24\textwidth]{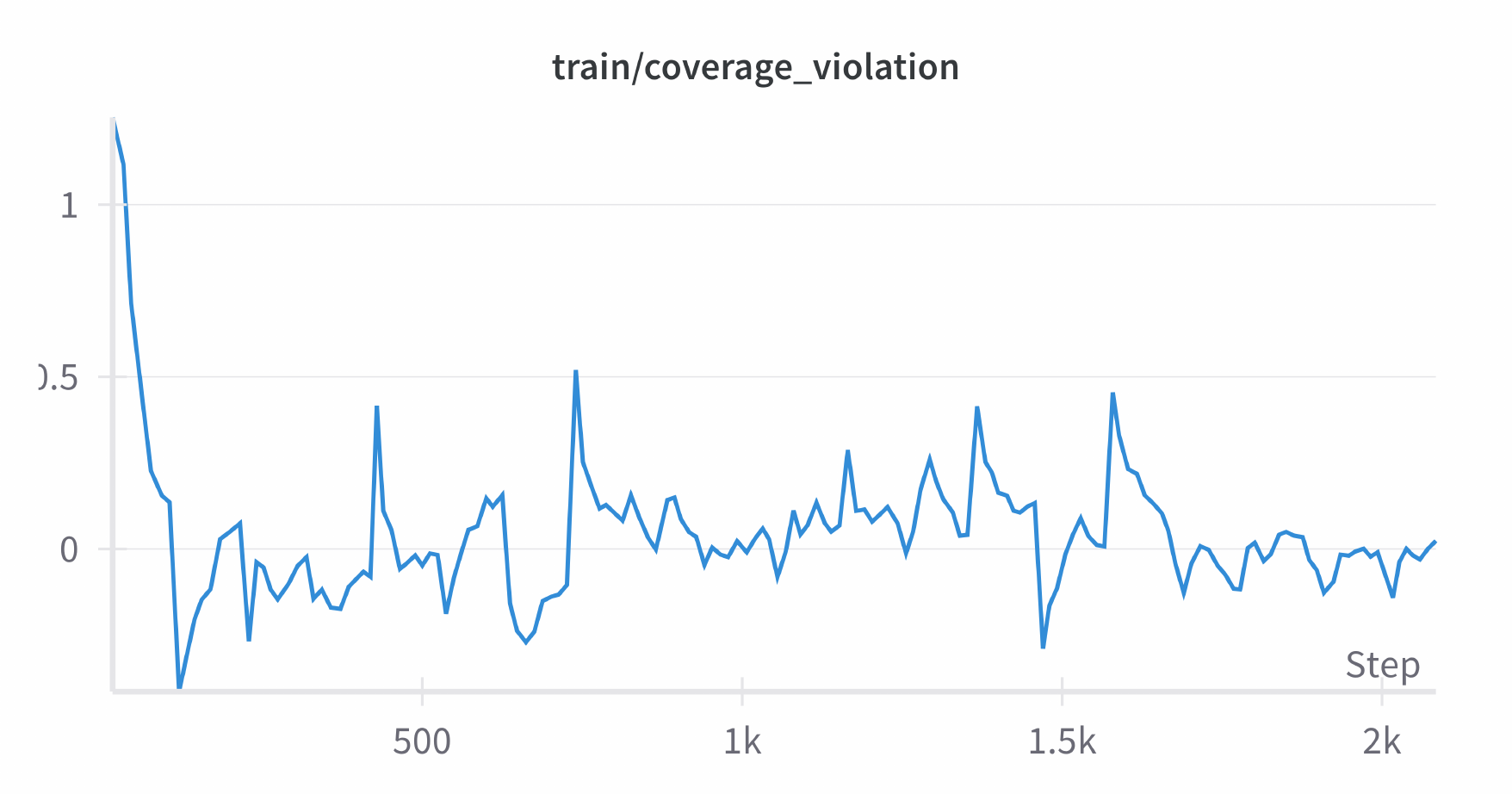}
    \includegraphics[width=0.24\textwidth]{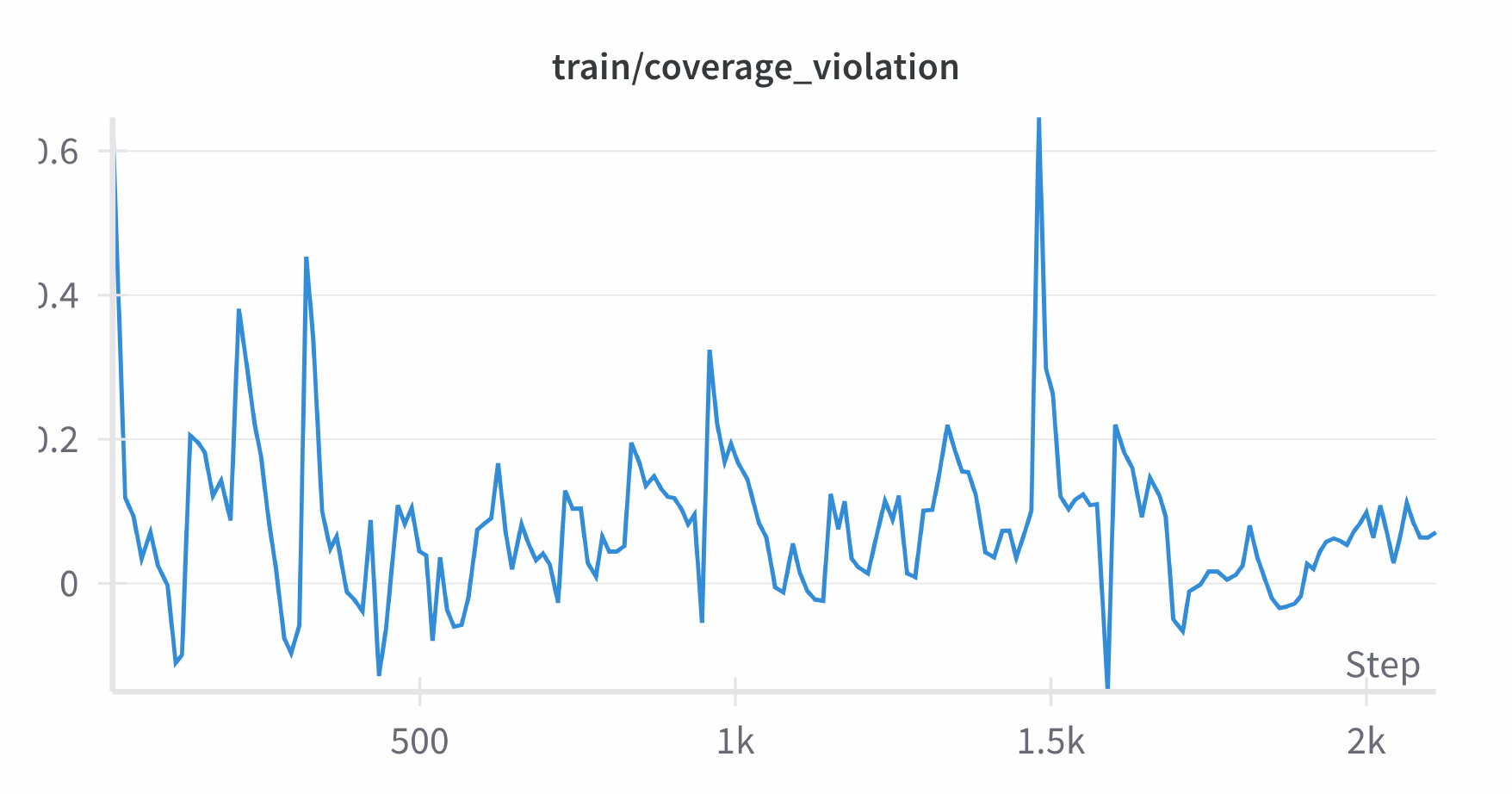}
    \vspace{1em}
    \text{Training Coverage}\\
    % Second row: 4 images with subcaptions
    \begin{subfigure}[b]{0.24\textwidth}
        \includegraphics[width=\textwidth]{ 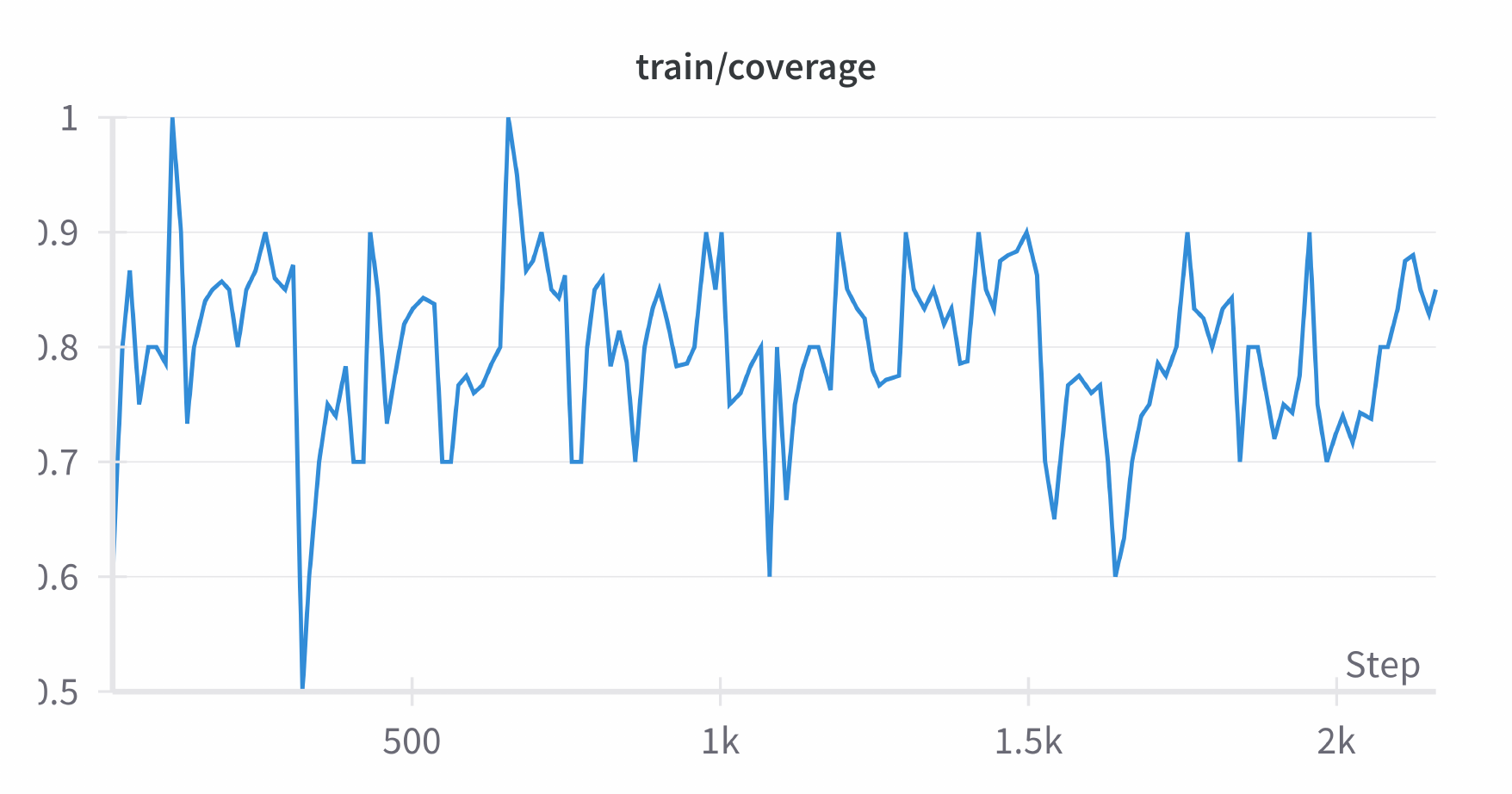}
        \caption{LLaMA-2-7b, $\alpha=0.2$,\\ $\lambda=1e^{-4}$.}
    \end{subfigure}
    \begin{subfigure}[b]{0.24\textwidth}
        \includegraphics[width=\textwidth]{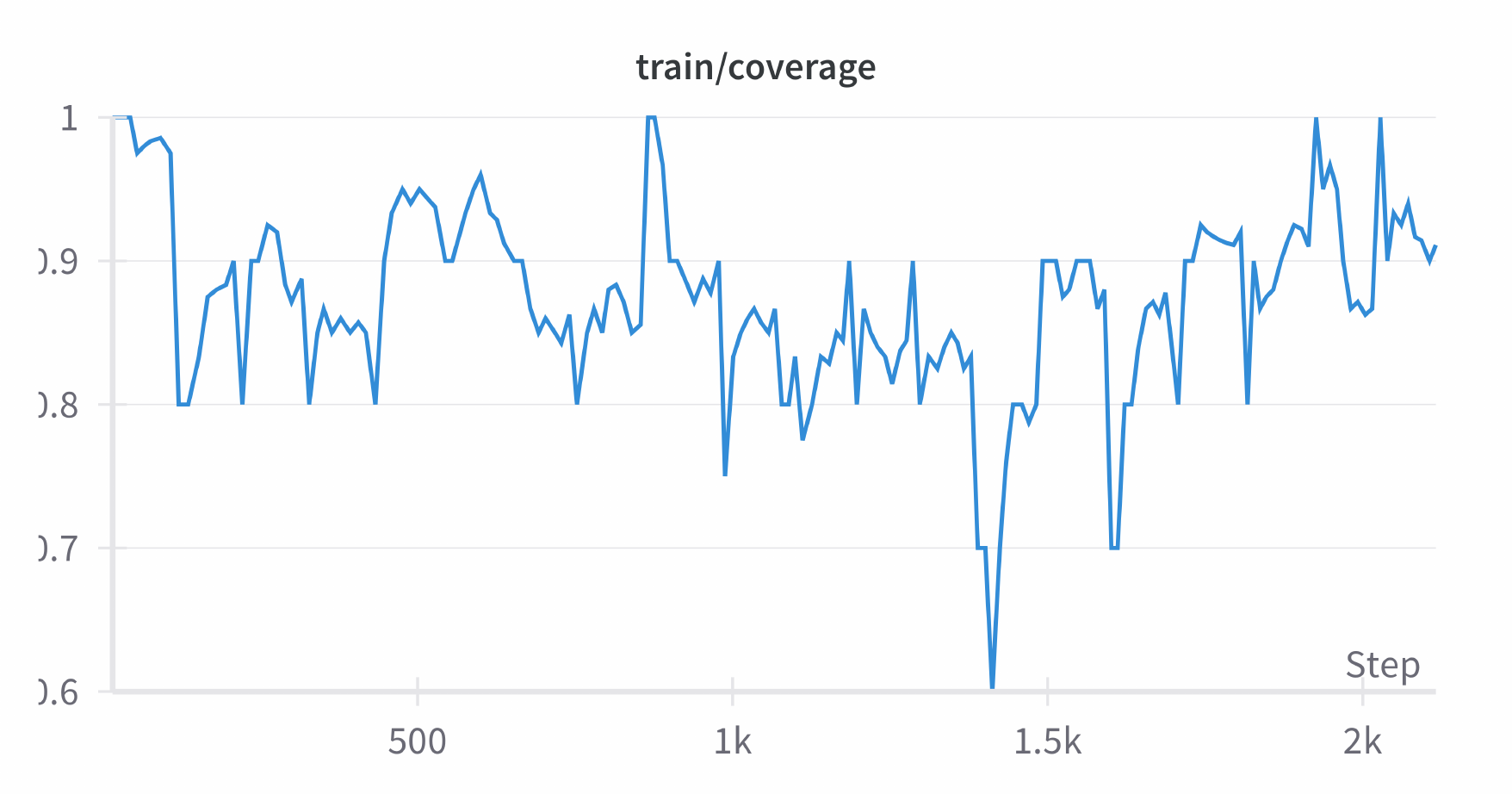}
        \caption{LLaMA-2-7b, $\alpha=0.1$,\\ $\lambda=2e^{-4}$.}
    \end{subfigure}
    \begin{subfigure}[b]{0.24\textwidth}
        \includegraphics[width=\textwidth]{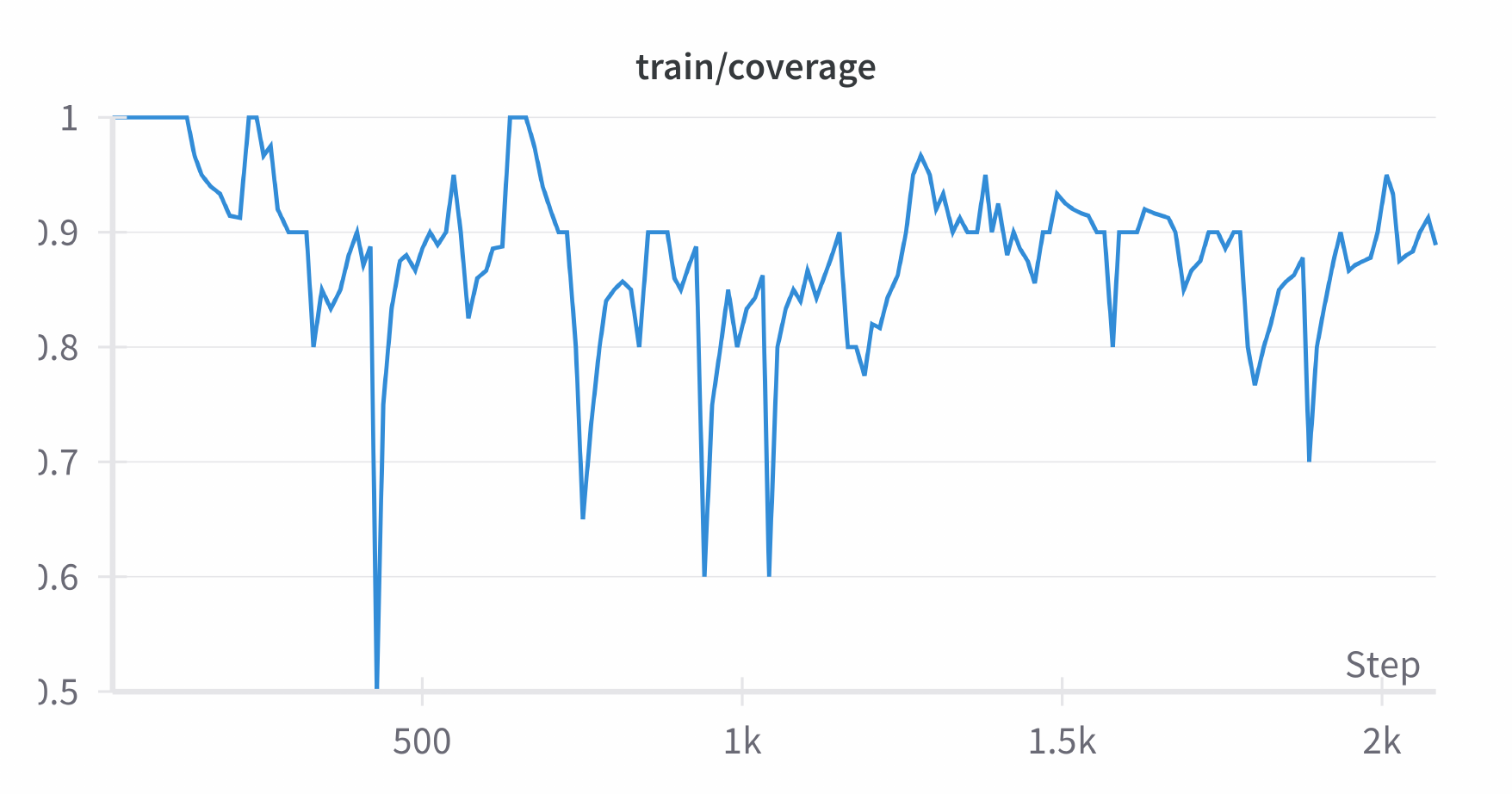}
        \caption{LLaMA-3.2-3b, $\alpha=0.1$,\\ $\lambda=1e^{-4}$.}
    \end{subfigure}
    \begin{subfigure}[b]{0.24\textwidth}
        \includegraphics[width=\textwidth]{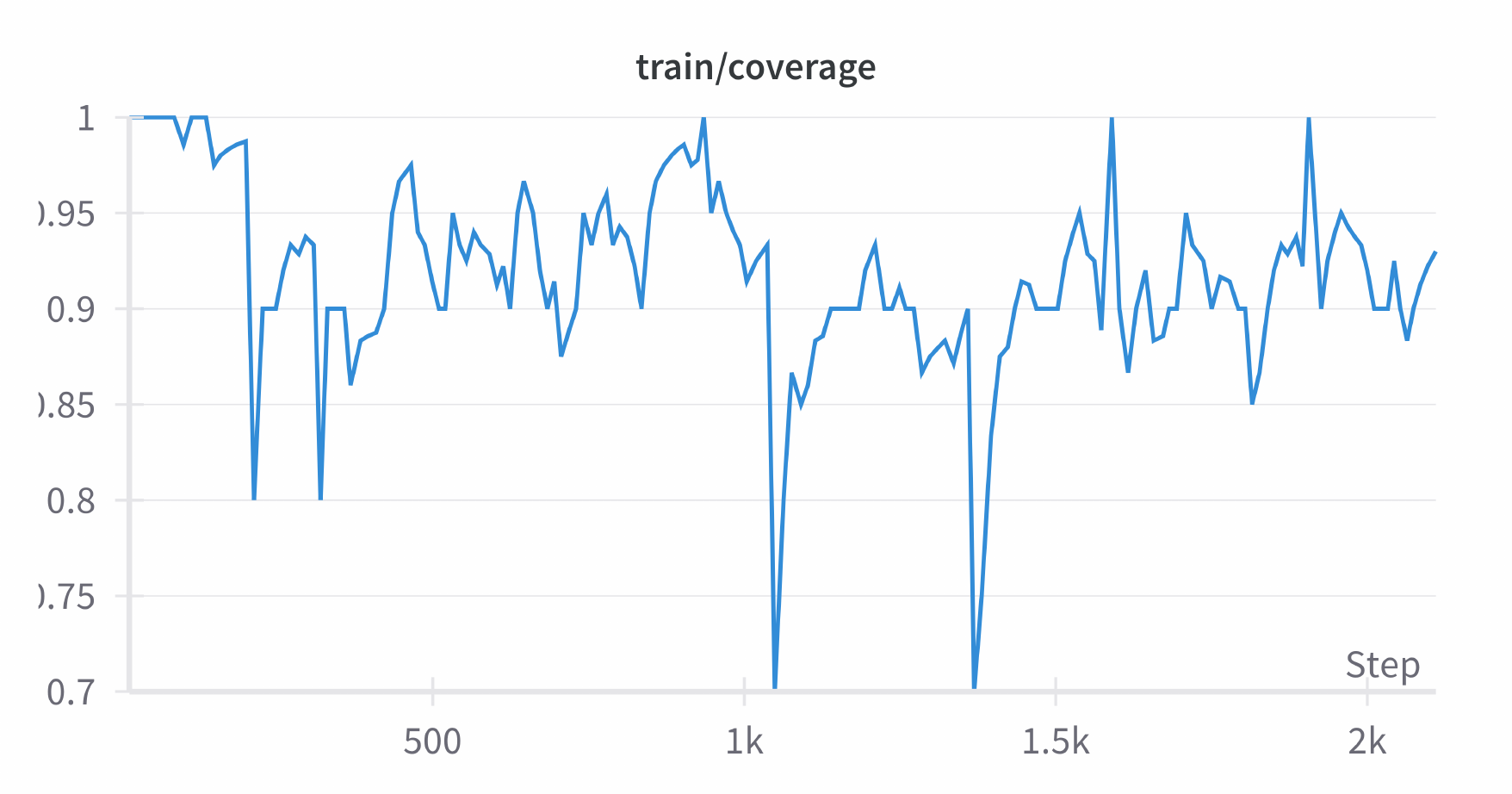}
        \caption{LLaMA-3.2-3b, $\alpha=0.05$,\\ $\lambda=1e^{-4}$.}
    \end{subfigure}
    
    \caption{Overall CCPO training performance on HotpotQA dataset with varying $\lambda$.}
    \label{fig:perf_lambda}
\end{figure*}

\begin{figure*}[htbp]
    \centering
    \vspace{1em}
    Coverage\\
    % First row: 4 images without subcaptions
    \includegraphics[width=0.24\textwidth]{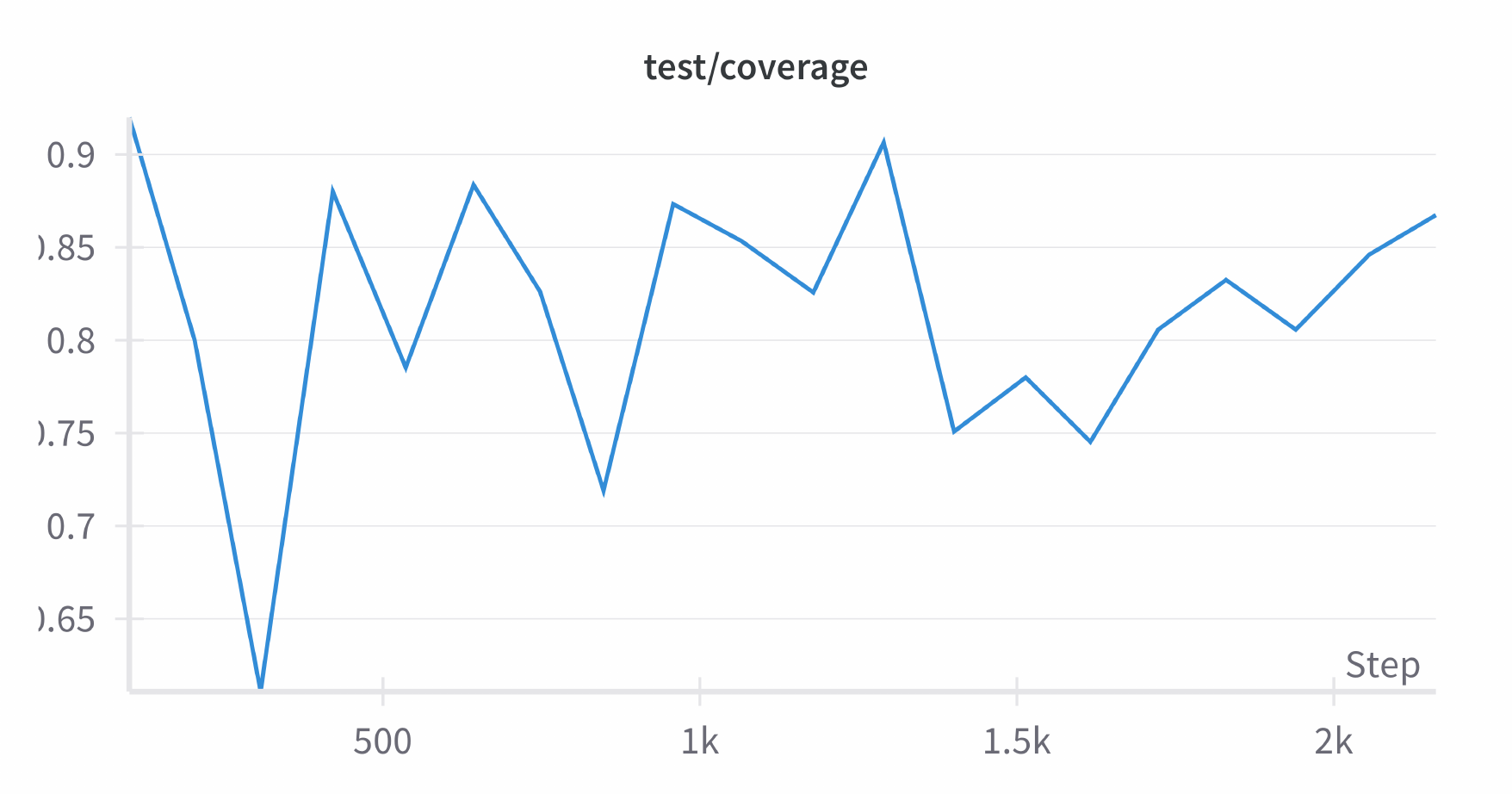}
    \includegraphics[width=0.24\textwidth]{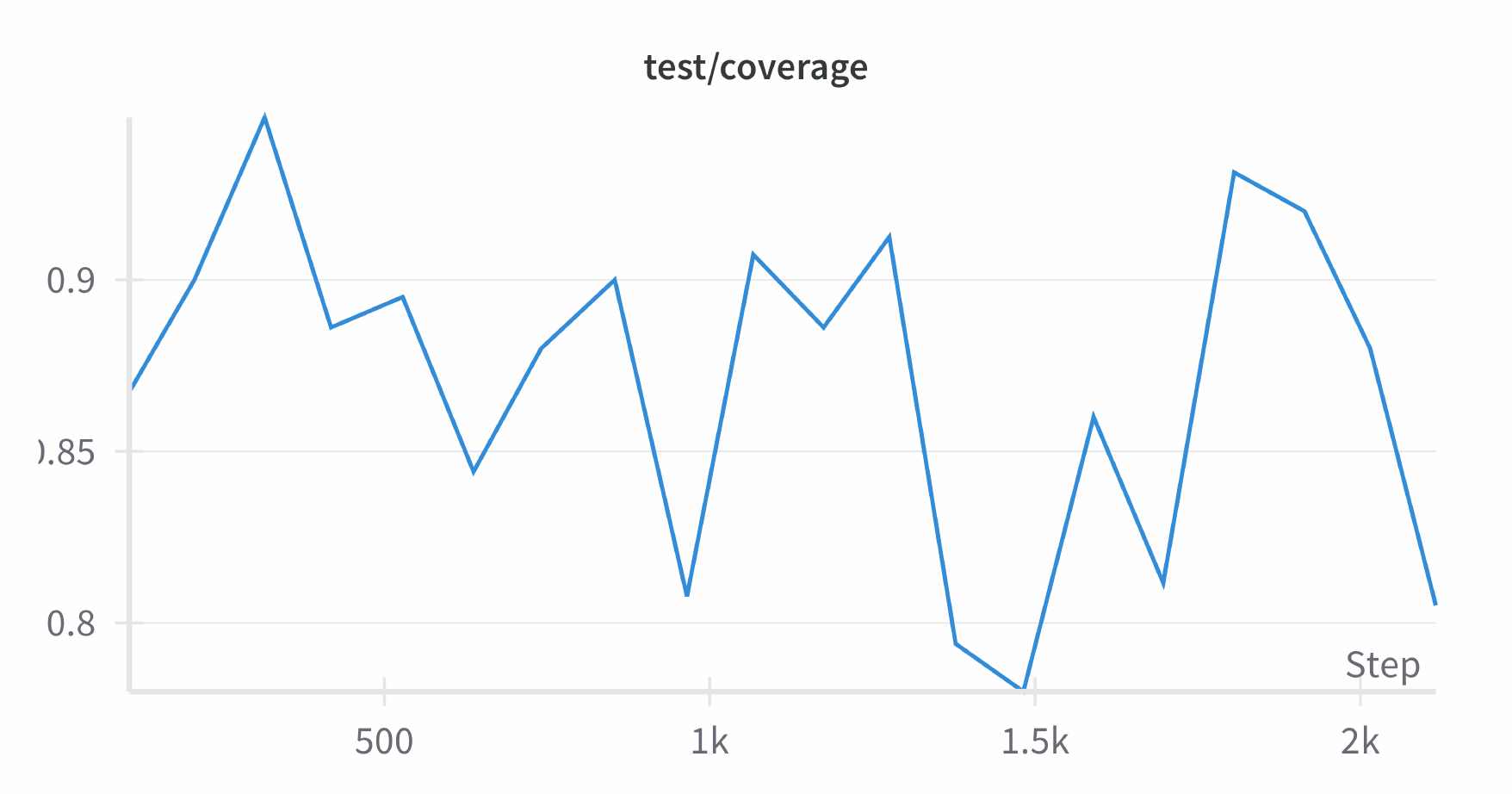}
    \includegraphics[width=0.24\textwidth]{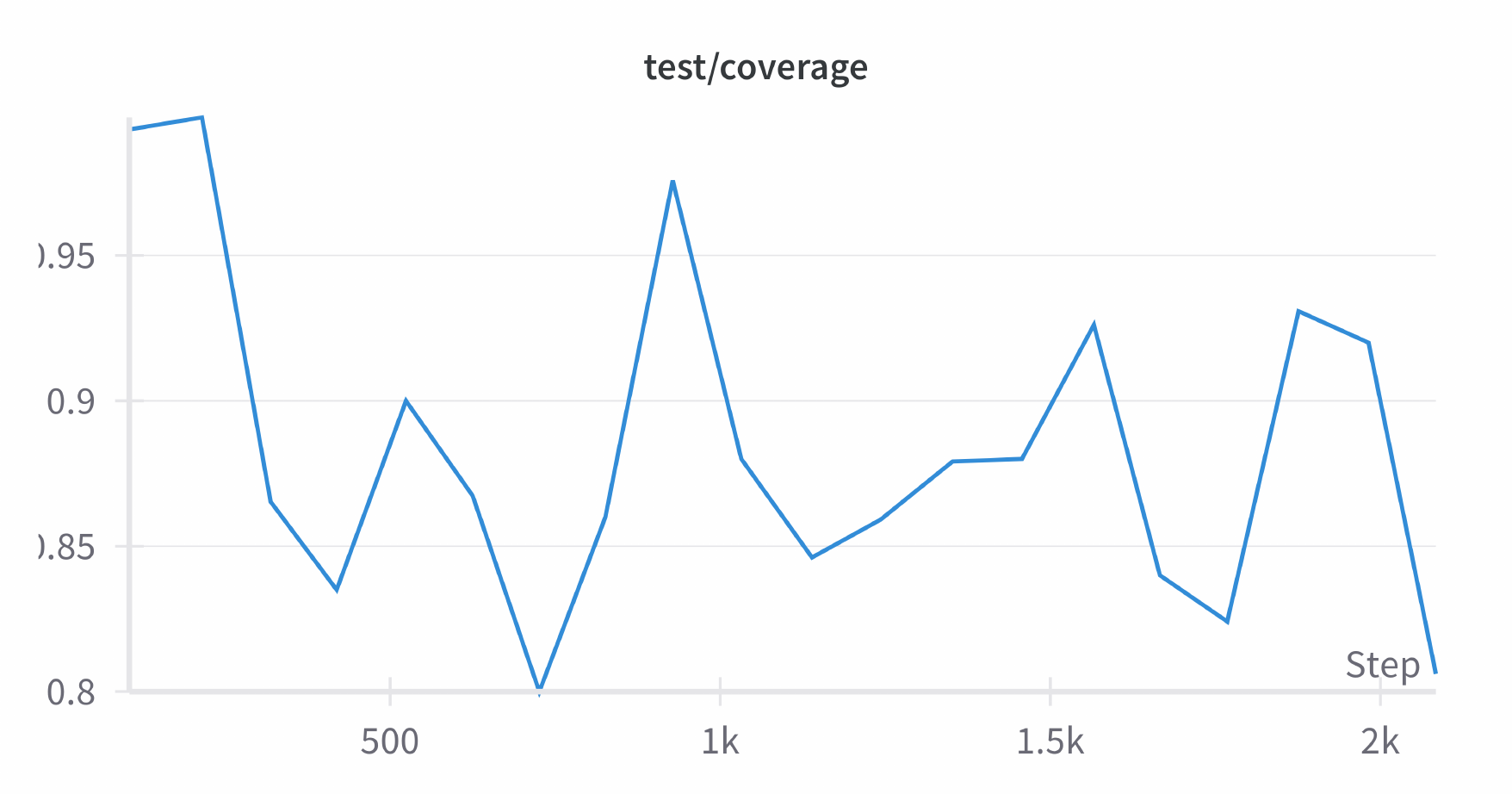}
    \includegraphics[width=0.24\textwidth]{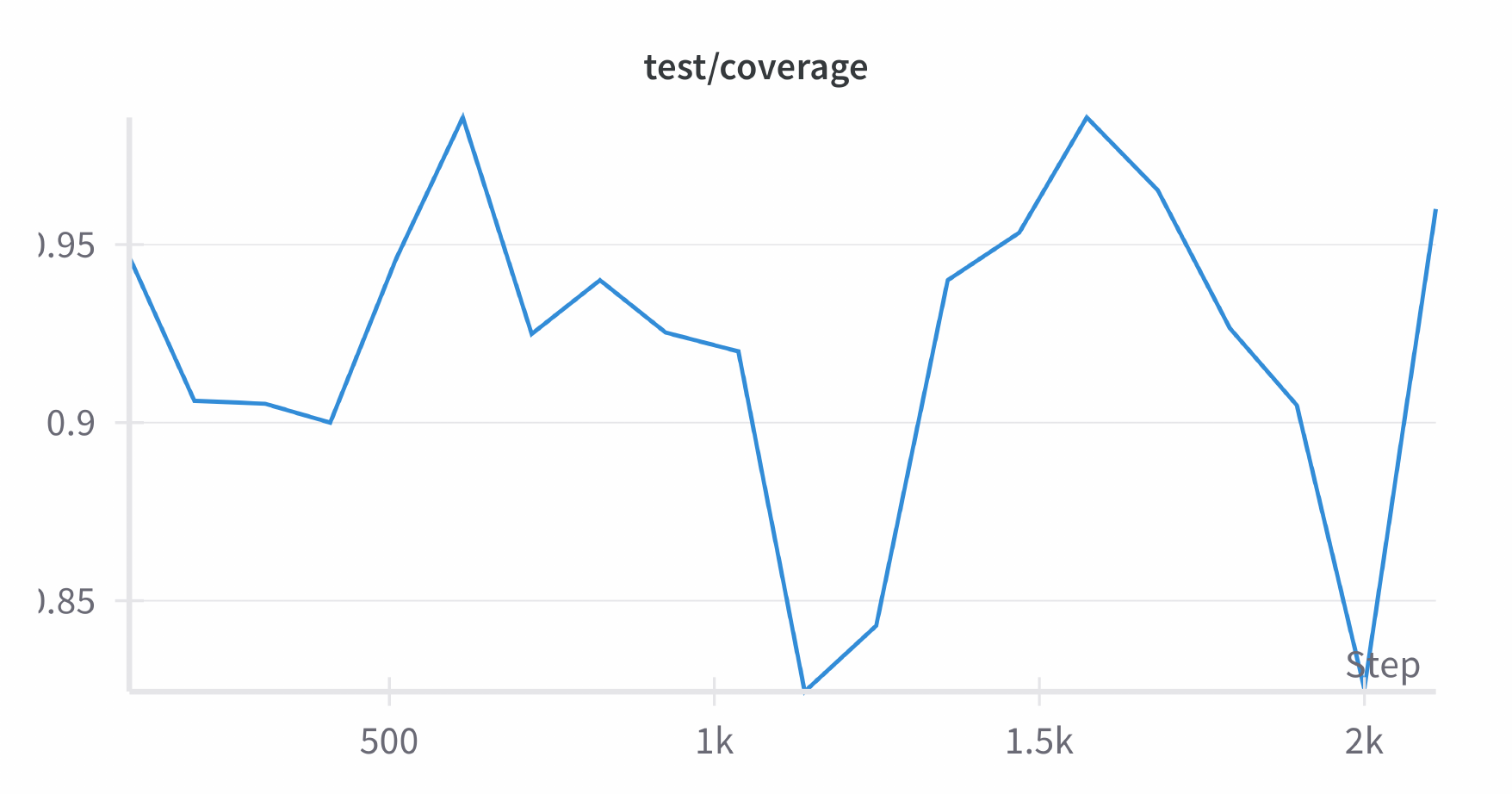}
    \vspace{1em}
    \text{Average Length}\\
    \begin{subfigure}[b]{0.24\textwidth}
        \includegraphics[width=\textwidth]{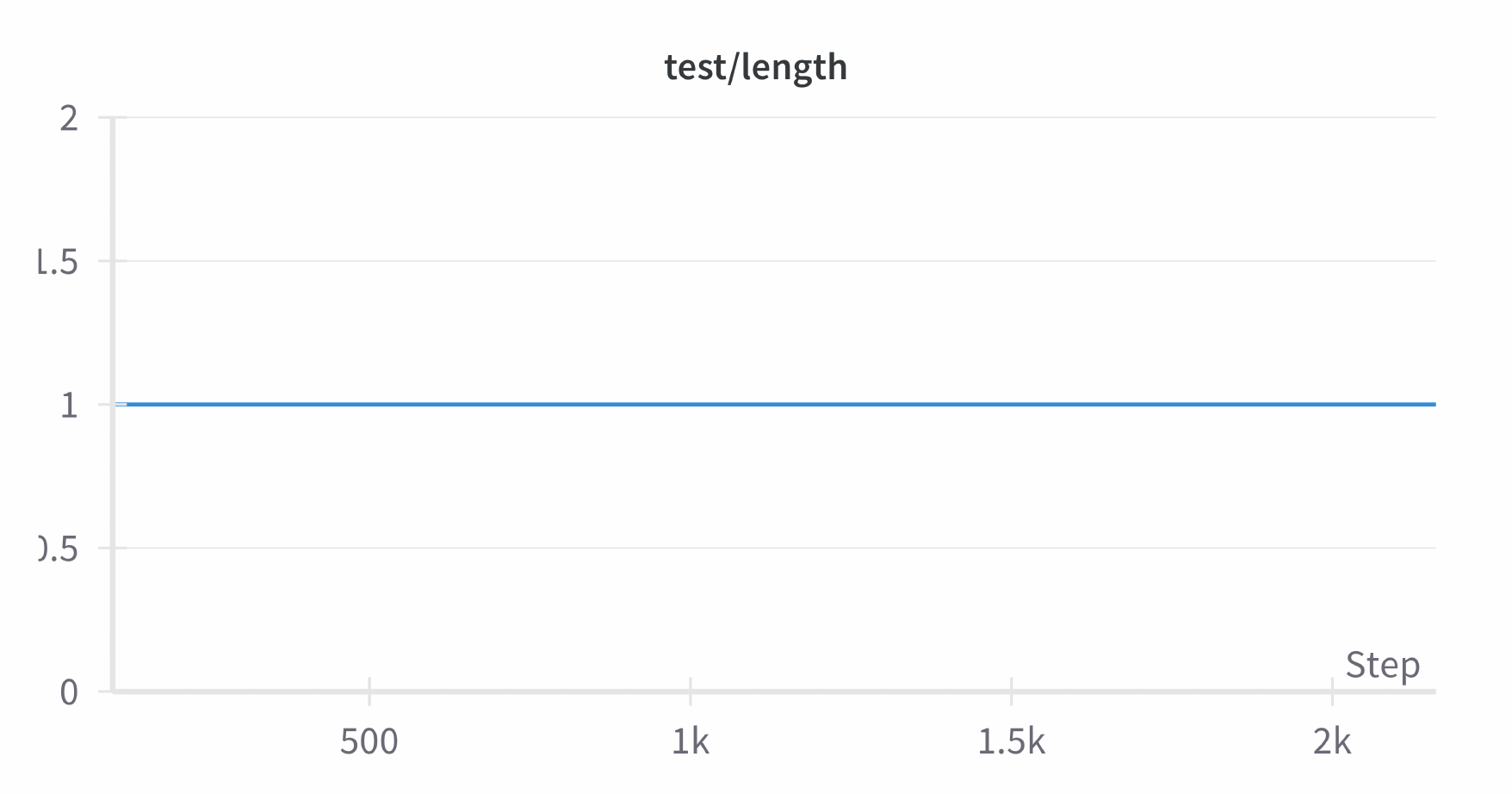}
        \caption{LLaMA-2-7b, $\alpha=0.2$,\\ $\lambda=1e^{-4}$.}
    \end{subfigure}
    \begin{subfigure}[b]{0.24\textwidth}
        \includegraphics[width=\textwidth]{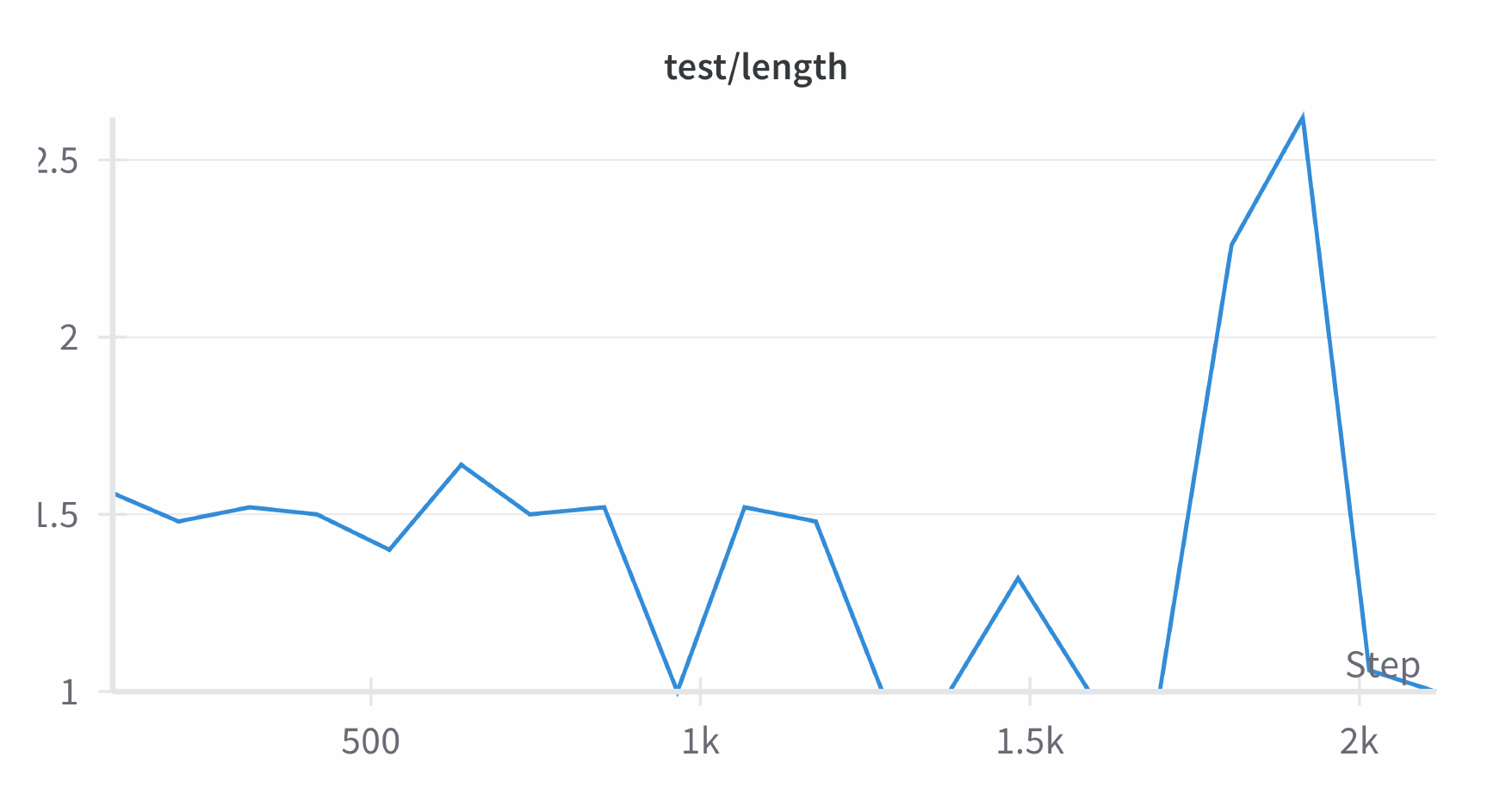}
        \caption{LLaMA-2-7b, $\alpha=0.1$,\\ $\lambda=2e^{-4}$.}
    \end{subfigure}
    \begin{subfigure}[b]{0.24\textwidth}
        \includegraphics[width=\textwidth]{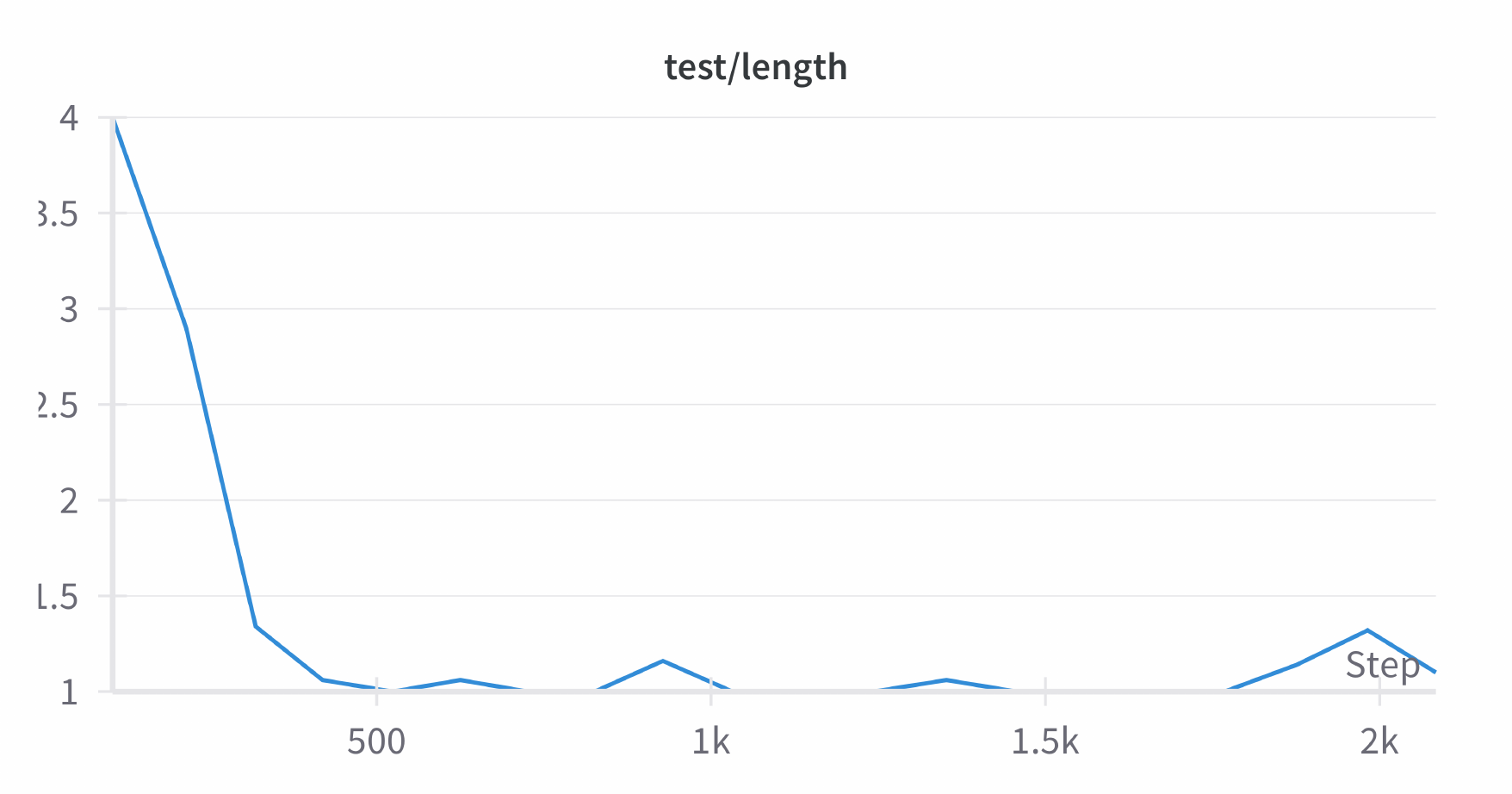}
        \caption{LLaMA-3.2-3b, $\alpha=0.1$,\\ $\lambda=1e^{-4}$.}
    \end{subfigure}
    \begin{subfigure}[b]{0.24\textwidth}
        \includegraphics[width=\textwidth]{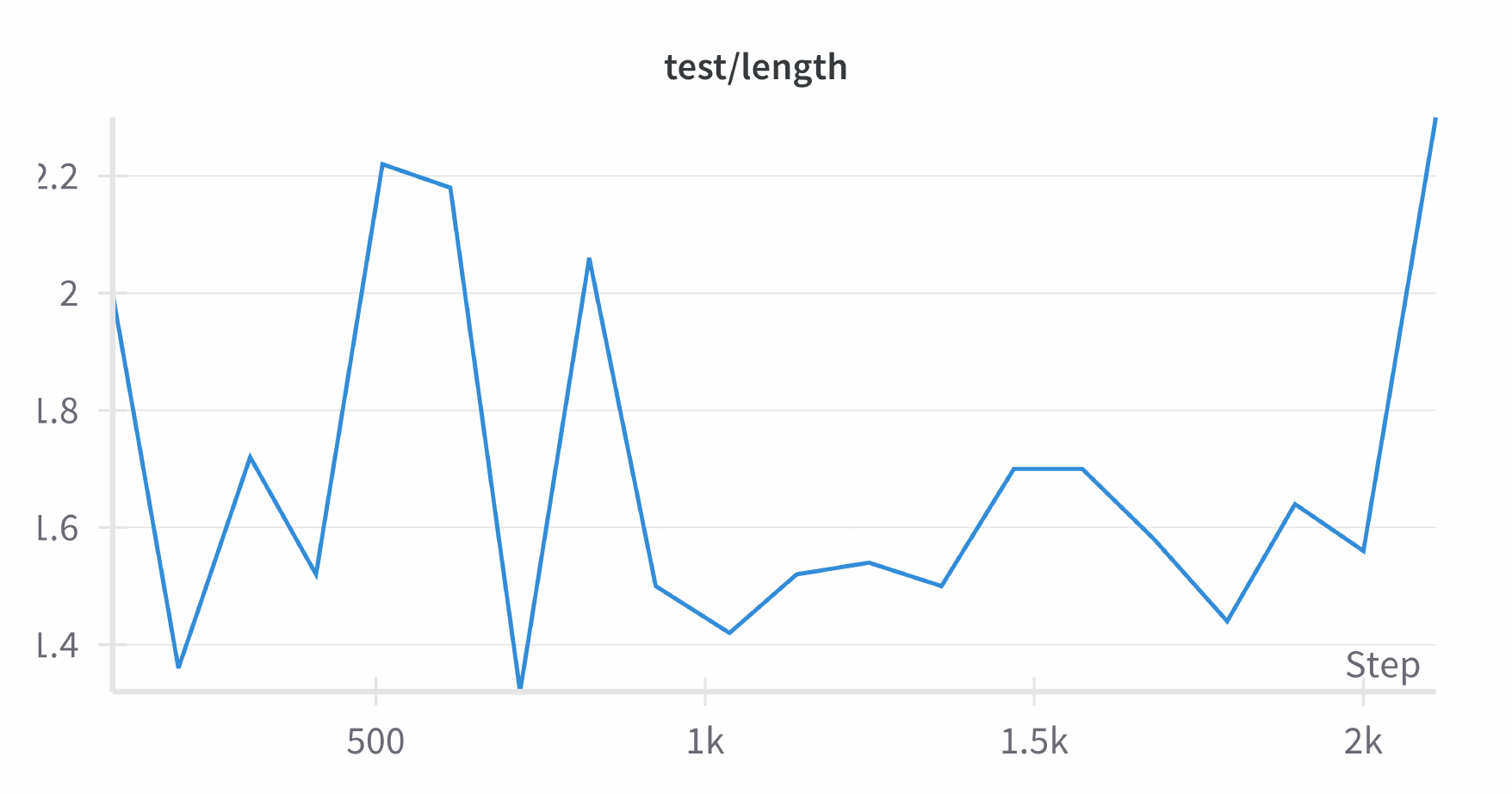}
        \caption{LLaMA-3.2-3b, $\alpha=0.05$,\\ $\lambda=1e^{-4}$.}
    \end{subfigure}
    
    \caption{CCPO validation performance on HotpotQA dataset with varying $\lambda$.}
    \label{fig:val_lambda}
\end{figure*}

\begin{figure*}[htbp]
    \centering
    \vspace{1em}
    Costs\\
    % First row: 4 images without subcaptions
    \includegraphics[width=0.24\textwidth]{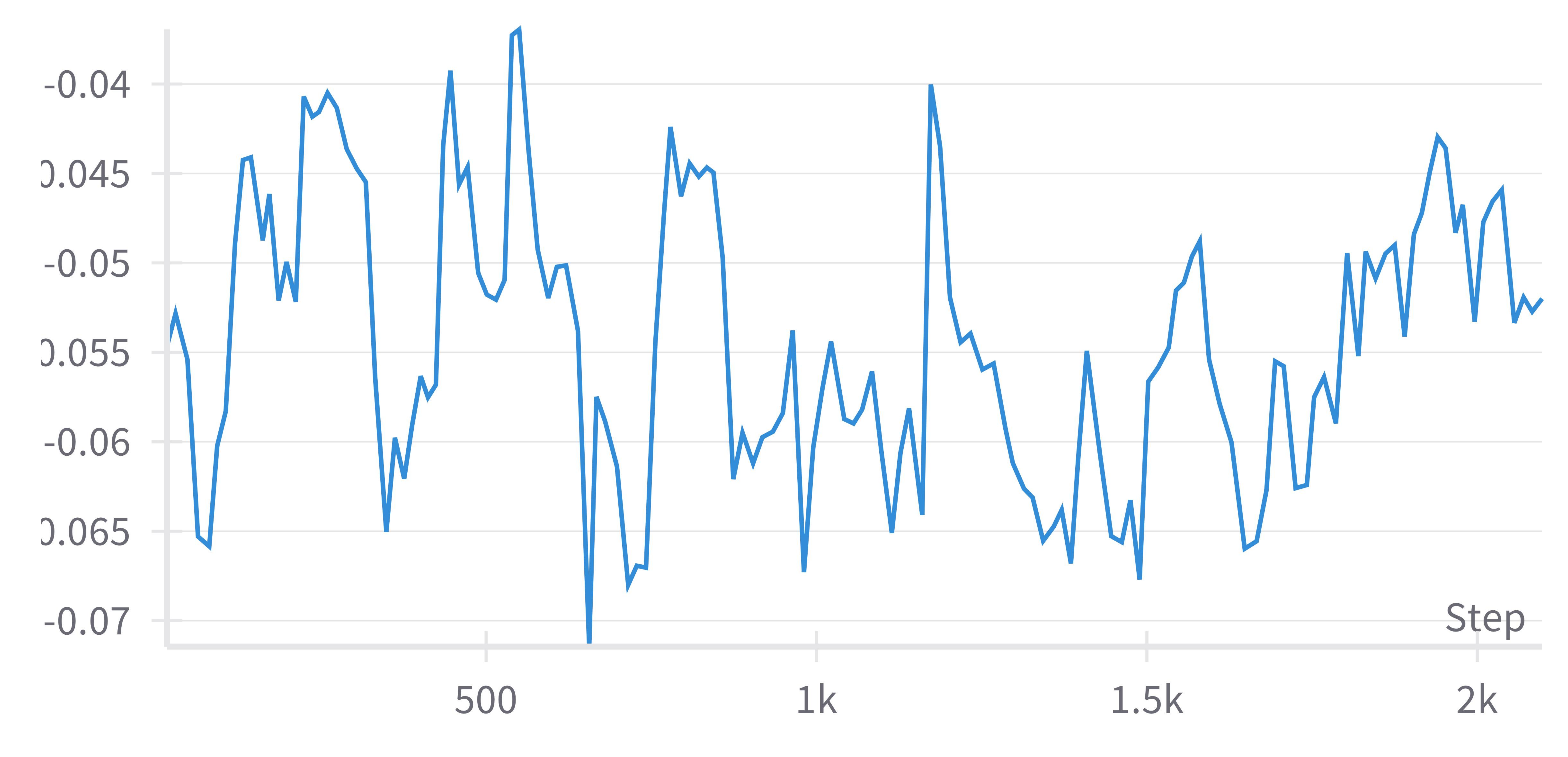}
    \includegraphics[width=0.24\textwidth]{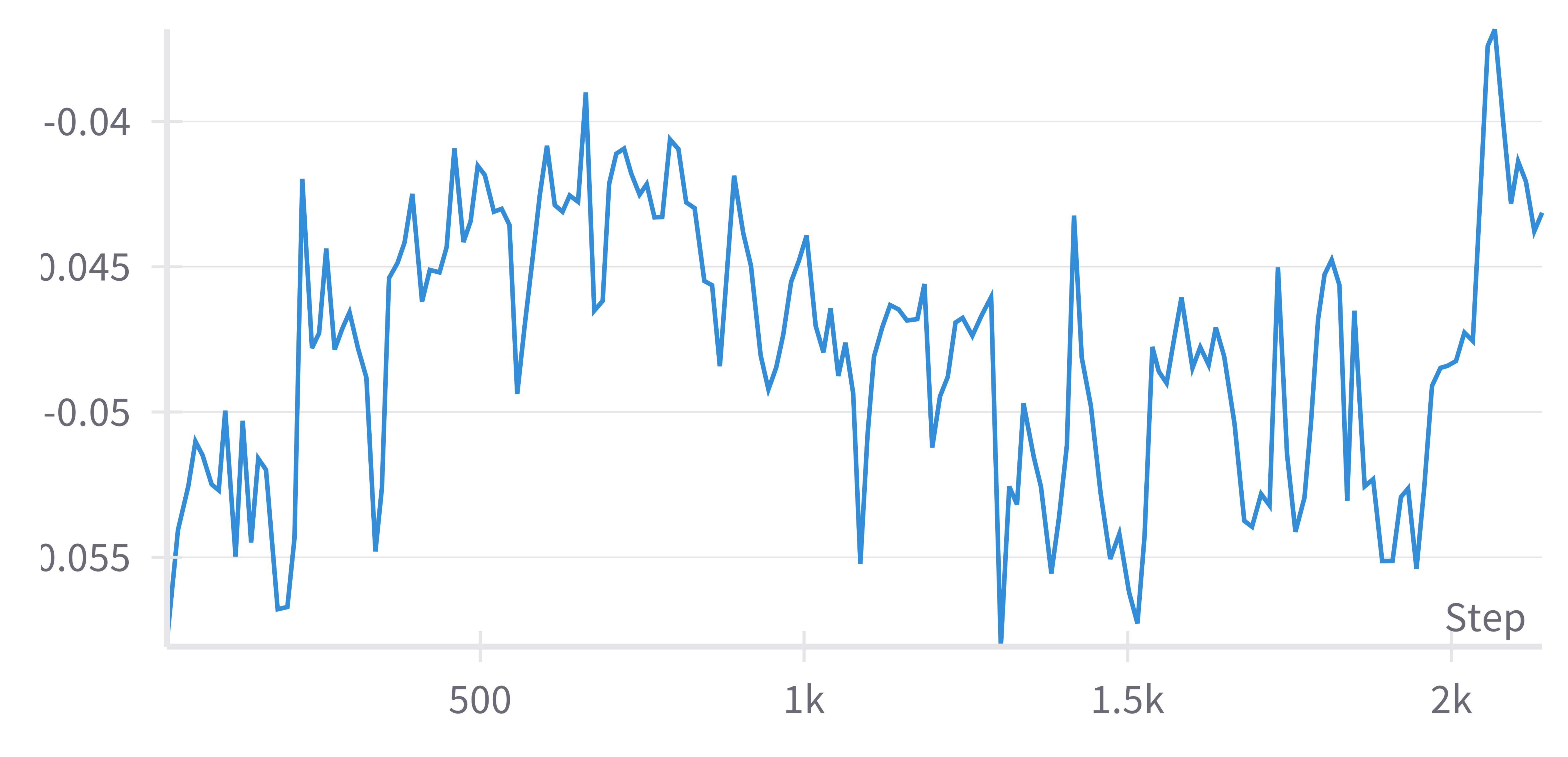}
    \includegraphics[width=0.24\textwidth]{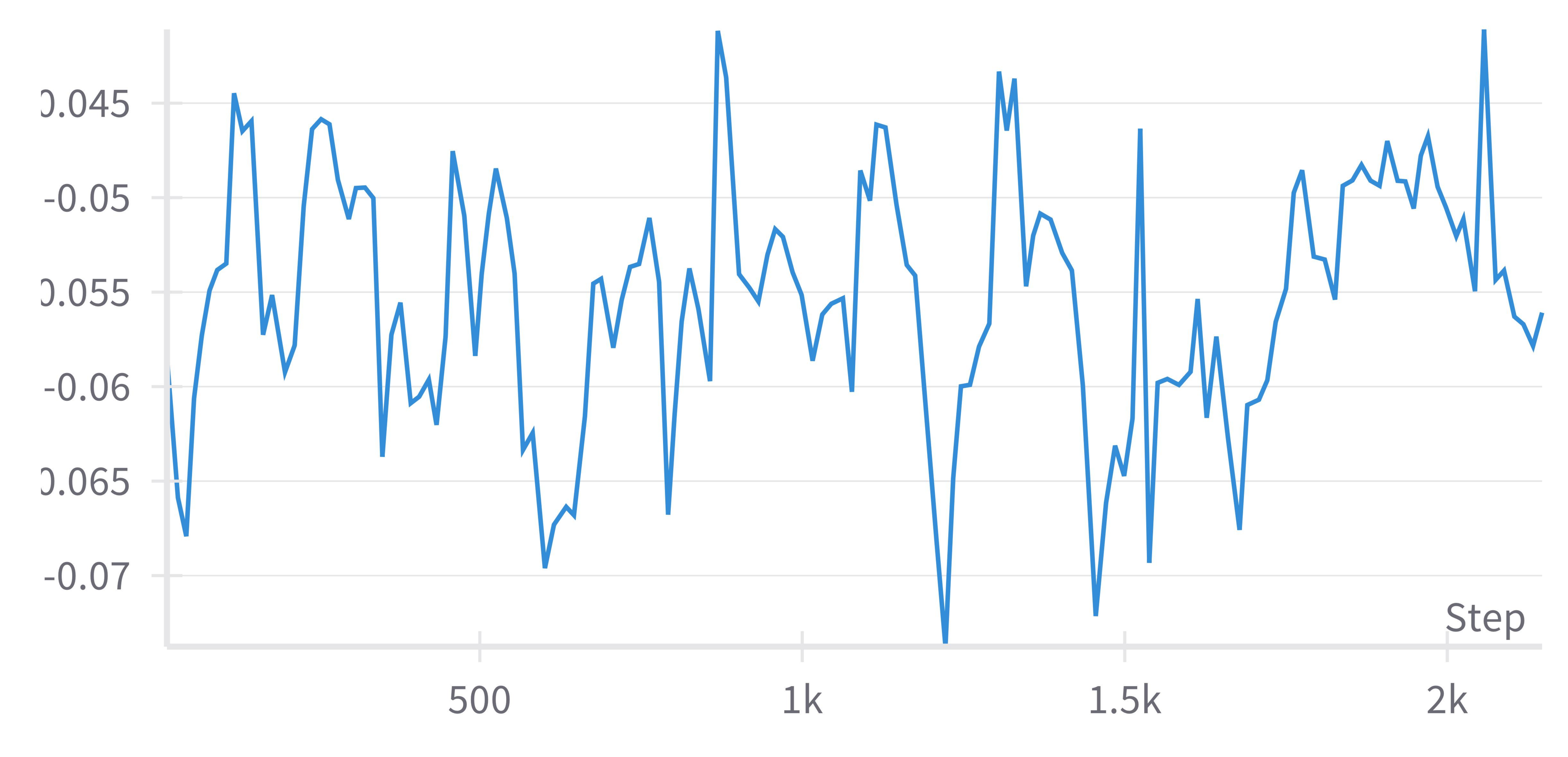}
    \includegraphics[width=0.24\textwidth]{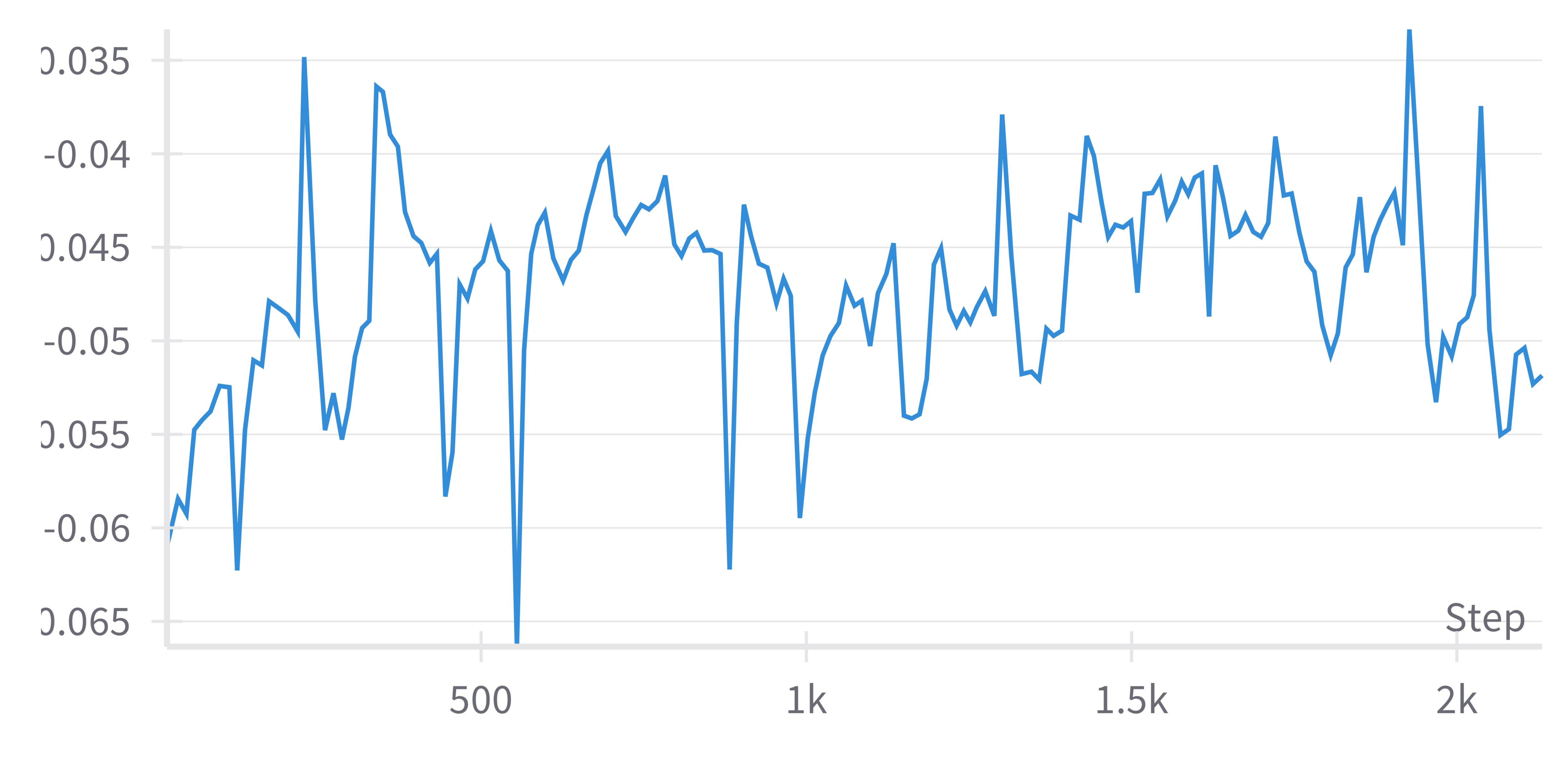}
    \vspace{1em}
    \text{Coverage Surrogate Violation}\\
    % \vspace{0.5cm} % vertical space between rows
    \includegraphics[width=0.24\textwidth]{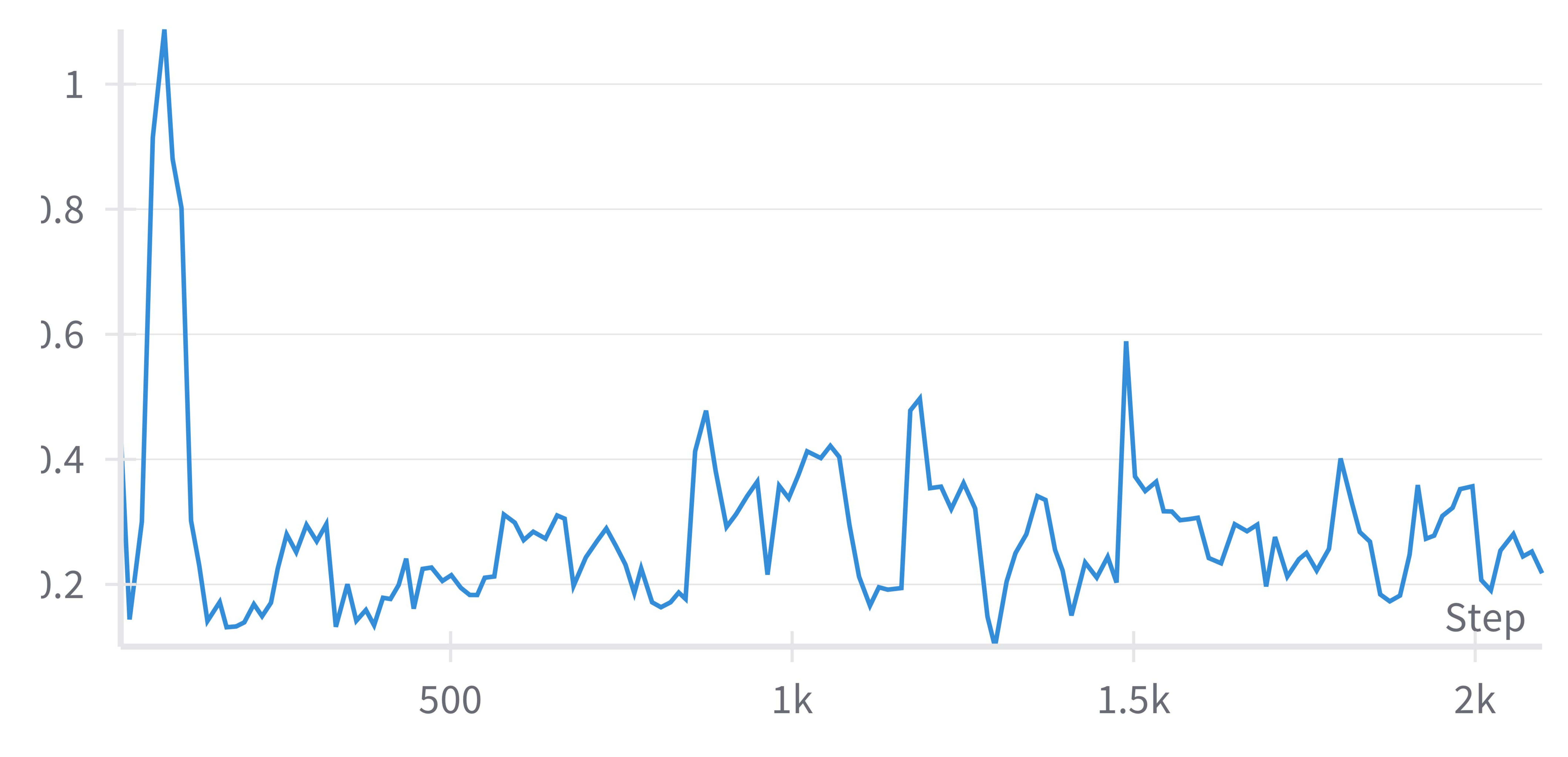}
    \includegraphics[width=0.24\textwidth]{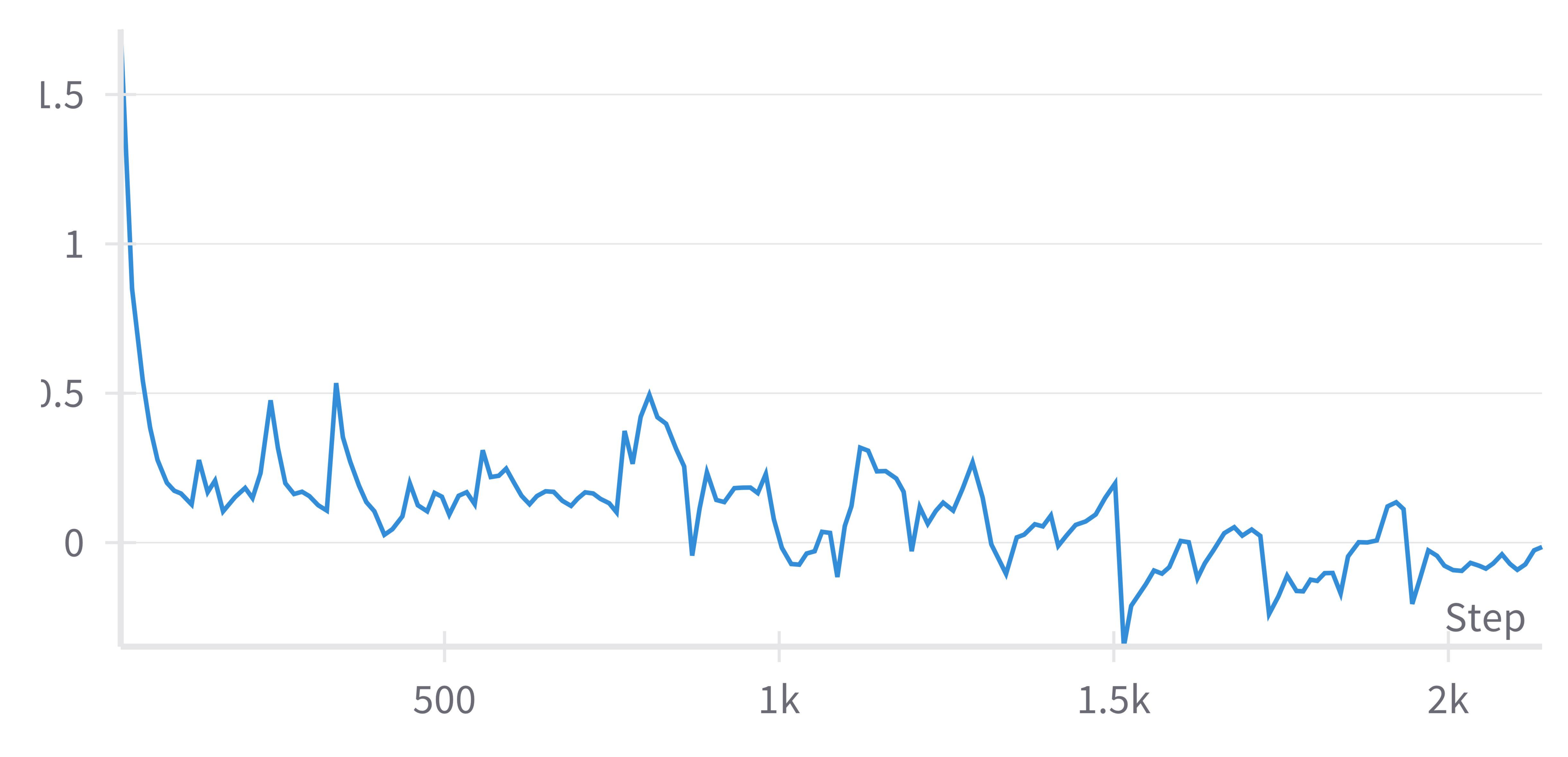}
    \includegraphics[width=0.24\textwidth]{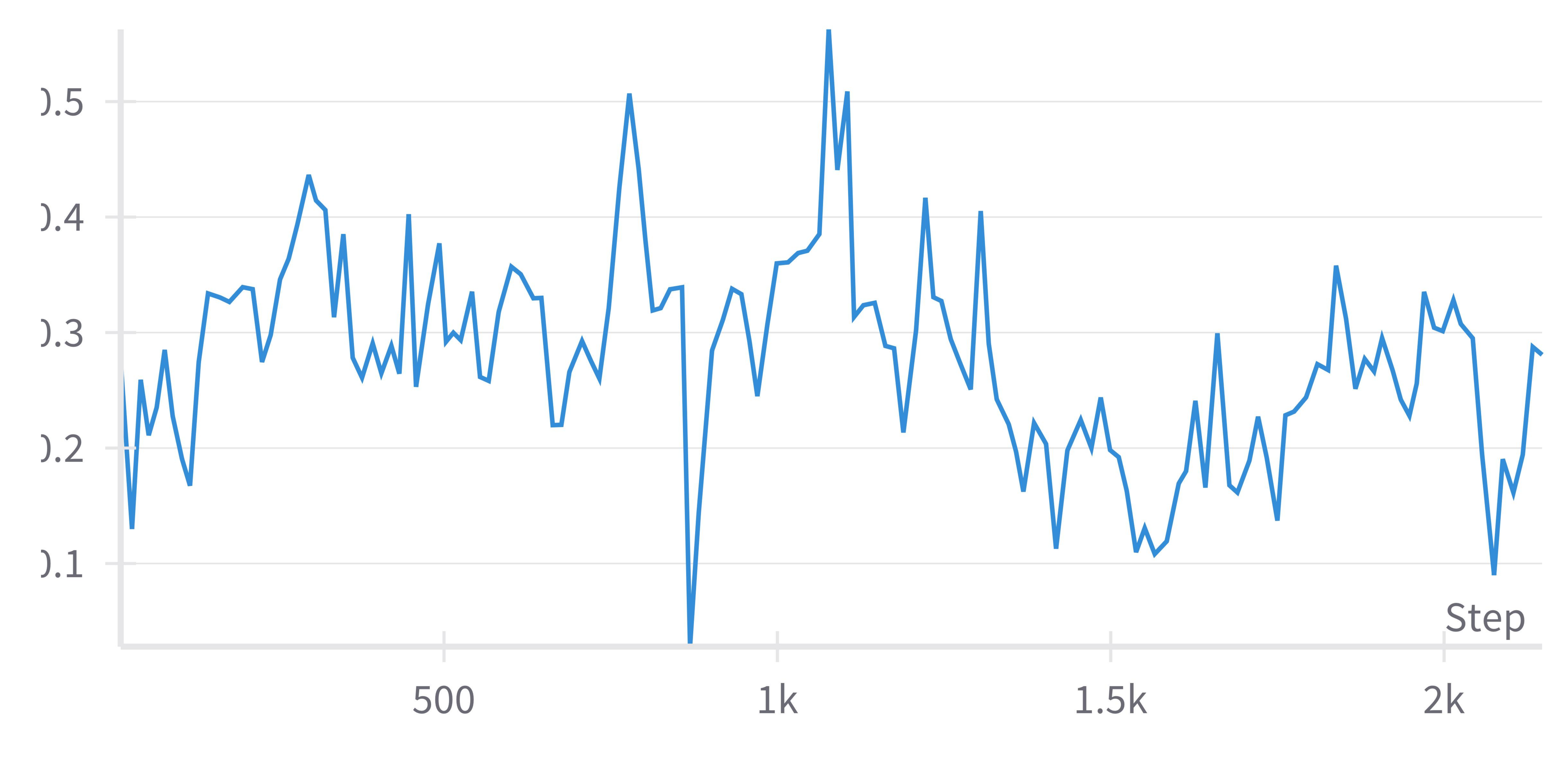}
    \includegraphics[width=0.24\textwidth]{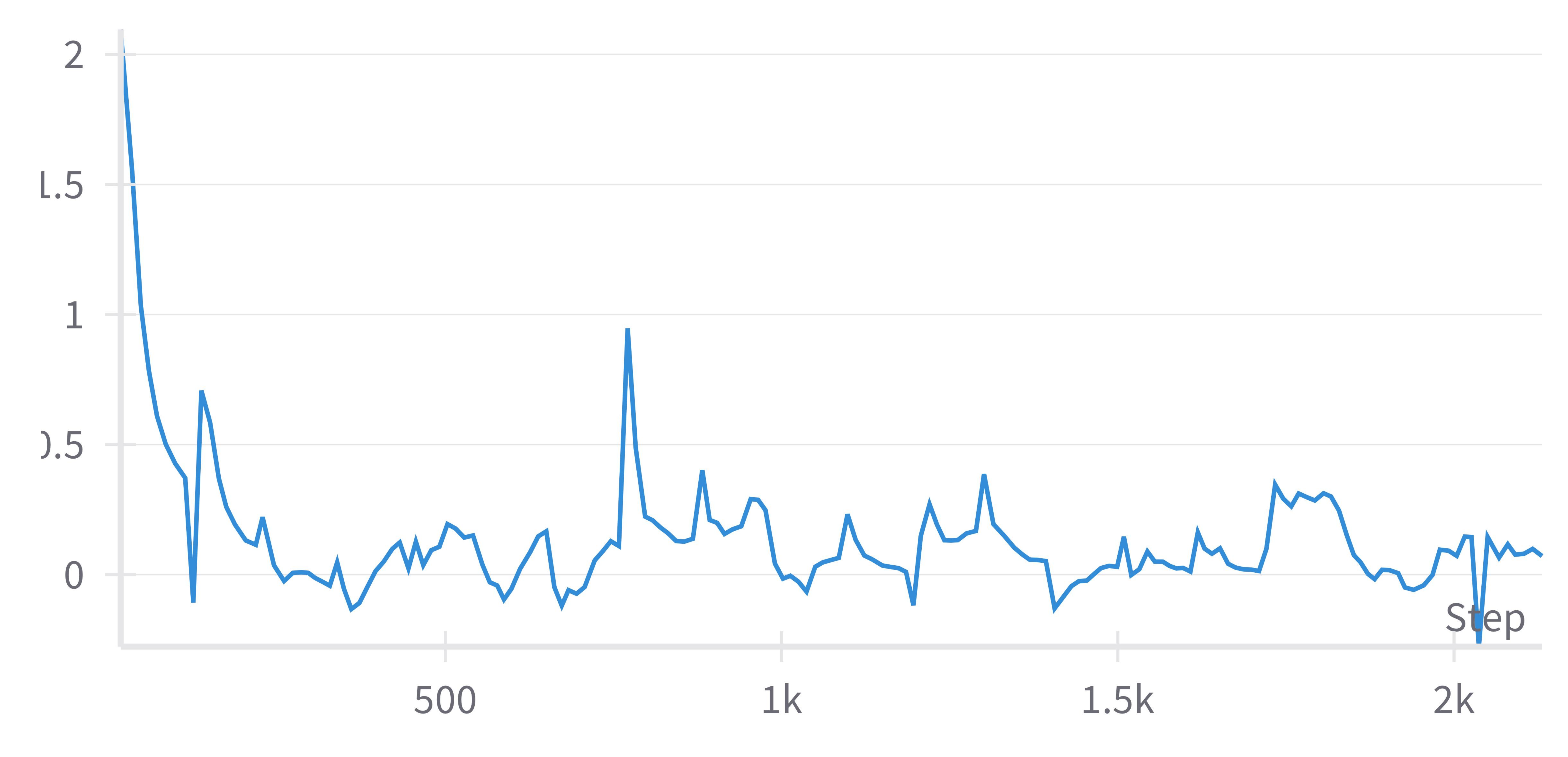}
    \vspace{1em}
    \text{Training Coverage}\\
    % Second row: 4 images with subcaptions
    \begin{subfigure}[b]{0.24\textwidth}
        \includegraphics[width=\textwidth]{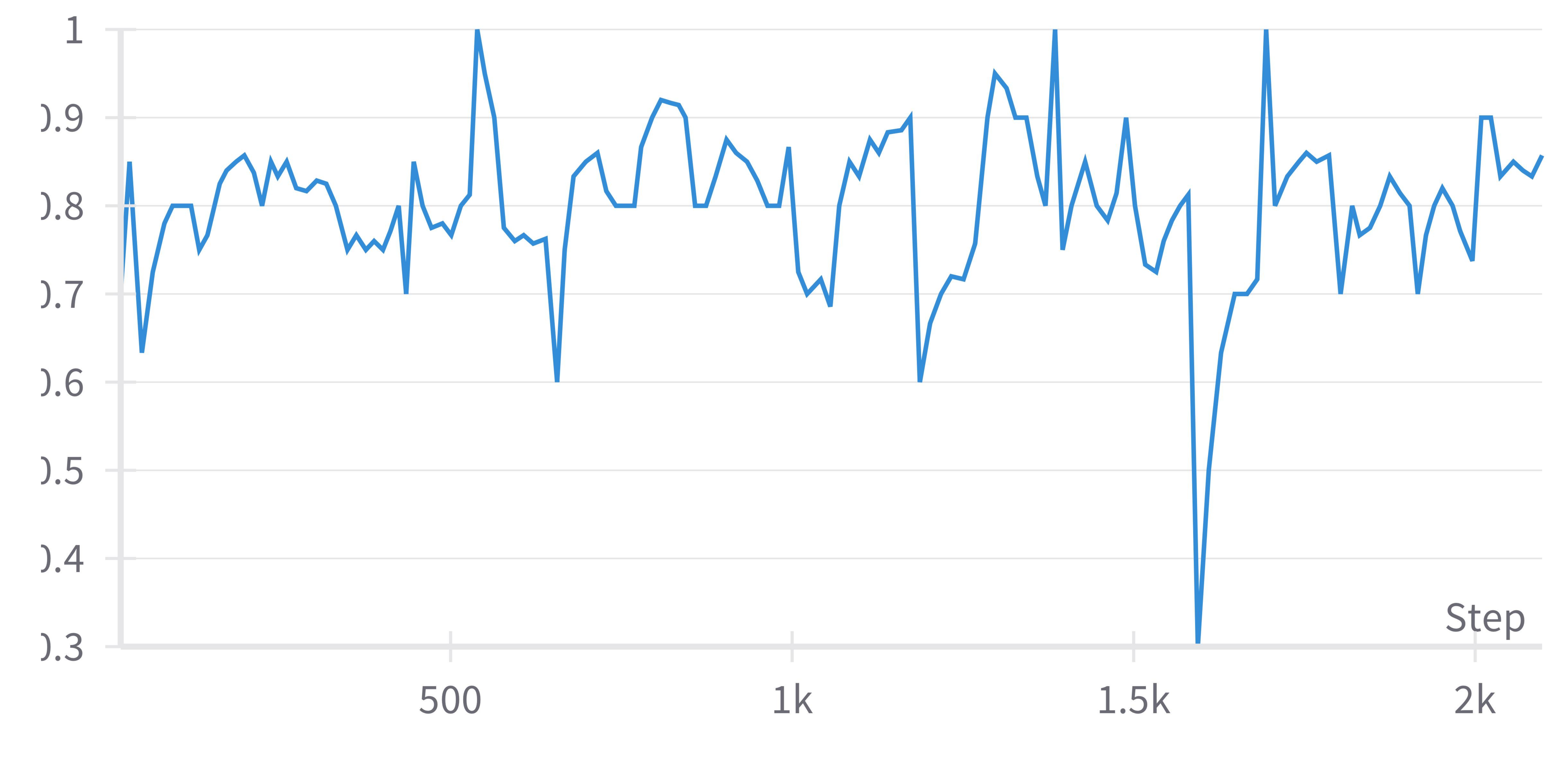}
        \caption{LLaMA-2-7b, $\alpha=0.2$}
    \end{subfigure}
    \begin{subfigure}[b]{0.24\textwidth}
        \includegraphics[width=\textwidth]{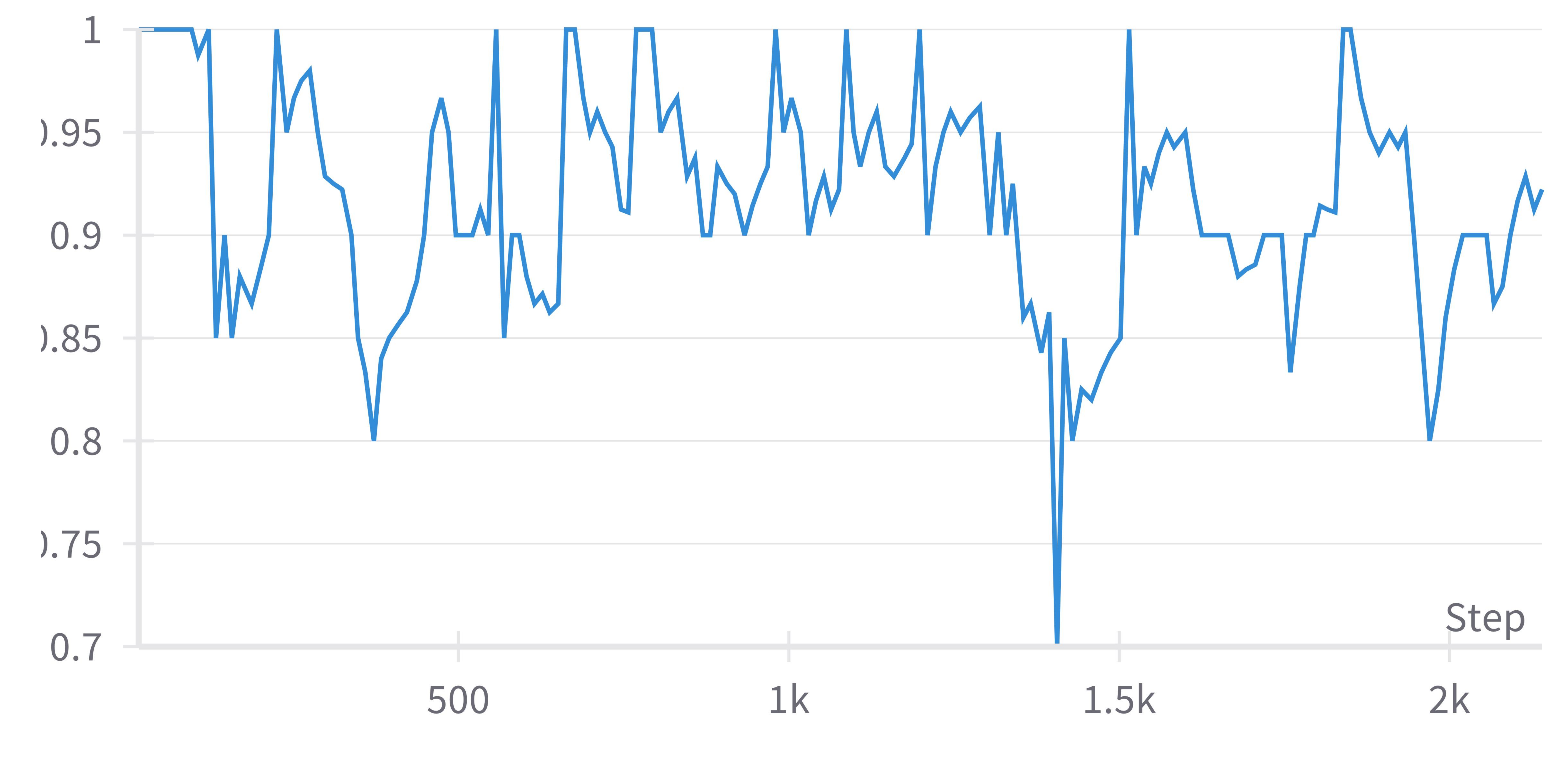}
        \caption{LLaMA-2-7b, $\alpha=0.1$}
    \end{subfigure}
    \begin{subfigure}[b]{0.24\textwidth}
        \includegraphics[width=\textwidth]{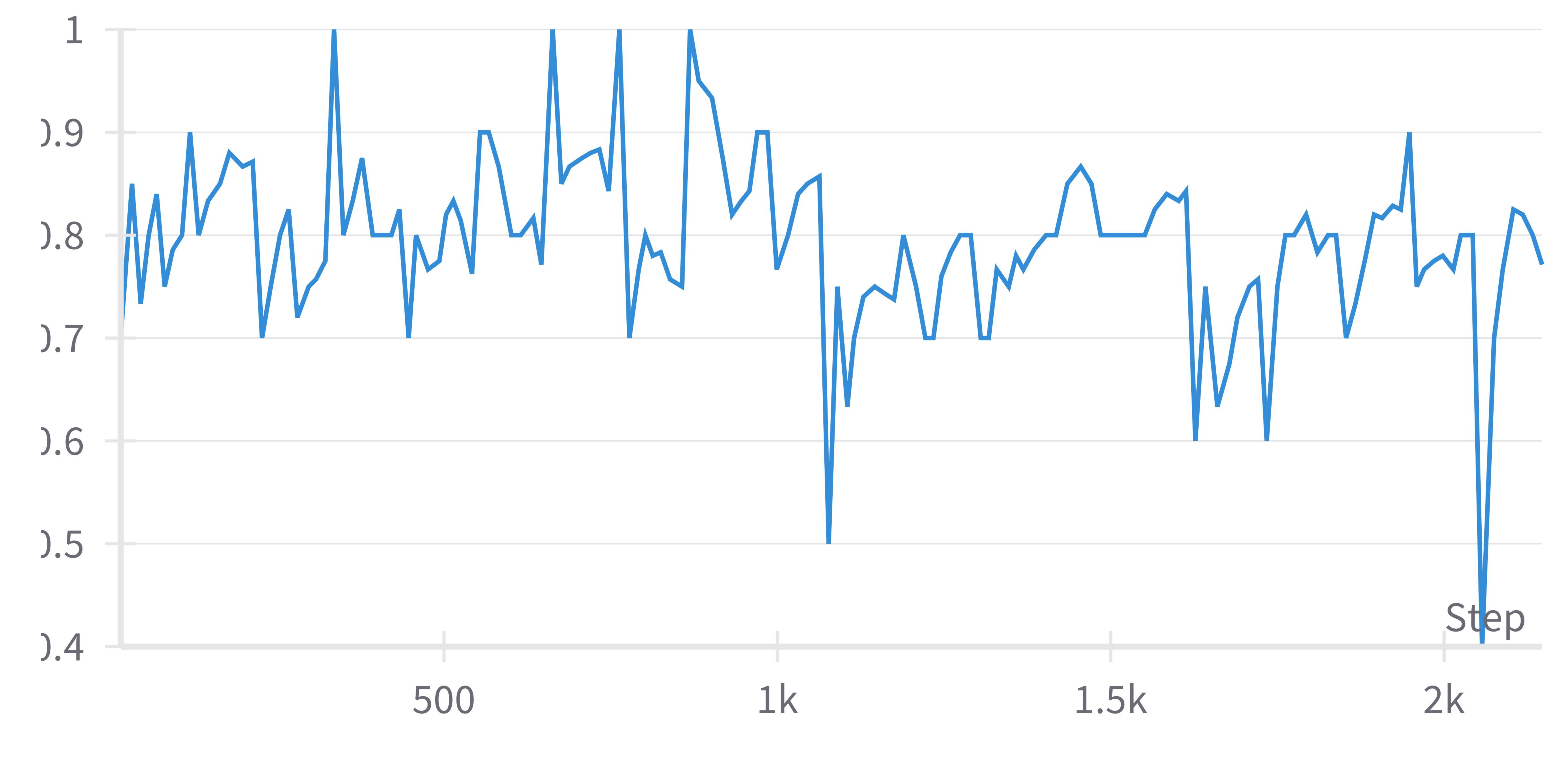}
        \caption{LLaMA-3.2-3b, $\alpha=0.2$}
    \end{subfigure}
    \begin{subfigure}[b]{0.24\textwidth}
        \includegraphics[width=\textwidth]{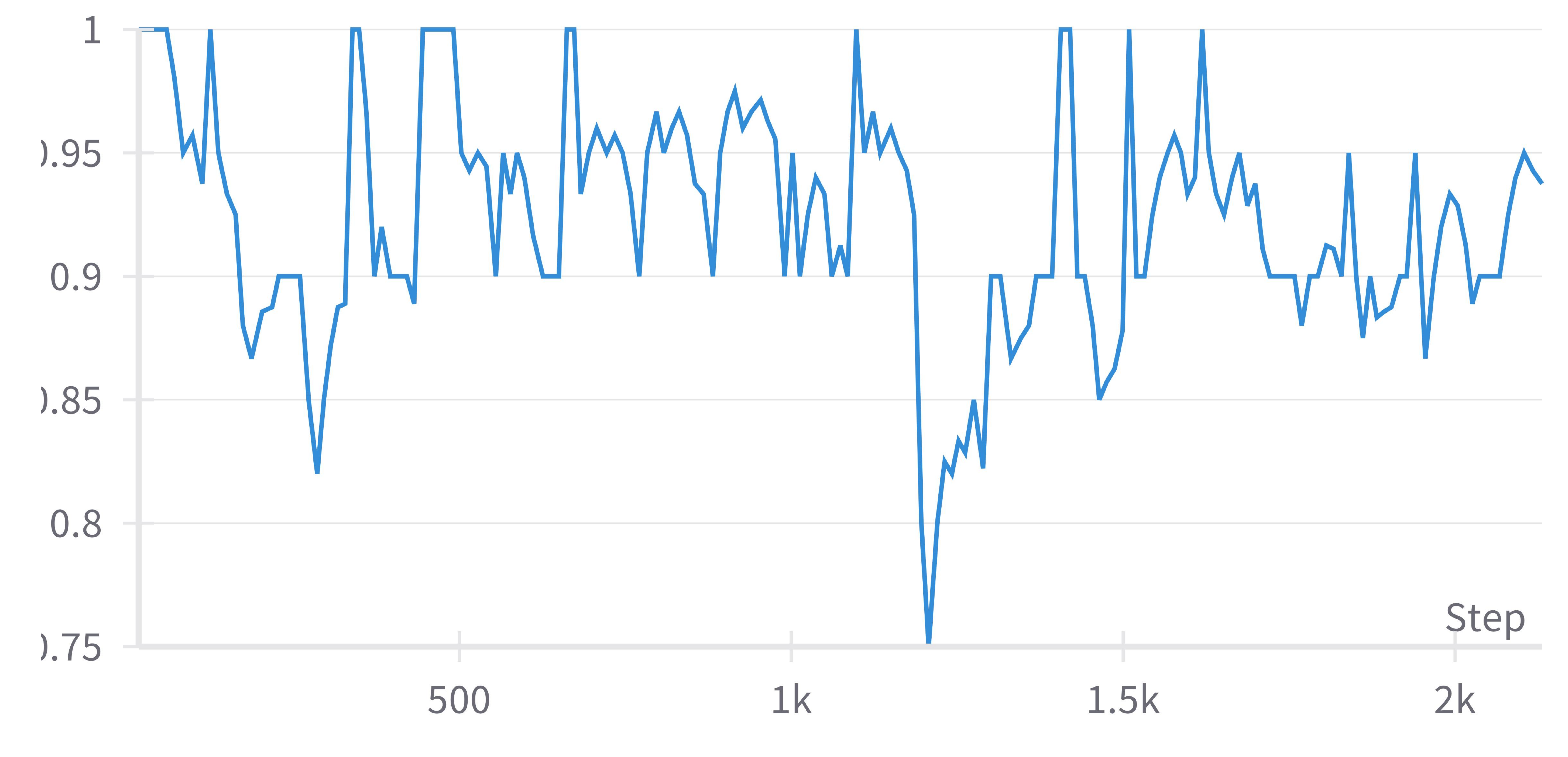}
        \caption{LLaMA-3.2-3b, $\alpha=0.1$}
    \end{subfigure}
    
    \caption{Overall CCPO training performance on MMLU dataset, $\lambda=0$.}
    \label{fig:perf_mmlu}
\end{figure*}

\begin{figure*}[htbp]
    \centering
    \vspace{1em}
    Coverage\\
    % First row: 4 images without subcaptions
    \includegraphics[width=0.24\textwidth]{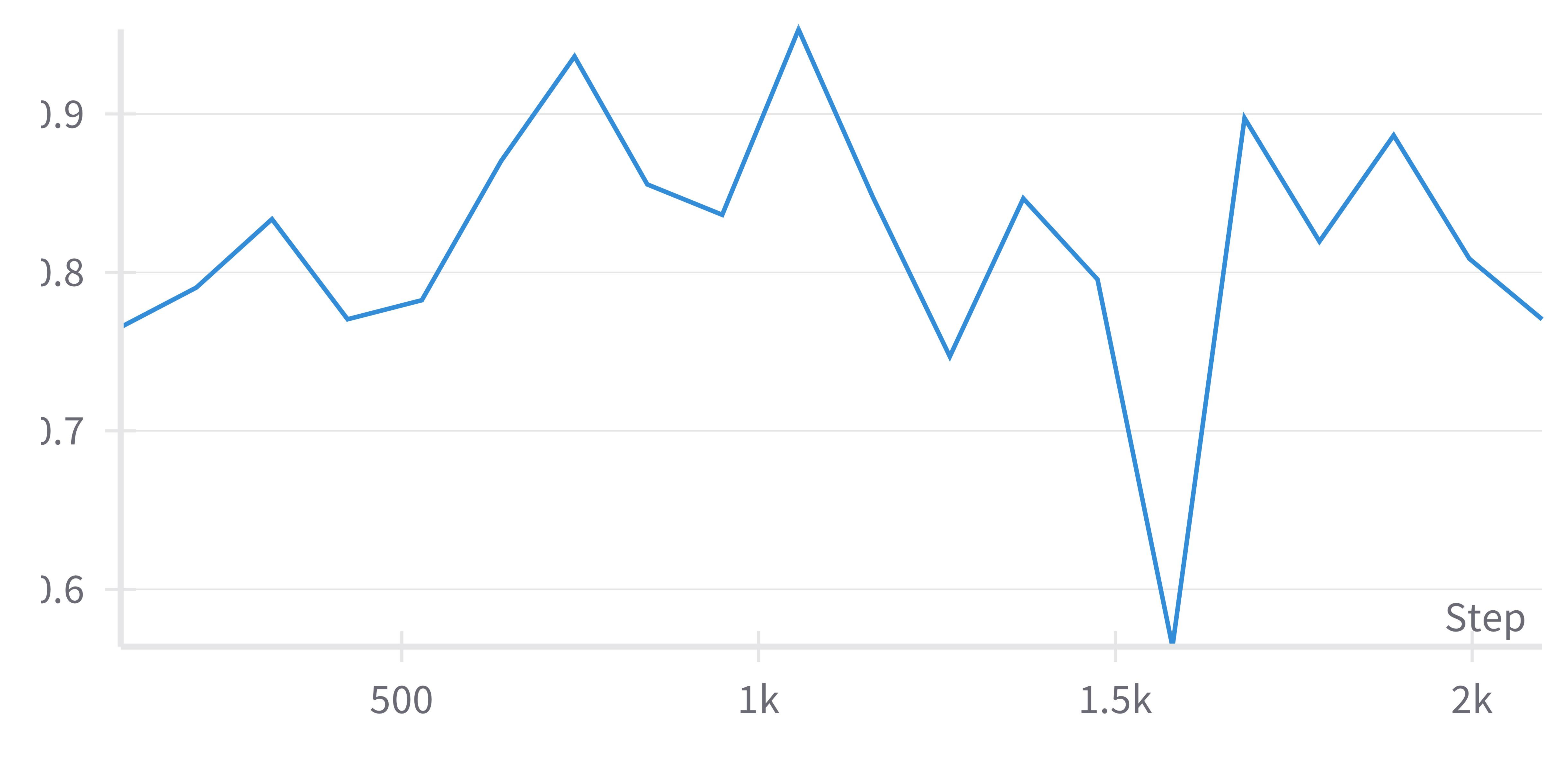}
    \includegraphics[width=0.24\textwidth]{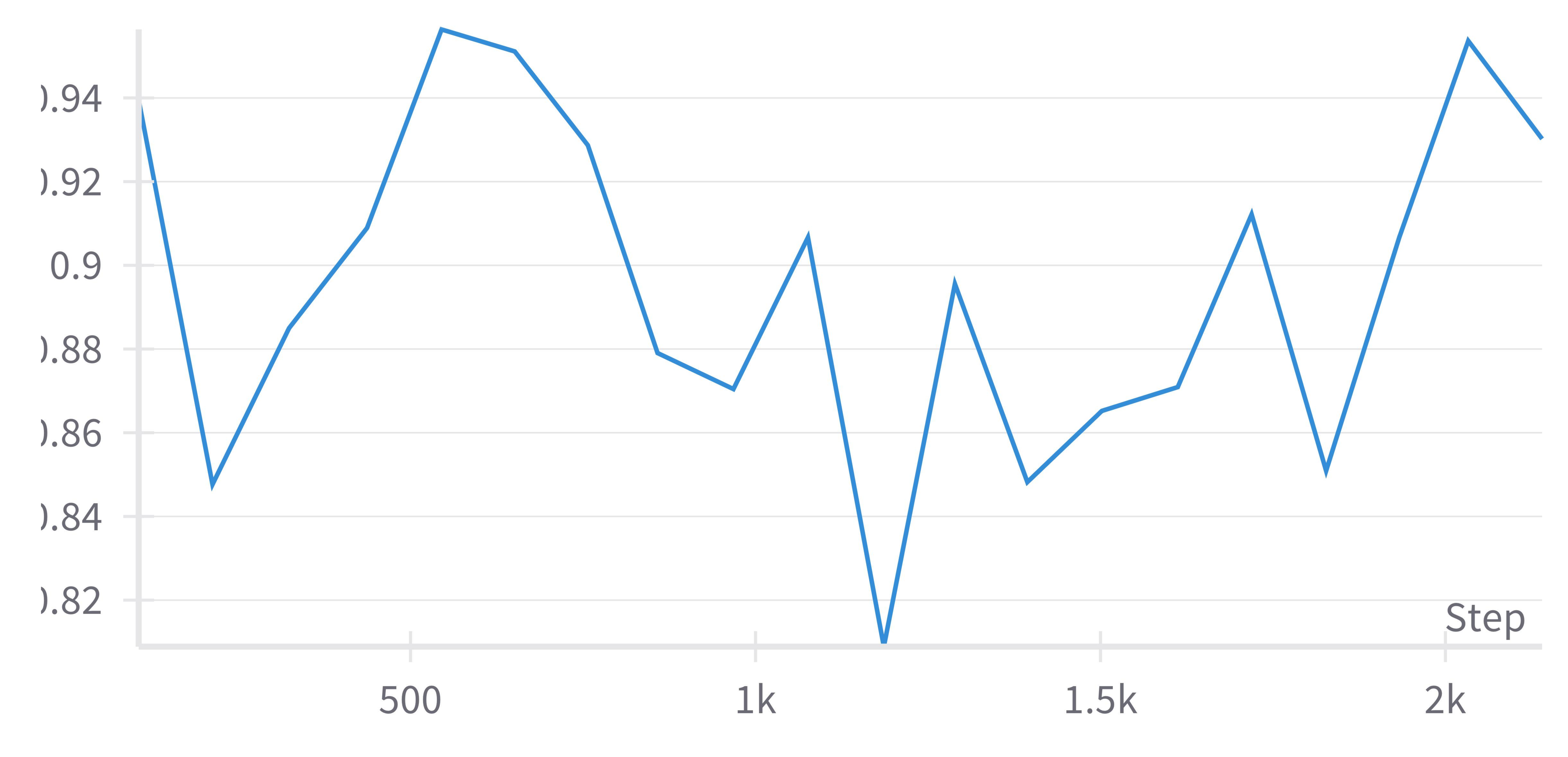}
    \includegraphics[width=0.24\textwidth]{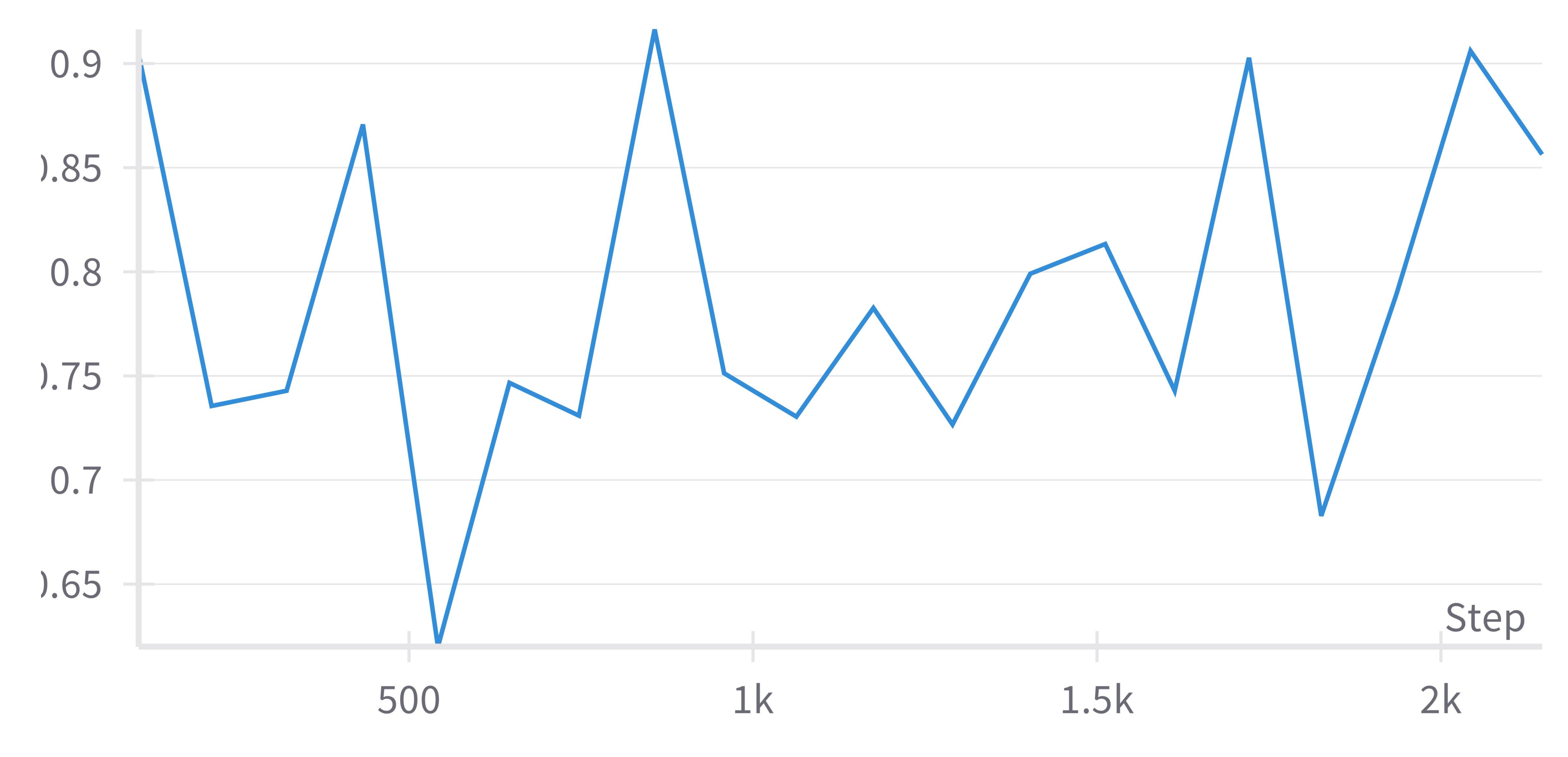}
    \includegraphics[width=0.24\textwidth]{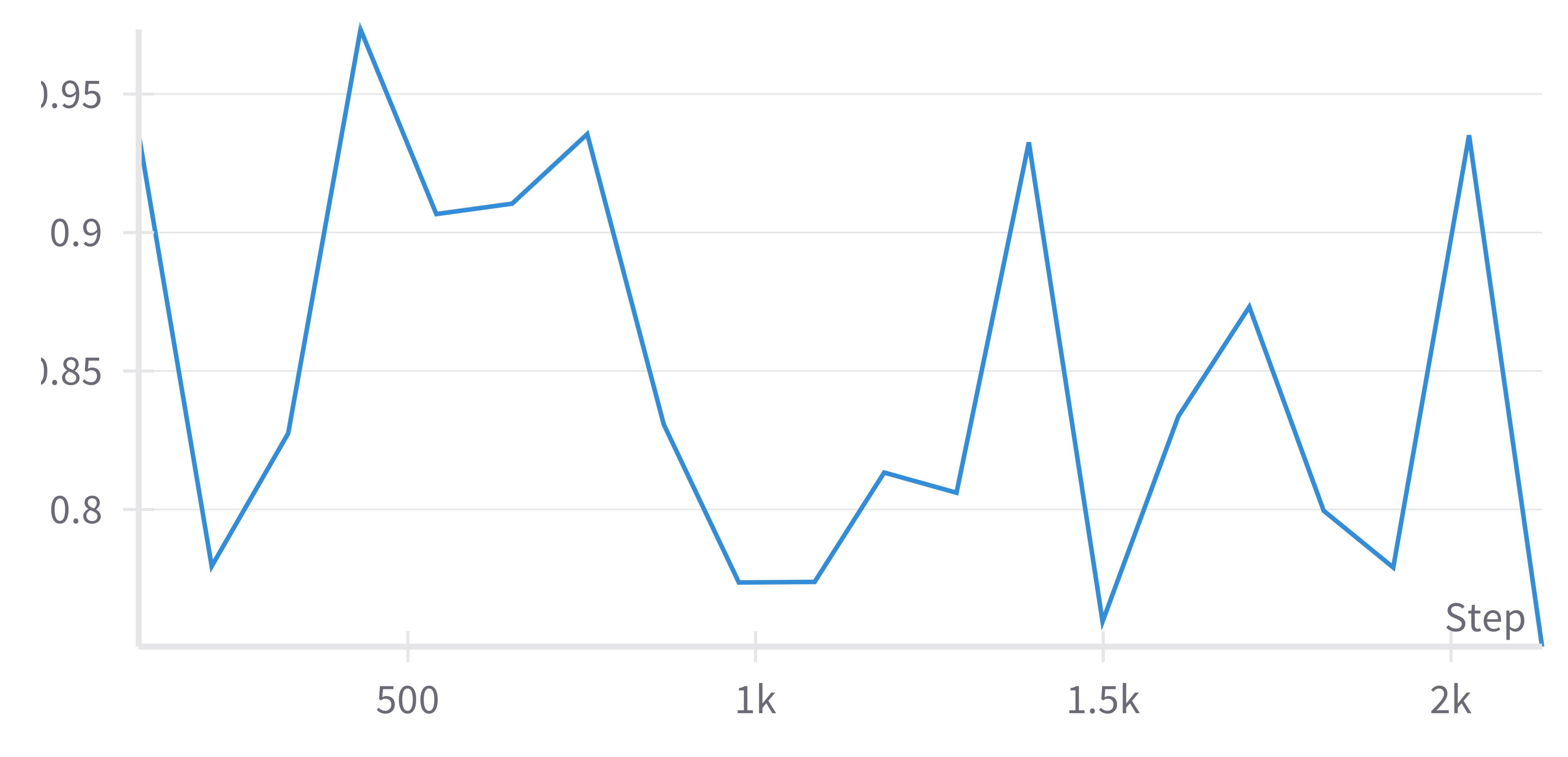}
    \vspace{1em}
    \text{Average Length}\\
    \begin{subfigure}[b]{0.24\textwidth}
        \includegraphics[width=\textwidth]{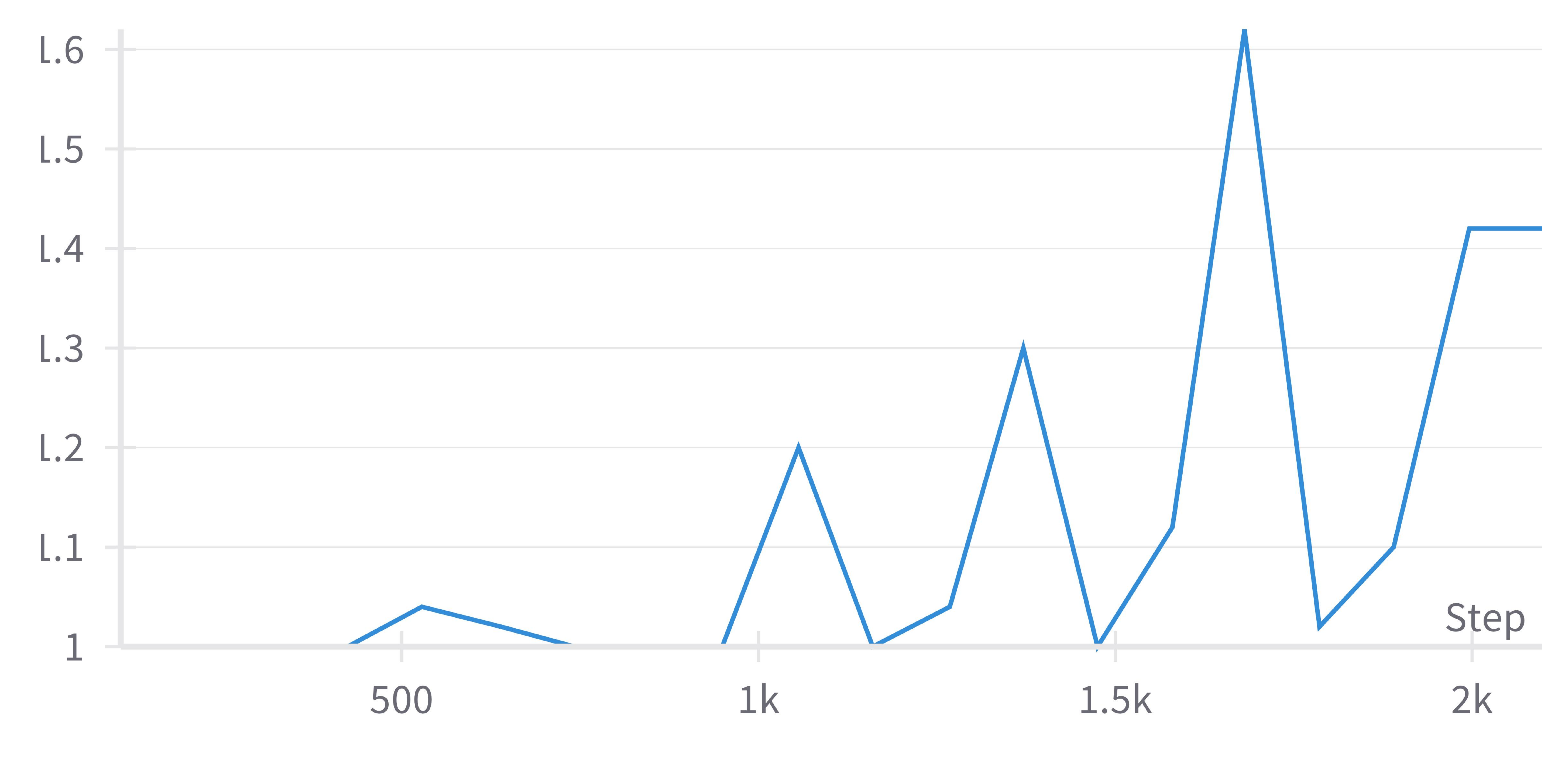}
        \caption{LLaMA-2-7b, $\alpha=0.2$}
    \end{subfigure}
    \begin{subfigure}[b]{0.24\textwidth}
        \includegraphics[width=\textwidth]{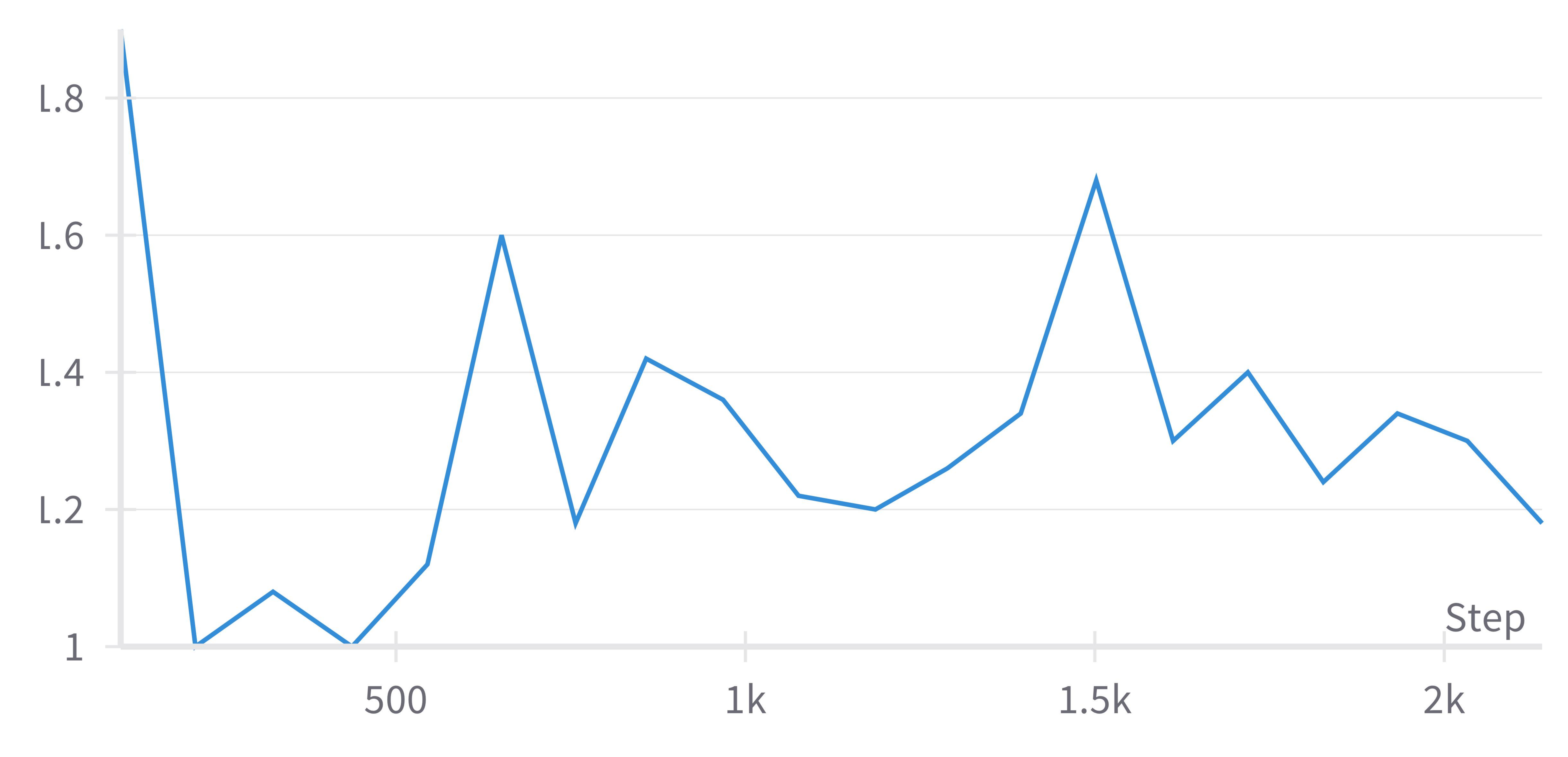}
        \caption{LLaMA-2-7b, $\alpha=0.1$}
    \end{subfigure}
    \begin{subfigure}[b]{0.24\textwidth}
        \includegraphics[width=\textwidth]{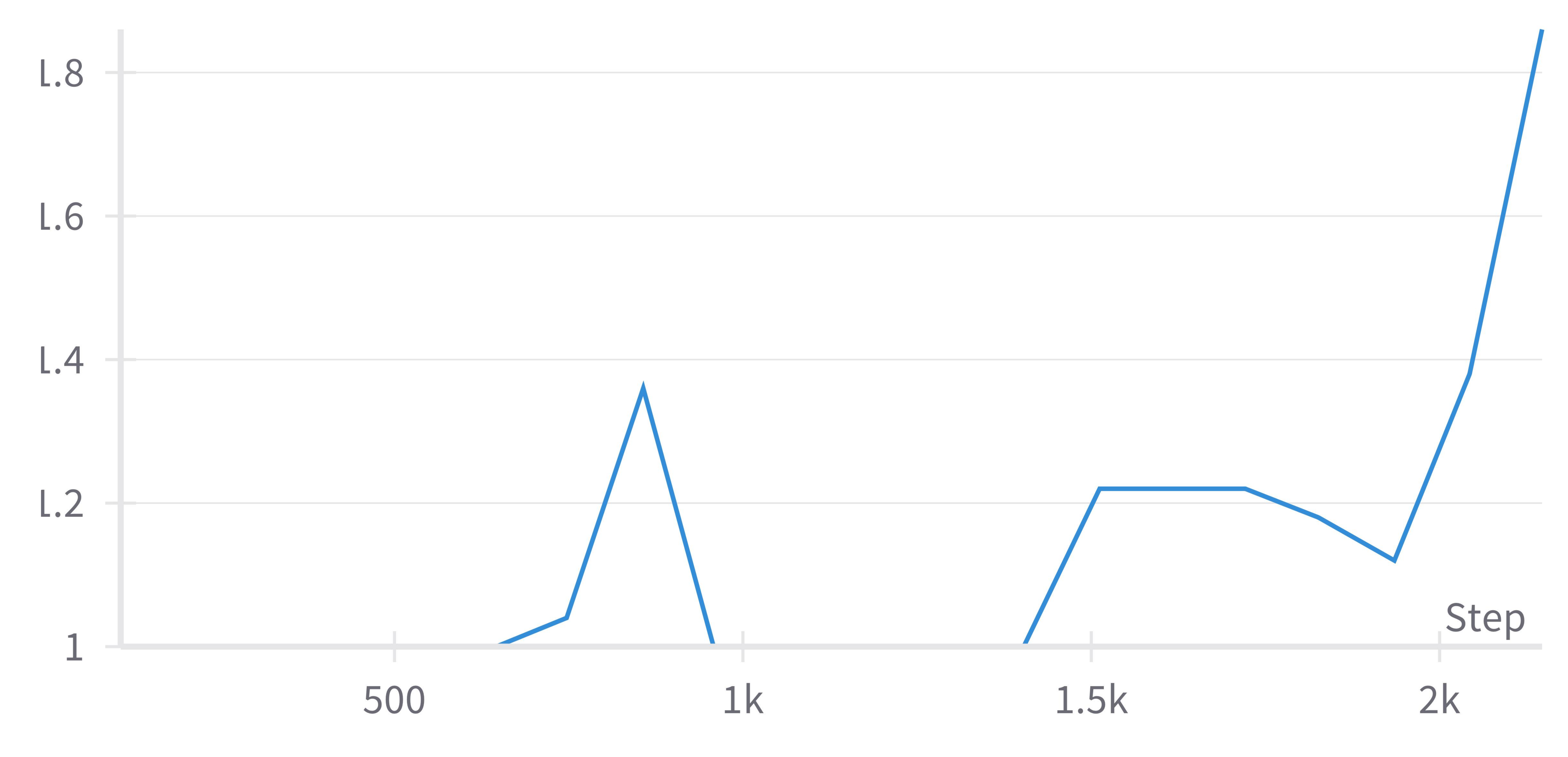}
        \caption{LLaMA-3.2-3b, $\alpha=0.2$}
    \end{subfigure}
    \begin{subfigure}[b]{0.24\textwidth}
        \includegraphics[width=\textwidth]{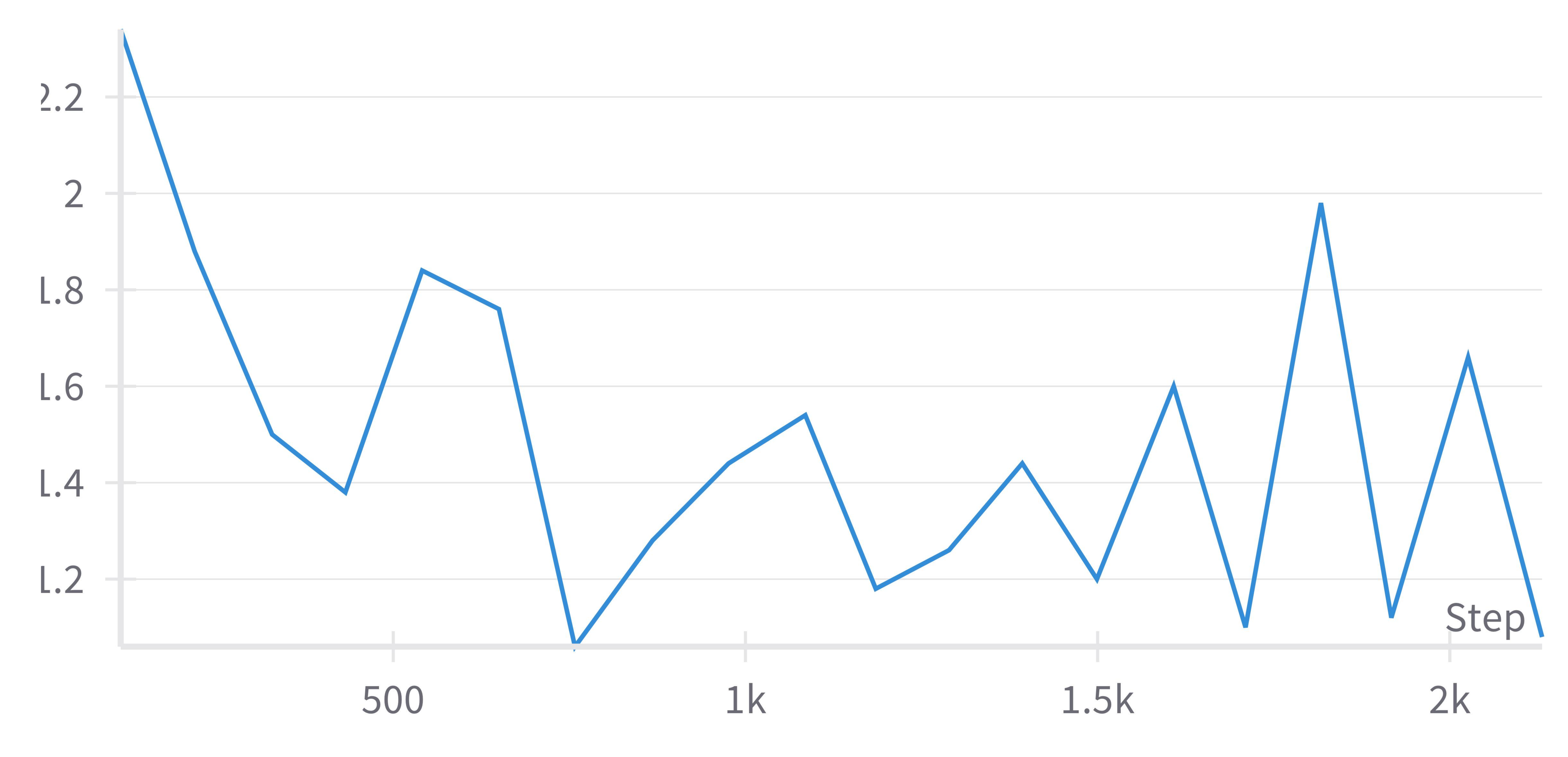}
        \caption{LLaMA-3.2-3b, $\alpha=0.1$}
    \end{subfigure}
    
    \caption{CCPO validation performance on MMLU, $\lambda=0$.}
    \label{fig:val_mmlu}
\end{figure*}

\begin{figure*}[htbp]
    \centering
    \vspace{1em}
    Train Coverage\\
    % First row: 4 images without subcaptions
    \includegraphics[width=0.24\textwidth]{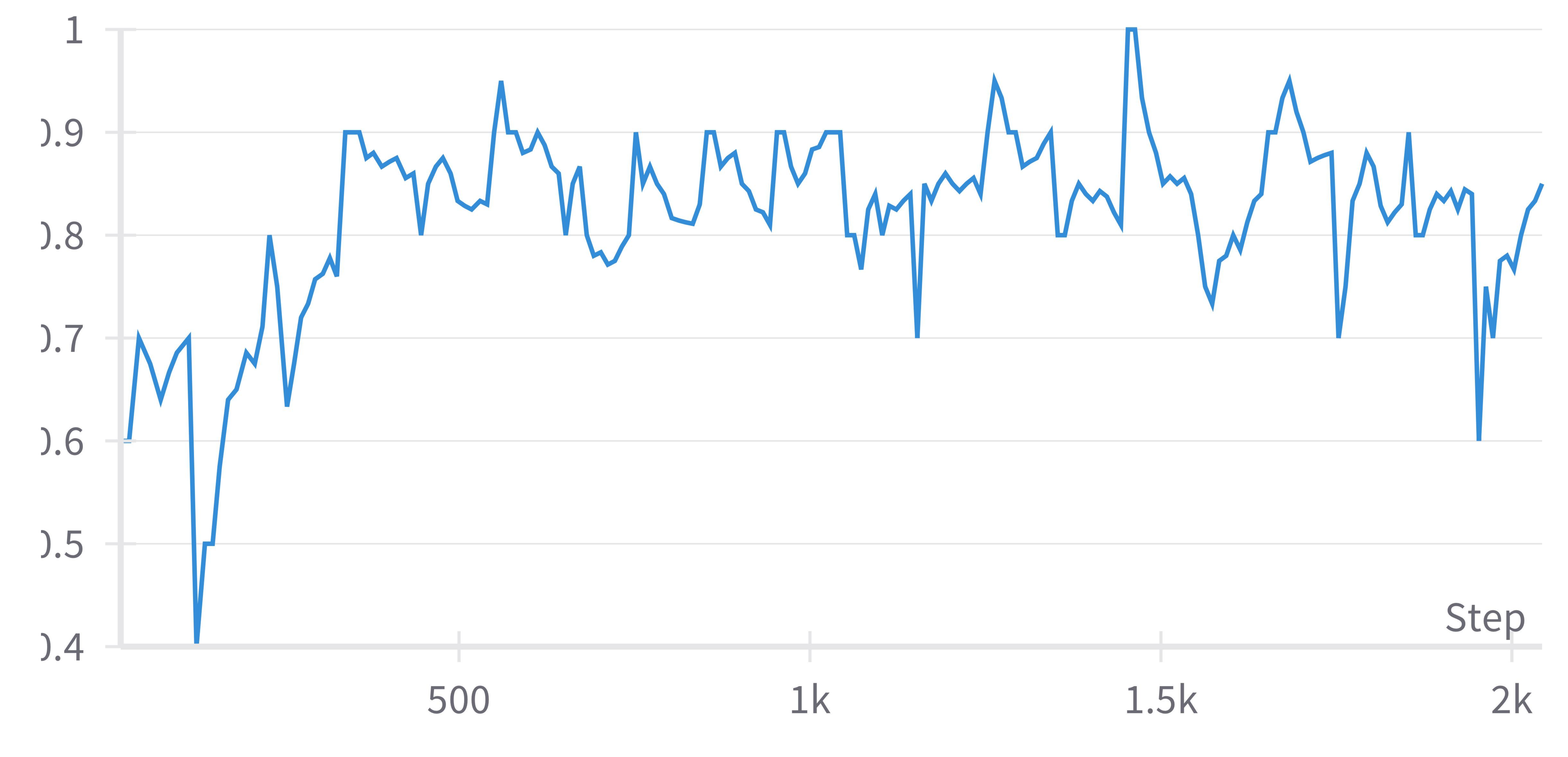}
    \includegraphics[width=0.24\textwidth]{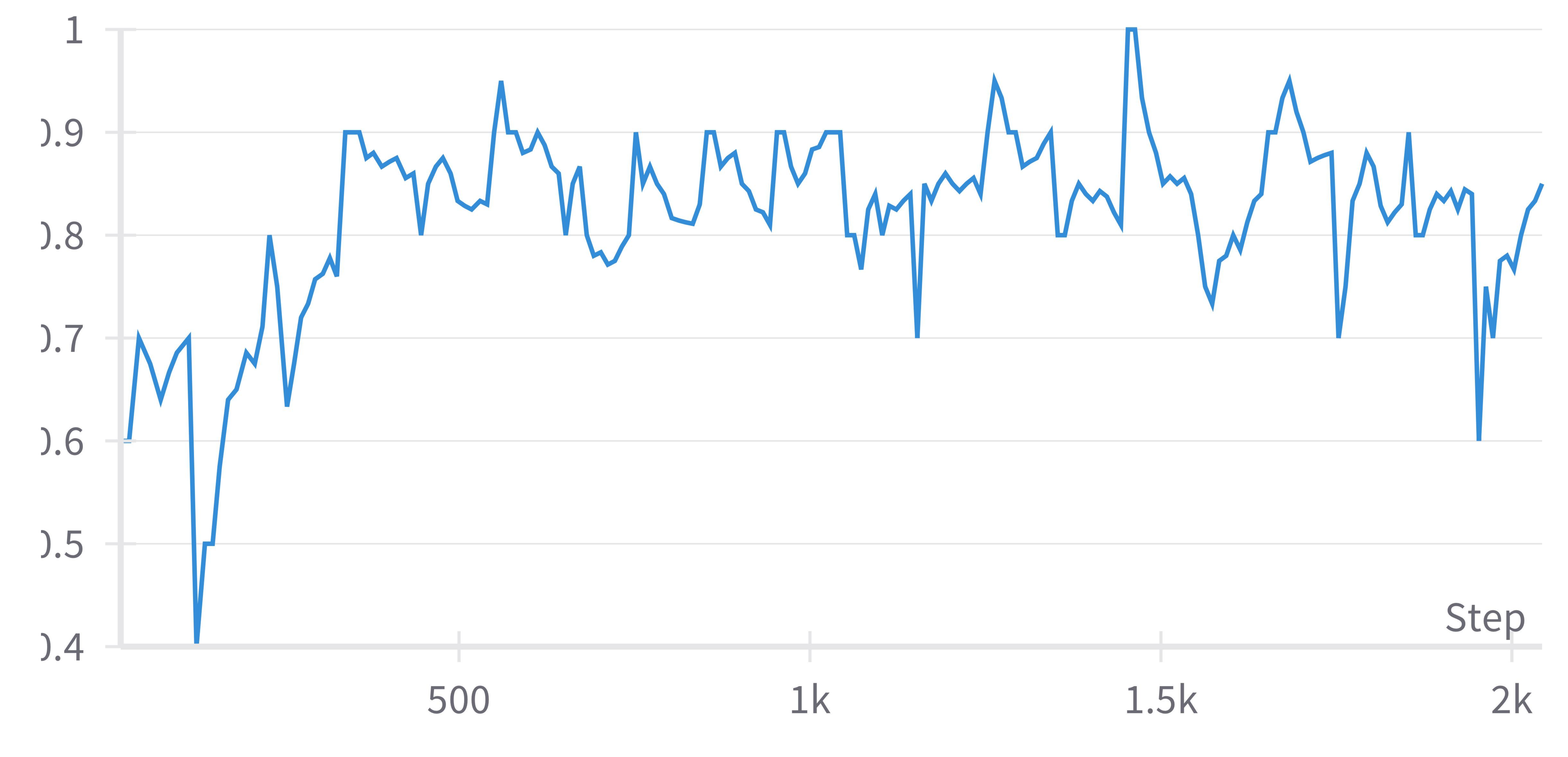}
    \includegraphics[width=0.24\textwidth]{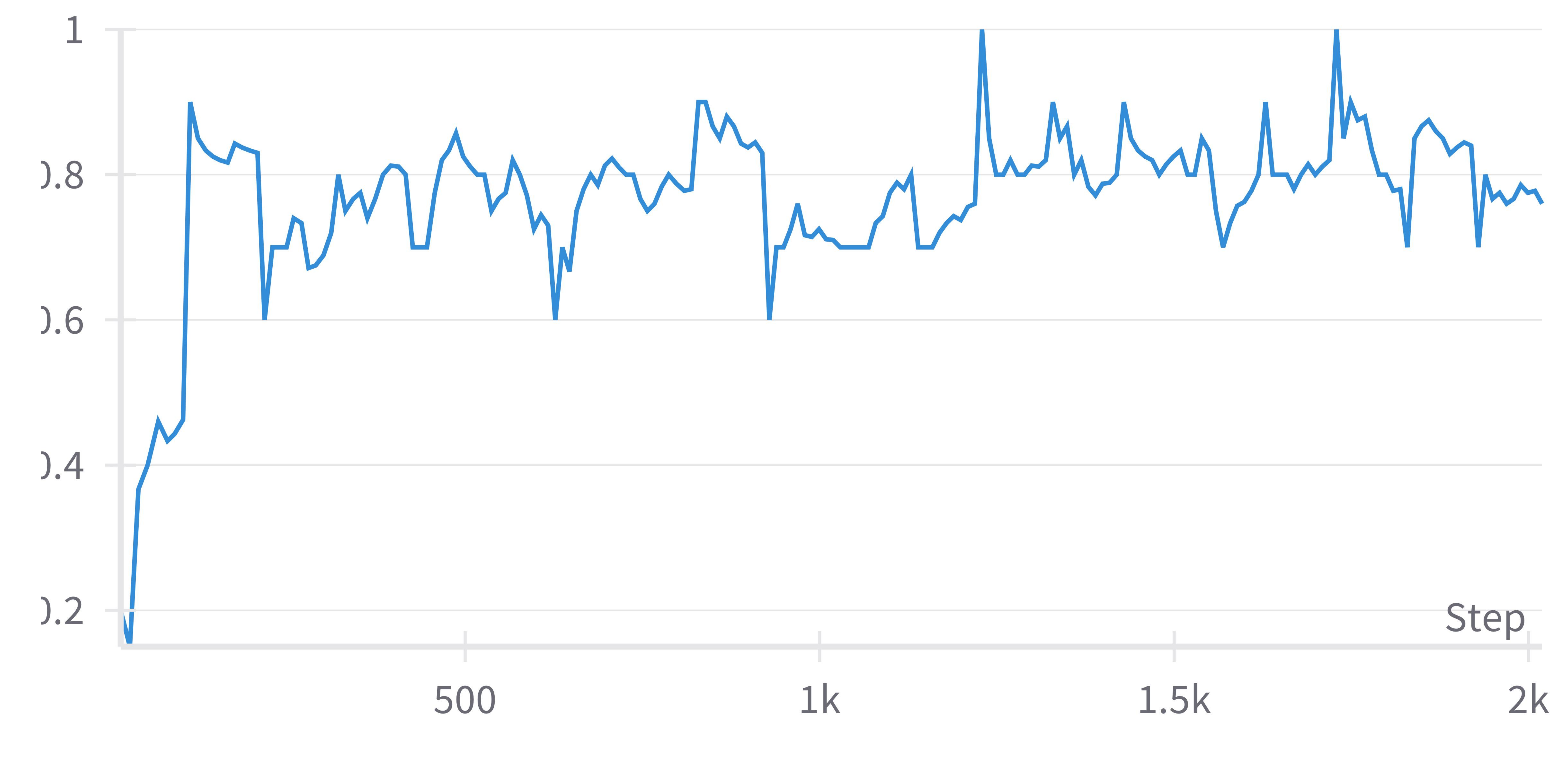}
    \includegraphics[width=0.24\textwidth]{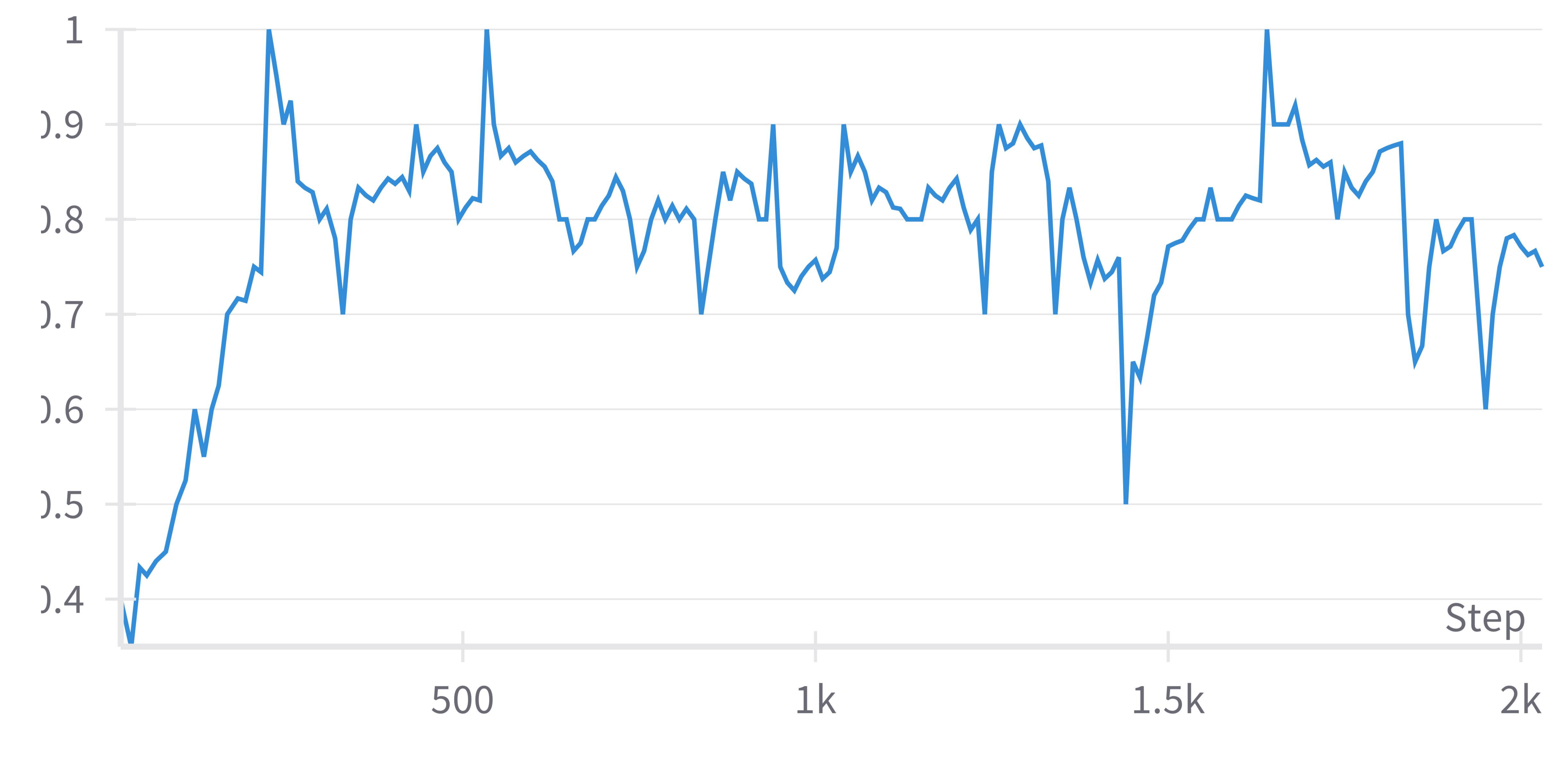}
    \vspace{1em}
    \text{Validation Coverage}\\
    % \vspace{0.5cm} % vertical space between rows
    \includegraphics[width=0.24\textwidth]{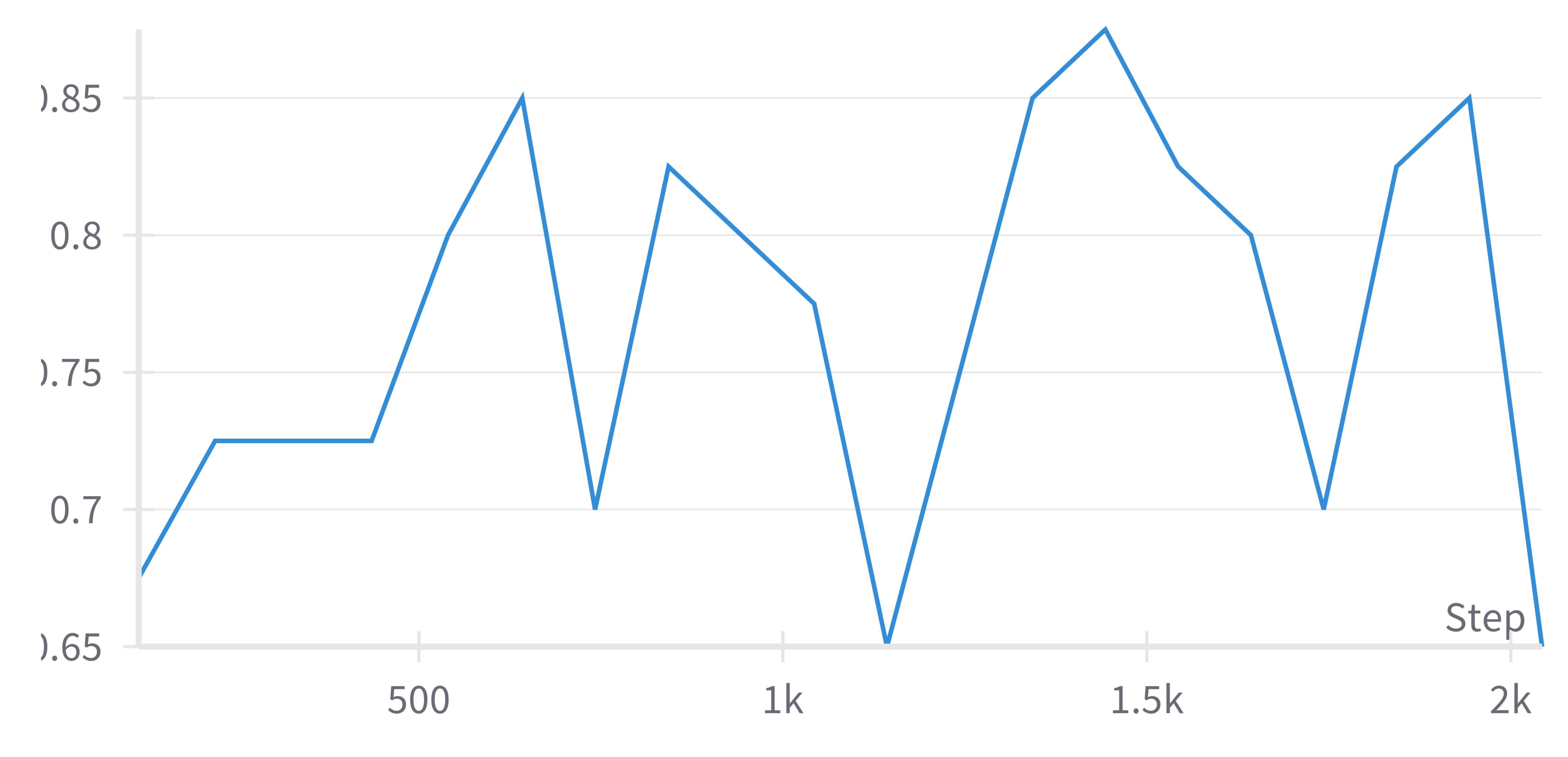}
    \includegraphics[width=0.24\textwidth]{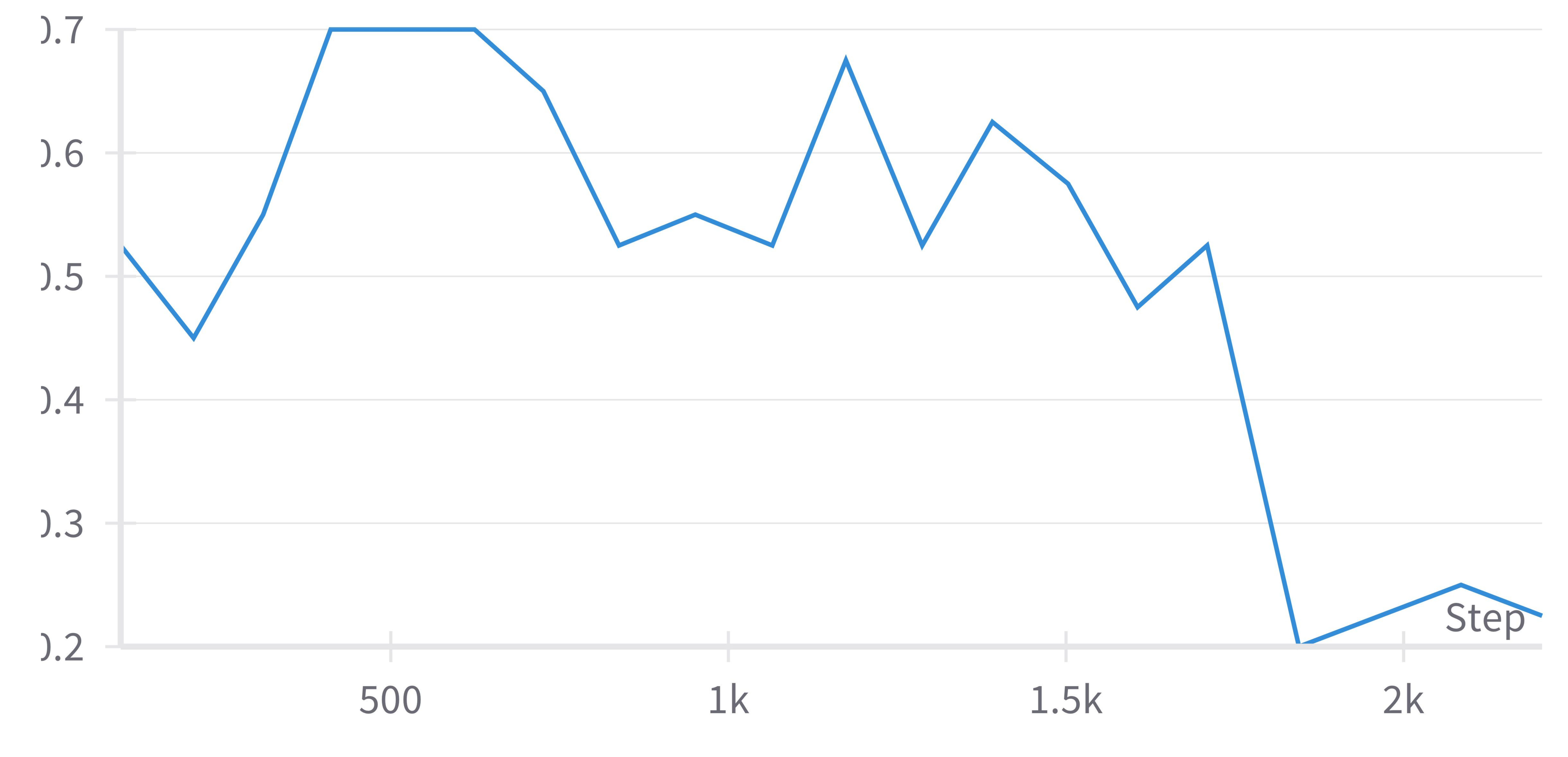}
    \includegraphics[width=0.24\textwidth]{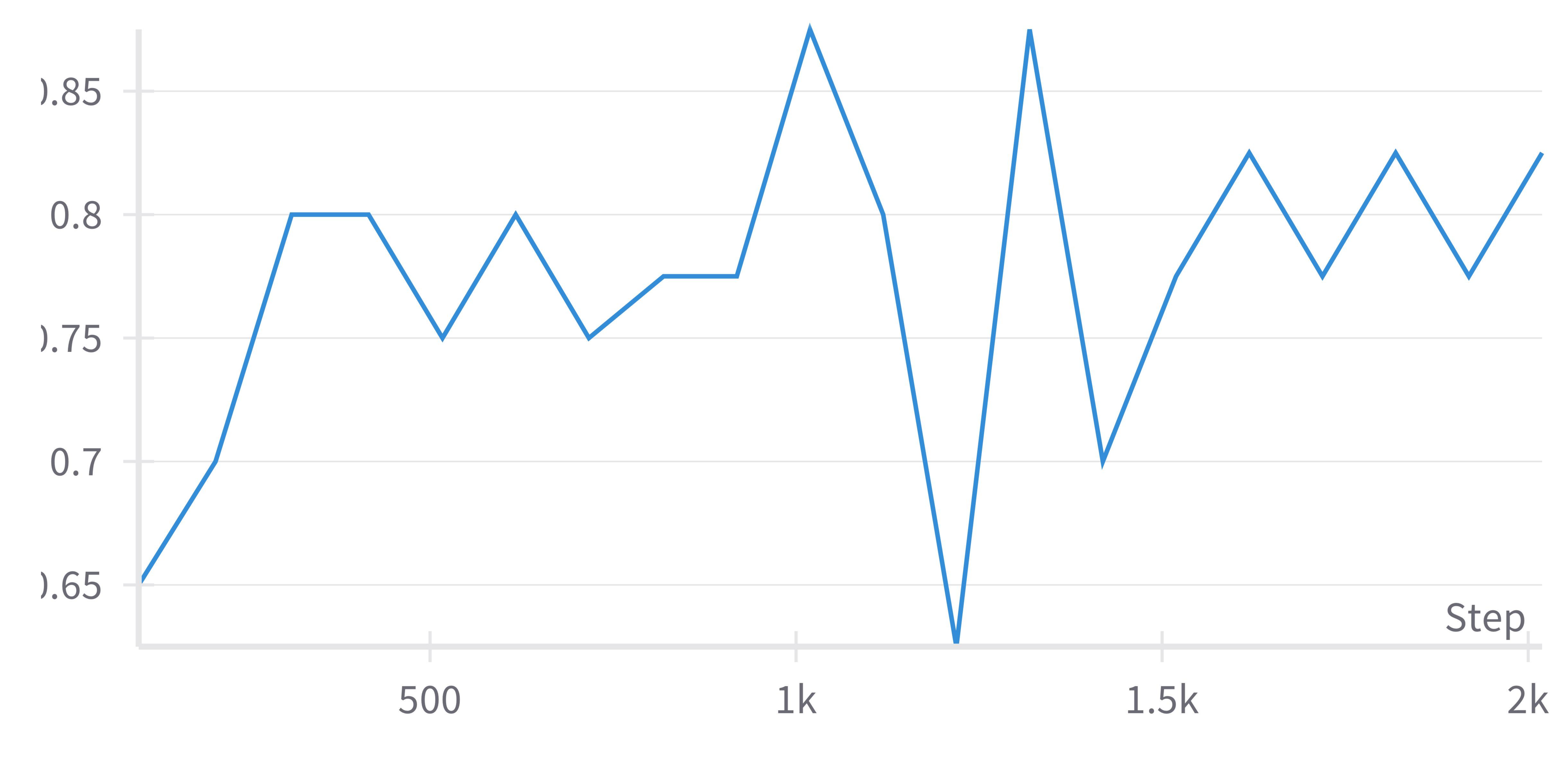}
    \includegraphics[width=0.24\textwidth]{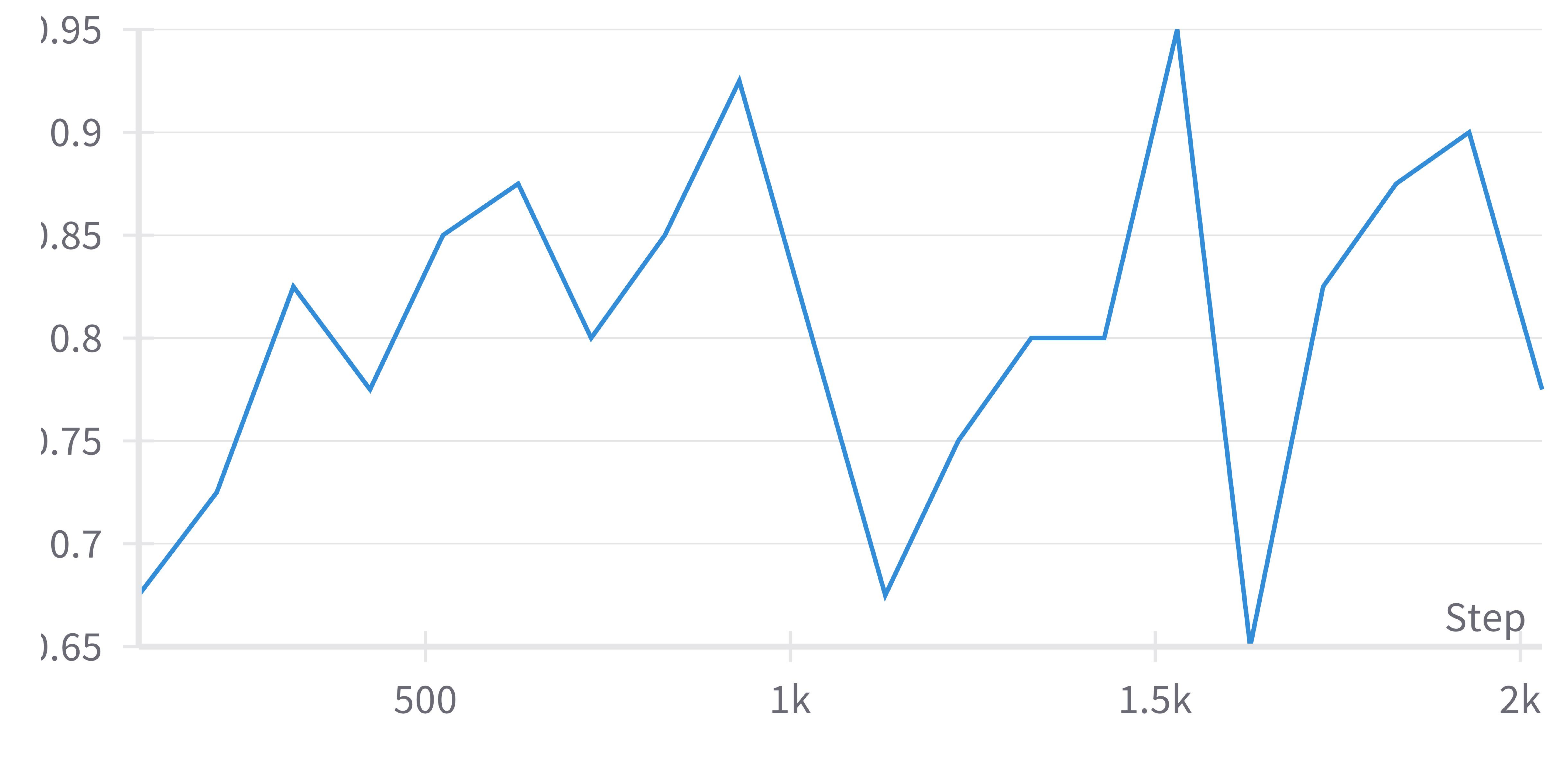}
    \vspace{1em}
    \text{Validation Length}\\
    % Second row: 4 images with subcaptions
    \begin{subfigure}[b]{0.24\textwidth}
        \includegraphics[width=\textwidth]{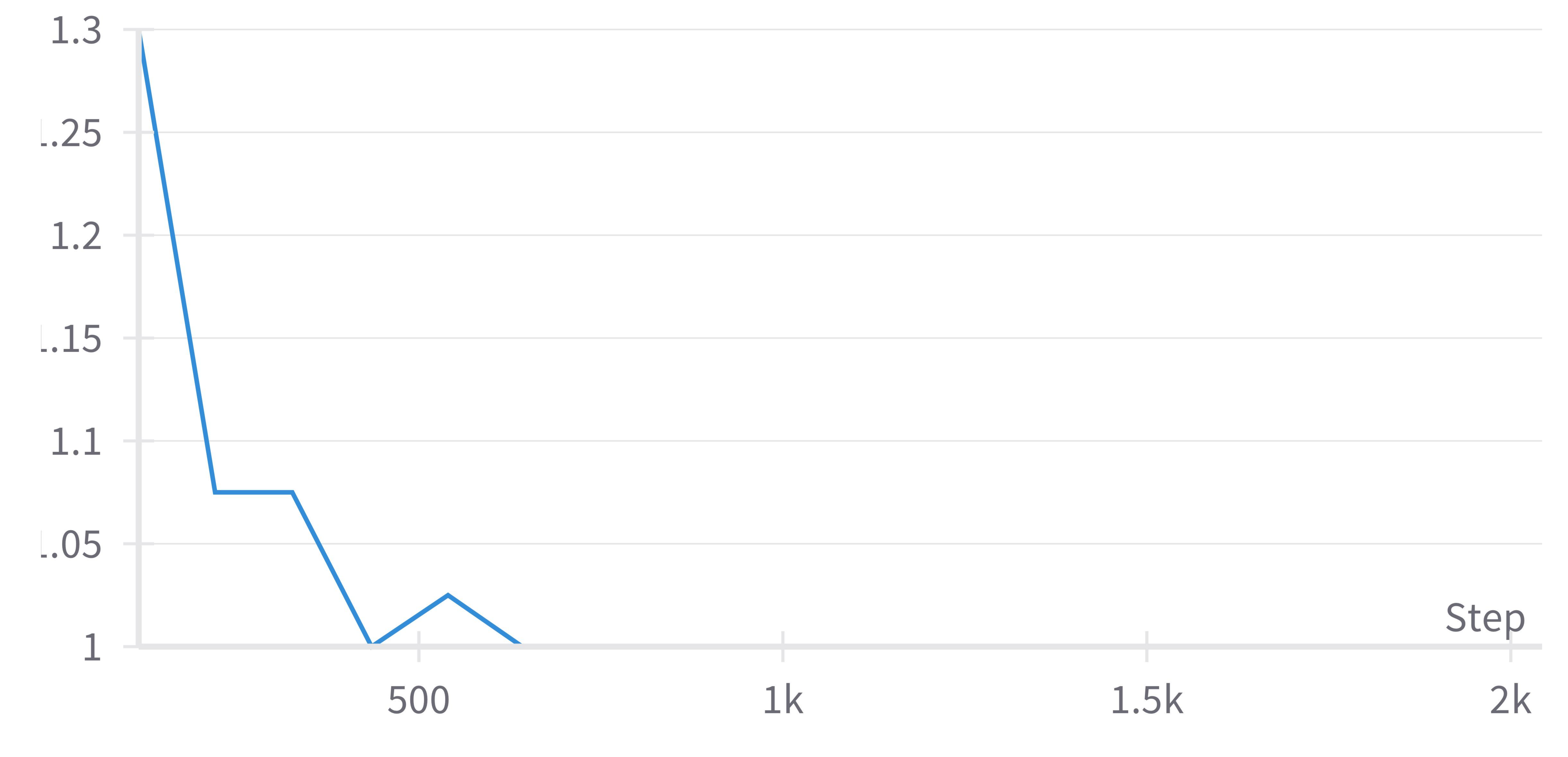}
        \caption{LLaMA-2-7b, $\alpha=0.2$}
    \end{subfigure}
    \begin{subfigure}[b]{0.24\textwidth}
        \includegraphics[width=\textwidth]{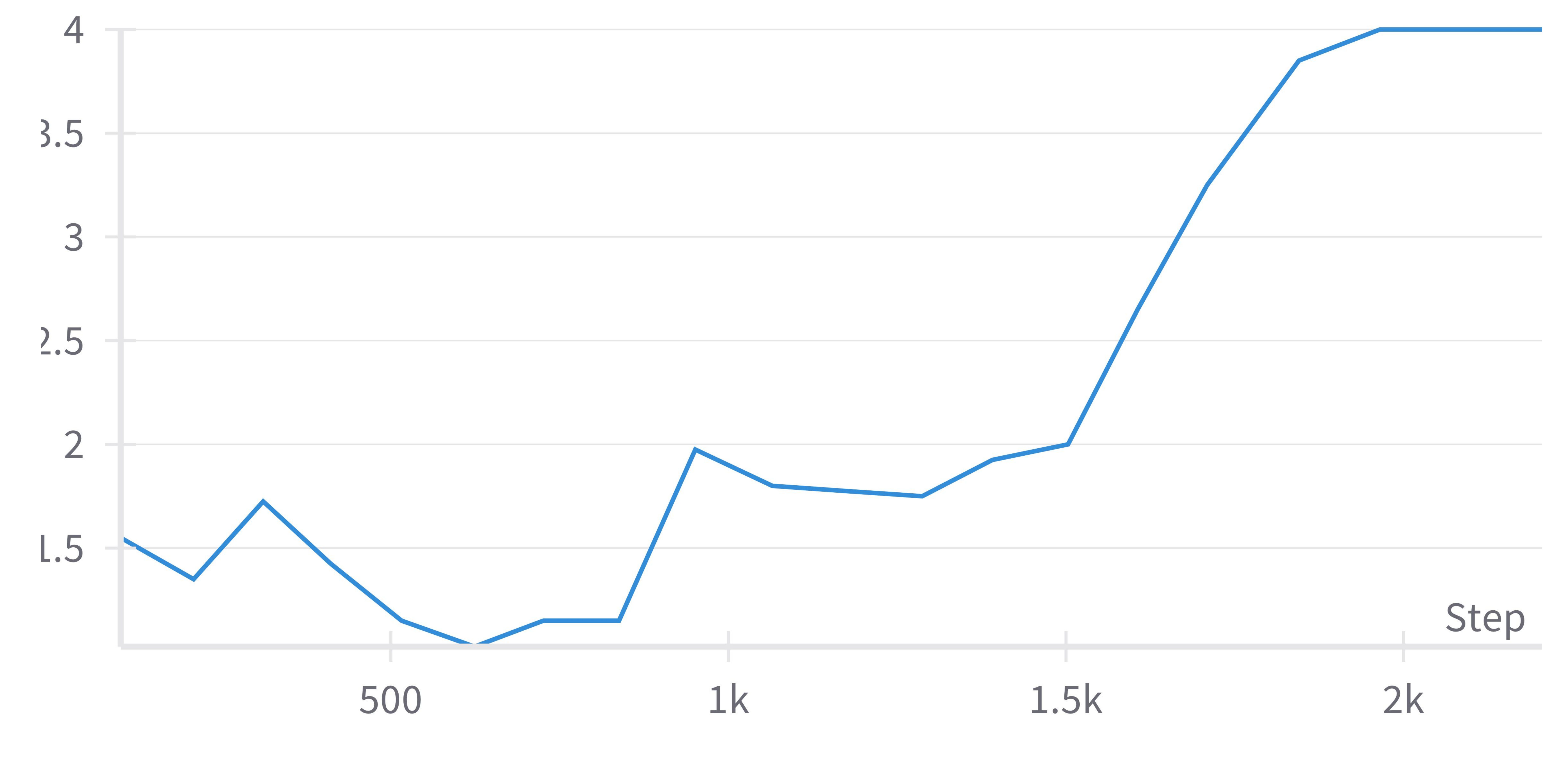}
        \caption{LLaMA-2-7b, $\alpha=0.1$}
    \end{subfigure}
    \begin{subfigure}[b]{0.24\textwidth}
        \includegraphics[width=\textwidth]{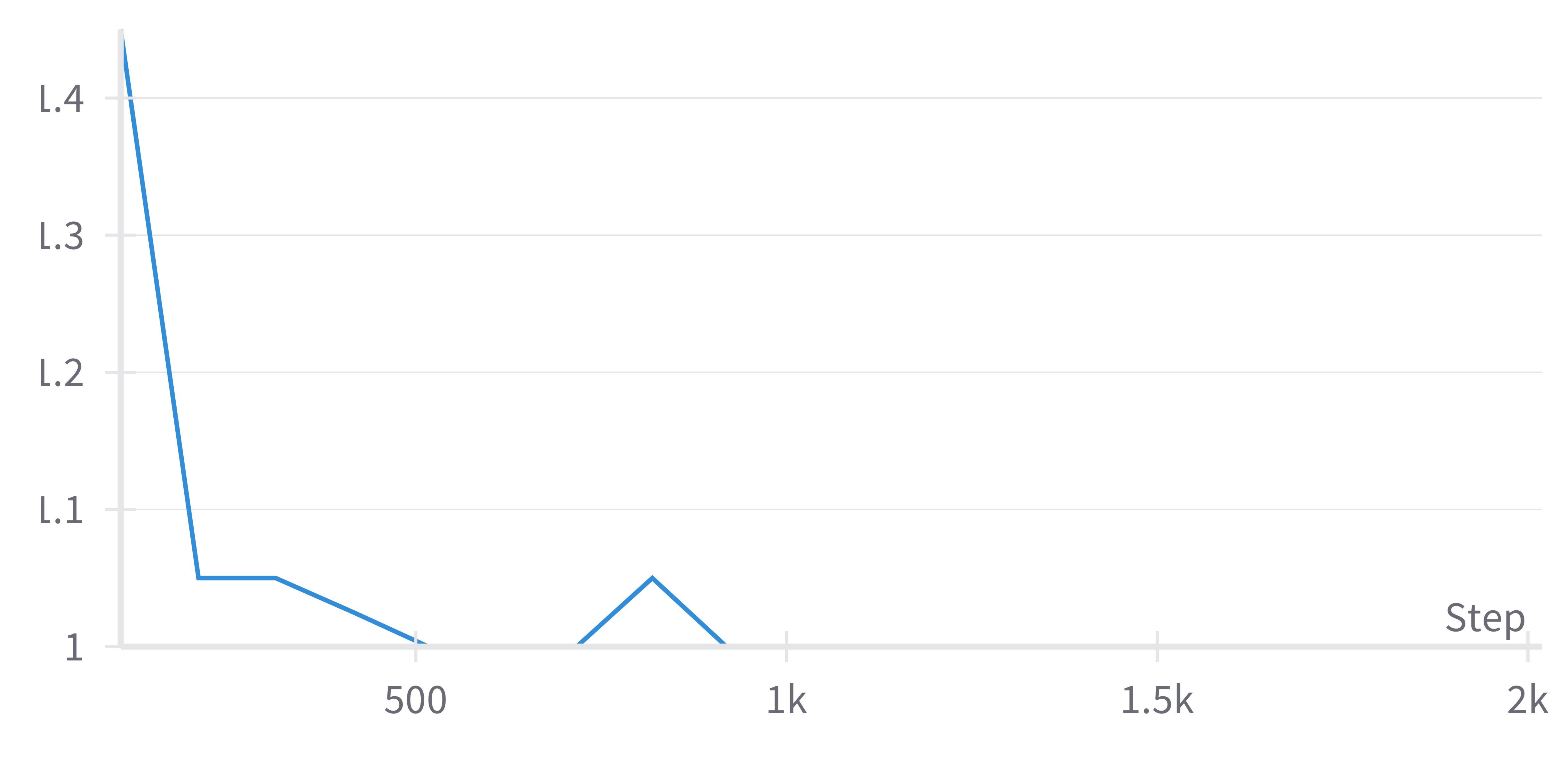}
        \caption{LLaMA-3.2-3b, $\alpha=0.2$}
    \end{subfigure}
    \begin{subfigure}[b]{0.24\textwidth}
        \includegraphics[width=\textwidth]{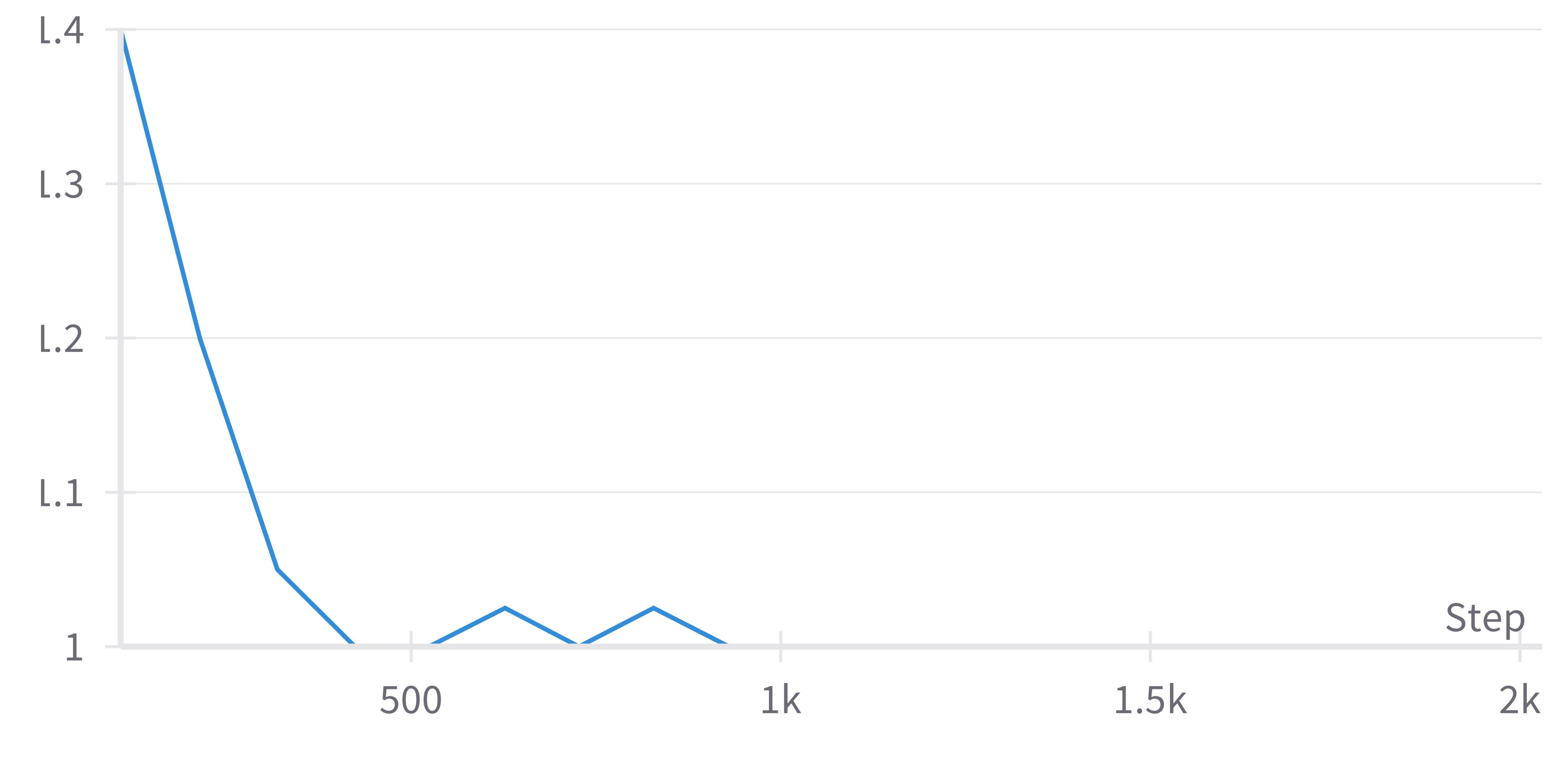}
        \caption{LLaMA-3.2-3b, $\alpha=0.1$}
    \end{subfigure}
    
    \caption{CPO performance on MMLU dataset.}
    \label{fig:cpo_perf_mmlu}
\end{figure*}

\begin{figure*}[htbp]
    \centering
    \vspace{1em}
    Costs\\
    % First row: 4 images without subcaptions
    \includegraphics[width=0.24\textwidth]{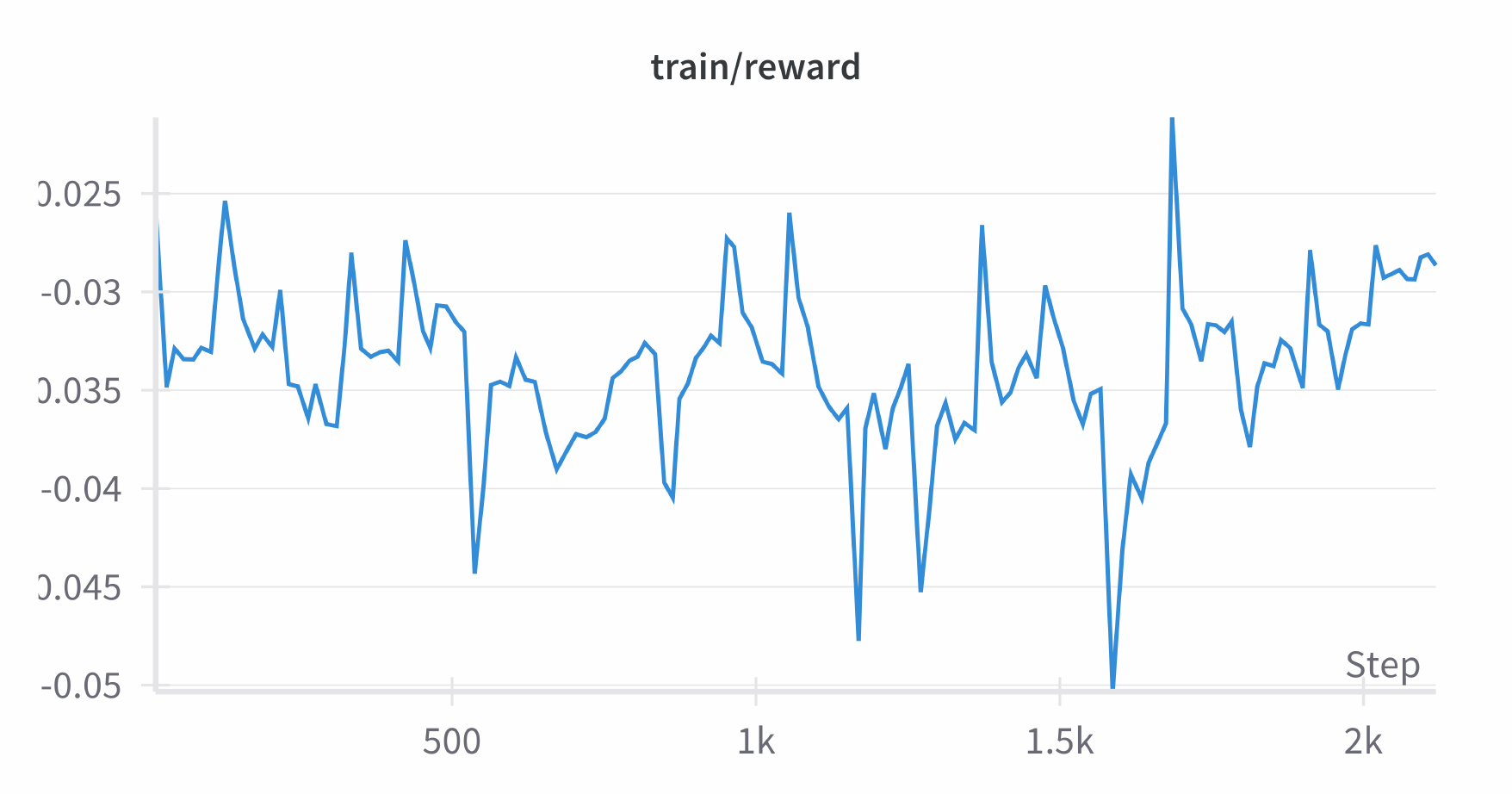}
    \includegraphics[width=0.24\textwidth]{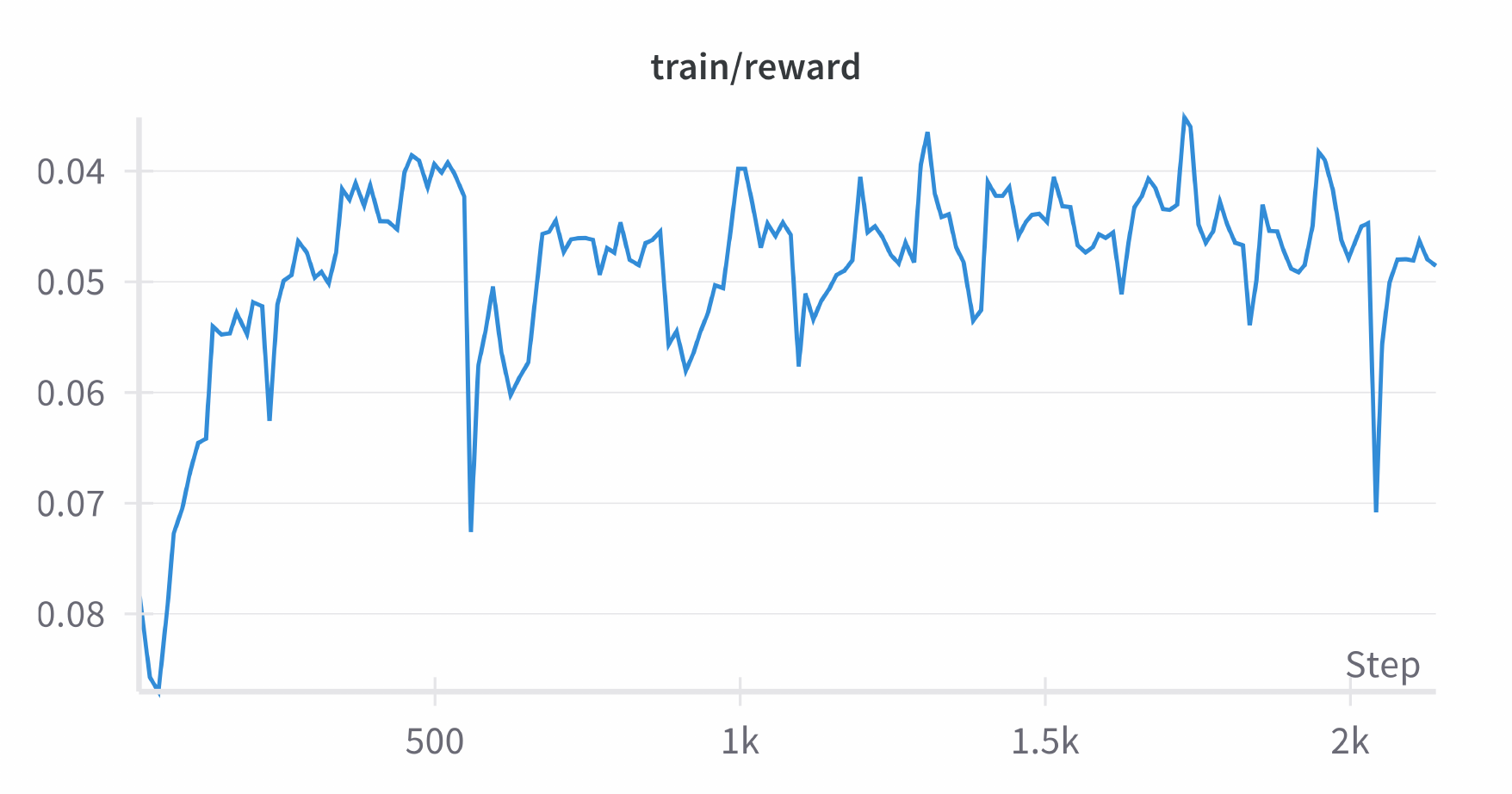}
    \includegraphics[width=0.24\textwidth]{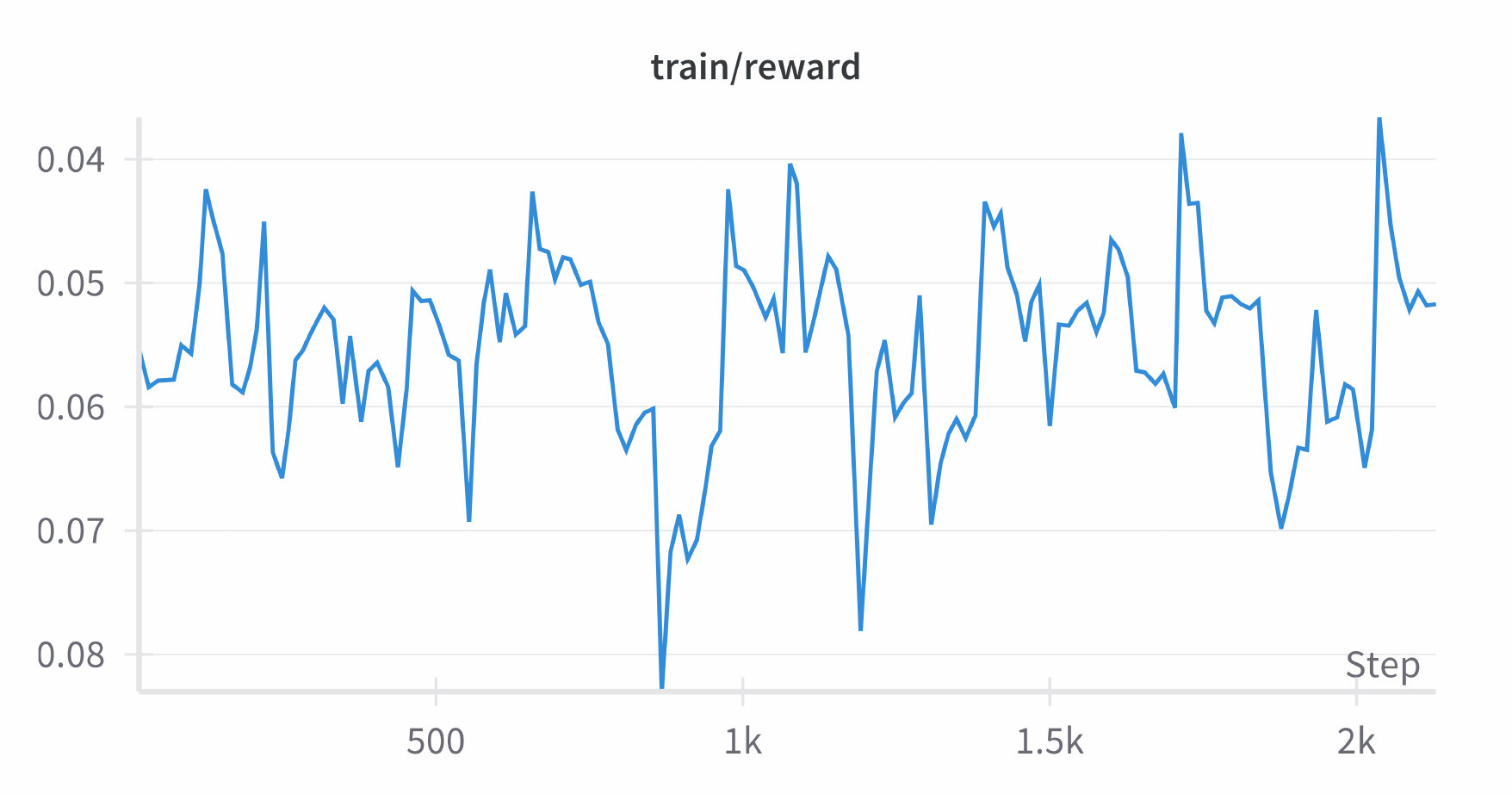}
    \includegraphics[width=0.24\textwidth]{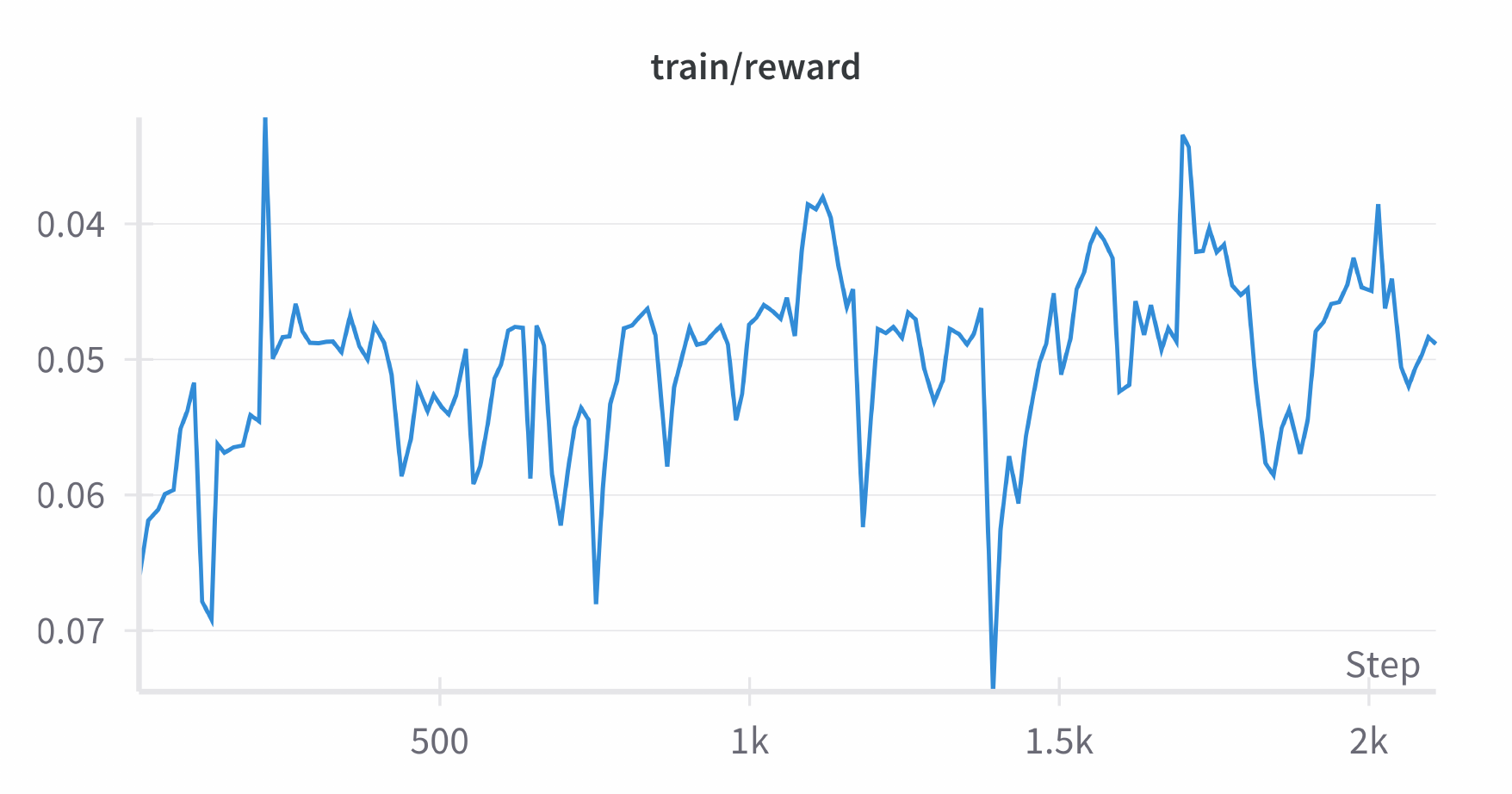}
    \vspace{1em}
    \text{Coverage Surrogate Violation}\\
    % \vspace{0.5cm} % vertical space between rows
    \includegraphics[width=0.24\textwidth]{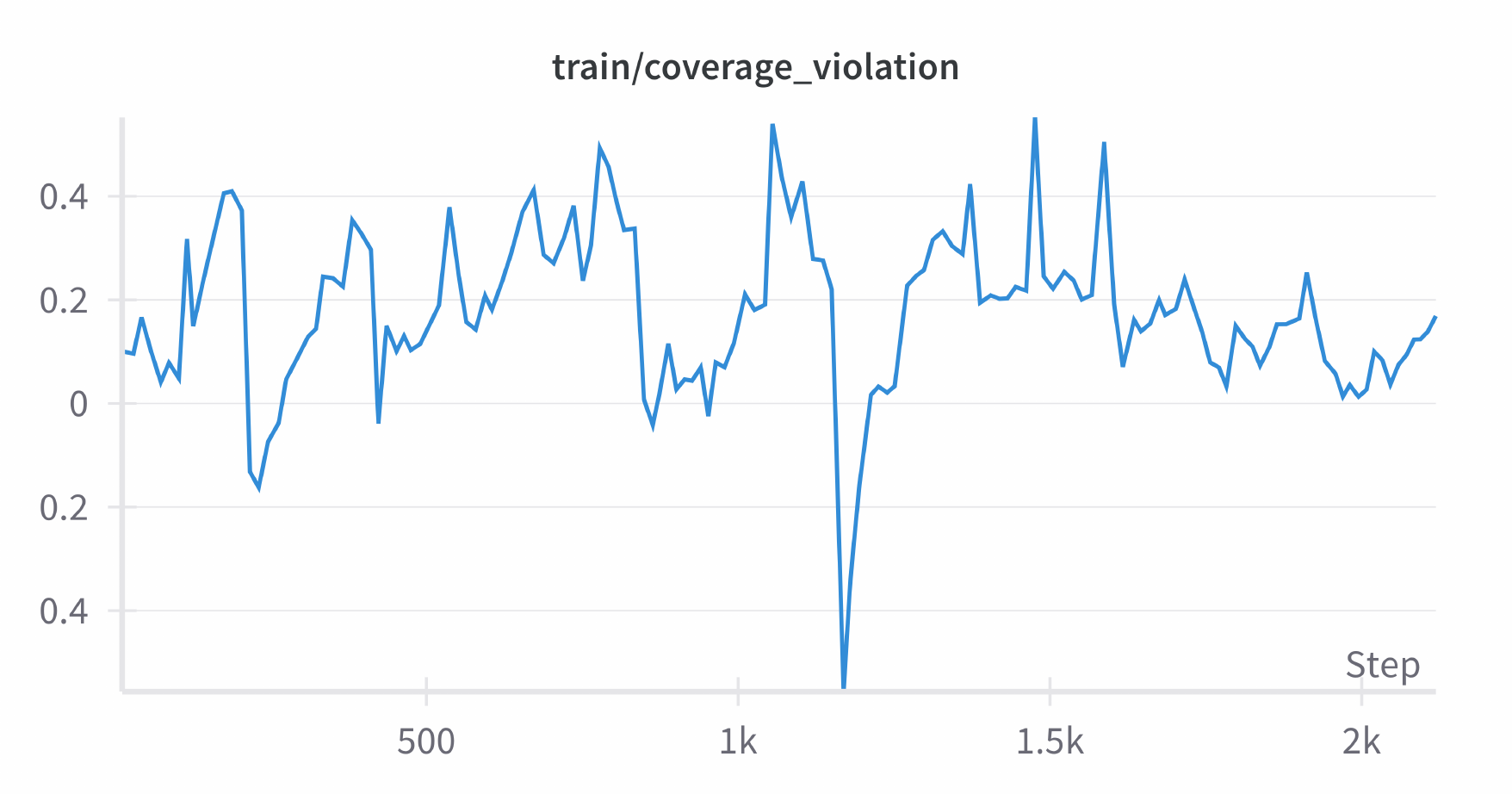}
    \includegraphics[width=0.24\textwidth]{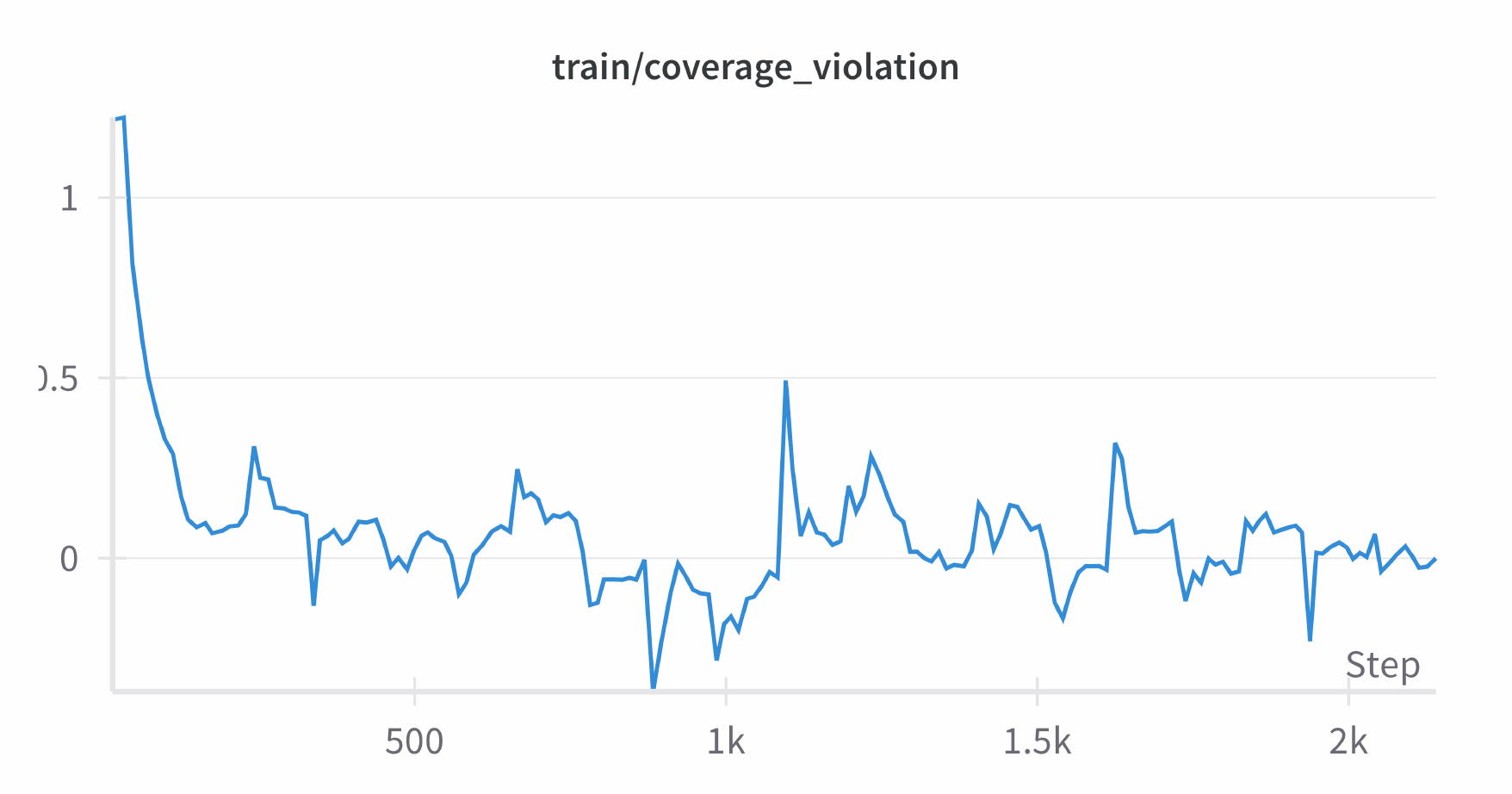}
    \includegraphics[width=0.24\textwidth]{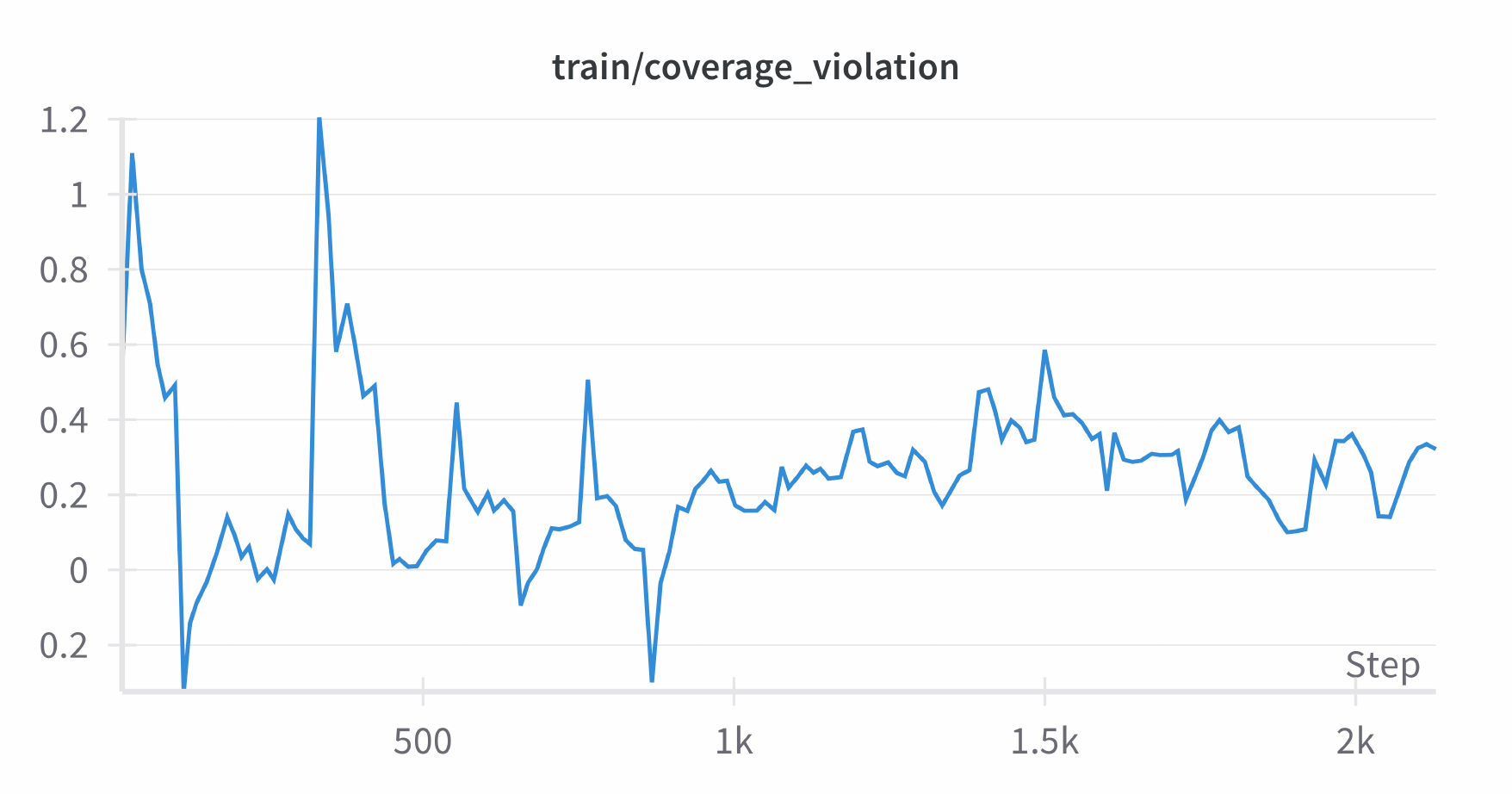}
    \includegraphics[width=0.24\textwidth]{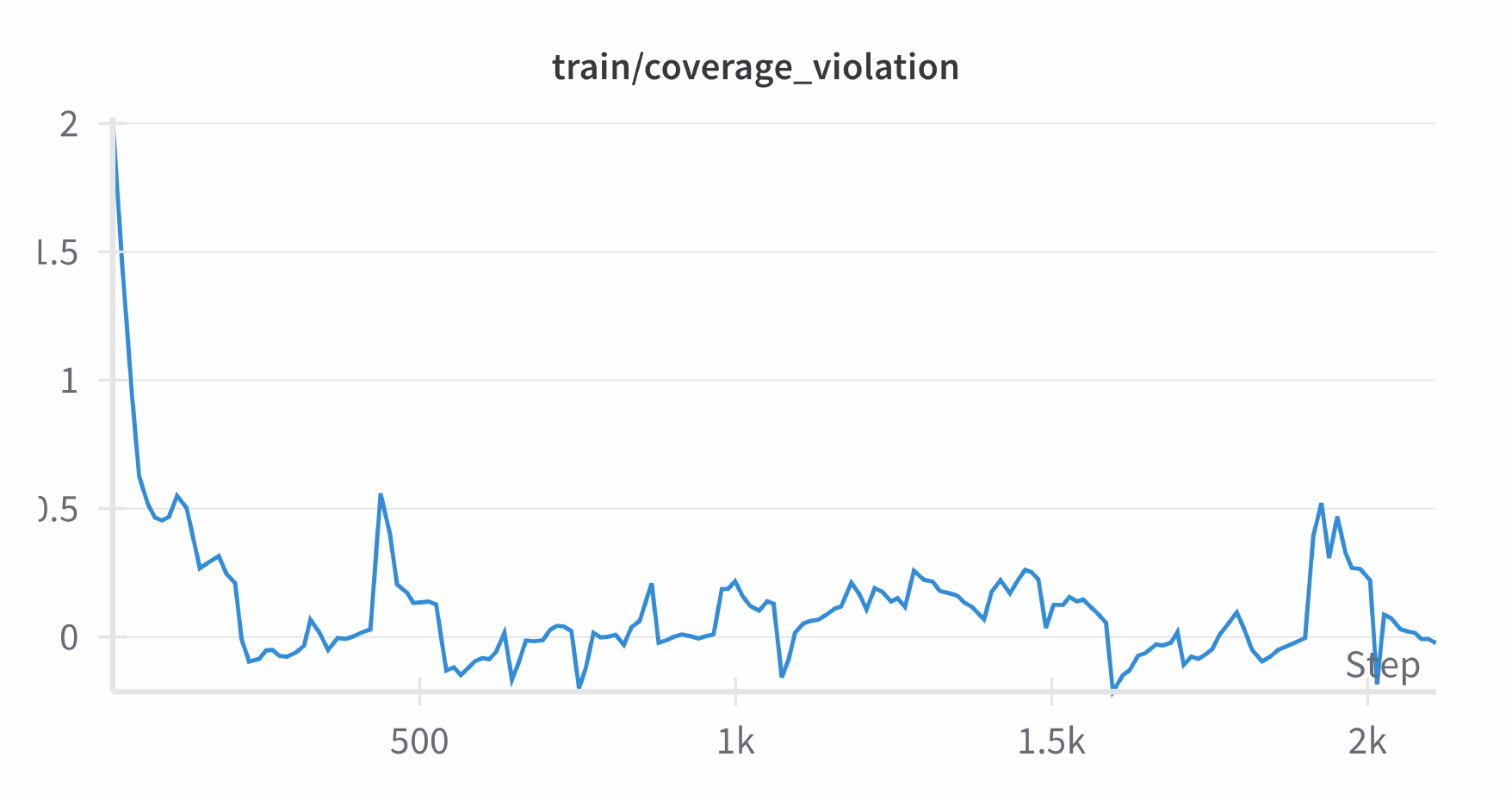}
    \vspace{1em}
    \text{Training Coverage}\\
    % Second row: 4 images with subcaptions
    \begin{subfigure}[b]{0.24\textwidth}
        \includegraphics[width=\textwidth]{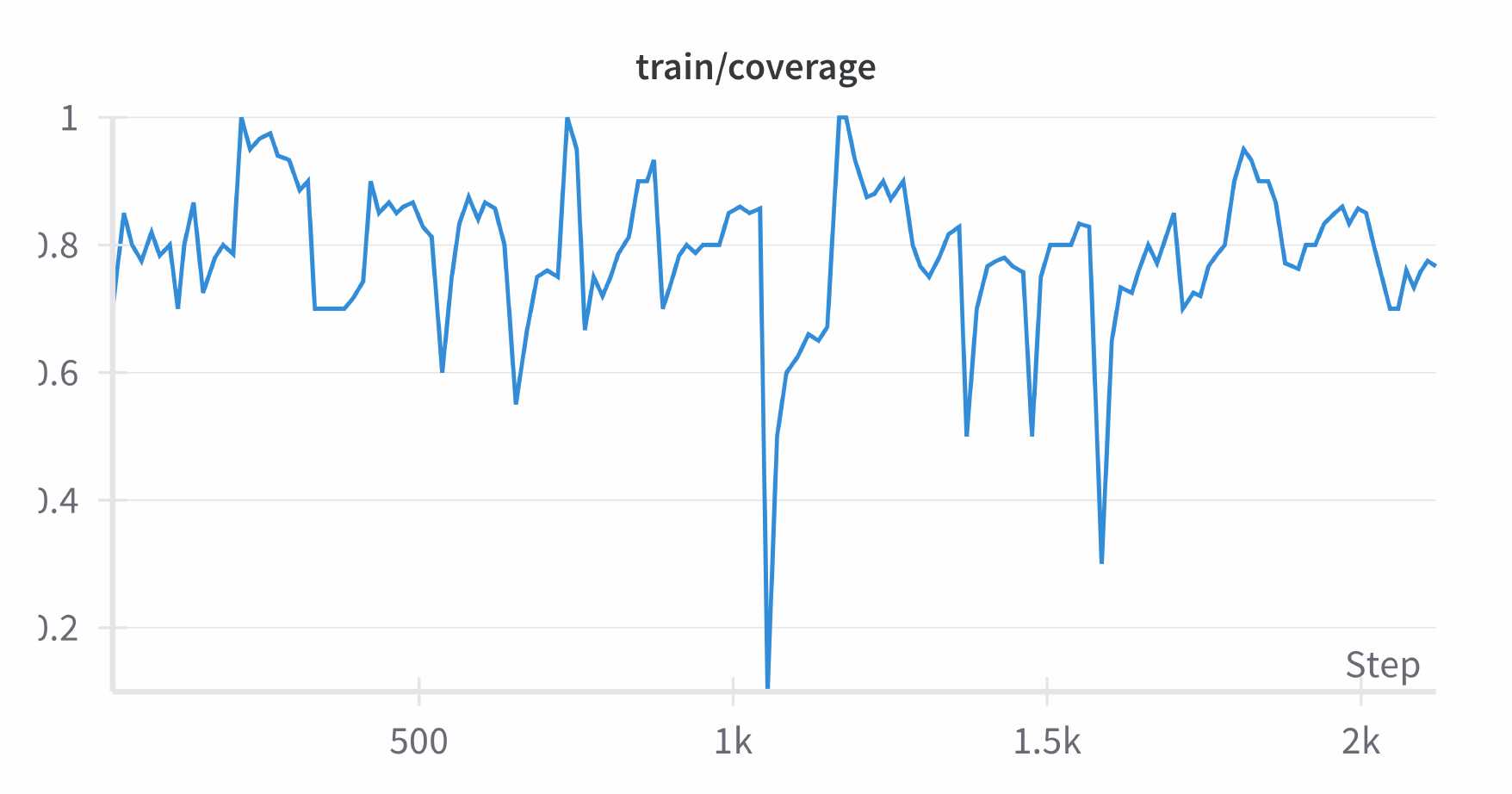}
        \caption{LLaMA-2-7b, $\alpha=0.2$, \\$\lambda=2e^{-4}$.}
    \end{subfigure}
    \begin{subfigure}[b]{0.24\textwidth}
        \includegraphics[width=\textwidth]{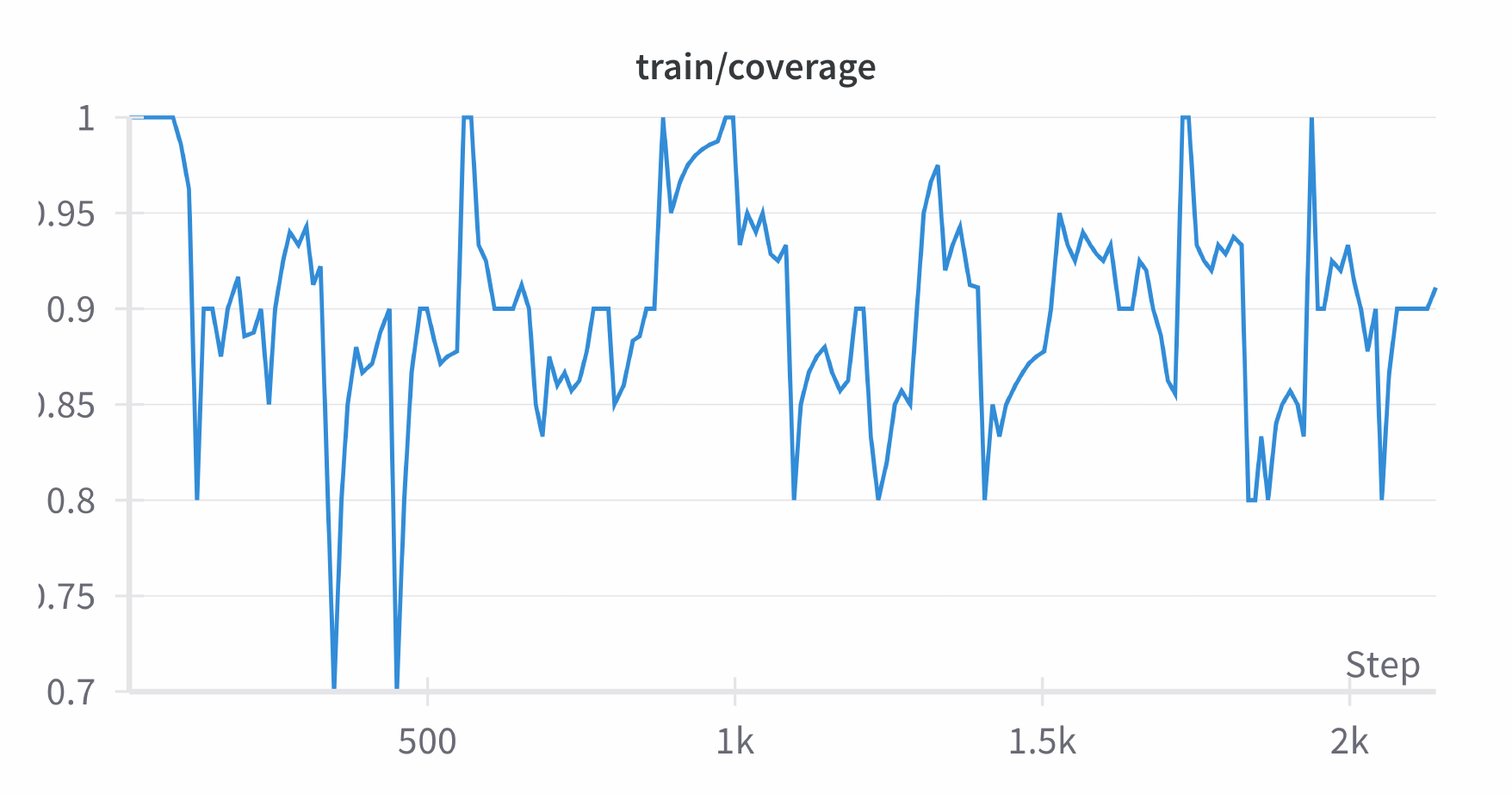}
        \caption{LLaMA-2-7b, $\alpha=0.1$, \\$\lambda=1e^{-4}$.}
    \end{subfigure}
    \begin{subfigure}[b]{0.24\textwidth}
        \includegraphics[width=\textwidth]{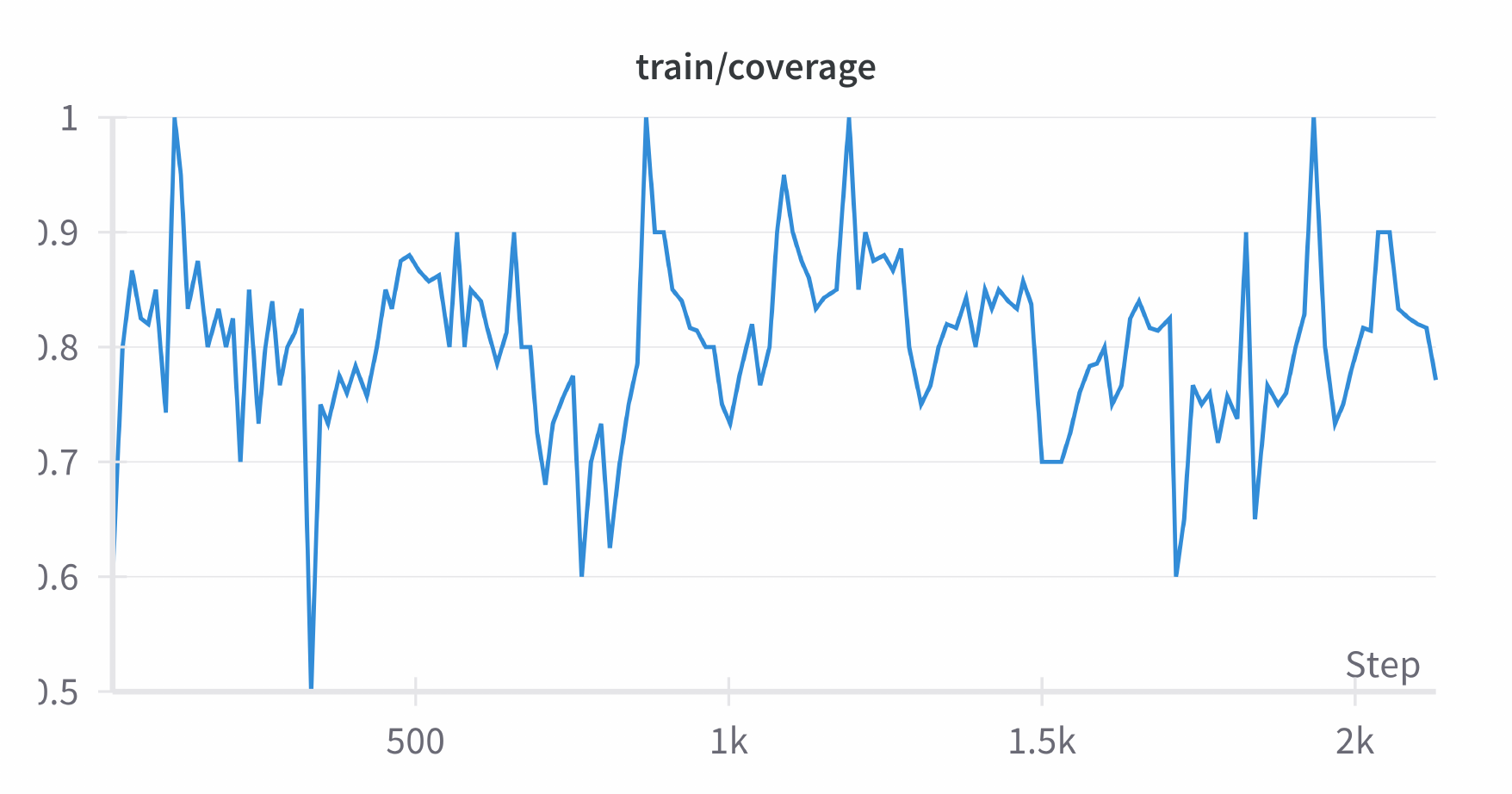}
        \caption{LLaMA-3.2-3b, $\alpha=0.2$, \\$\lambda=1e^{-4}$.}
    \end{subfigure}
    \begin{subfigure}[b]{0.24\textwidth}
        \includegraphics[width=\textwidth]{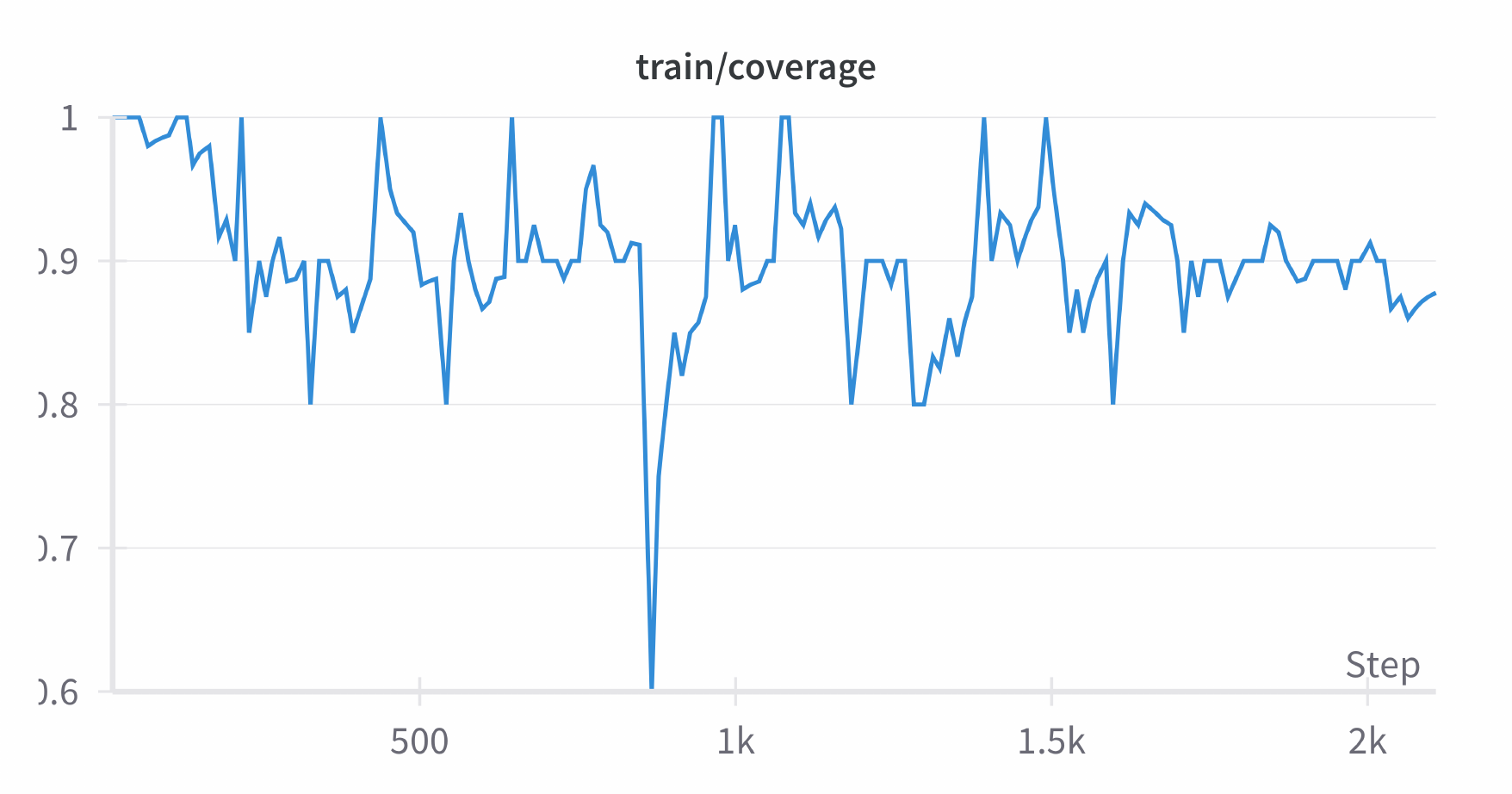}
        \caption{LLaMA-3.2-3b, $\alpha=0.1$, \\$\lambda=1e^{-4}$.}
    \end{subfigure}
    
    \caption{Overall CCPO training performance on MMLU dataset with varying $\lambda$.}
    \label{fig:perf_mmlu_lambda}
\end{figure*}

\begin{figure*}[htbp]
    \centering
    \vspace{1em}
    Coverage\\
    % First row: 4 images without subcaptions
    \includegraphics[width=0.24\textwidth]{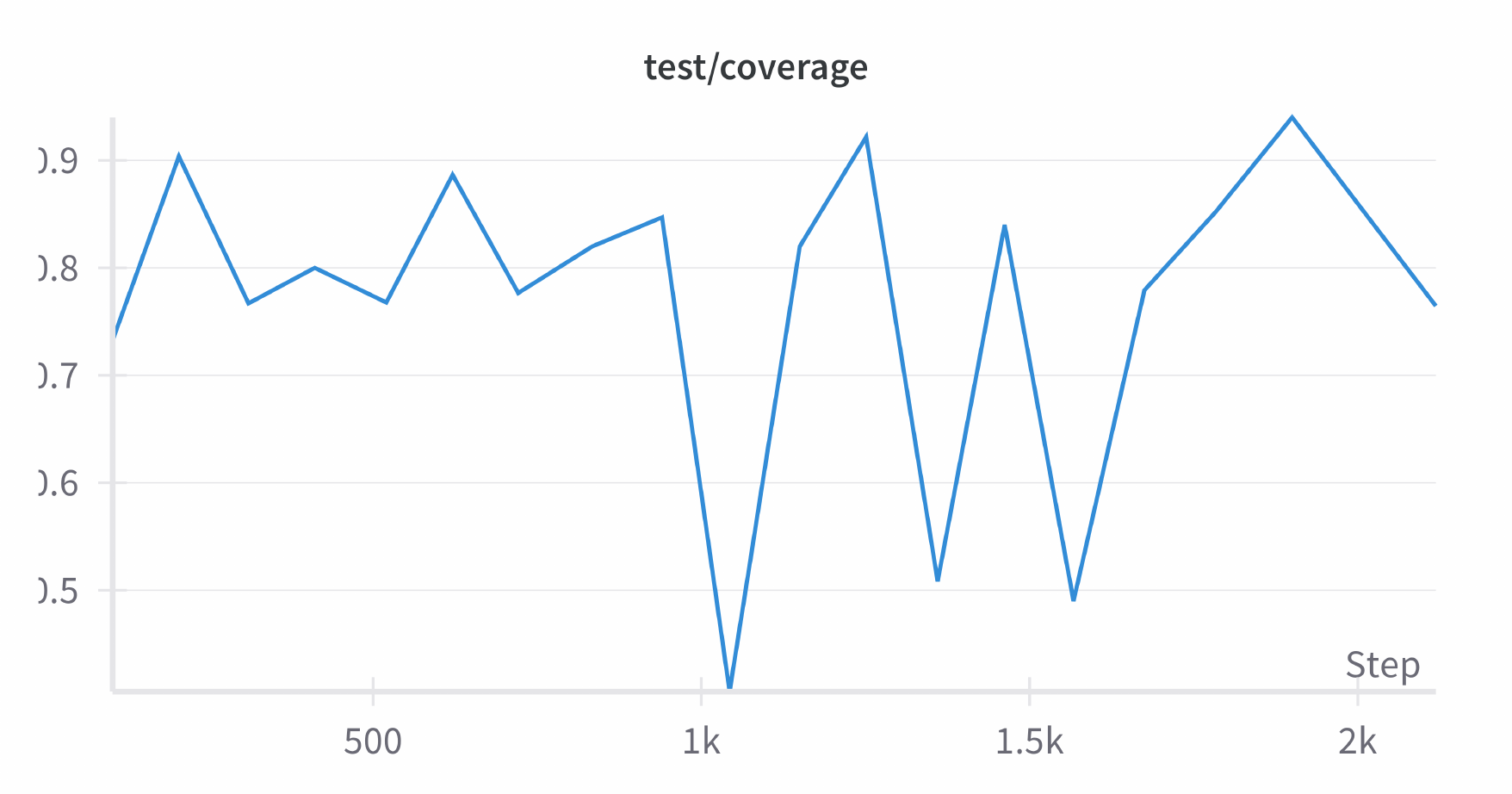}
    \includegraphics[width=0.24\textwidth]{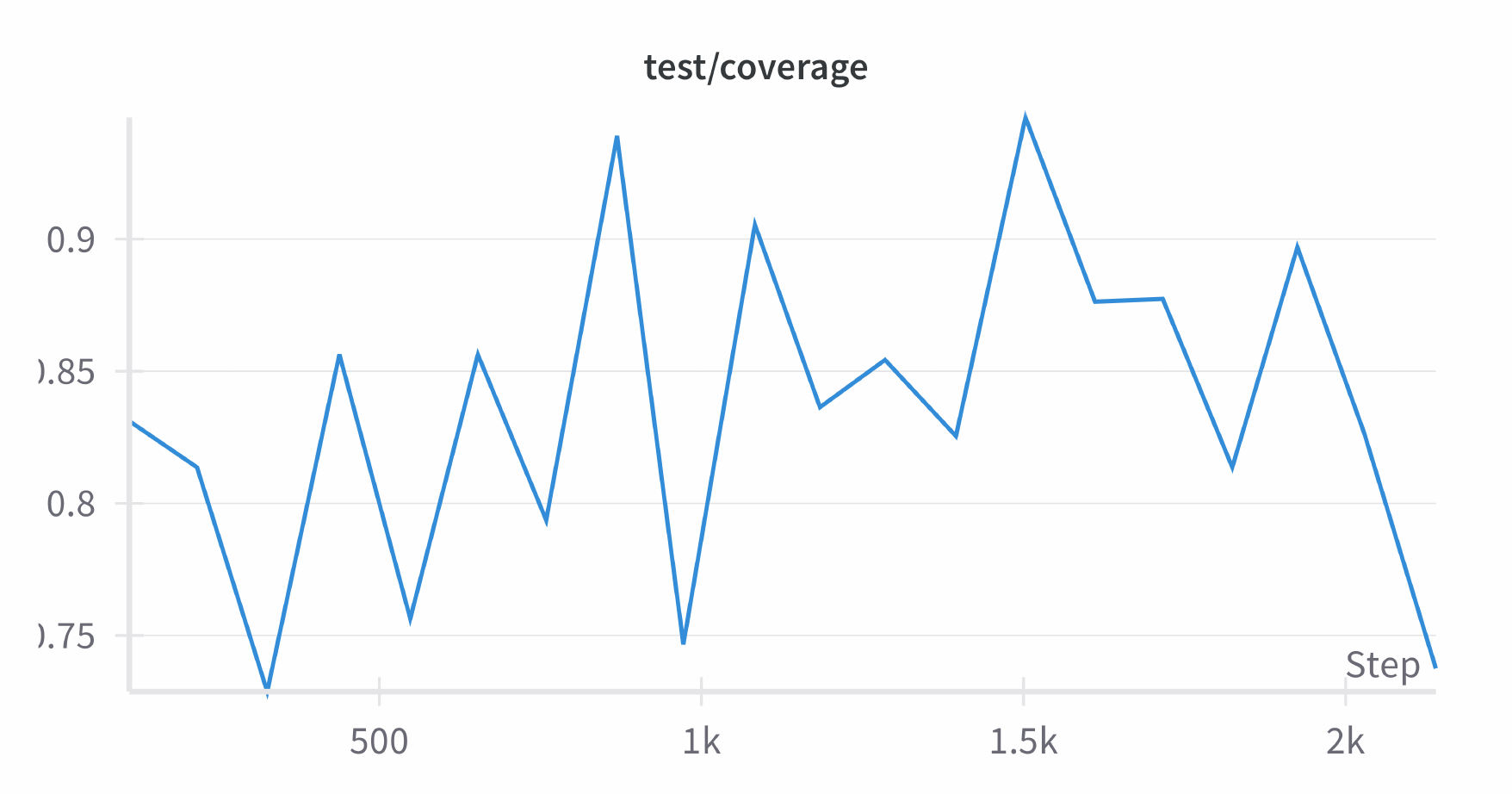}
    \includegraphics[width=0.24\textwidth]{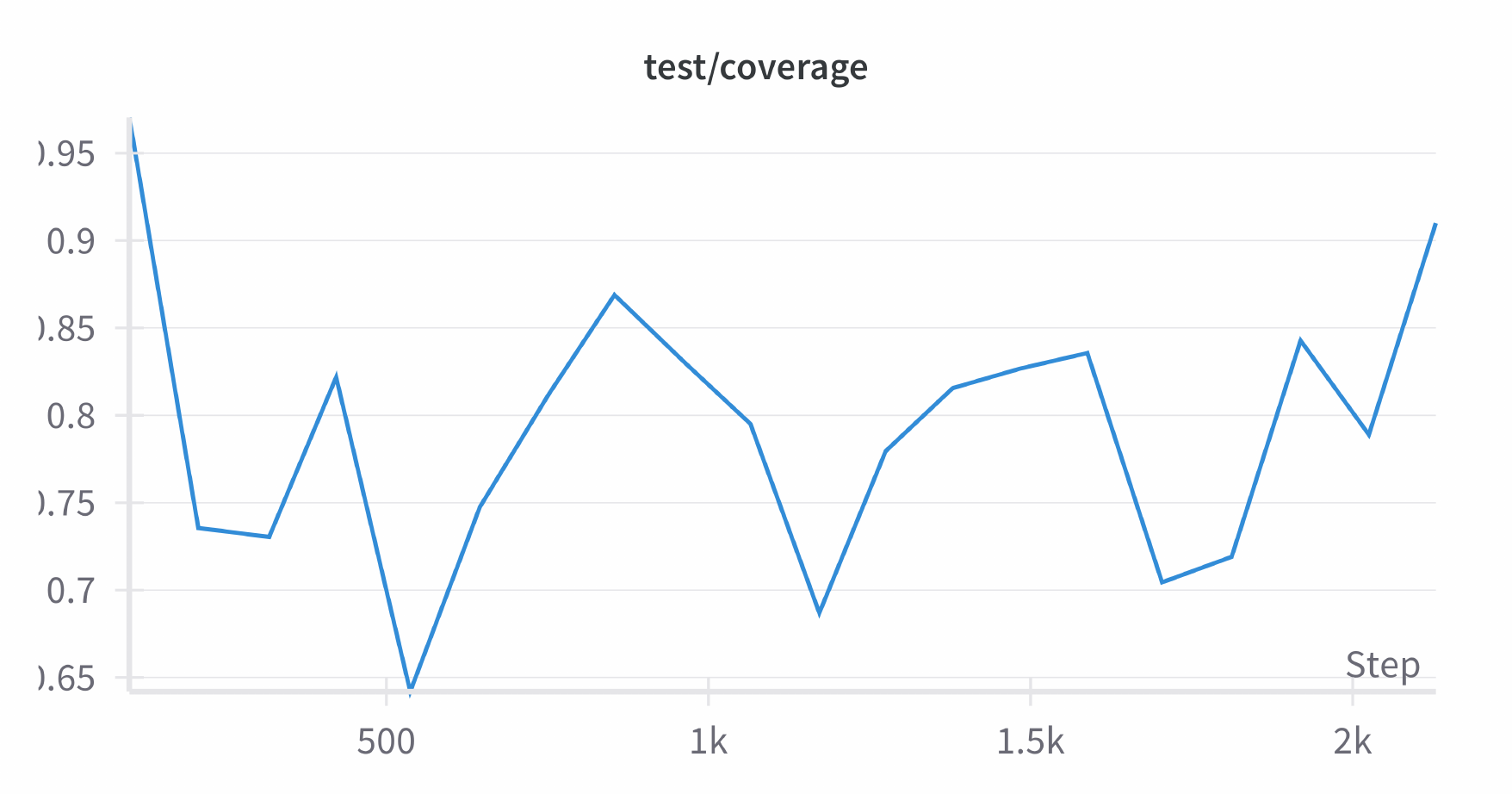}
    \includegraphics[width=0.24\textwidth]{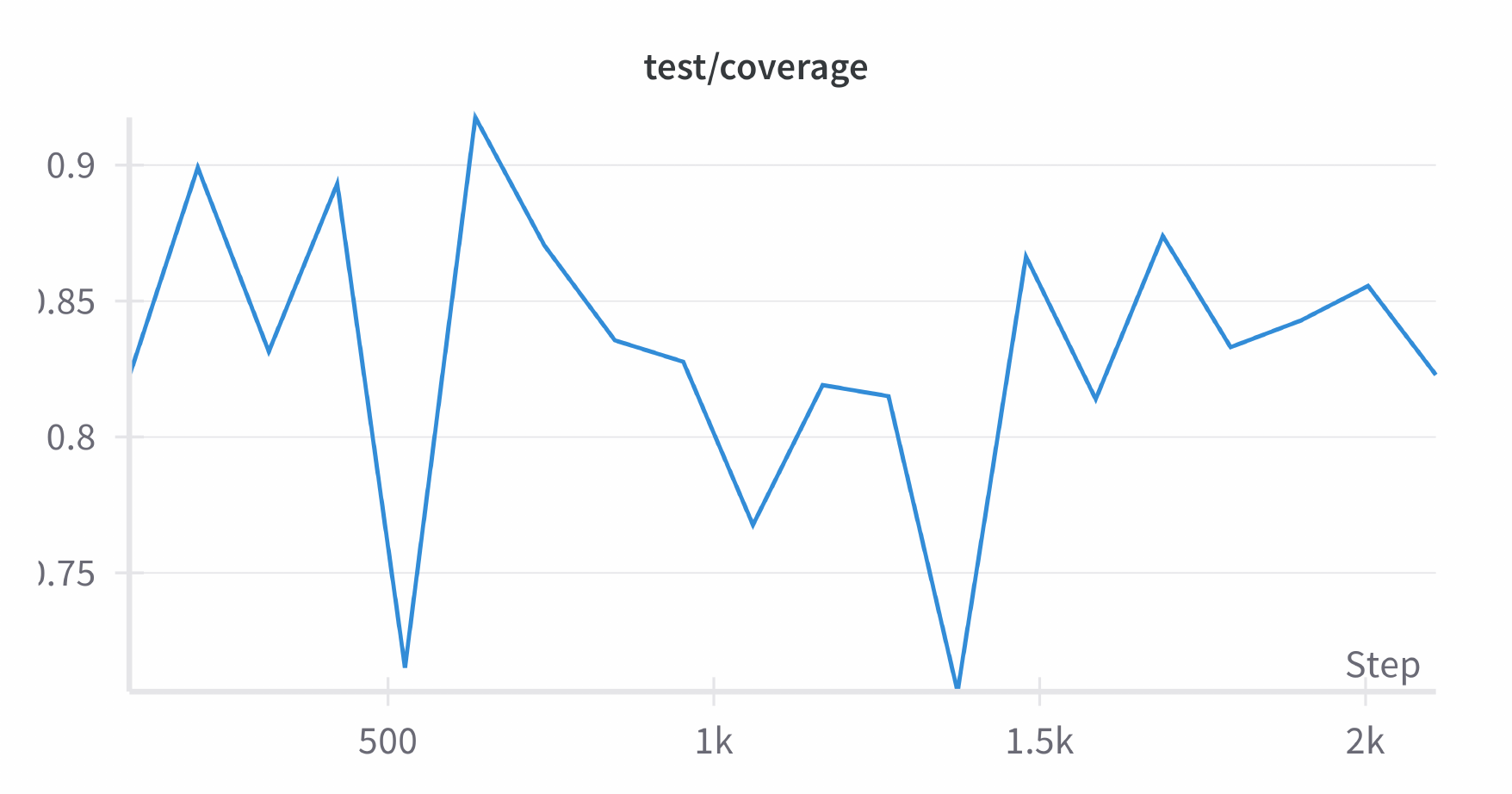}
    \vspace{1em}
    \text{Average Length}\\
    \begin{subfigure}[b]{0.24\textwidth}
        \includegraphics[width=\textwidth]{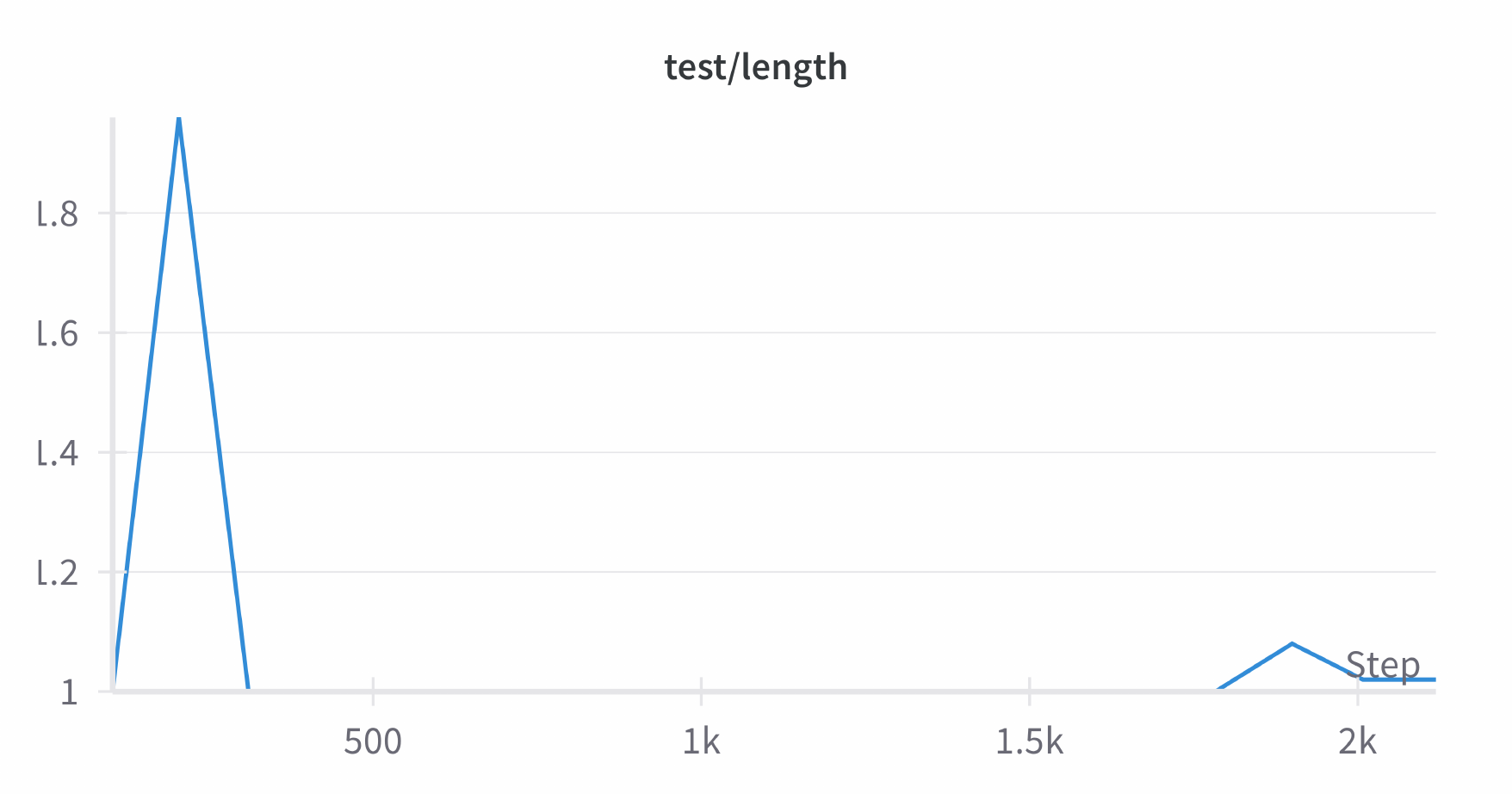}
        \caption{LLaMA-2-7b, $\alpha=0.2$, \\$\lambda=2e^{-4}$.}
    \end{subfigure}
    \begin{subfigure}[b]{0.24\textwidth}
        \includegraphics[width=\textwidth]{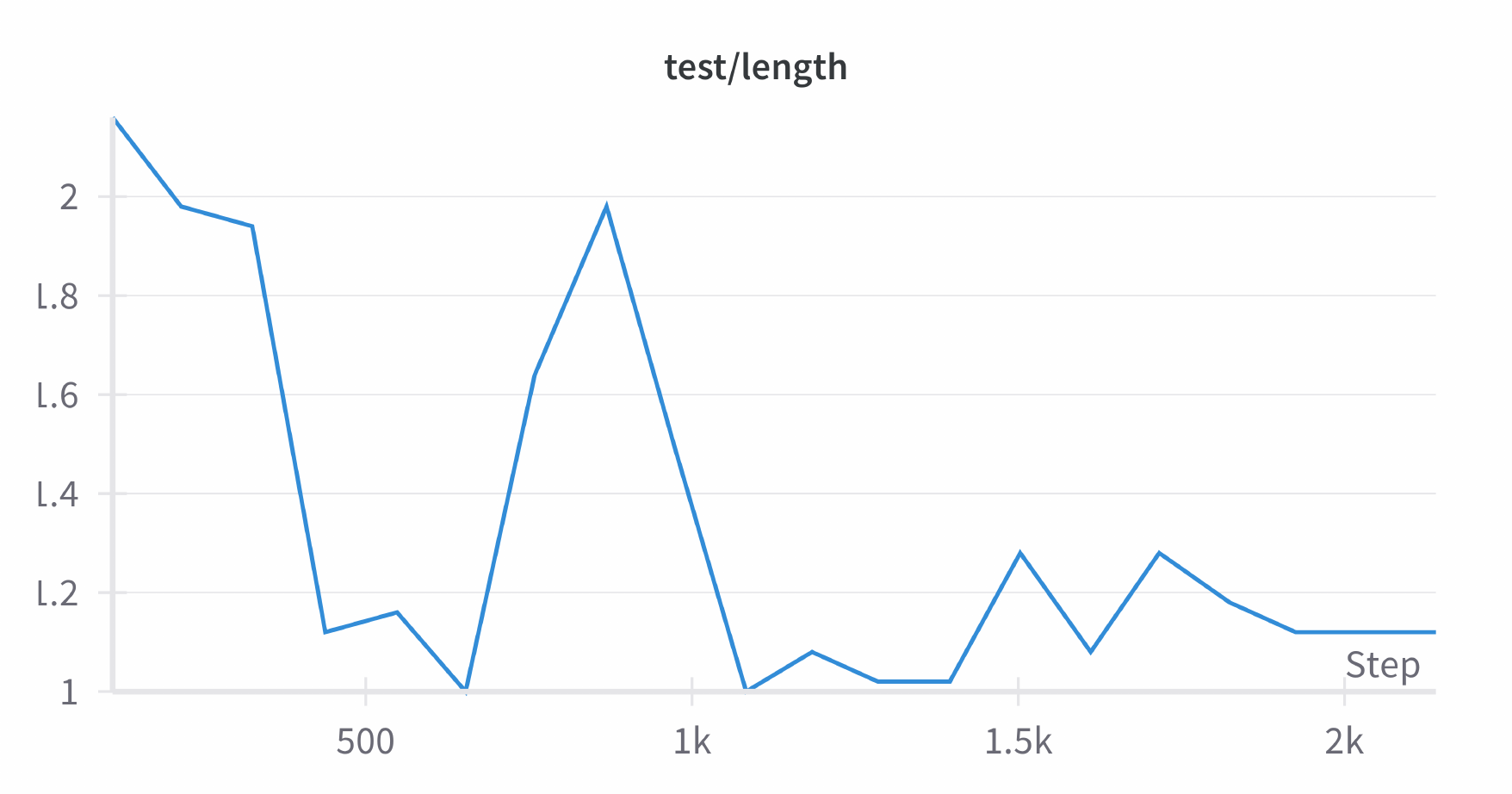}
        \caption{LLaMA-2-7b, $\alpha=0.1$, \\$\lambda=1e^{-4}$.}
    \end{subfigure}
    \begin{subfigure}[b]{0.24\textwidth}
        \includegraphics[width=\textwidth]{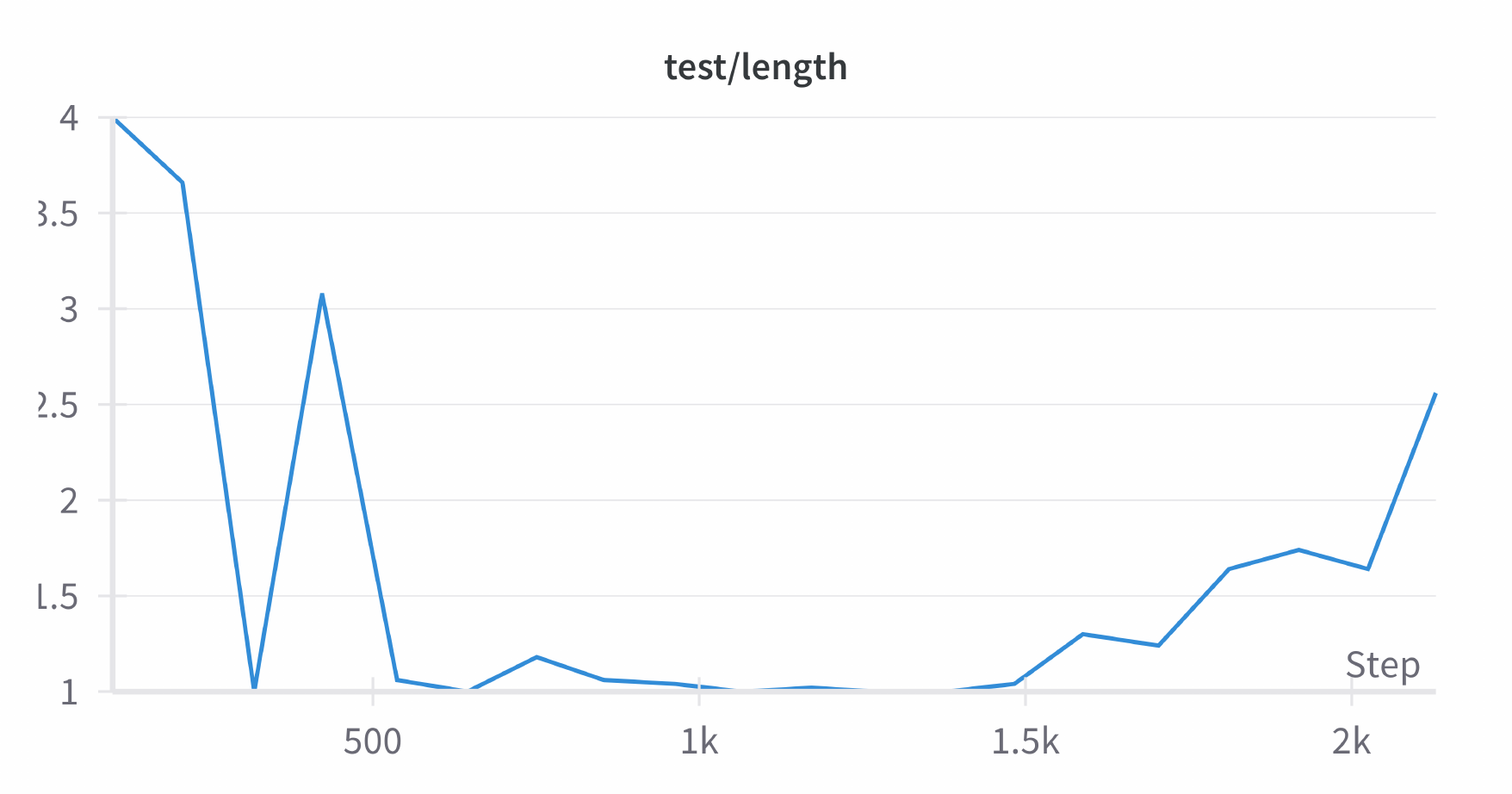}
        \caption{LLaMA-3.2-3b, $\alpha=0.2$, \\$\lambda=1e^{-4}$.}
    \end{subfigure}
    \begin{subfigure}[b]{0.24\textwidth}
        \includegraphics[width=\textwidth]{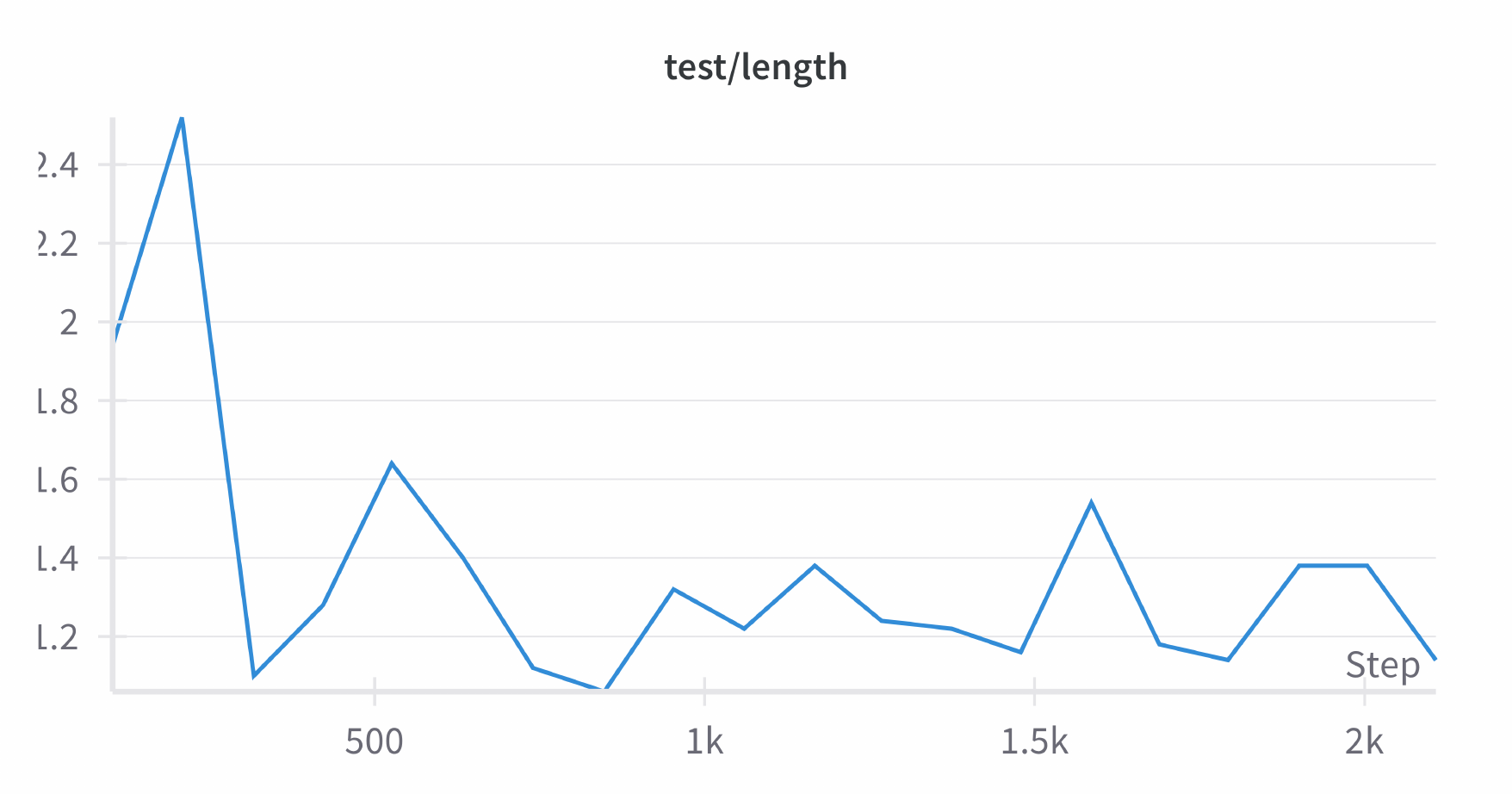}
        \caption{LLaMA-3.2-3b, $\alpha=0.1$, \\$\lambda=1e^{-4}$.}
    \end{subfigure}
    
    \caption{CCPO validation performance on MMLU with varying $\lambda$.}
    \label{fig:val_mmlu_lambda}
\end{figure*}

\end{document}